\documentclass[acmtog,nonacm]{acmart}
\usepackage{multirow}
\usepackage{bm}

\acmJournal{TOG}

\makeatletter
\DeclareRobustCommand\onedot{\futurelet\@let@token\@onedot}
\def\@onedot{\ifx\@let@token.\else.\null\fi\xspace}

\makeatother

\newcommand{\boldparagraph}[1]{\vspace{0.2cm}\noindent{\bf #1:} }

\definecolor{darkgreen}{rgb}{0,0.7,0}

\begin{document}

\title{Latent Dynamics for Full Body Avatar Animation}
\titlenote{Supplementary video available at \href{https://youtu.be/xjnr3YM0yIE}{this link}.}

\author{Shichong Peng}
\authornote{Work was done during an internship at Meta.}
\email{shichong_peng@sfu.ca}
\affiliation{%
  \institution{Simon Fraser University}
  \country{Canada}
}
\author{Chengxiang Yin}
\email{chengxiangyin@meta.com}
\affiliation{%
  \institution{Codec Avatars Lab, Meta}
  \country{USA}
}
\author{Fei Jiang}
\email{feij@meta.com}
\affiliation{%
  \institution{Codec Avatars Lab, Meta}
  \country{USA}
}
\author{Zhongshi Jiang}
\email{jzs@meta.com}
\affiliation{%
  \institution{Codec Avatars Lab, Meta}
  \country{USA}
}
\author{Lingchen Yang}
\email{lcyang@meta.com}
\affiliation{%
  \institution{Codec Avatars Lab, Meta}
  \country{USA}
}
\author{Qingyang Tan}
\email{qytan@outlook.com}
\affiliation{%
  \institution{Codec Avatars Lab, Meta}
  \country{USA}
}
\author{Amin Jourabloo}
\email{jourabloo@meta.com}
\affiliation{%
  \institution{Codec Avatars Lab, Meta}
  \country{USA}
}
\author{Jason Saragih}
\email{jason.saragih@gmail.com}
\affiliation{%
  \institution{Codec Avatars Lab, Meta}
  \country{USA}
}
\author{Ke Li}
\email{keli@sfu.ca}
\affiliation{%
  \institution{Simon Fraser University}
  \country{Canada}
}
\author{Christian Häne}
\email{chaene@meta.com}
\affiliation{%
  \institution{Codec Avatars Lab, Meta}
  \country{USA}
}
\renewcommand{\shortauthors}{Peng et al.}
\begin{abstract}
Pose-driven full-body avatars built on neural rendering produce high-quality novel views of a captured subject.
Yet loose clothing and other dynamic elements deform in ways pose alone cannot explain: the same pose can correspond to many different states, because their motion depends on history, inertia, and contact.
Explicit simulation and layered-garment methods can model such dynamics, but they require either a dedicated garment template, which raw multi-view capture does not naturally provide, or a test-time physics simulator with non-trivial runtime cost. A parallel line of work learns data-driven clothing avatars that avoid explicit garment layers. These methods add an auxiliary latent for variation beyond pose; at inference, they fix it, regress it from pose, or retrieve it from training data, without explicitly modeling how the latent evolves with its own dynamics.
Additionally, even in everyday motion with loose clothing, existing architectures often struggle to capture fine-grained detail, producing blurry renderings and temporal artifacts.
We augment a pose-conditioned 3D Gaussian avatar with a transformer-based decoder and a \emph{dynamics residual latent}. The decoder takes two inputs: user-supplied driving signals, and the residual latent, which captures temporal appearance and geometry variation beyond the driving signals.
At inference, a learned \emph{latent dynamics model} evolves the residual latent from a short pose history and the previous latent state. The model decomposes each update into driving, restoring, and dissipative forces, producing temporally coherent, history-dependent rollouts with negligible added cost.
Different initial conditions yield diverse yet plausible motion trajectories, and the force decomposition exposes controls such as stiffness. Across nine captured sequences of everyday motion with diverse loose garments, quantitative metrics and a perceptual user study show improved animation quality over recent data-driven baselines.
\end{abstract}


\begin{teaserfigure}
  \includegraphics[width=\textwidth]{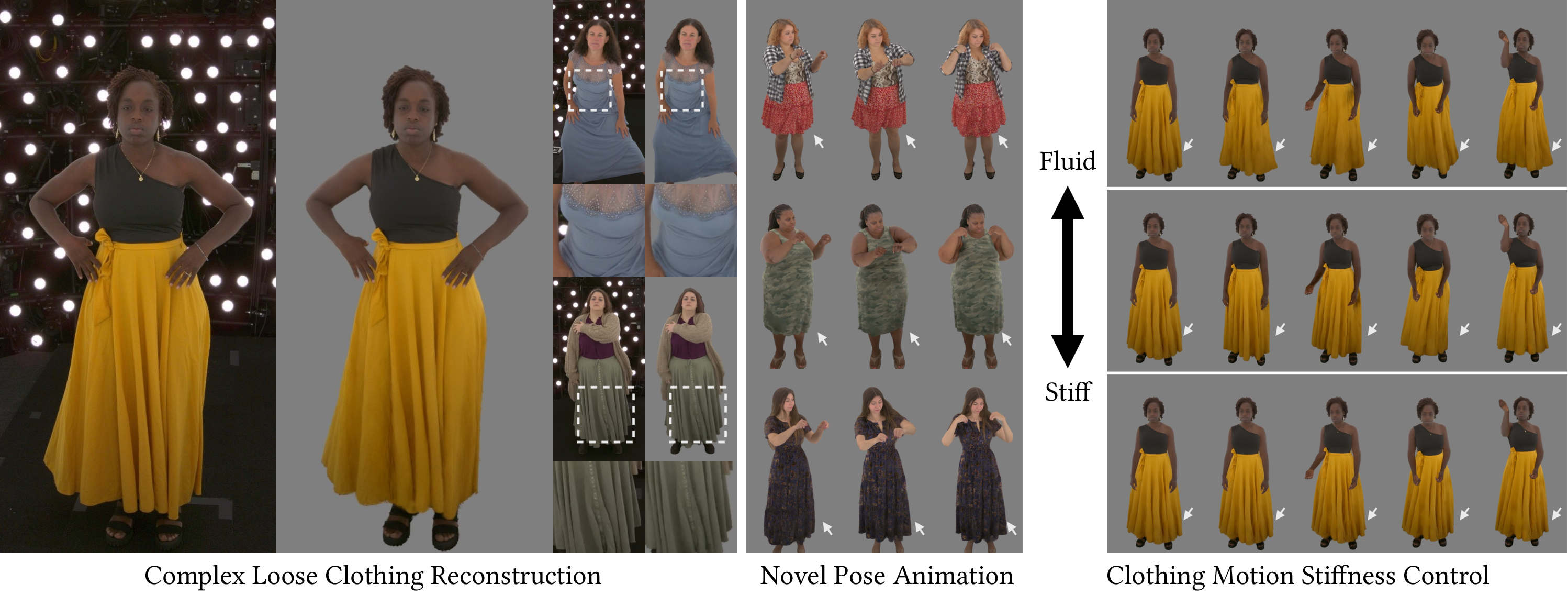}
  \caption{We present a data-driven full-body Gaussian avatar that models dynamic components such as loose clothing, whose deformations are not fully determined by body pose. Our method enables faithful reconstruction of complex garment deformations (left), pose-driven animation with temporally coherent, history-dependent clothing motion (middle), and control over garment behavior such as stiffness (right).}
  \label{fig:teaser}
\end{teaserfigure}


\maketitle

\section{Introduction}
\label{sec:intro}

Pose-driven animatable human avatars turn captured footage of a person into a model that can be rendered from new viewpoints and animated from body poses and facial expressions, enabling telepresence and interactive content creation.
Early systems largely relied on mesh- and template-based representations that integrate naturally with graphics pipelines (e.g., SMPL)~\cite{smpl}.
With the rise of neural rendering, implicit radiance field models sharply improved visual fidelity by learning pose-conditioned deformations and appearance in canonical spaces~\cite{Mildenhall2020NeRFRS,Peng2020NeuralBI,Weng2022HumanNeRFFR,Wang2022ARAHAV,Peng2021AnimatableNR,Peng2022AnimatableIN}, but require costly volumetric rendering.
More recently, 3D Gaussian Splatting (3DGS)~\cite{Kerbl20233DGS} has enabled efficient novel-view synthesis with high quality, and has quickly become a strong foundation for pose-conditioned full-body avatars~\cite{zielonka25dega,Hu2023GauHumanAG,Qian20233DGSAvatarAA,Zheng2024PhysAvatarLT}.

Real subjects, however, include dynamic components such as loose clothing whose motion depends on history, inertia, and contact, which pose-driven models struggle to reproduce.
One line of work addresses loose clothing with explicit garment layers or physics-based simulation~\cite{xiang2022dressing,xiang2021modeling,lee2025mpmavatar,Zheng2024PhysAvatarLT}. These methods improve physical plausibility and controllability, but they require either a garment template with an explicit body--clothing split, which raw multi-view capture does not naturally provide, or a physics simulator at test time with non-trivial runtime cost.
A parallel line of work learns \emph{data-driven clothing avatars} directly from capture data, avoiding explicit garment layers and test-time simulation.
Because the same pose can correspond to many different garment states, these methods condition the avatar on pose together with an auxiliary latent code that captures the appearance and geometry variation beyond what pose specifies.
At test time, prior work treats this latent narrowly: fixing it~\cite{Zhu2025UMAUH}, regressing it from pose history~\cite{Chen2024WithinTD,xu2025seqavatar,kwon2026dynavatar}, or retrieving it from training sequences~\cite{Zhan2025R3AvatarRA}.
In each case, the latent has no temporal dynamics of its own, limiting history-dependent garment behavior.
Even methods that do learn a transition over the latent~\cite{Li2025RealityAvatarTR} predict updates in a feature space that entangles pose and clothing rather than explicitly modeling the latent's evolution, making garment behavior less controllable.

We study this gap in a practical setting common to everyday human activity: motion such as reaching, casual stepping, and conversational gesturing, where loose garments move dynamically but not at extreme amplitude. We capture nine subjects performing such motions to study this setting. Even in this regime, existing pose-driven architectures are limited: they struggle to reconstruct fine garment deformations, producing blurred renderings and temporal artifacts where garment motion changes quickly.

We propose a data-driven full-body avatar that models dynamic deformations, especially those arising from loose clothing. We augment a 3DGS avatar~\cite{Wang2025RelightableFG} with a transformer-based decoder and a \emph{dynamics residual latent}. The transformer architecture captures fine deformation details in the data.
The decoder takes two inputs: user-supplied driving signals (body and face keypoints), and the residual latent that captures temporal appearance and geometry variation beyond the driving signals.
At the core of our method, a \emph{latent dynamics model} predicts how the residual latent evolves under pose driving.
The model decomposes each update into pose-dependent driving, restoring, and dissipative forces, predicted from a short pose history together with the previous latent state. We then integrate these forces over time to update the latent.
During inference, we roll out this dynamics model autoregressively along an input pose sequence to produce temporally coherent loose-clothing motion with minimal computational overhead.
Varying the initial residual latent produces diverse yet plausible garment motions for the same pose sequence. The force-component parameterization also provides control over garment behavior, such as stiffness, damping, and responsiveness to pose changes.
Across nine subjects with diverse loose garments, quantitative and perceptual evaluations show improved animation quality over recent data-driven baselines.

\noindent\textbf{Contributions.}
We
(i) propose a \emph{latent dynamics model} for test-time evolution of a \emph{dynamics residual latent} in pose-driven avatars, producing temporally coherent, history-dependent rollouts;
(ii) introduce a transformer-based architecture to model fine garment deformations, and
(iii) enable controllable and diverse loose clothing motion for everyday human activity, validated on real-world captures.

\section{Related Work}
\label{sec:related_work}

\subsection{Animatable Human Reconstruction}

\boldparagraph{Mesh-based avatars}
Mesh and template-based avatars are a long-standing foundation, compatible with established rendering and animation pipelines.
Early methods reconstruct person-specific templates and animate them using simulation or example-based strategies~\cite{stoll2010video,xu2011video,Guan2012DRAPE}.
More recent approaches learn motion-dependent texture and deformation on tracked templates from multi-view capture~\cite{Bagautdinov2021DrivingsignalAF,habermann2021real}.
These pipelines are effective under dense observations and reliable tracking, yet they demand substantial capture and tracking effort, and limited template resolution constrains fine-grained clothing detail.

\boldparagraph{Implicit neural avatars}
Implicit-field neural rendering, such as NeRF~\cite{Mildenhall2020NeRFRS} and SDF variants, represents geometry and appearance in a canonical space that is warped to the target pose for animation. NeuralBody~\cite{Peng2020NeuralBI} and HumanNeRF~\cite{Weng2022HumanNeRFFR} pioneered pose-conditioned implicit avatars, and later extensions added articulated SDFs~\cite{Wang2022ARAHAV}, deformation models for video supervision~\cite{Peng2021AnimatableNR,Peng2022AnimatableIN}, and 4D feature grids for long-motion fidelity~\cite{Isik2023HumanRF}. These methods deliver high fidelity, but ray-marching is sampling- and evaluation-heavy, inflating inference latency.

\boldparagraph{Point-based and 3DGS avatars}
Point-based renderers, especially 3D Gaussian Splatting (3DGS)~\cite{Kerbl20233DGS}, have enabled high-quality rendering with favorable speed--quality trade-offs. Recent work combines 3DGS with articulated deformation models such as LBS, cages, or templates for pose-driven avatars~\cite{zielonka25dega,Hu2023GauHumanAG,Qian20233DGSAvatarAA,Lei2023GARTGA}. Beyond per-subject fitting, generalizable formulations such as GPS-Gaussian~\cite{Zheng2024GPSGaussian} regress pixel-aligned Gaussians from sparse views for real-time novel-view synthesis of unseen humans. Most such formulations remain pose-conditioned, which yields a single deformation per pose and cannot represent history-dependent loose-clothing motion.

\subsection{Modeling Dynamic Motion in Pose-Driven Avatars}

Real-subject capture exhibits dynamic motion underdetermined by pose. These include hair and soft-tissue motion, but loose clothing has received the most attention in pose-driven avatar work and is our focus here. We organize the discussion into: model-based approaches, which introduce explicit garment structure or run physics simulation at test time, and data-driven approaches, which learn deformation behavior directly from capture data.

\boldparagraph{Physics-based cloth simulation}
Physics simulation is the most established approach to clothing dynamics. Classic formulations include position-based methods~\cite{xpbd}, energy-based methods with robust contact handling~\cite{li2020incremental,li2020codimensional}, and continuum methods such as the material point method (MPM)~\cite{jiang2016material}.
Recent work brings such physical priors into avatar reconstruction and dynamics estimation~\cite{Zheng2024PhysAvatarLT,lee2025mpmavatar}, and the simulation community continues to push toward higher resolution and better run-time performance, e.g., via domain-decomposed projective dynamics on CPU~\cite{Lu2025DDPD} and physics-inspired estimation of optimal mesh resolution~\cite{Zhang2025PIE}.
Physics-based pipelines offer strong physical plausibility, yet running a simulator at test time, with its iterative solvers and contact handling, introduces non-trivial computational overhead.

\boldparagraph{Structured garment modeling}
A complementary line of work introduces explicit garment structure into either the representation or the learned dynamics.
On the representation side, mesh-based formulations treat clothing as a separate entity from the body to capture garment-specific deformation~\cite{xiang2021modeling,xiang2022dressing,halimi2022pattern}, while neural-field variants such as ULNeF~\cite{Santesteban2022ULNeF} achieve analogous garment separation and untangling using implicit fields rather than explicit meshes.
A parallel branch amortizes cloth dynamics on garment meshes given the body geometry, e.g., via hierarchical graph networks trained with physics-based self-supervision~\cite{Grigorev2023HOOD} or bone-driven networks supervised on simulated cloth sequences with run-time control over physical parameters such as bending stiffness~\cite{Pan2022BoneDriven}.
Supervision and primitives differ across these methods, yet they share a common reliance on per-garment structure (a known mesh template or pre-defined neural-field layering) that must be authored or registered up front.
Our setting is complementary: we learn directly from raw multi-view capture, without an underlying garment mesh, hand annotation, or per-garment template.

\boldparagraph{Data-driven approaches}
Data-driven avatars learn dynamic behavior of loose clothing directly from capture, without garment layers or test-time simulators.
Some works enrich the driving signal; others use latent variables or temporal context for history-dependent clothing motion.
ToMiE~\cite{zhan2024tomiemodulargrowthenhanced} grows the kinematic scaffold with extra joints, making the driving signal more expressive so additional joints explain clothing deformations; richer rigging alone, however, provides no explicit evolution model at test time.
Dyco~\cite{Chen2024WithinTD} conditions neural avatars on delta-pose sequences with a localized context encoder to capture inertia-induced appearance changes, Seq-Avatar~\cite{xu2025seqavatar} predicts 3DGS deformations from hierarchical motion context, and DynaAvatar~\cite{kwon2026dynavatar} feeds a fixed-length pose-history window into a feed-forward transformer decoder; all three, however, infer deformation directly from pose history rather than exposing an explicit residual-latent with a rolloutable transition model.
UMA~\cite{Zhu2025UMAUH} introduces learnable per-frame latents to absorb pose-to-surface stochasticity, but zeros them during novel-pose animation, so the latent remains static under driving.
RealityAvatar~\cite{Li2025RealityAvatarTR} applies an LSTM to predict latent transitions in an aggregated feature space that mixes pose cues with a randomly initialized, fixed latent code; because the LSTM outputs only this mixed feature, the method neither explicitly models latent evolution nor supports motion control.
$R^3$-Avatar~\cite{Zhan2025R3AvatarRA} uses a temporal codebook and retrieves entries at test time, restricting latents to those seen during training rather than rolling out a residual latent under pose driving.
These approaches leave the latent's evolution either absent or implicit; we instead introduce an explicit \emph{latent dynamics model} that predicts how it evolves under pose driving, yielding diverse and controllable garment motion at inference.

\section{Method}
\label{sec:method}

\begin{figure*}[t]
    \centering
    \includegraphics[width=\textwidth]{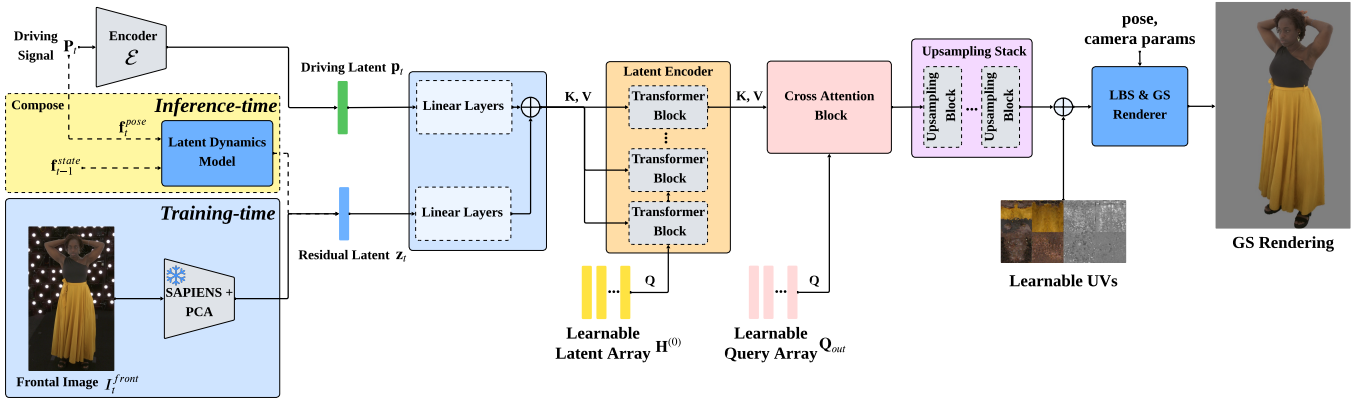}
    \caption{\textbf{Overview of our pipeline.}
    The driving signal $\mathbf{P}_t$ is encoded into a driving latent $\mathbf{p}_t$, which together with a residual latent $\mathbf{z}_t$ feeds a transformer decoder whose output is upsampled via convolutional blocks into Gaussian corrective UV maps; the corrected Gaussians are then posed via LBS and splatted to render the avatar.
    During training (bottom), $\mathbf{z}_t$ is extracted from a frontal-view image via a SAPIENS+PCA pipeline (Sec.~\ref{sec:method_arch}).
    At inference (top-dotted), $\mathbf{z}_t$ is instead synthesized by our latent dynamics model (Sec.~\ref{sec:method_dynamics}) from pose history and the previous-state feature $\mathbf{f}^{\text{state}}_{t-1}$ (the previous latent and its velocity).}
    \label{fig:decoder_arch}
\end{figure*}

We build on the UV-parameterized 3D Gaussian avatar of Wang et al.~\cite{Wang2025RelightableFG} (Sec.~\ref{sec:method_background}) and replace its decoder with a transformer-based architecture that takes two inputs: a \emph{primary} driving signal supplied by the user, and a \emph{secondary} residual latent $\mathbf{z}_t$ supplied by our pipeline (Sec.~\ref{sec:method_arch}). At training time we extract $\mathbf{z}_t$ from a frontal capture image. At inference our learned latent dynamics model (Sec.~\ref{sec:method_dynamics}) produces $\mathbf{z}_t$ from pose history and prior latent state, so the test-time input from the user is the driving signal $\mathbf{P}_t$ alone. Fig.~\ref{fig:decoder_arch} gives an overview of the full pipeline. The avatar and dynamics model are both trained per subject.

\subsection{Background: UV-Parameterized Gaussian Avatar}
\label{sec:method_background}

We build on a UV-parameterized 3D Gaussian avatar representation~\cite{Wang2025RelightableFG} rendered with 3DGS~\cite{Kerbl20233DGS}. 
The avatar pipeline takes a target driving signal $\mathbf{P}_t$ (body and face keypoints, following Wang et al.~\cite{Wang2025RelightableFG}) as input and encodes it into a driving latent $\mathbf{p}_t$ via a lightweight encoder $\mathcal{E}$. A decoder then maps $\mathbf{p}_t$ to UV-space Gaussian correctives $\Delta\mathbf{U}_t$, which are applied in the canonical frame and posed via linear blend skinning (LBS) for rendering. We follow a similar geometry setup but use a simplified appearance model based on standard 3DGS spherical harmonics (details below).

\boldparagraph{UV-anchored Gaussian primitives}
Let a tracked template mesh define a UV parameterization. We associate each UV texel index $k \in \{1,\dots,M\}$ with one 3D Gaussian primitive, storing geometric parameters (translation $\mathbf{t}_k \in \mathbb{R}^3$, rotation $\mathbf{R}_k \in SO(3)$, anisotropic scale $\mathbf{s}_k \in \mathbb{R}^3$, opacity $o_k \in [0,1]$) and appearance parameters (SH coefficients for color). Per-texel parameters are stored as UV maps and updated by predicting per-texel corrective maps from the driving signal and residual latent (Sec.~\ref{sec:method_arch}).

\boldparagraph{Gaussian splatting renderer}
Each primitive is modeled as an unnormalized 3D Gaussian $\mathcal{G}_k(\mathbf{x}) = \exp\!\left(-\tfrac{1}{2}(\mathbf{x}-\mathbf{t}_k)^{\top}\Sigma_k^{-1}(\mathbf{x}-\mathbf{t}_k)\right)$ with $\Sigma_k = \mathbf{R}_k \,\mathrm{diag}(\mathbf{s}_k)^2\, \mathbf{R}_k^{\top}$. Following 3DGS~\cite{Kerbl20233DGS}, primitives are projected to the image plane via a screen-space footprint~\cite{ewaSplatting} and composited back-to-front with standard alpha blending to yield the rendered pixel color $\mathbf{C}(\mathbf{u})$.
We parameterize per-Gaussian color with 3DGS-style spherical harmonics up to degree 2: a learnable UV texture stores the degree-0 coefficient $\rho_0(k)\in\mathbb{R}^3$ for each texel, and the network predicts the remaining coefficients $\{\mathbf{c}_{k,lm}\}_{l\ge 1}$. The view direction $\omega$ is only used to evaluate these coefficients during rendering; our networks do not take $\omega$ as input.

\boldparagraph{Pose deformation (canonical-to-posed)}
We model the avatar in a canonical template space and deform it to a target pose using standard skeletal skinning. Each UV texel $k$ inherits a surface point $\mathbf{v}_k$ and a local tangent frame $\mathbf{TBN}_k=[\bar{\mathbf{t}}_k,\bar{\mathbf{b}}_k,\bar{\mathbf{n}}_k]$ from the tracked mesh. The decoder predicts per-texel local correctives (e.g., a translation offset $\delta\mathbf{t}_k$ and a rotation $\delta\mathbf{R}_k$) in this local frame, which are mapped to 3D as
\begin{align}
    \mathbf{t}_k &= \mathbf{v}_k + \mathbf{TBN}_k\, \delta \mathbf{t}_k, \qquad
    \mathbf{R}_k = \mathbf{TBN}_k\, \delta \mathbf{R}_k.
\end{align}
The posed avatar is obtained by applying the pose-dependent skinning transform to these canonical primitives (details below).

\subsection{Residual-Latent-Augmented Decoder}
\label{sec:method_arch}

Alongside the driving latent $\mathbf{p}_t$, the decoder consumes a residual latent $\mathbf{z}_t\in\mathbb{R}^{d_z}$ ($d_z=128$). The two latents are fed jointly to a transformer-based architecture that produces the Gaussian corrective UV maps $\Delta \mathbf{U}_t$.

\boldparagraph{Residual latent extraction}
At training time, we extract $\mathbf{z}_t$ from a fixed frontal-view RGB capture $I_t^{\text{front}}$ at the same timestamp $t$. We first compute a feature vector $\mathbf{f}_t=\Phi(I_t^{\text{front}})$ using a frozen pretrained SAPIENS~\cite{khirodkar2024sapiens} feature extractor $\Phi(\cdot)$. We then apply PCA and standardize the result:
\begin{align}
    \mathbf{z}'_t = \mathbf{W}\big(\mathbf{f}_t-\boldsymbol{\mu}_f\big), \qquad
    \mathbf{z}_t = \frac{\mathbf{z}'_t - \boldsymbol{\mu}_z}{\boldsymbol{\sigma}_z + \epsilon},
\end{align}
where $\mathbf{W}$ are the PCA projection weights and $\boldsymbol{\mu}_f$ is the mean of $\mathbf{f}_t$ in feature space. Dataset statistics are computed over all PCA features from the (per-subject) training set, with per-dimension mean $\boldsymbol{\mu}_z$ and standard deviation $\boldsymbol{\sigma}_z$ (the division is element-wise; $\epsilon$ is a small constant). We use $\mathbf{z}_t$ as a conditioning coordinate: frames with similar frontal appearance get nearby latents, and the decoder learns to read this neighborhood structure alongside $\mathbf{p}_t$. We never invert $\mathbf{z}_t$ back to pixels; the coordinate is all the decoder consumes. The SAPIENS encoder $\Phi$ and the PCA projection run at training time only. At test time the user supplies $\mathbf{P}_t$; the dynamics model (Sec.~\ref{sec:method_dynamics}) supplies $\mathbf{z}_t$. No frontal image is encoded at inference, and no per-frame retrieval against training latents is performed.

\boldparagraph{Transformer-based decoder architecture}
Inspired by the Perceiver-IO design~\cite{Jaegle2022PerceiverIO}, we first map the two inputs into a set of input tokens $\mathbf{X}_t$ and project them to keys and values:
\begin{align}
    \mathbf{X}_t &= \phi_{\text{in}}([\mathbf{p}_t;\mathbf{z}_t]) \in \mathbb{R}^{N_{\text{in}}\times d}, \\
    \mathbf{K}_t &= \mathbf{X}_t \mathbf{W}_K, \qquad \mathbf{V}_t = \mathbf{X}_t \mathbf{W}_V,
\end{align}
where $\phi_{\text{in}}$ is a lightweight MLP that produces $N_{\text{in}}$ input tokens and $d$ is the model width. We use a learnable latent array $\mathbf{H}^{(0)}\in\mathbb{R}^{N_{\text{latent}}\times d}$ whose rows are the initial queries (Perceiver ``latent array''). Each transformer block $\ell=1,\dots,L$ applies (i) cross-attention from the latent array to the input tokens, followed by (ii) self-attention within the latent array:
\begin{align}
    \mathbf{H}^{(\ell)} &= \mathrm{SelfAttn}\!\left(\mathrm{CrossAttn}(\mathbf{H}^{(\ell-1)}, \mathbf{K}_t, \mathbf{V}_t)\right).
\end{align}
To produce spatially dense outputs, we introduce a second learnable query array $\mathbf{Q}_{\text{out}}\in\mathbb{R}^{N_{\text{out}}\times d}$ (Perceiver ``output query array'') and perform a final cross-attention where keys/values come from the processed latent array:
\begin{align}
    \mathbf{O}_t = \mathrm{CrossAttn}(\mathbf{Q}_{\text{out}}, \mathbf{H}^{(L)}\mathbf{W}'_K, \mathbf{H}^{(L)}\mathbf{W}'_V) \in \mathbb{R}^{N_{\text{out}}\times d}.
\end{align}
We reshape $\mathbf{O}_t$ into a 2D feature map $\mathbf{F}_t \in \mathbb{R}^{H \times W \times d}$ (with $N_{\text{out}} = HW$) and feed it through a series of convolutional upsampling blocks. Each block consists of bilinear upsampling followed by convolution layers with skip connections, and the network outputs a set of UV maps $\Delta \mathbf{U}_t = \text{UpsampleBlocks}(\mathbf{F}_t)$.
Each texel $k$ in $\Delta \mathbf{U}_t$ parameterizes per-Gaussian correctives applied in the local tangent frame (Sec.~\ref{sec:method_background}), including a translation offset $\delta\mathbf{t}_k$, a rotation offset $\delta\mathbf{R}_k$, opacity $o_k$, scale $\mathbf{s}_k$, and residual SH coefficients (degree $\ge 1$) for appearance. These corrected Gaussian parameters are then rendered using the splatting pipeline described in Sec.~\ref{sec:method_background}.

\boldparagraph{Avatar model training objective}
We train the Gaussian avatar parameters together with the decoder using multi-view supervision from the captured sequences. For each training frame, we render the predicted Gaussians into each camera view and optimize a weighted combination of photometric reconstruction loss (i.e., $\ell_1$), perceptual loss~\cite{Zhang2018TheUE}, and a foreground mask loss computed from the rendered alpha map. During this stage, we condition on the training residual latents $\mathbf{z}_t$, while test-time animation uses synthesized latents from the dynamics model (Sec.~\ref{sec:method_dynamics}).

\subsection{Latent Dynamics Model}
\label{sec:method_dynamics}

At inference, our latent dynamics model autoregressively produces the residual-latent trajectory ${\mathbf{z}_t}$ from pose history and the previous latent state. Because the per-frame latent targets are precomputed, we train one dynamics model per subject in parallel with the avatar.

\boldparagraph{Model inputs}
We operate in the standardized latent space defined in Sec.~\ref{sec:method_arch}. The dynamics model maintains a second-order state $(\mathbf{z}_t,\mathbf{v}_t)$, where $\mathbf{v}_t\in\mathbb{R}^{d_z}$ is the latent velocity. In addition to a compact pose descriptor $\mathbf{f}^{\text{pose}}_t$ computed from a short body-keypoint history (see supplementary material for construction details), the model takes as input a previous-state feature
$\mathbf{f}^{\text{state}}_{t-1}=[\mathbf{z}_{t-1};\,\mathbf{v}_{t-1};\,\|\mathbf{v}_{t-1}\|_2]\in\mathbb{R}^{2d_z+1}$.

\boldparagraph{Latent update rule}
We parameterize the latent update as a spring-damper ordinary differential equation, treating the latent as a particle attached to a reference (rest) latent $\mathbf{z}_{\text{ref}}$. Intuitively, loose garments often swing around a resting state and respond to body motion, so the latent is pulled toward $\mathbf{z}_{\text{ref}}$ while being driven by pose changes. This structure is designed to promote stable rollouts and aid generalization, as we evaluate in Sec.~\ref{sec:experiment}.
Concretely, the net force is decomposed into a pose driving force, a damping force, and a spring force:
\begin{align}
\label{eq:latent_dynamics}
\mathbf{F}_{\text{pose},t} &= \mathbf{g}_t,\quad
\mathbf{F}_{\text{damping},t} = \mathbf{c}_t \odot \mathbf{v}_t,\quad
\mathbf{F}_{\text{spring},t} = \boldsymbol{\kappa}_t \odot (\mathbf{z}_t - \mathbf{z}_{\text{ref}}), \notag\\
\mathbf{a}_t &= \bigl(\mathbf{F}_{\text{pose},t}
- \mathbf{F}_{\text{damping},t}
- \mathbf{F}_{\text{spring},t}\bigr) \oslash \mathbf{m}_t, \notag\\
\mathbf{v}_{t+1} &= \mathbf{v}_t + \Delta t\, \mathbf{a}_t, \notag\\
\mathbf{z}_{t+1} &= \mathbf{z}_t + \Delta t\, \mathbf{v}_{t+1}.
\end{align}
where $\odot$ and $\oslash$ denote element-wise product and division. Here $\mathbf{g}_t, \boldsymbol{\kappa}_t,\mathbf{c}_t,\mathbf{m}_t \in \mathbb{R}^{d_z}$ are per-dimension pose driving force, stiffness, damping, and mass, respectively, and we set $\Delta t = 1$.
The reference $\mathbf{z}_{\text{ref}}$ represents a rest latent; in our experiments, we set $\mathbf{z}_{\text{ref}}$ per subject to the standardized SAPIENS-PCA latent extracted from the first frame of that subject's training sequence, which is roughly stationary.
Although $\mathbf{g}_t, \boldsymbol{\kappa}_t, \mathbf{c}_t, \mathbf{m}_t$ are learned per-timestep and do not correspond to physical quantities with SI units, the three forces play distinct roles (pose-driven, restoring, and dissipative), which we validate by scaling each individually (Sec.~\ref{sec:experiment}). The same structure exposes user controls over stiffness, damping, and responsiveness.

\boldparagraph{Neural parameterization}
All force parameters are predicted by a lightweight neural network that takes the concatenated features $[\mathbf{f}^{\text{pose}}_t;\mathbf{f}^{\text{state}}_{t-1}]$ as input. Concretely, we use multiple small MLPs that output each of $\mathbf{g}_t, \boldsymbol{\kappa}_t,\mathbf{c}_t,\mathbf{m}_t$. We enforce positivity of $\boldsymbol{\kappa}_t,\mathbf{c}_t,\mathbf{m}_t$ via a softplus layer at the end, so the parameters remain non-negative throughout training.

\boldparagraph{Training}
During training, we supervise the dynamics using the standardized SAPIENS-PCA latent sequence $\{\mathbf{z}^\star_t\}$ extracted from training frames. We train with teacher forcing and multi-step rollouts to reduce exposure bias: we gradually increase the rollout horizon while decreasing the teacher forcing probability. We minimize an MSE loss in latent space,
\begin{align}
    \mathcal{L}_{\text{dyn}} = \frac{1}{T}\sum_{t=1}^{T}\left\|\mathbf{z}_t - \mathbf{z}^\star_t\right\|_2^2,
\end{align}
computed over rollout steps within each motion clip.

\boldparagraph{Inference}
At test time, given a driving signal sequence $\{\mathbf{P}_t\}$ and an initial state $(\mathbf{z}_0,\mathbf{v}_0)$ (with $\mathbf{v}_0=\mathbf{0}$), we autoregressively roll out Eq.~\eqref{eq:latent_dynamics} to obtain $\{\mathbf{z}_t\}$. By default, we initialize $\mathbf{z}_0=\mathbf{z}_{\text{ref}}$. Alternatively, $\mathbf{z}_0$ can be set to a different latent code for the same subject (e.g., selected from that subject's training latents) to generate diverse clothing trajectories under the same driving pose sequence.
The rollout adds only ${\sim}3\%$ to per-frame inference cost (Tab.~\ref{tab:runtime}, supplementary).

\begin{figure*}[t]
    \centering
    \small
    \setlength{\tabcolsep}{0pt}
    \newcommand{\imgw}{0.093\linewidth} 
    \newcommand{\imgwshort}{0.09\linewidth} 
    \newcommand{\evenraise}{-2pt} 
    
    \begin{tabular}{cc@{\hskip 6pt}cc@{\hskip 6pt}cc@{\hskip 6pt}cc@{\hskip 6pt}cc}
    &
    \raisebox{\evenraise}{\includegraphics[width=\imgwshort]{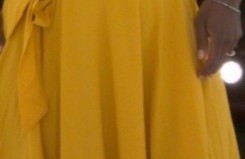}} &
    &
    \raisebox{\evenraise}{\includegraphics[width=\imgwshort]{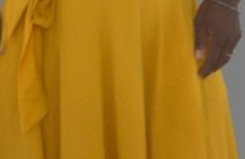}} &
    &
    \raisebox{\evenraise}{\includegraphics[width=\imgwshort]{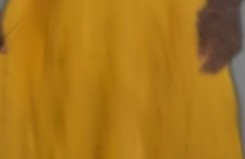}} &
    &
    \raisebox{\evenraise}{\includegraphics[width=\imgwshort]{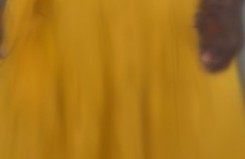}} &
    &
    \raisebox{\evenraise}{\includegraphics[width=\imgwshort]{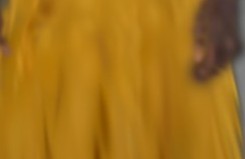}} \\[0pt]
    \multirow[t]{2}{*}{\includegraphics[width=\imgw]{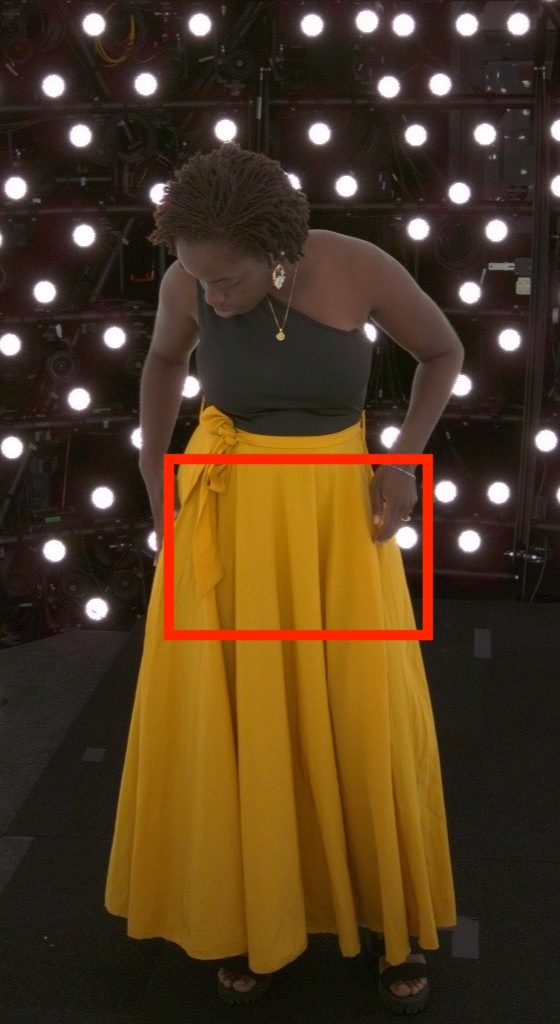}} &
    \includegraphics[width=\imgw]{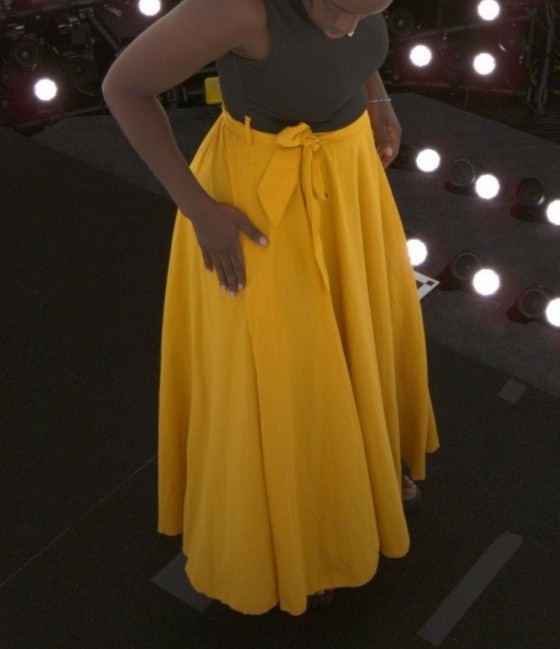} &
    \multirow[t]{2}{*}{\includegraphics[width=\imgw]{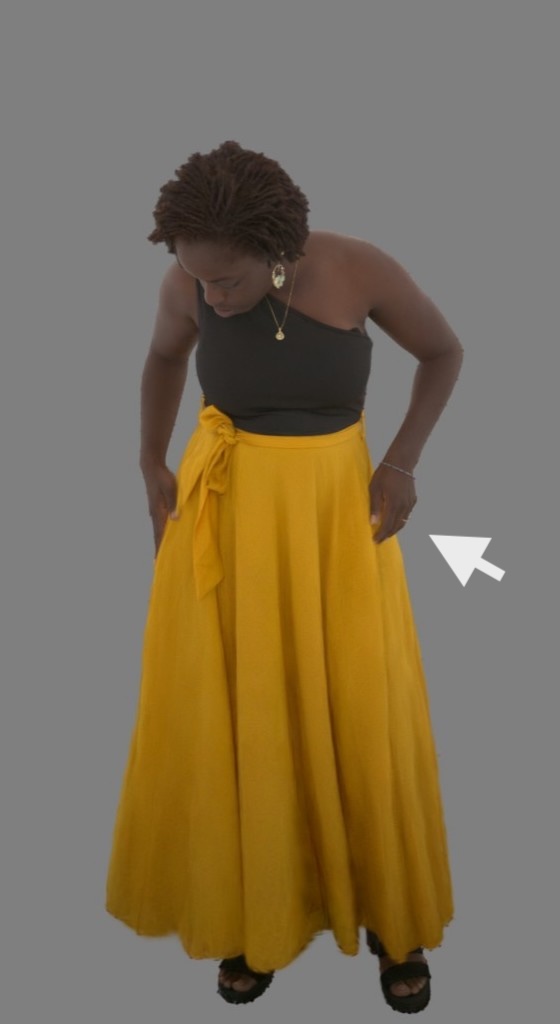}} &
    \includegraphics[width=\imgw]{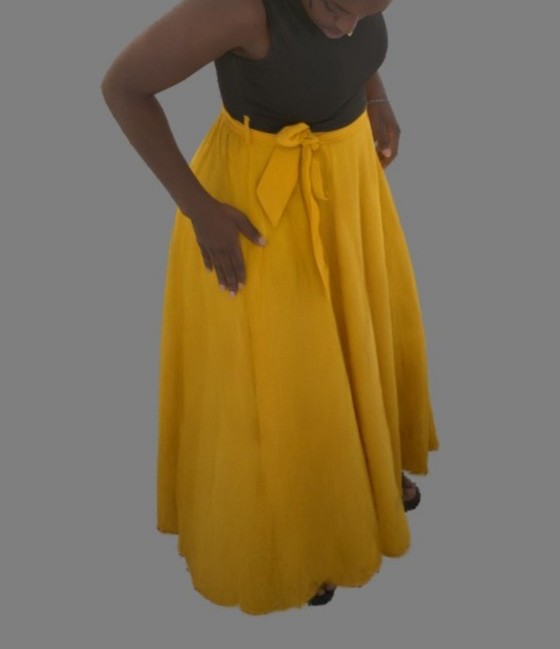} &
    \multirow[t]{2}{*}{\includegraphics[width=\imgw]{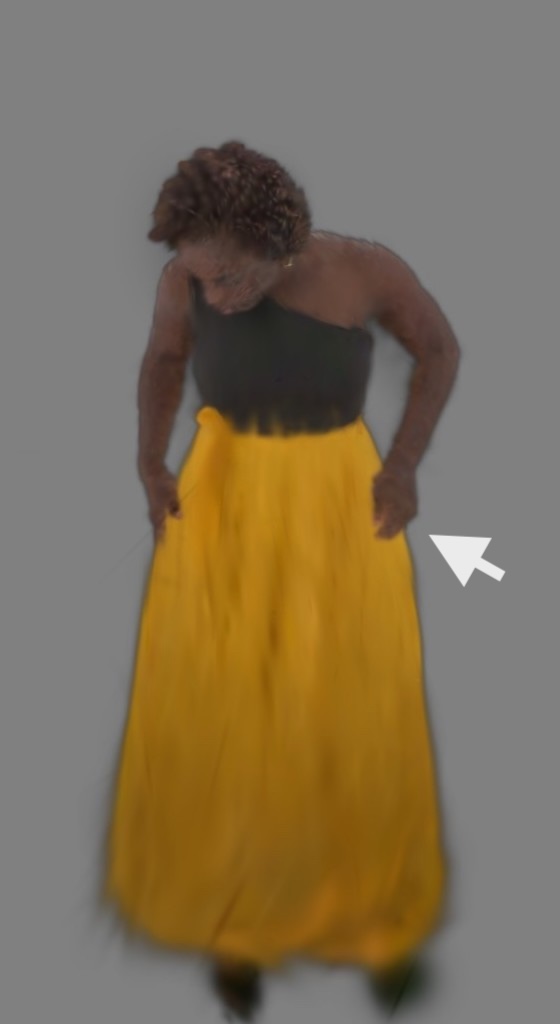}} &
    \includegraphics[width=\imgw]{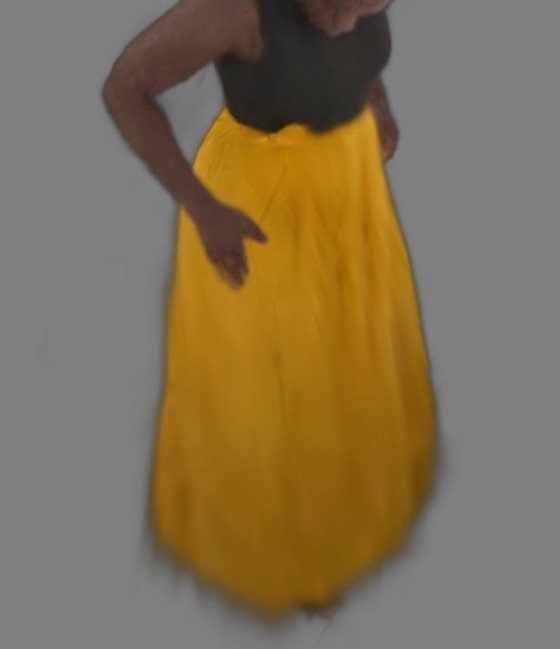} &
    \multirow[t]{2}{*}{\includegraphics[width=\imgw]{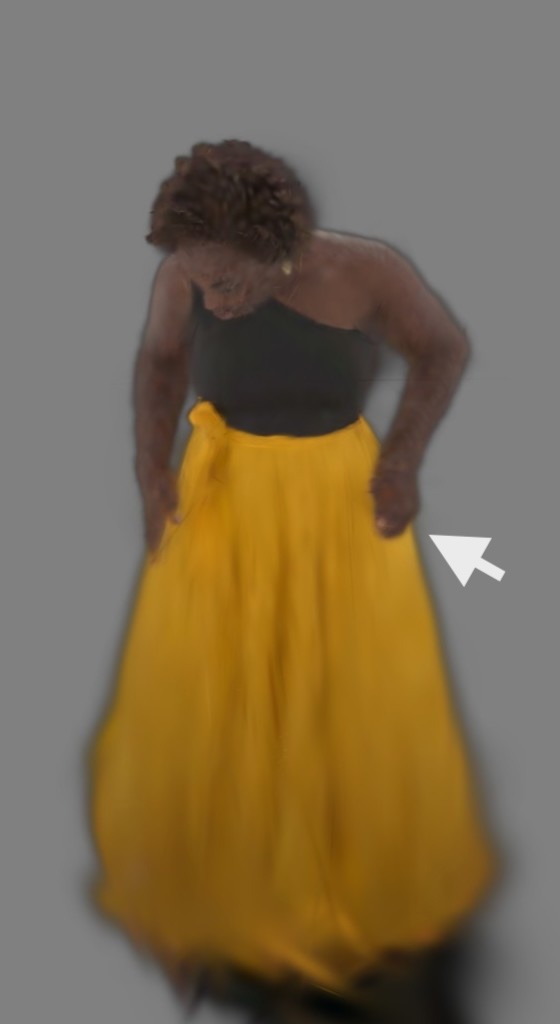}} &
    \includegraphics[width=\imgw]{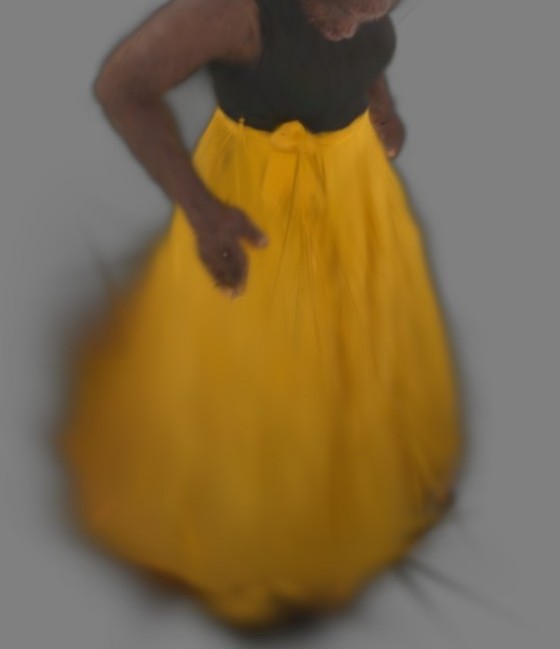} &
    \multirow[t]{2}{*}{\includegraphics[width=\imgw]{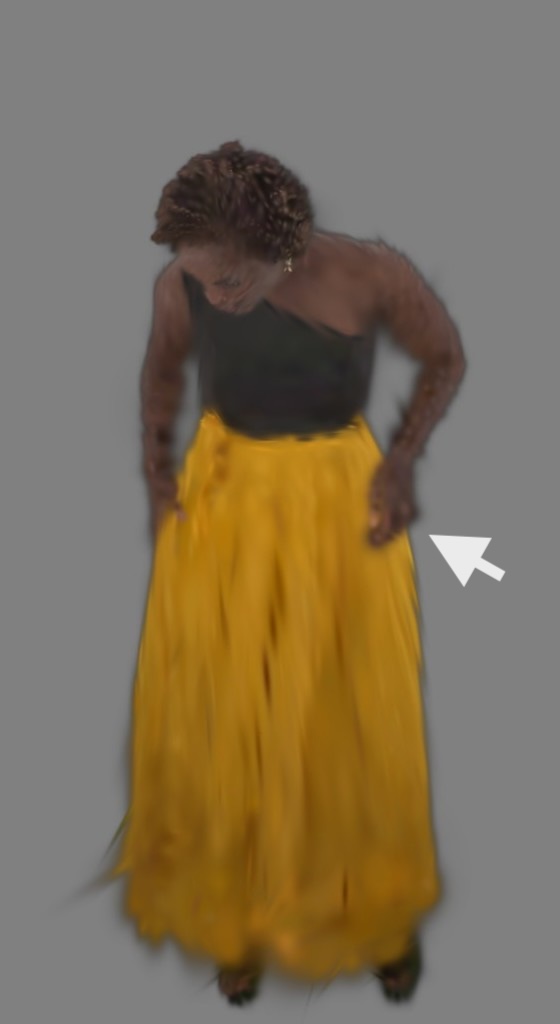}} &
    \includegraphics[width=\imgw]{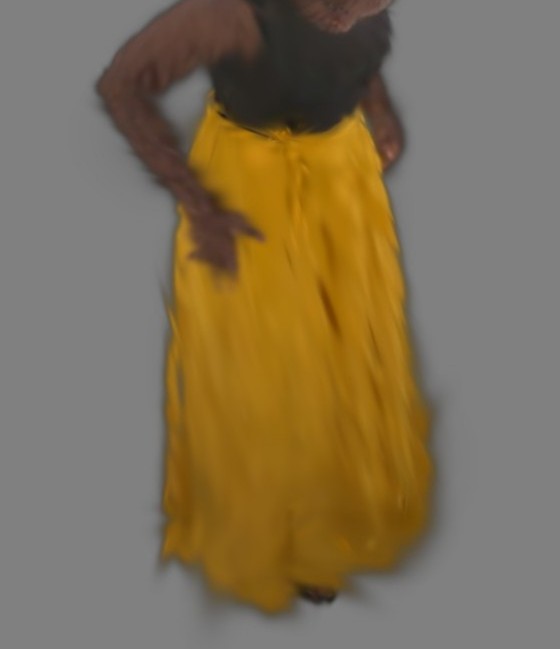} \\[0pt]
    &
    \raisebox{\evenraise}{\includegraphics[width=\imgwshort]{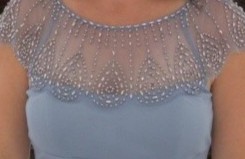}} &
    &
    \raisebox{\evenraise}{\includegraphics[width=\imgwshort]{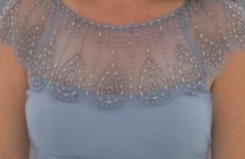}} &
    &
    \raisebox{\evenraise}{\includegraphics[width=\imgwshort]{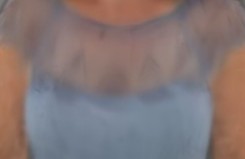}} &
    &
    \raisebox{\evenraise}{\includegraphics[width=\imgwshort]{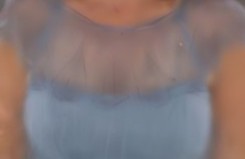}} &
    &
    \raisebox{\evenraise}{\includegraphics[width=\imgwshort]{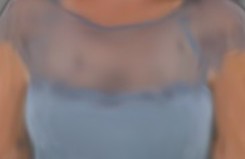}} \\[0pt]
    \multirow[t]{2}{*}{\includegraphics[width=\imgw]{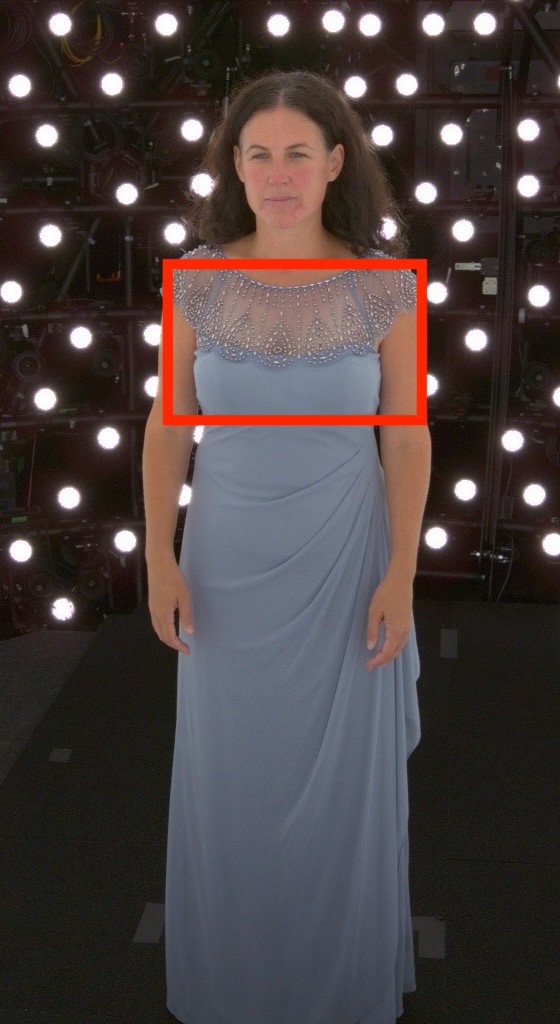}} &
    \includegraphics[width=\imgw]{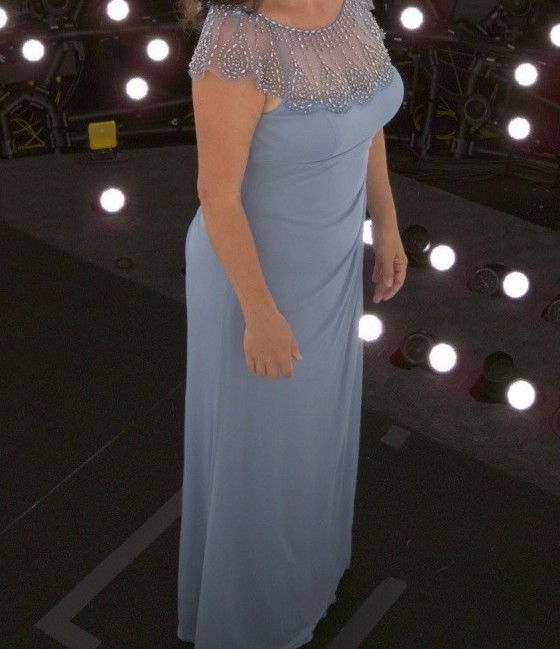} &
    \multirow[t]{2}{*}{\includegraphics[width=\imgw]{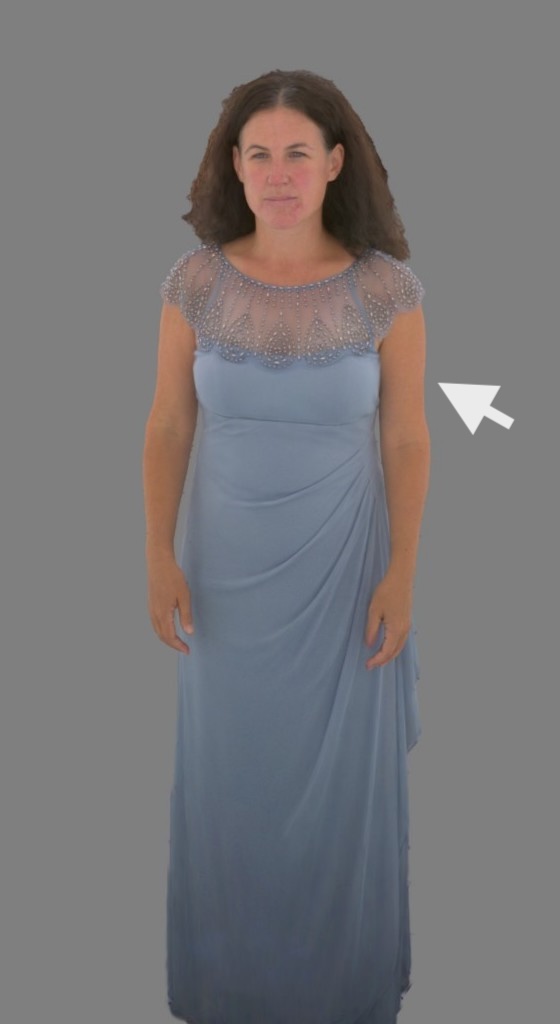}} &
    \includegraphics[width=\imgw]{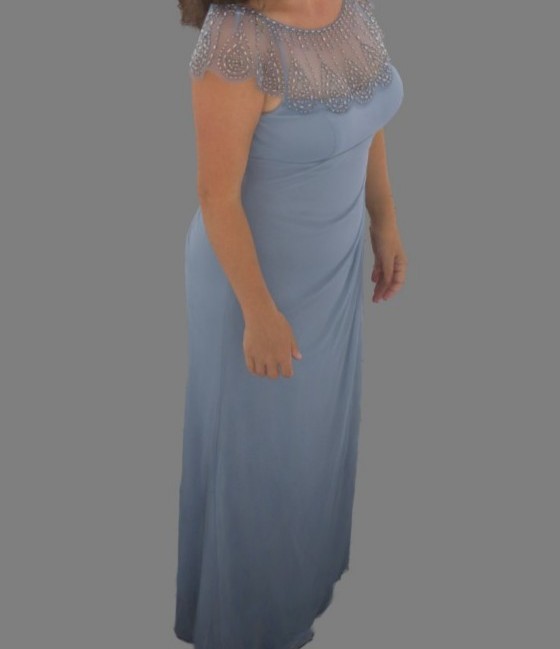} &
    \multirow[t]{2}{*}{\includegraphics[width=\imgw]{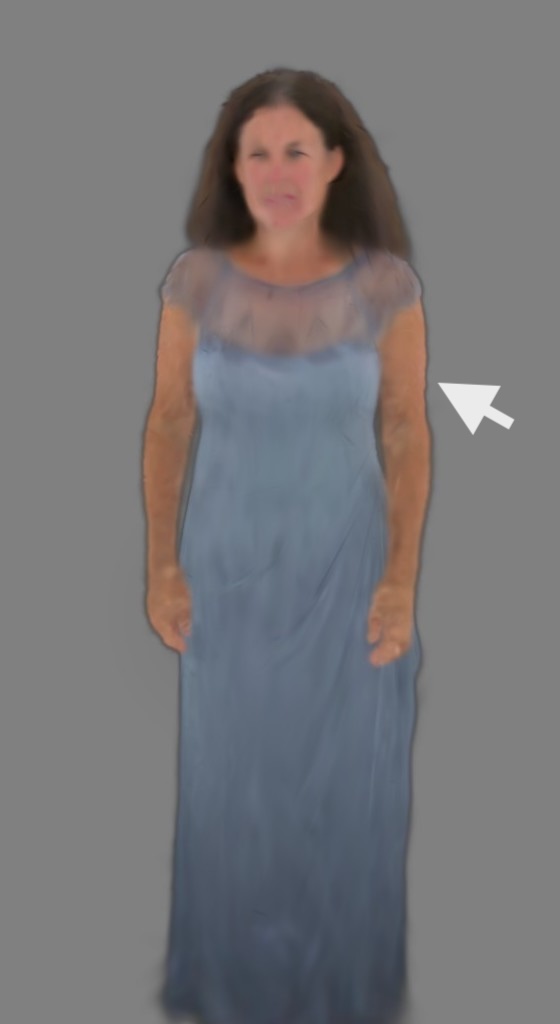}} &
    \includegraphics[width=\imgw]{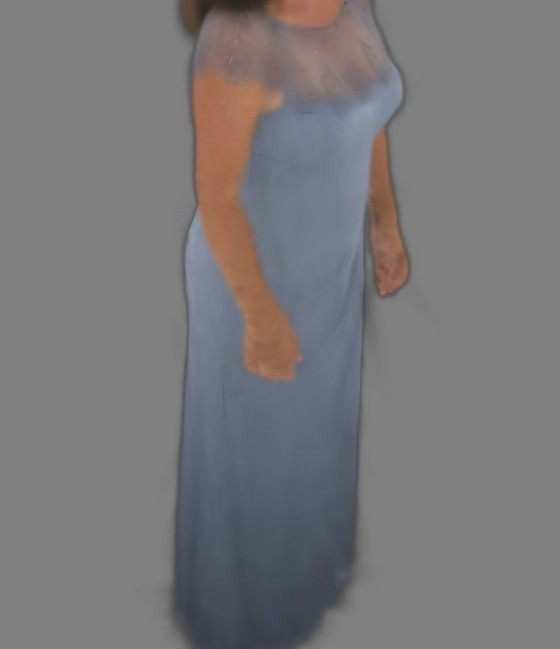} &
    \multirow[t]{2}{*}{\includegraphics[width=\imgw]{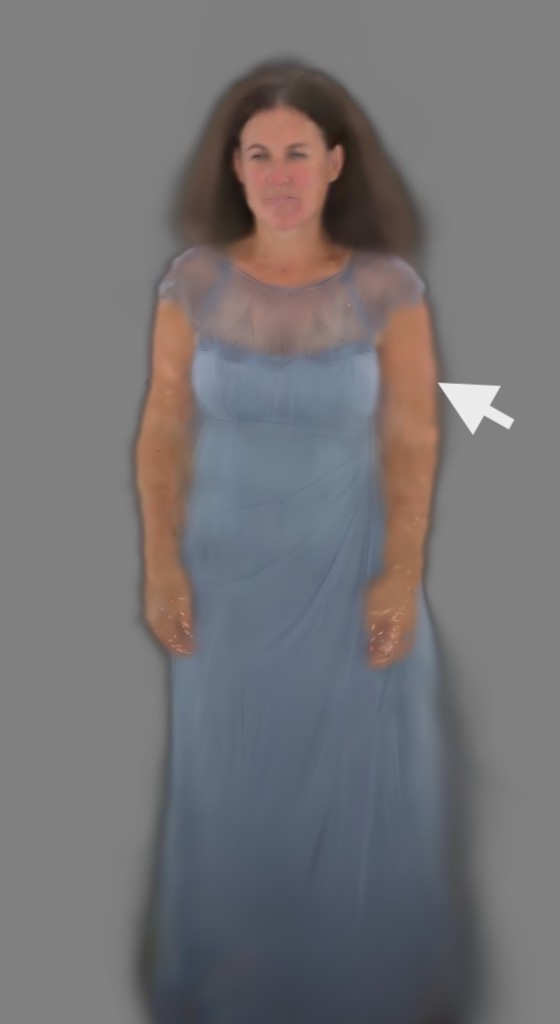}} &
    \includegraphics[width=\imgw]{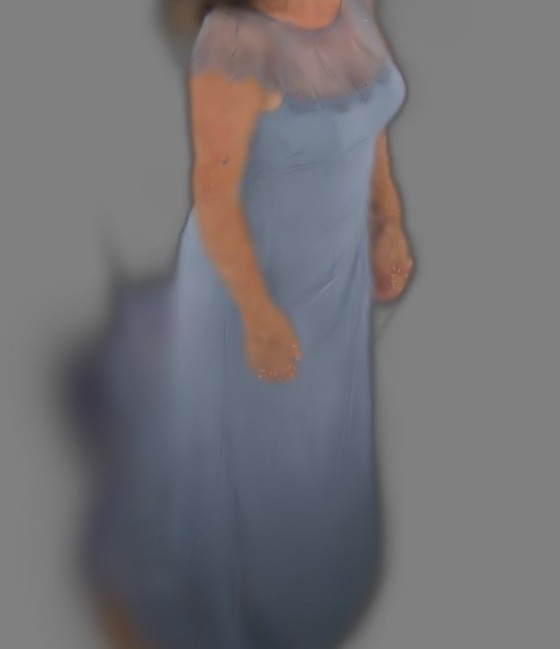} &
    \multirow[t]{2}{*}{\includegraphics[width=\imgw]{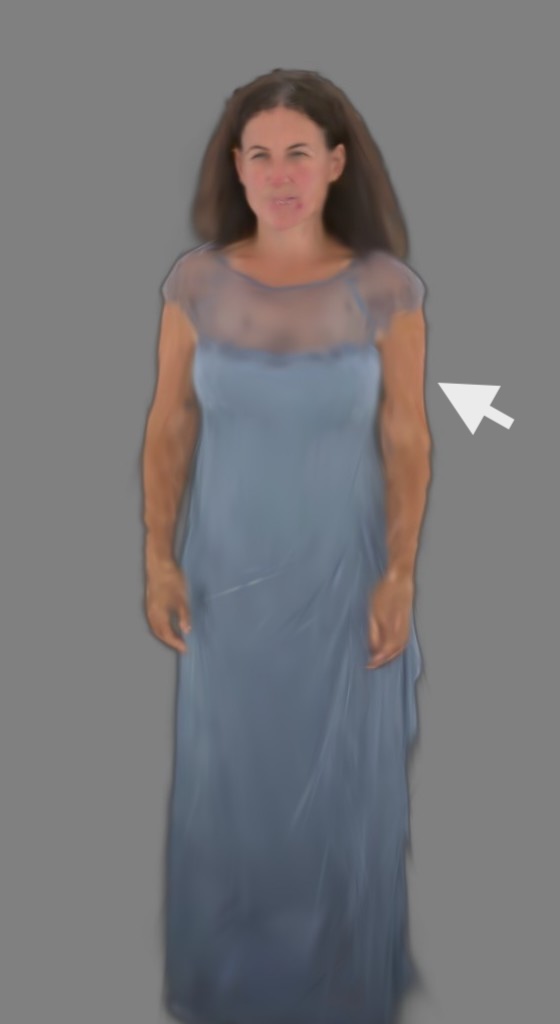}} &
    \includegraphics[width=\imgw]{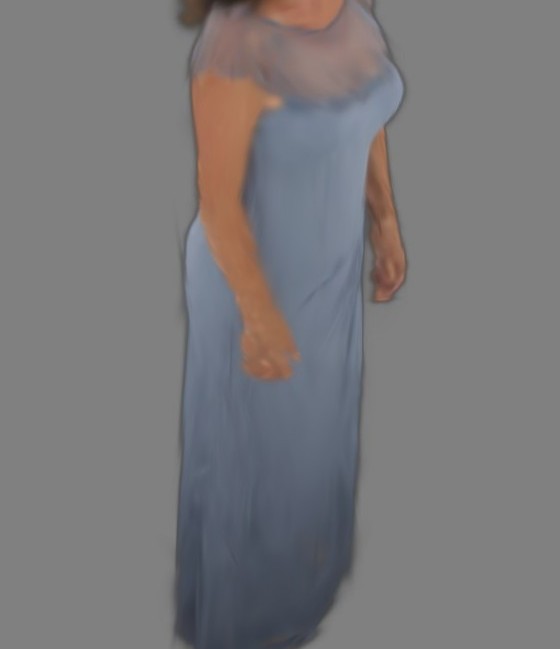} \\[0pt]
    &
    \raisebox{\evenraise}{\includegraphics[width=\imgwshort]{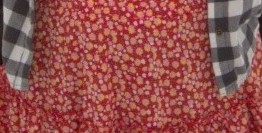}} &
    &
    \raisebox{\evenraise}{\includegraphics[width=\imgwshort]{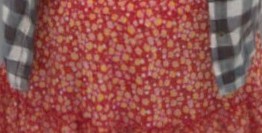}} &
    &
    \raisebox{\evenraise}{\includegraphics[width=\imgwshort]{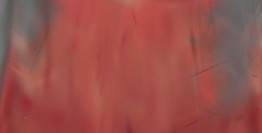}} &
    &
    \raisebox{\evenraise}{\includegraphics[width=\imgwshort]{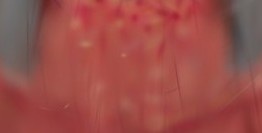}} &
    &
    \raisebox{\evenraise}{\includegraphics[width=\imgwshort]{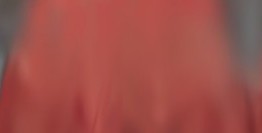}} \\[0pt]
    \multirow[t]{2}{*}{\includegraphics[width=\imgw]{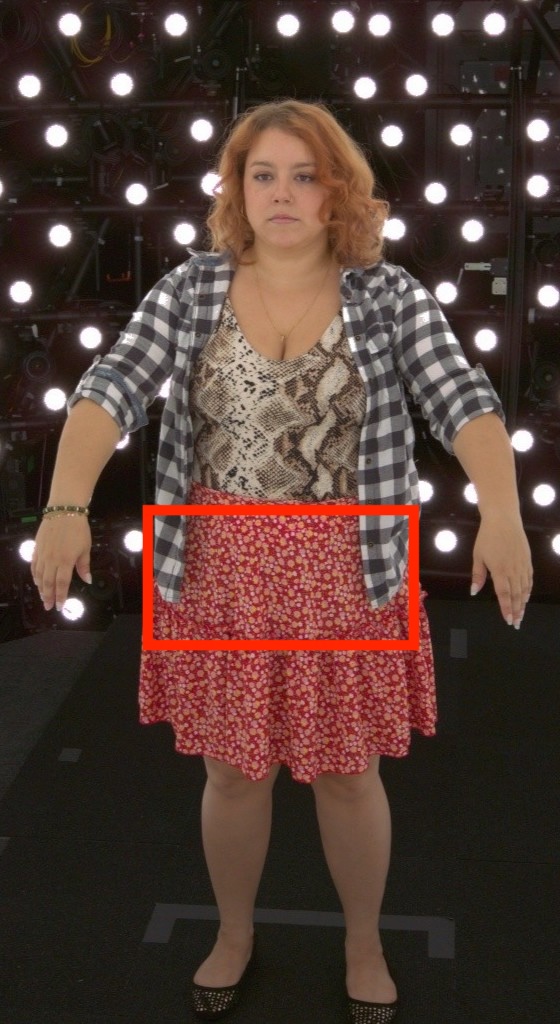}} &
    \includegraphics[width=\imgw]{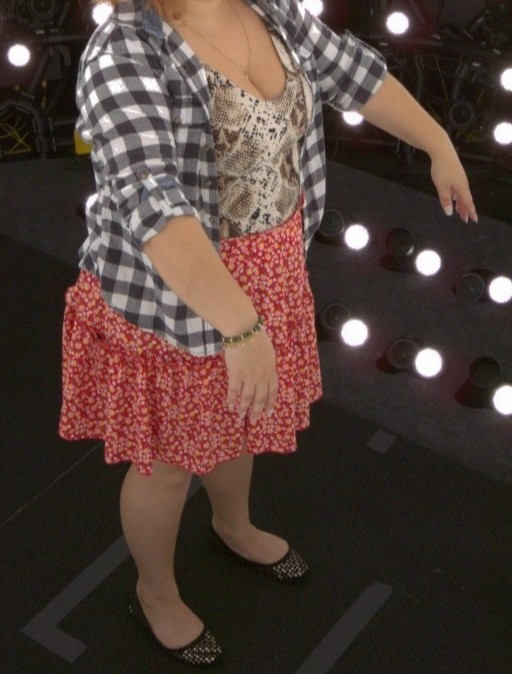} &
    \multirow[t]{2}{*}{\includegraphics[width=\imgw]{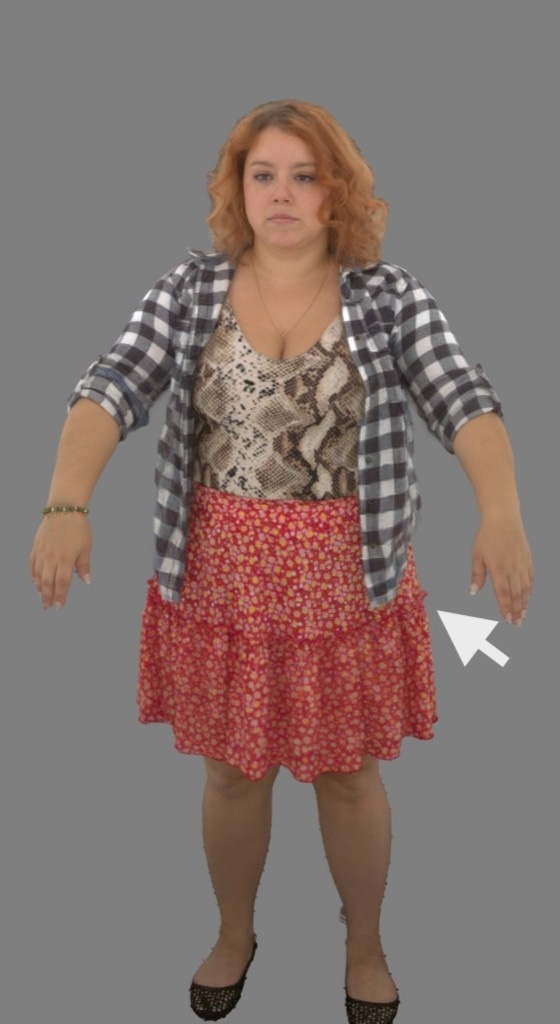}} &
    \includegraphics[width=\imgw]{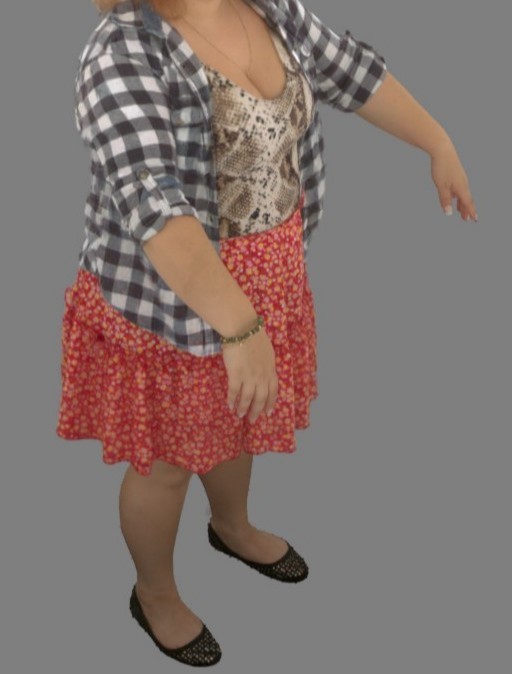} &
    \multirow[t]{2}{*}{\includegraphics[width=\imgw]{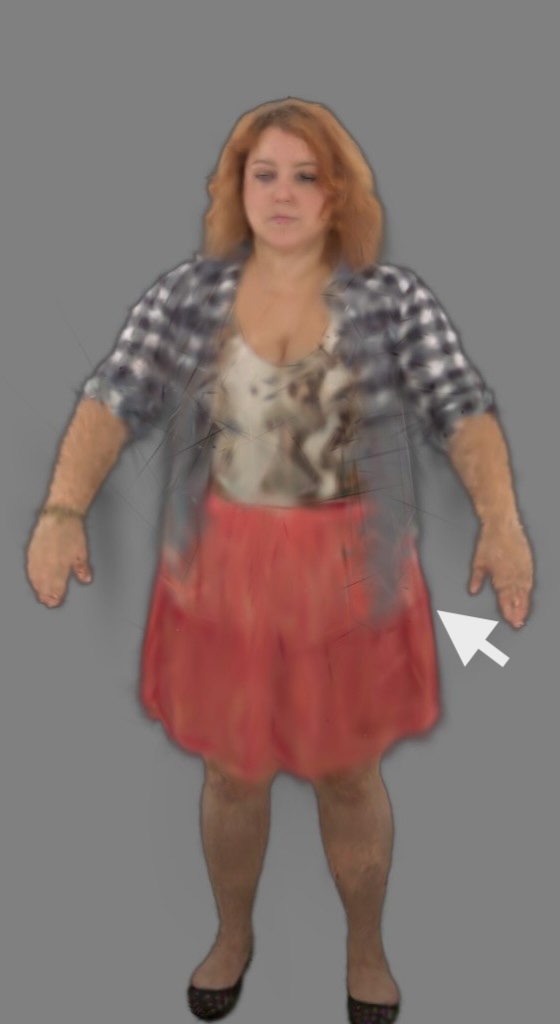}} &
    \includegraphics[width=\imgw]{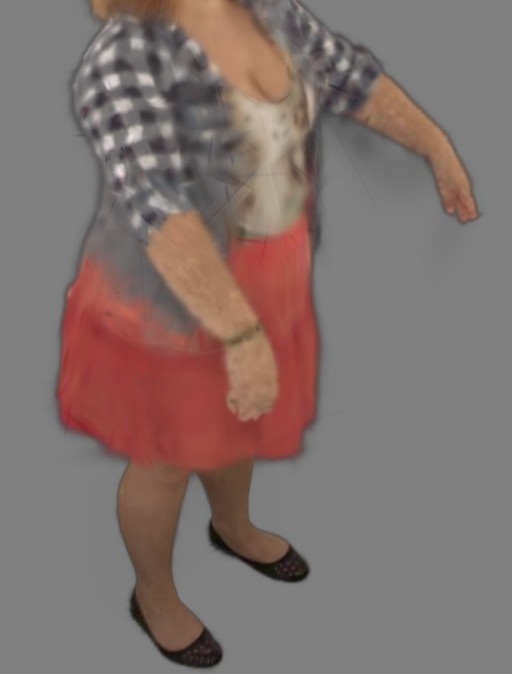} &
    \multirow[t]{2}{*}{\includegraphics[width=\imgw]{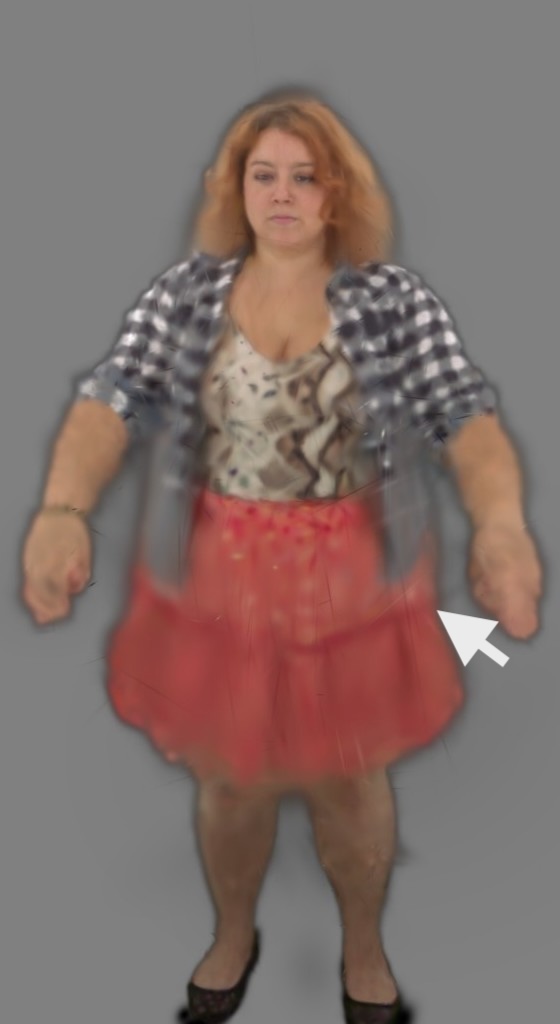}} &
    \includegraphics[width=\imgw]{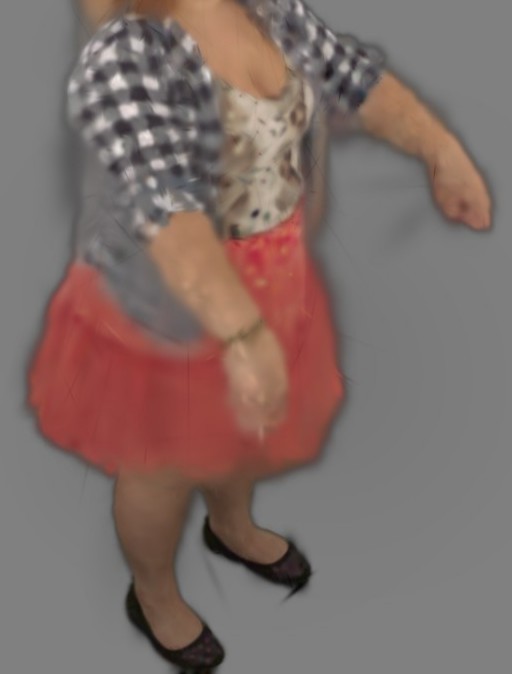} &
    \multirow[t]{2}{*}{\includegraphics[width=\imgw]{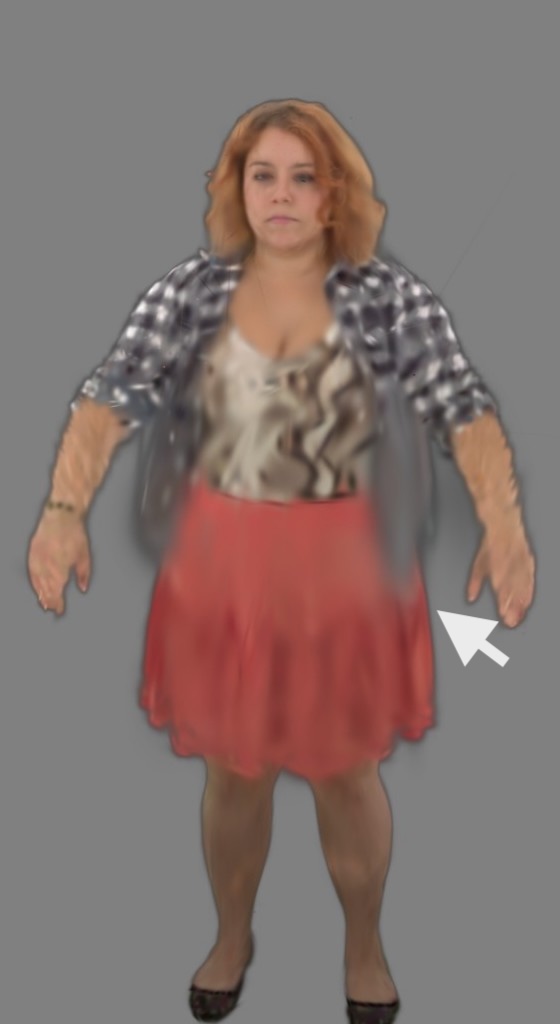}} &
    \includegraphics[width=\imgw]{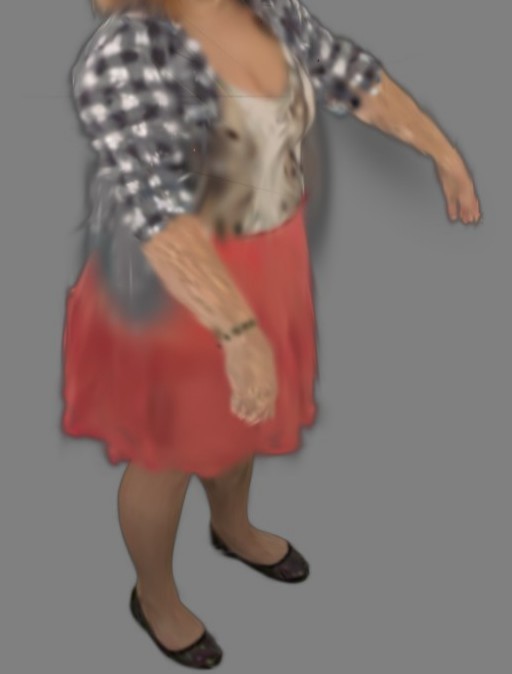} \\[6pt]
    \multicolumn{2}{c}{Reference Image} &
    \multicolumn{2}{c}{Ours} &
    \multicolumn{2}{c}{ToMiE~\cite{zhan2024tomiemodulargrowthenhanced}} &
    \multicolumn{2}{c}{Seq-Avatar~\cite{xu2025seqavatar}} &
    \multicolumn{2}{c}{$R^3$-Avatar~\cite{Zhan2025R3AvatarRA}} \\
    \end{tabular}
    \caption{\textbf{Qualitative comparison on novel-view synthesis.} Our method produces sharper renderings and better preserves fine garment details, while baseline methods tend to oversmooth regions undergoing large deformations, such as folds in the yellow dress (rows 1), embroidery around the upper chest (rows 2), and dotted or grid patterns on the garments (rows 3).}
    \label{fig:novel_view}
\end{figure*}

\begin{figure*}[ht]
    \centering
    \setlength{\tabcolsep}{0pt}
    \begin{tabular}{@{}*{6}{c}@{\hspace{6pt}}*{6}{c}@{}}
      \includegraphics[width=0.078\linewidth]{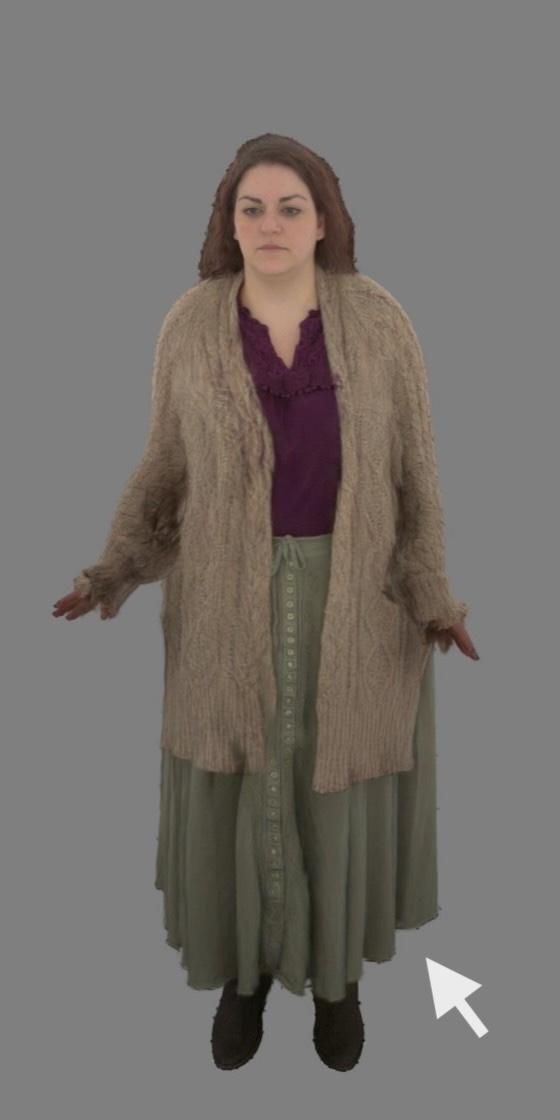} &
      \includegraphics[width=0.078\linewidth]{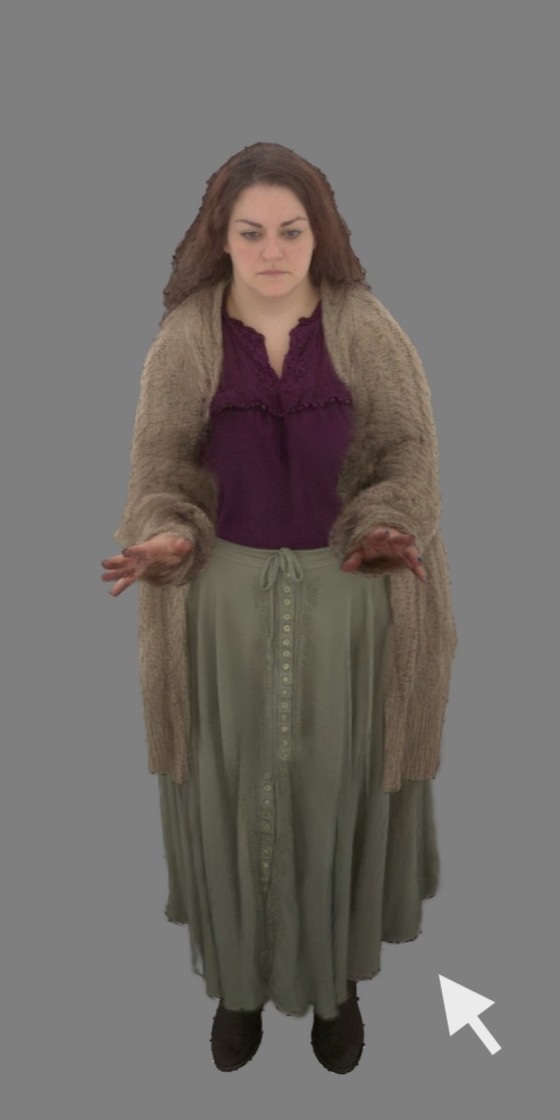} &
      \includegraphics[width=0.078\linewidth]{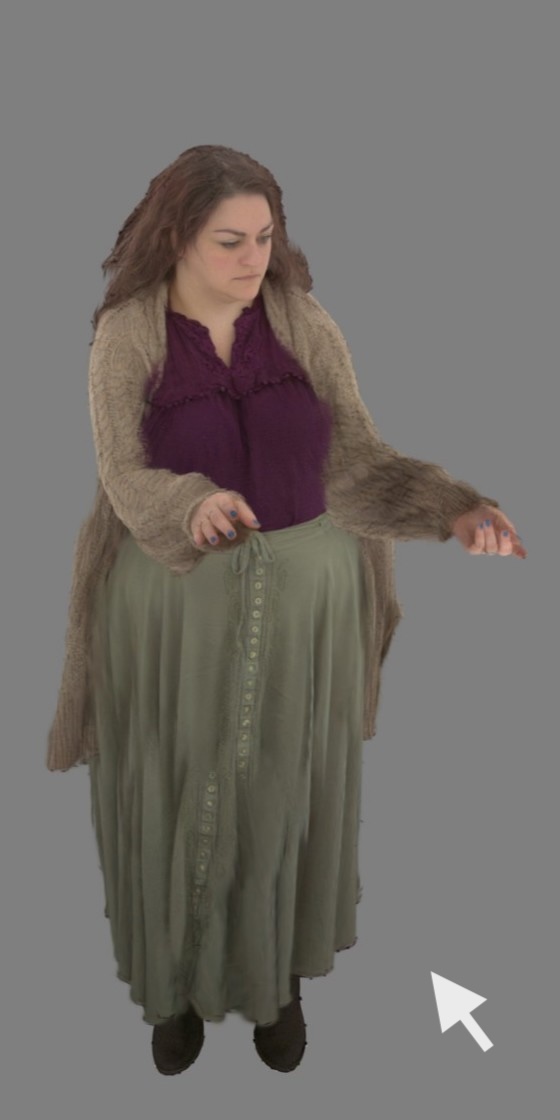} &
      \includegraphics[width=0.078\linewidth]{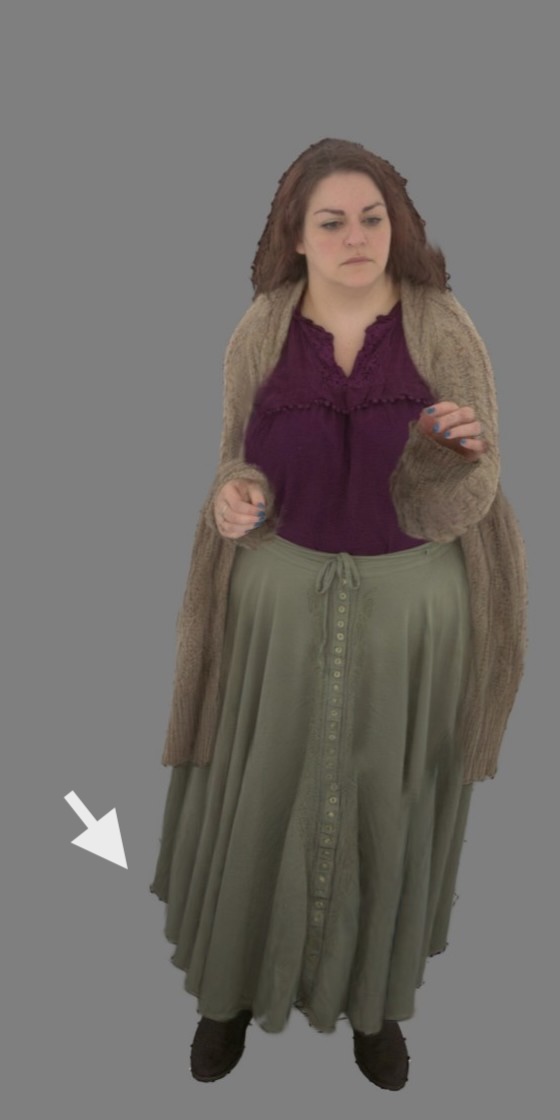} &
      \includegraphics[width=0.078\linewidth]{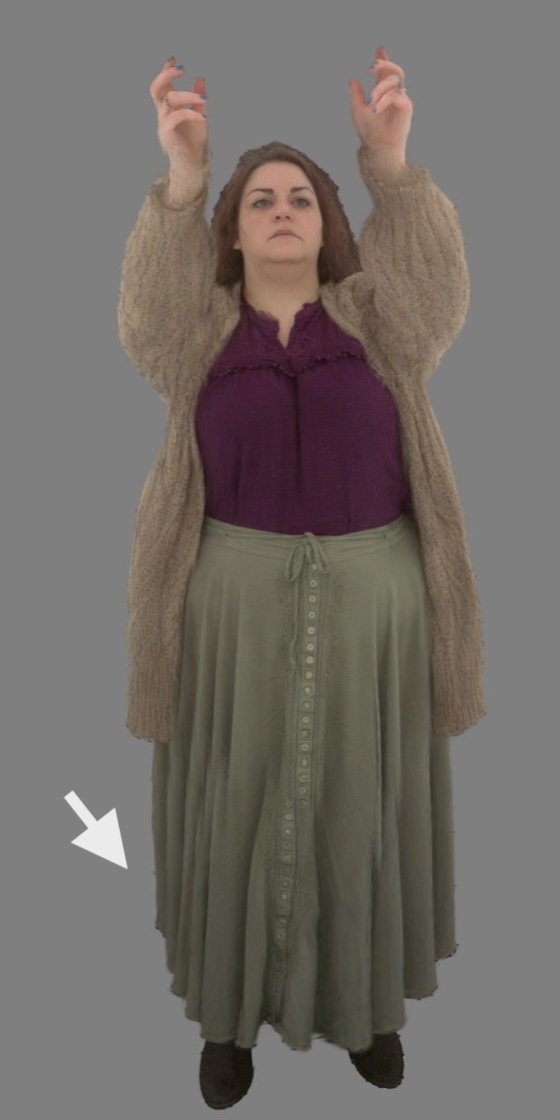} &
      \includegraphics[width=0.078\linewidth]{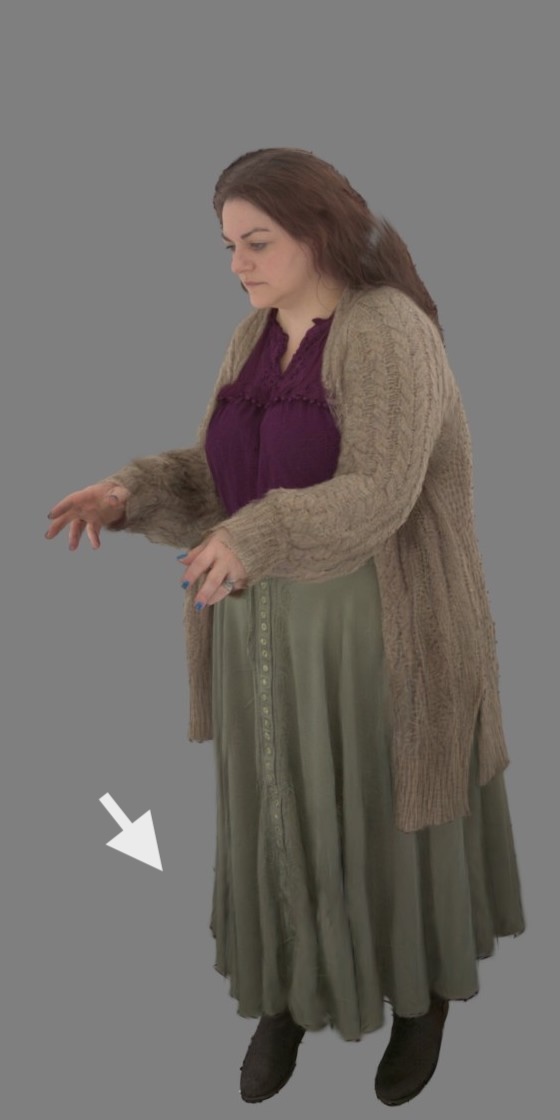} &
      \includegraphics[width=0.078\linewidth]{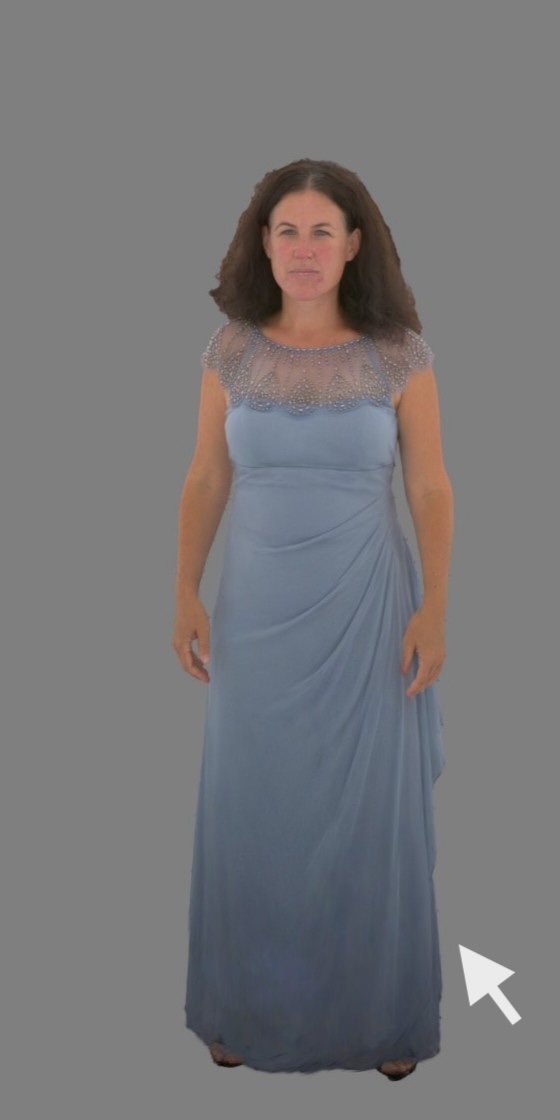} &
      \includegraphics[width=0.078\linewidth]{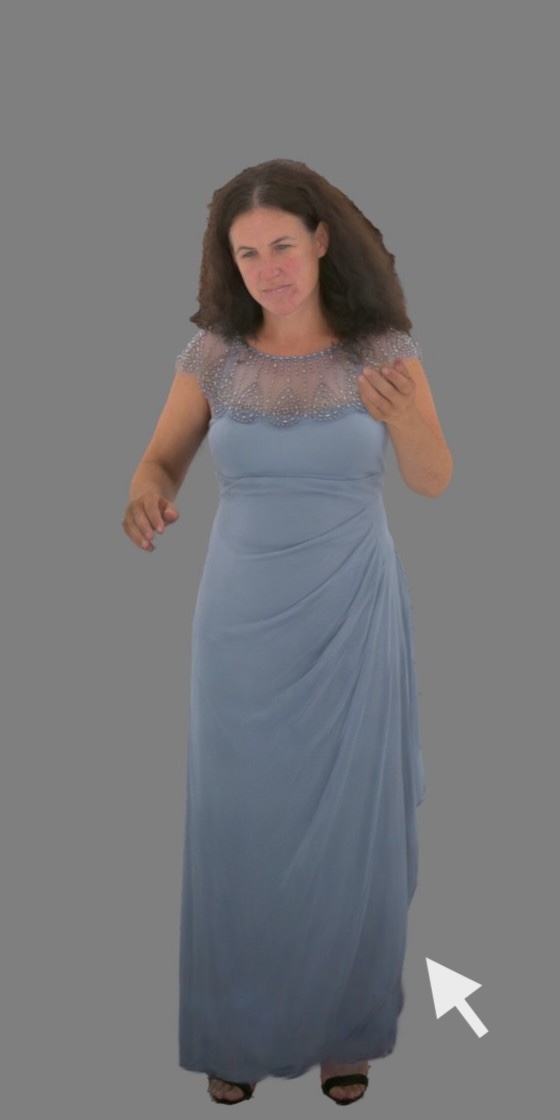} &
      \includegraphics[width=0.078\linewidth]{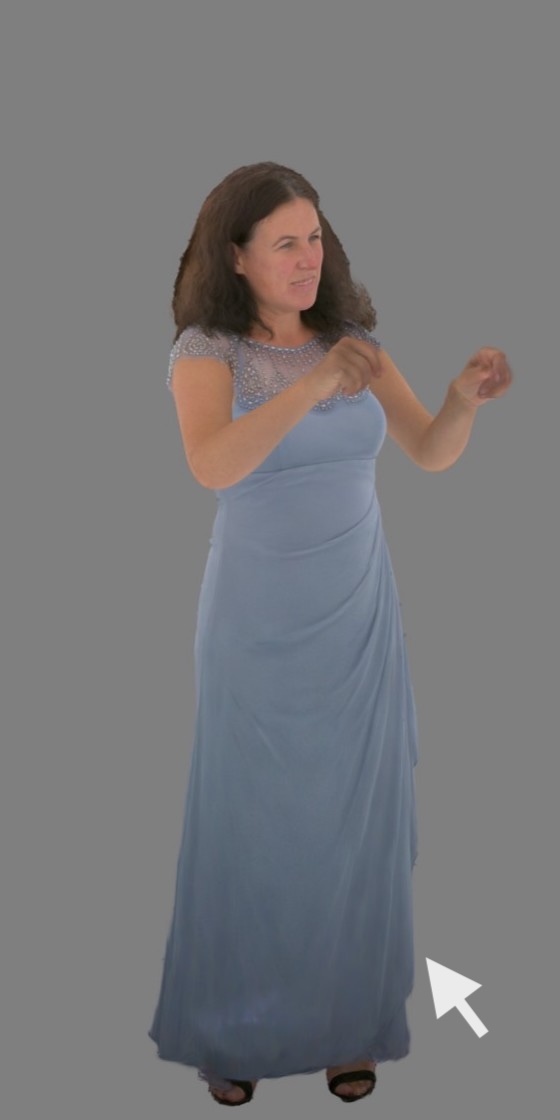} &
      \includegraphics[width=0.078\linewidth]{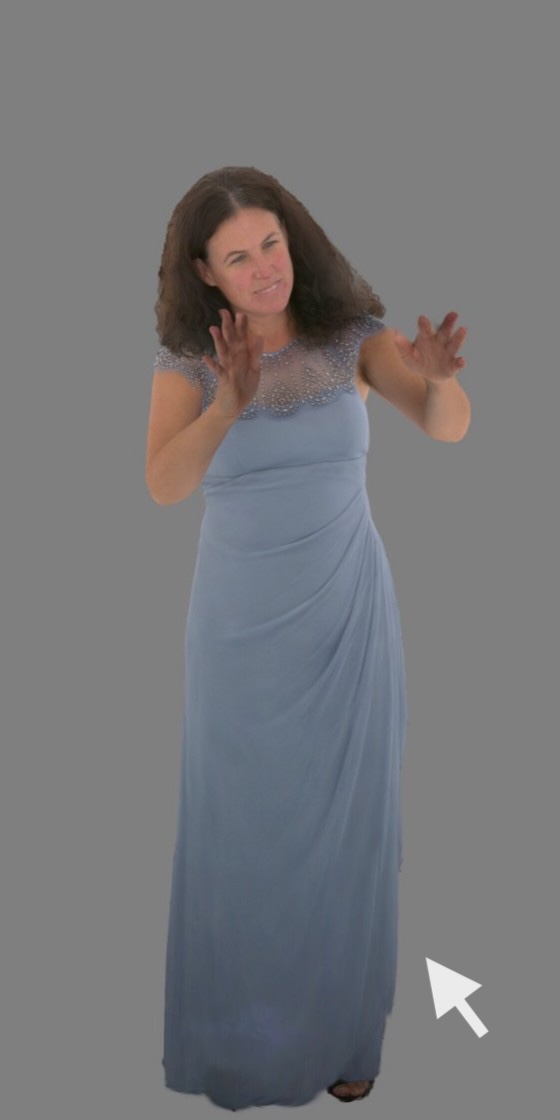} &
      \includegraphics[width=0.078\linewidth]{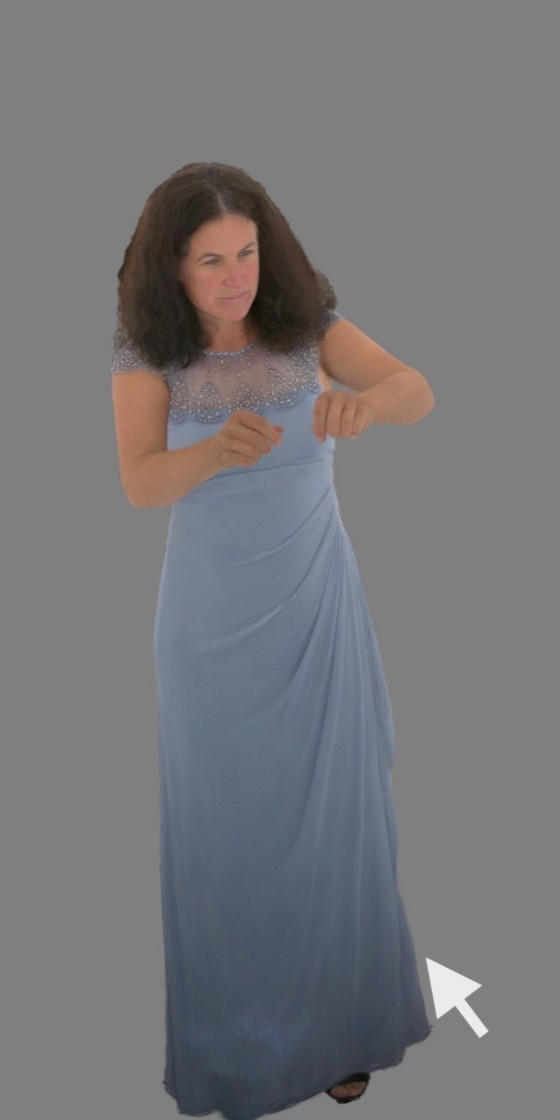} &
      \includegraphics[width=0.078\linewidth]{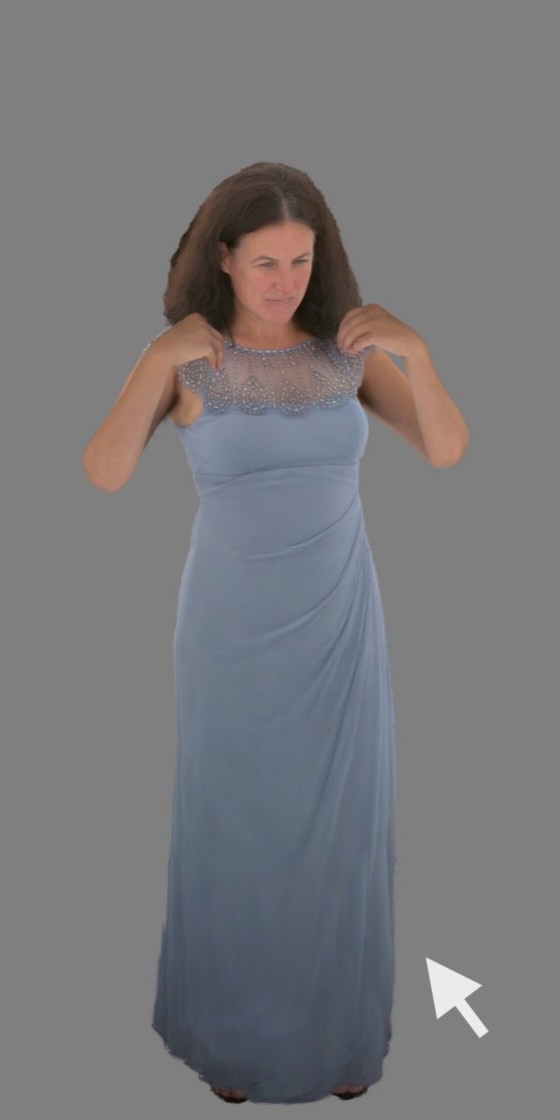} \\
      \includegraphics[width=0.078\linewidth]{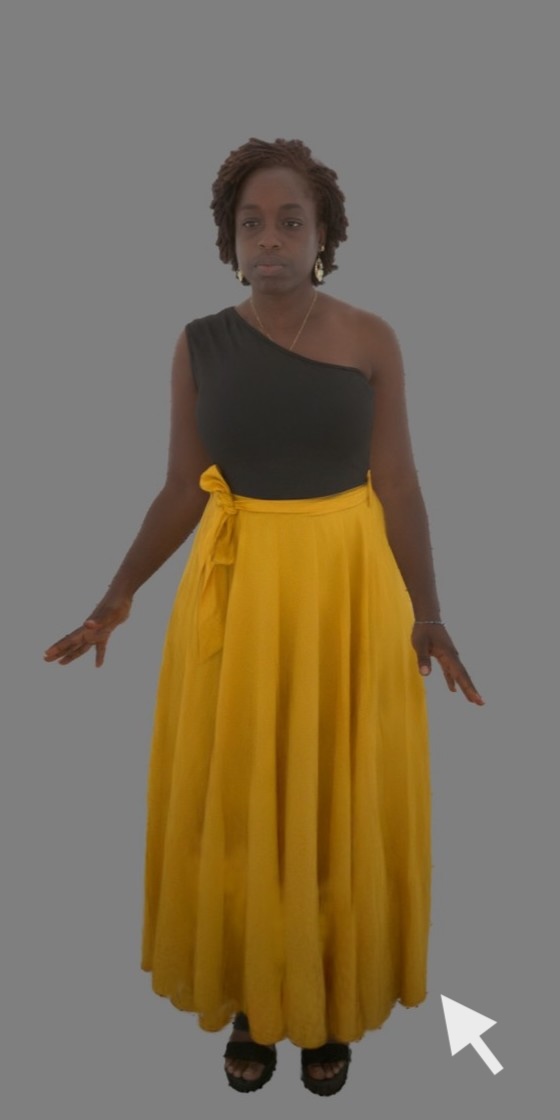} &
      \includegraphics[width=0.078\linewidth]{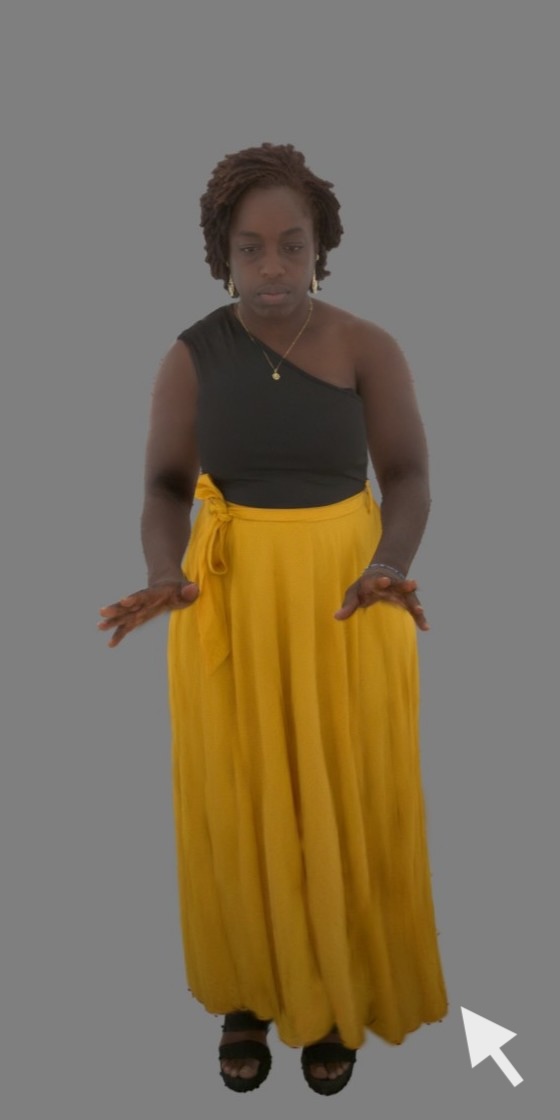} &
      \includegraphics[width=0.078\linewidth]{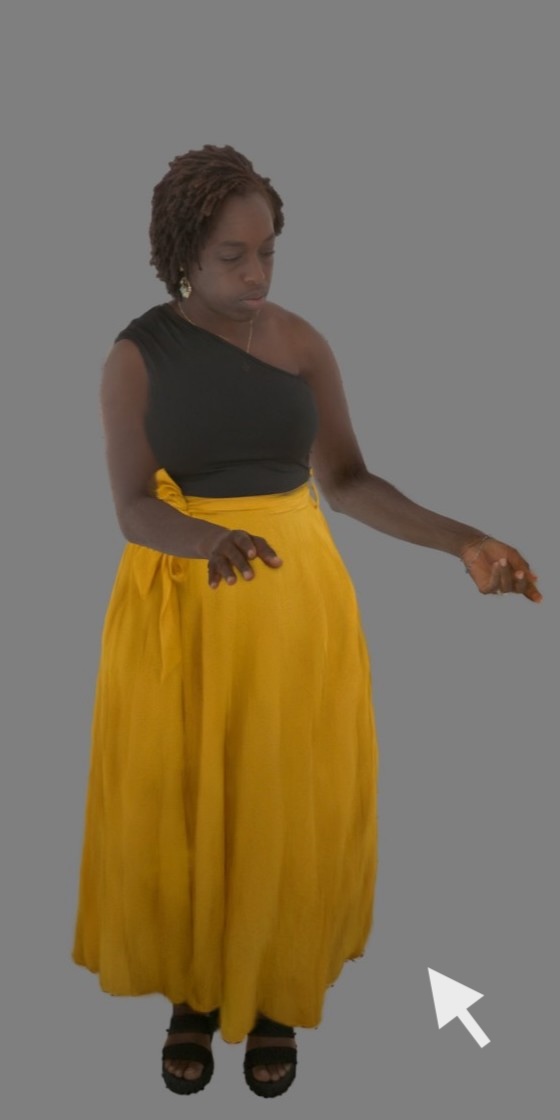} &
      \includegraphics[width=0.078\linewidth]{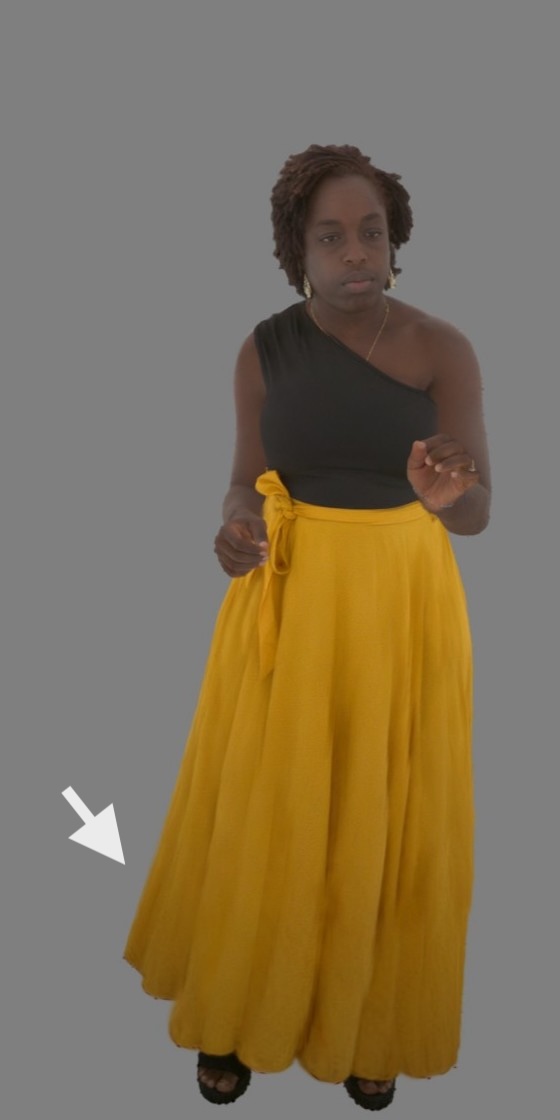} &
      \includegraphics[width=0.078\linewidth]{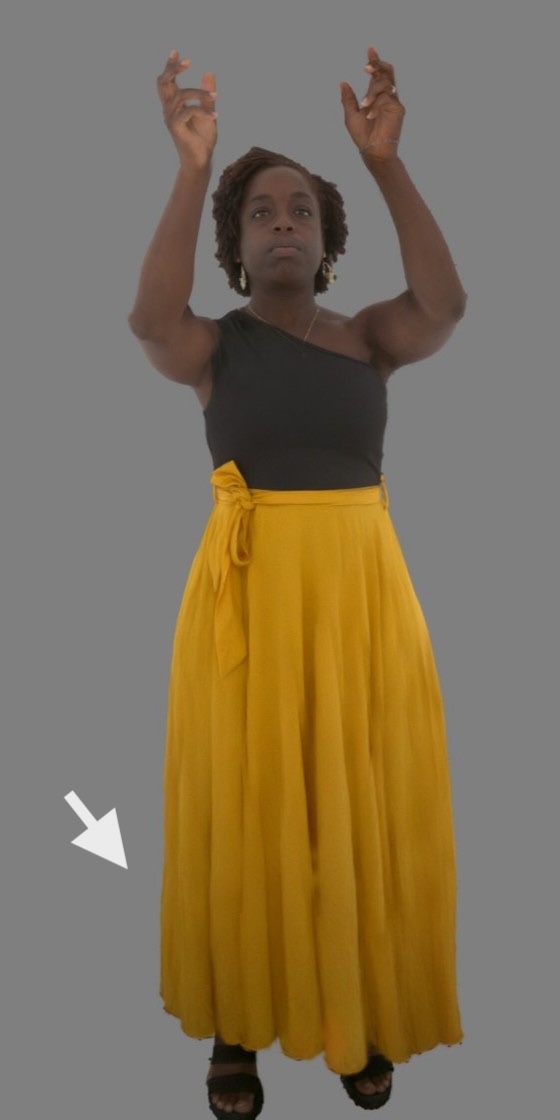} &
      \includegraphics[width=0.078\linewidth]{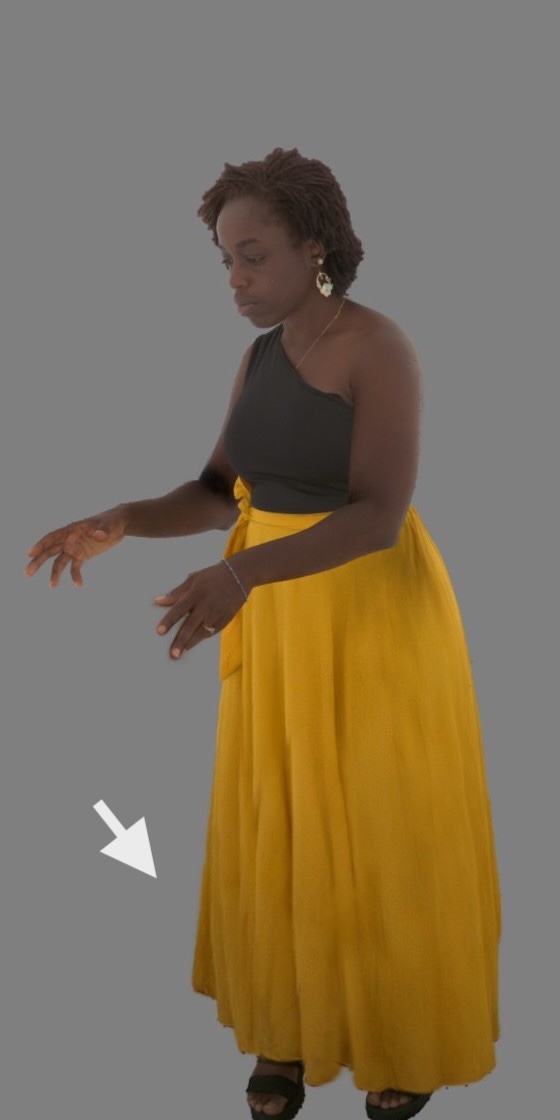} &
      \includegraphics[width=0.078\linewidth]{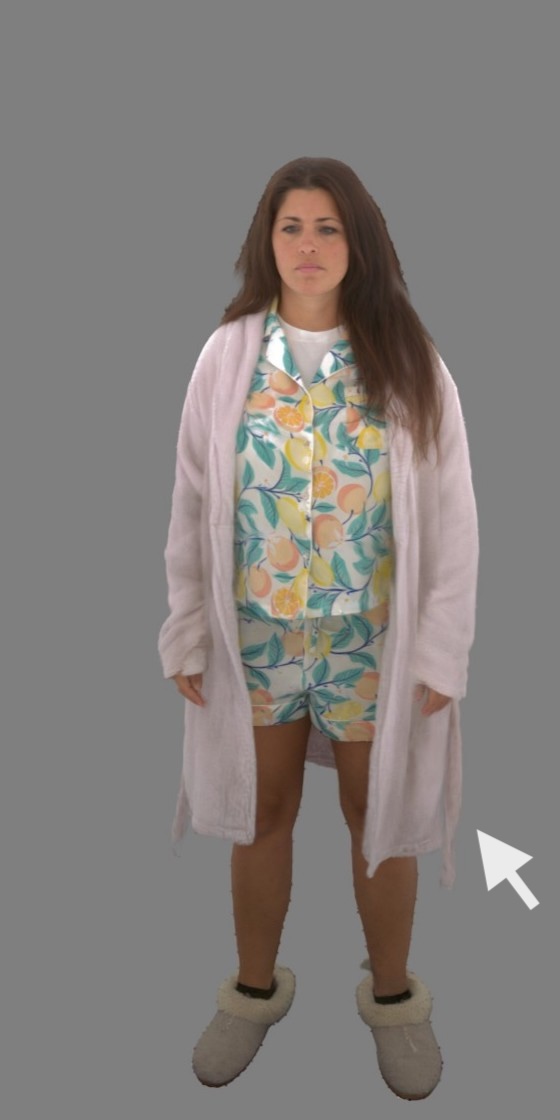} &
      \includegraphics[width=0.078\linewidth]{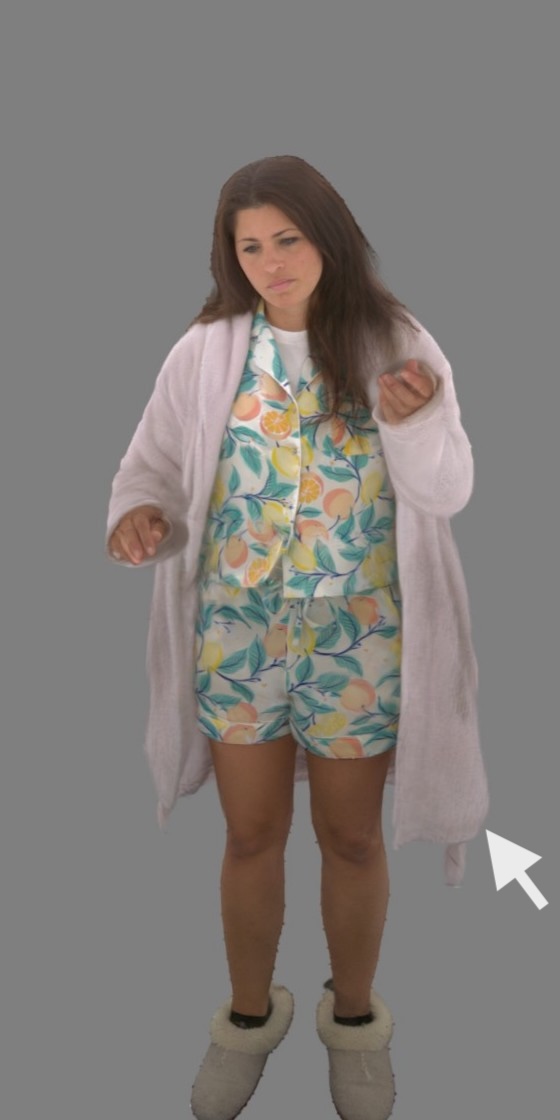} &
      \includegraphics[width=0.078\linewidth]{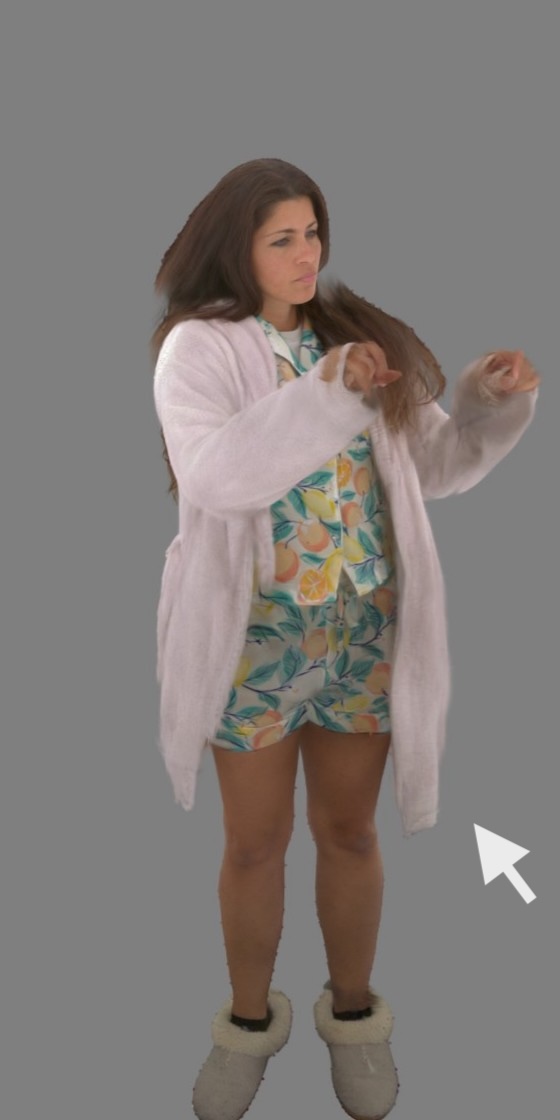} &
      \includegraphics[width=0.078\linewidth]{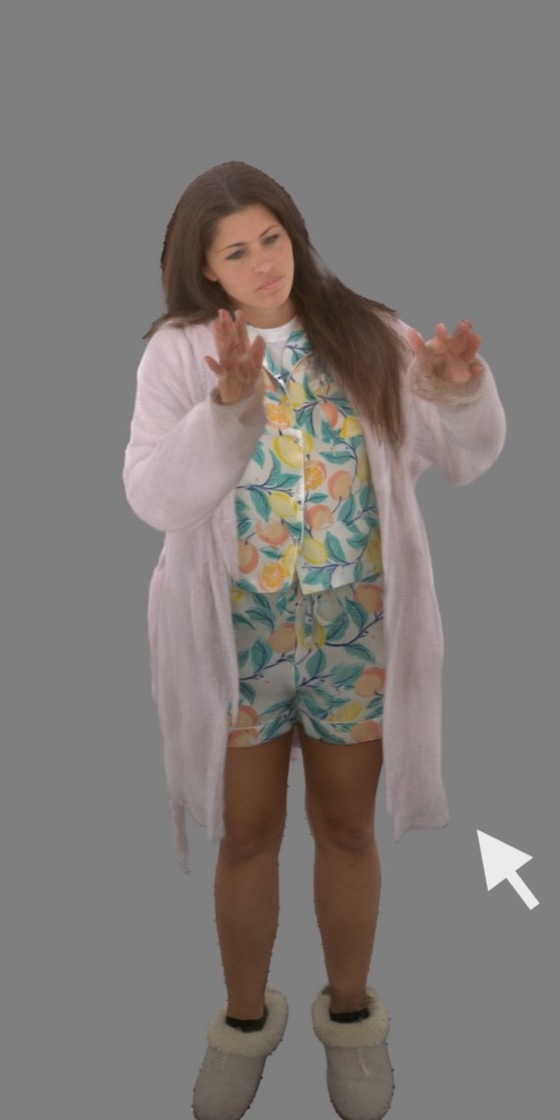} &
      \includegraphics[width=0.078\linewidth]{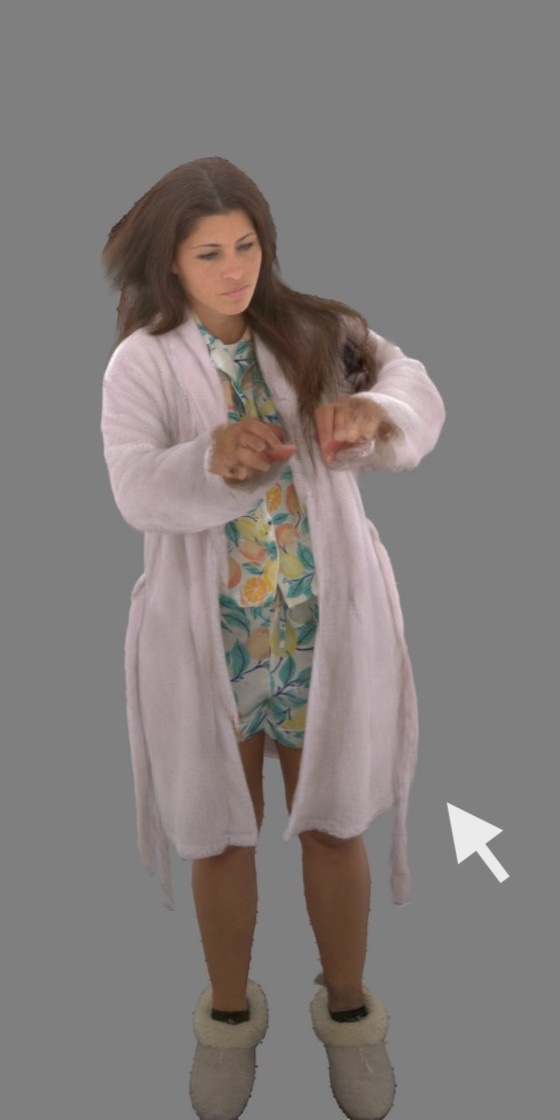} &
      \includegraphics[width=0.078\linewidth]{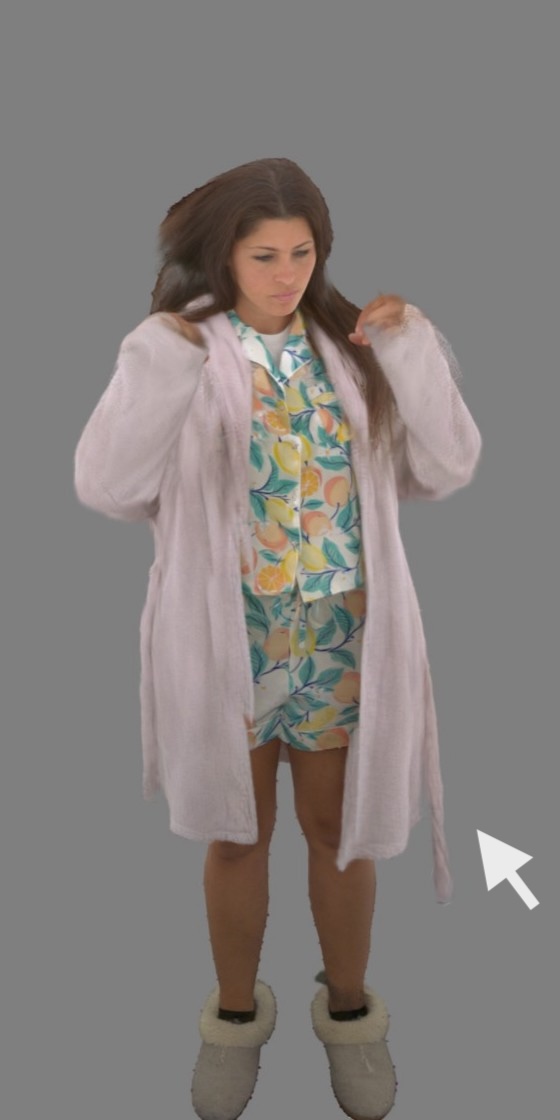} \\
      \includegraphics[width=0.078\linewidth]{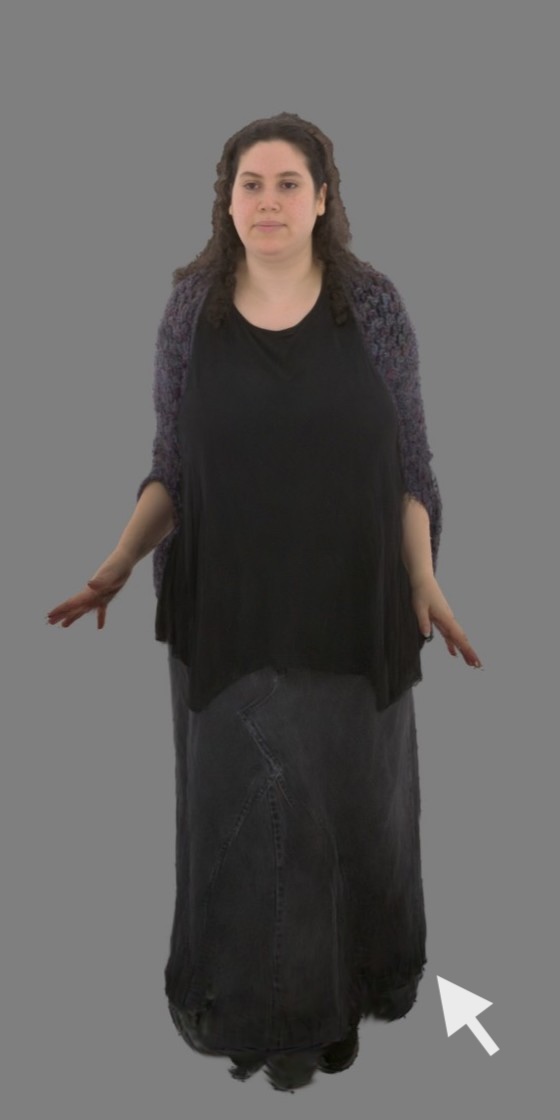} &
      \includegraphics[width=0.078\linewidth]{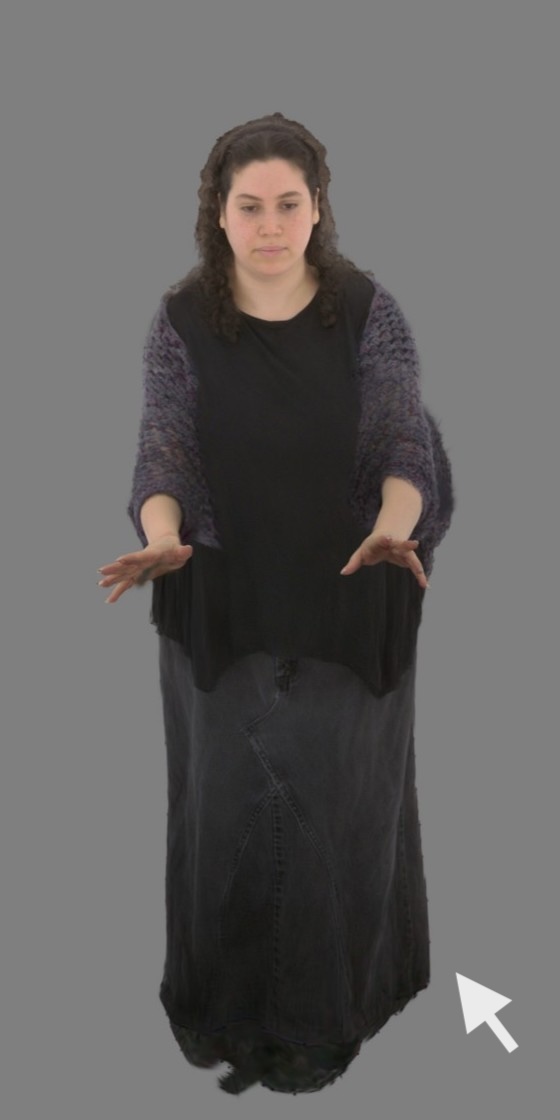} &
      \includegraphics[width=0.078\linewidth]{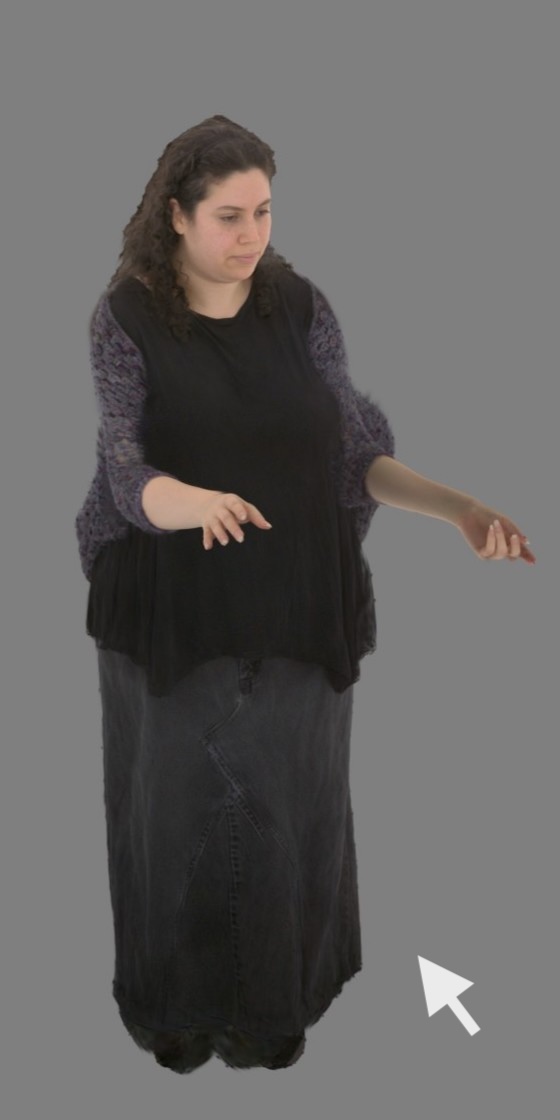} &
      \includegraphics[width=0.078\linewidth]{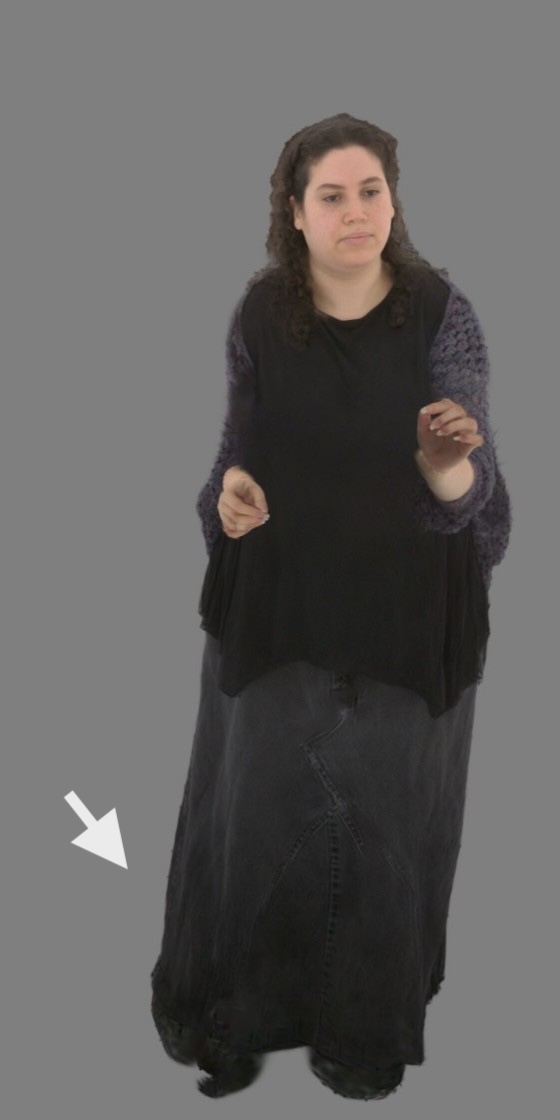} &
      \includegraphics[width=0.078\linewidth]{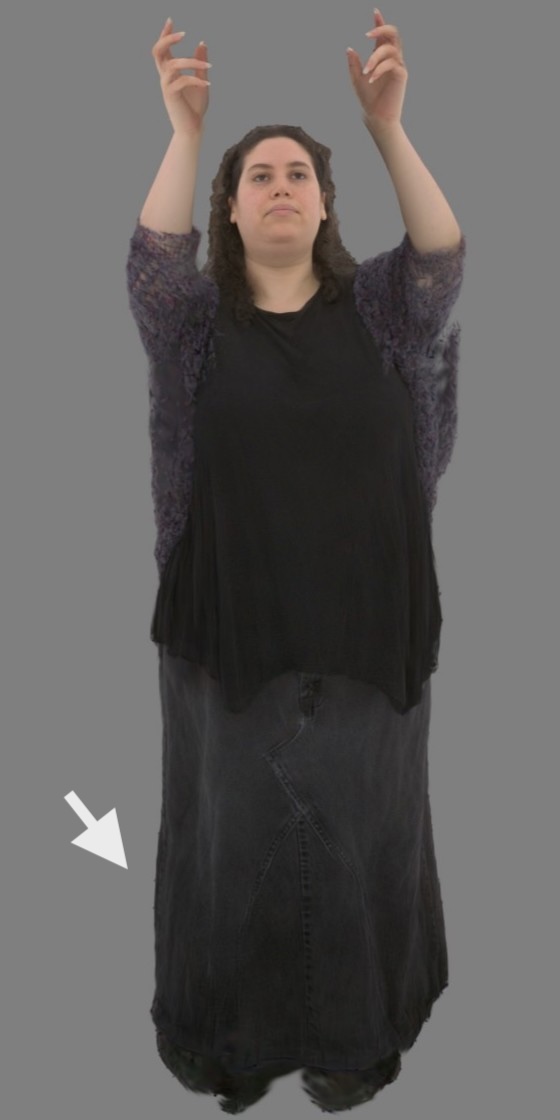} &
      \includegraphics[width=0.078\linewidth]{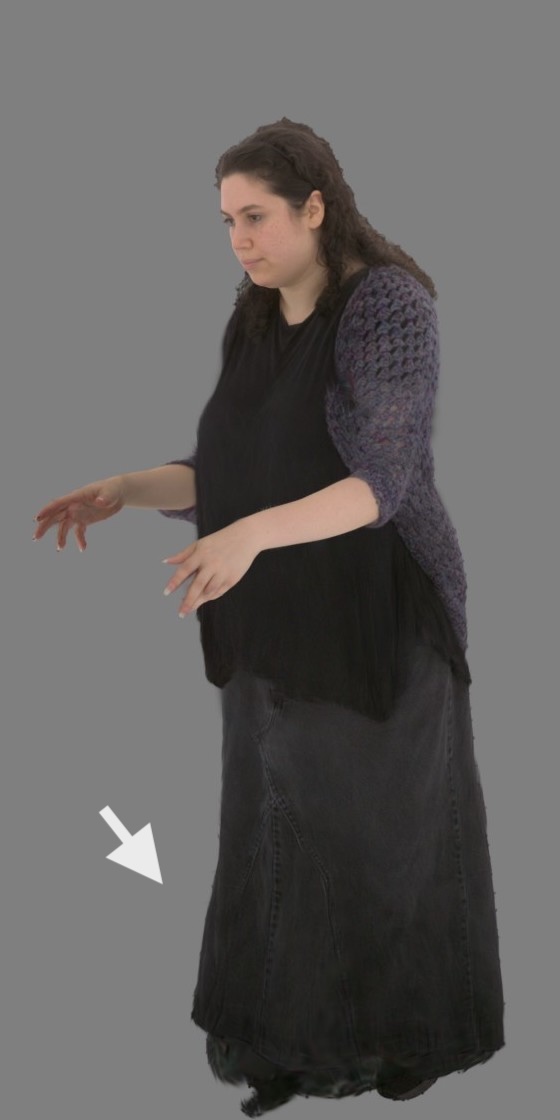} &
      \includegraphics[width=0.078\linewidth]{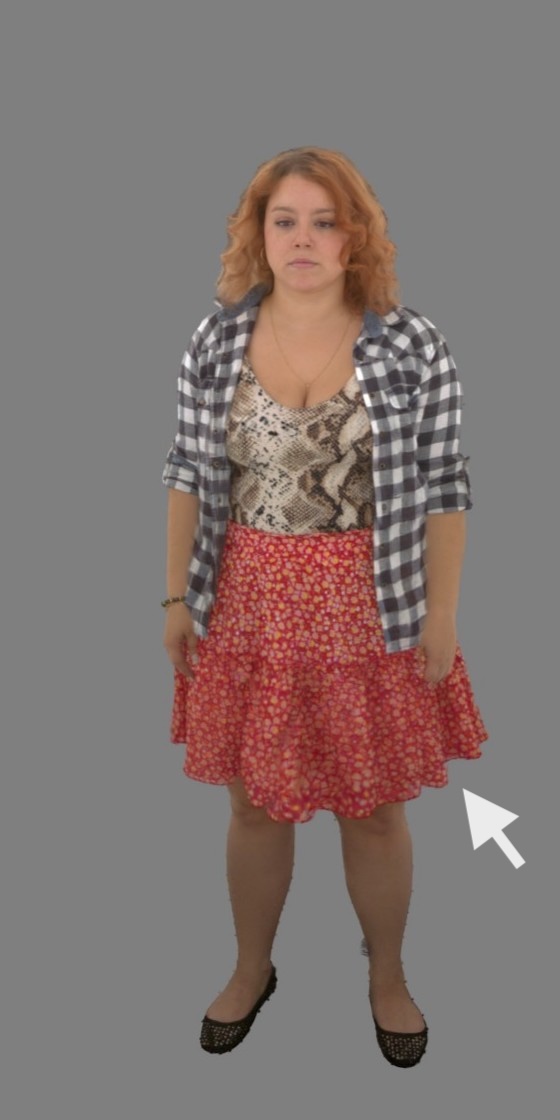} &
      \includegraphics[width=0.078\linewidth]{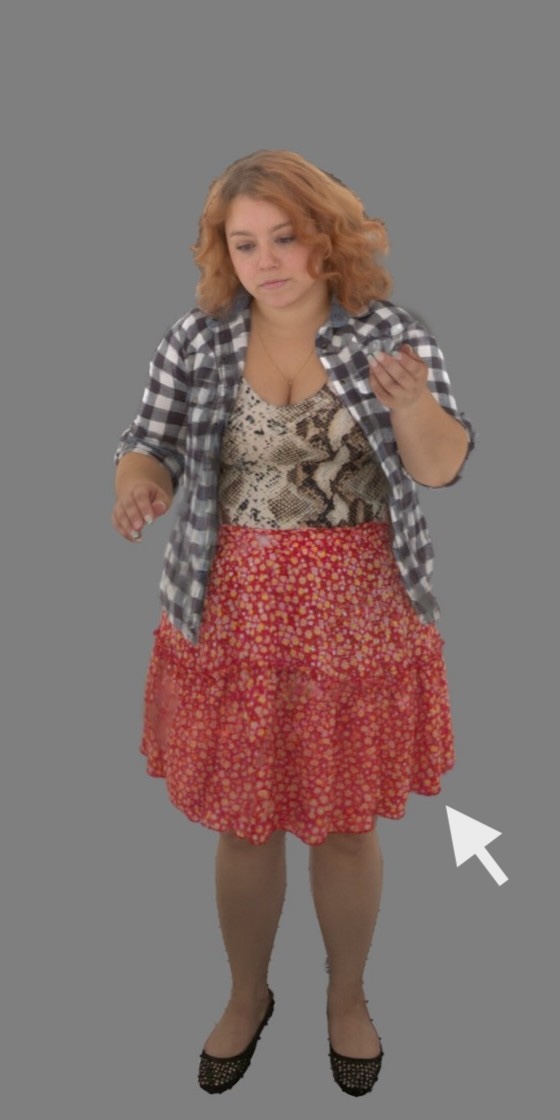} &
      \includegraphics[width=0.078\linewidth]{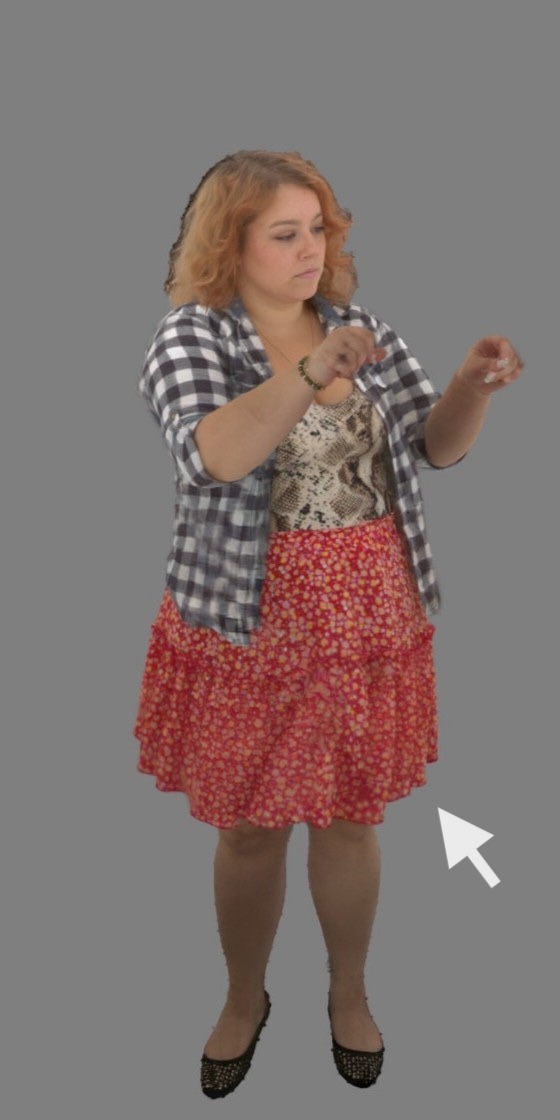} &
      \includegraphics[width=0.078\linewidth]{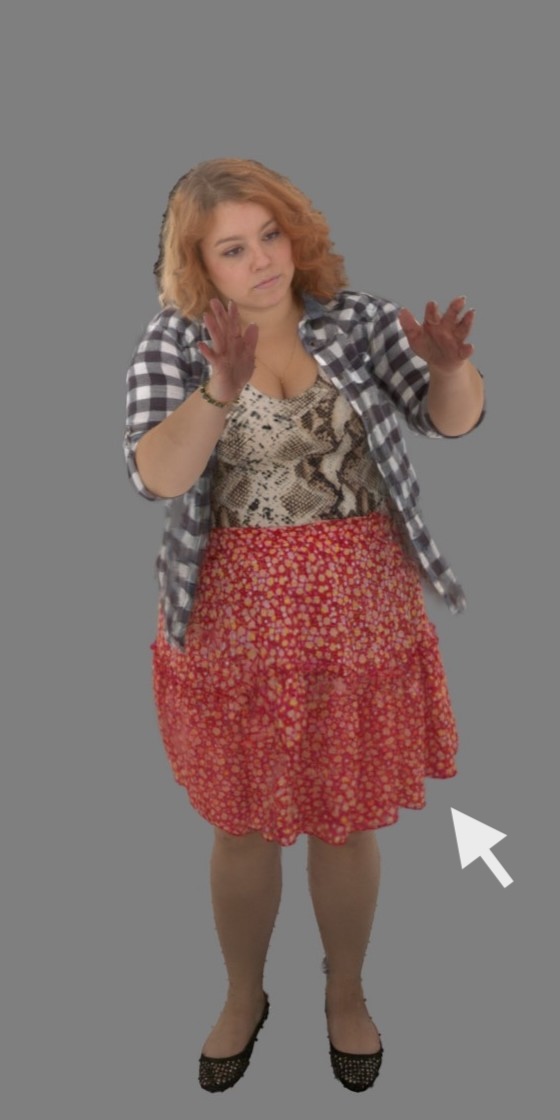} &
      \includegraphics[width=0.078\linewidth]{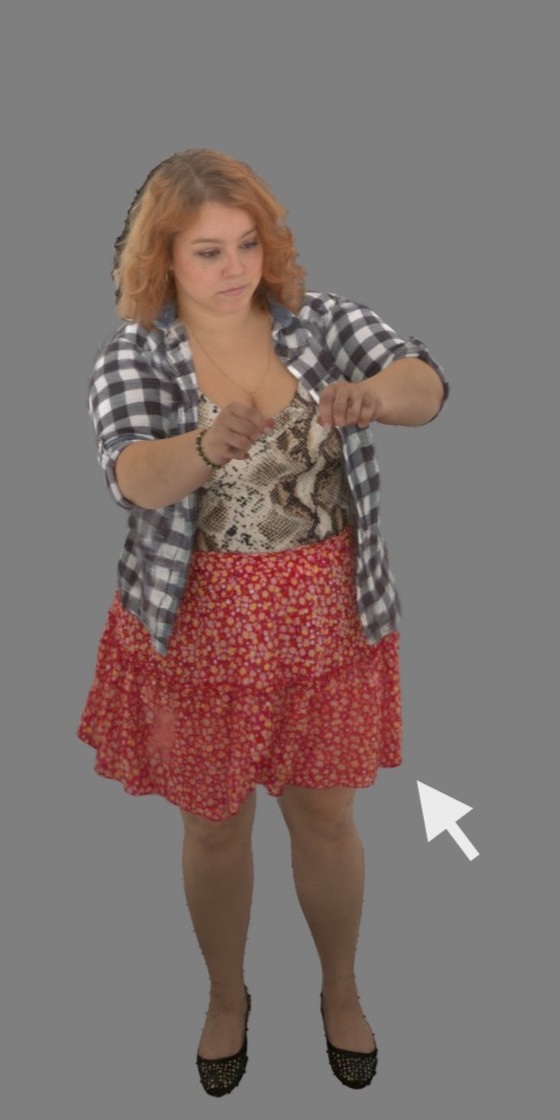} &
      \includegraphics[width=0.078\linewidth]{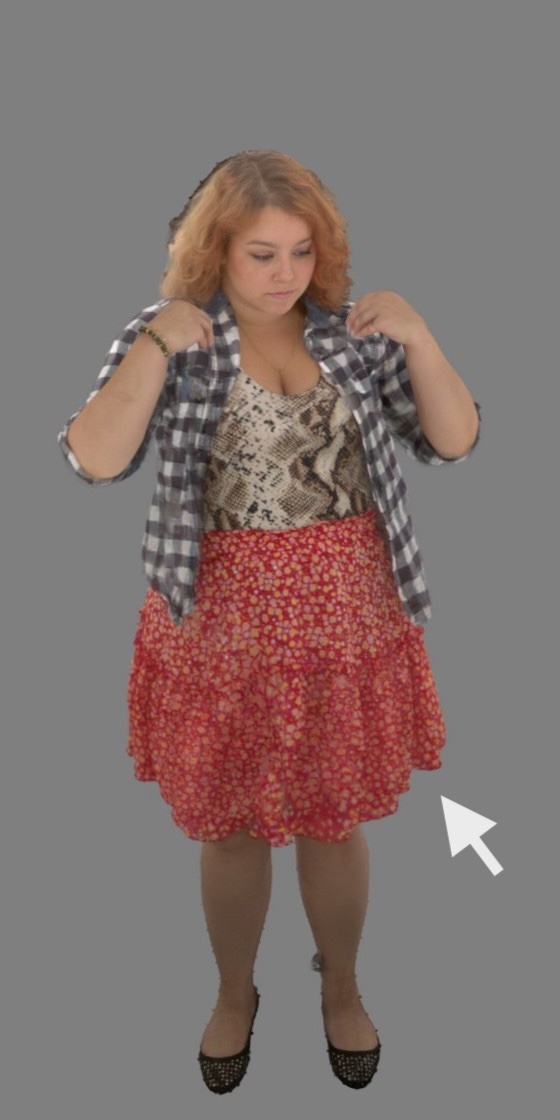} \\
    \end{tabular}
    \caption{\textbf{Cross-reenactment results across identities.} Source pose sequences are transferred to multiple target identities. Our method preserves identity-specific garment details while producing naturally-looking loose-clothing motion, including the swinging and folding of dresses and loose garments.}
    \label{fig:cross_grid}
\end{figure*}

\begin{figure*}[ht]
    \centering
    \small
    \setlength{\tabcolsep}{0pt}
    \newcommand{\imgw}{0.075\linewidth} 
    
    \begin{tabular}{ccc@{\hskip 6pt}ccc@{\hskip 6pt}ccc@{\hskip 6pt}ccc}
    \includegraphics[width=\imgw]{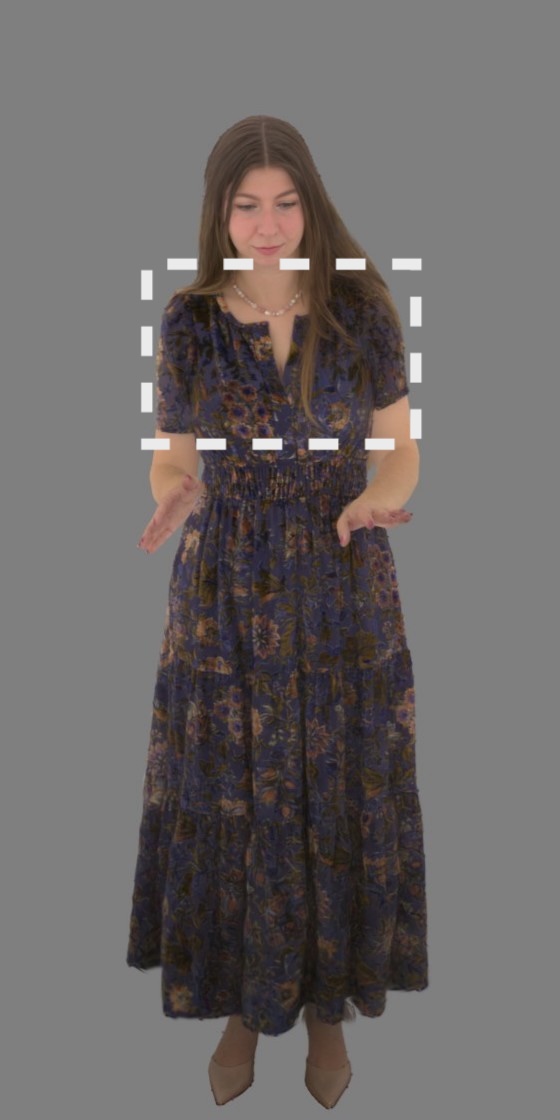} &
    \includegraphics[width=\imgw]{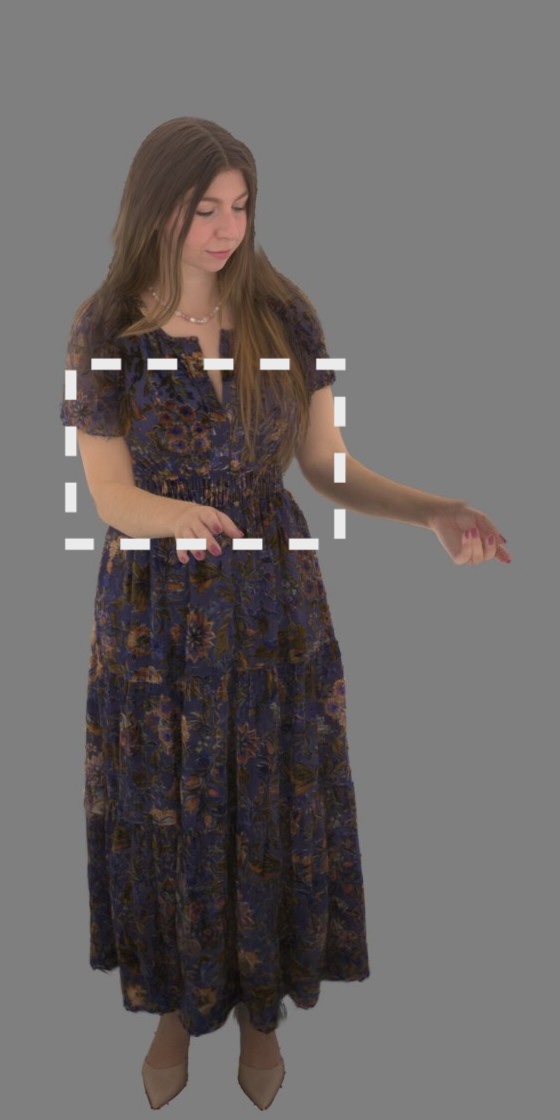} &
    \includegraphics[width=\imgw]{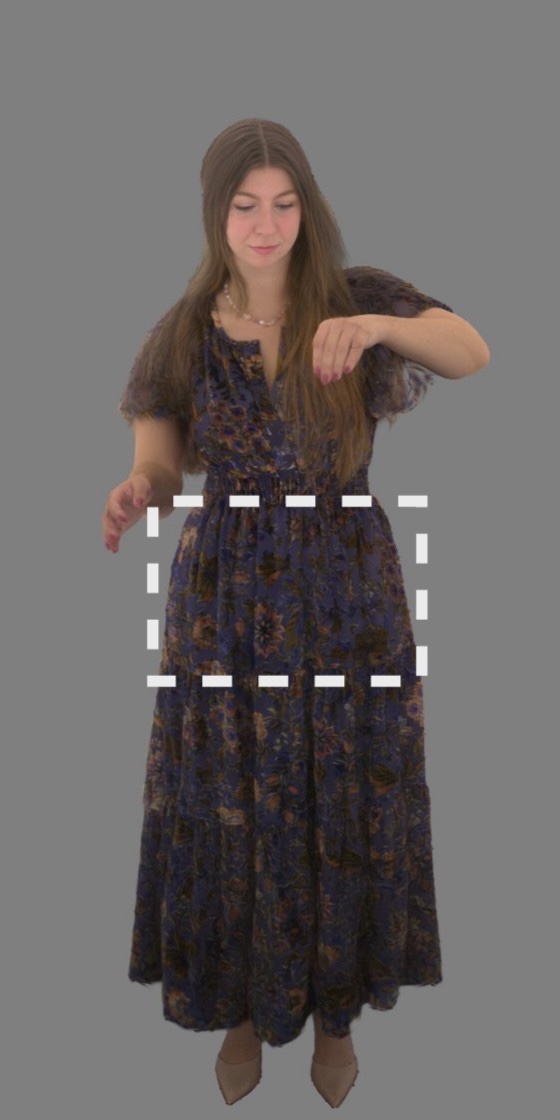} &
    \includegraphics[width=\imgw]{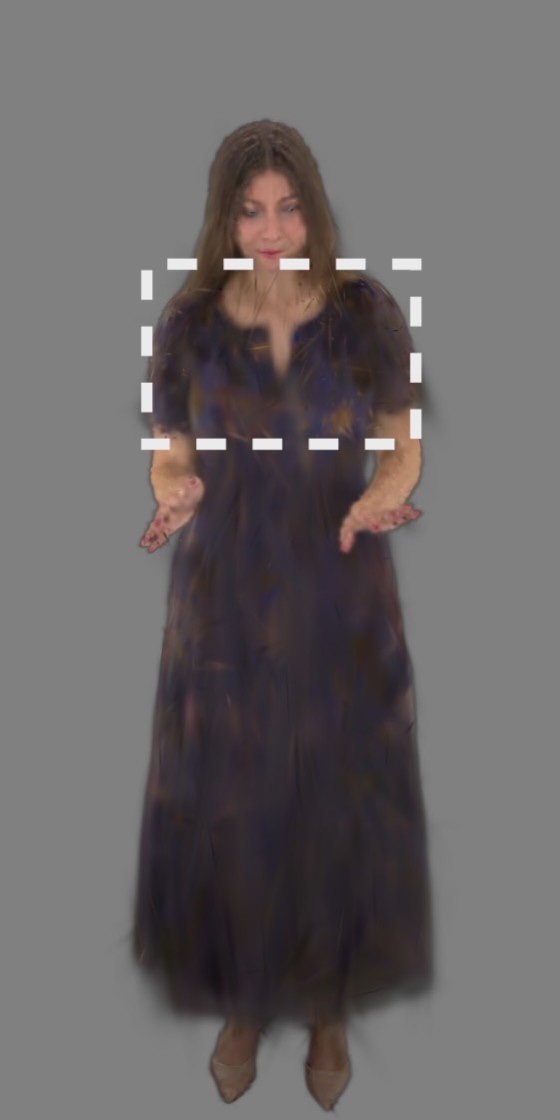} &
    \includegraphics[width=\imgw]{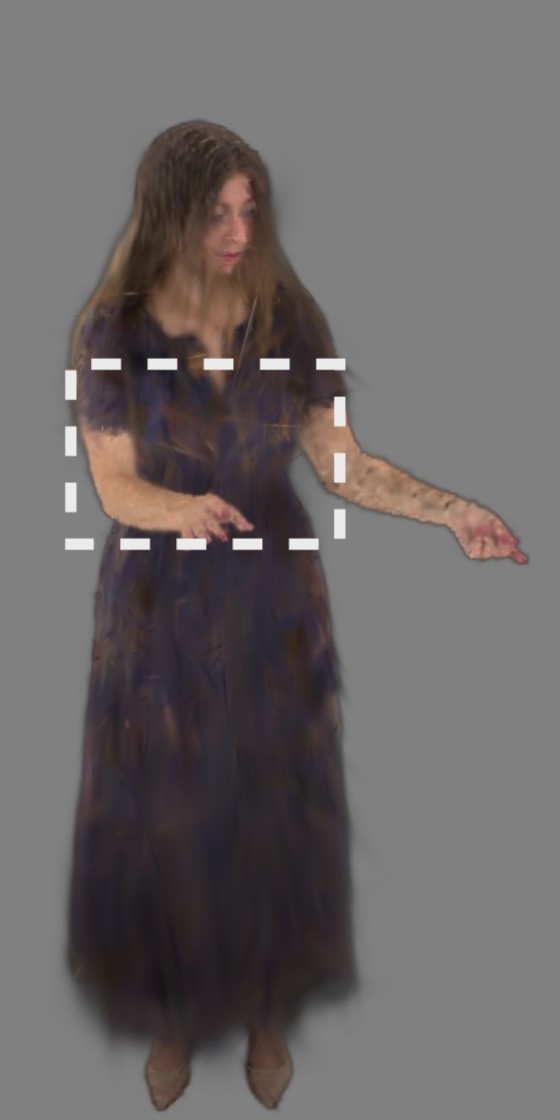} &
    \includegraphics[width=\imgw]{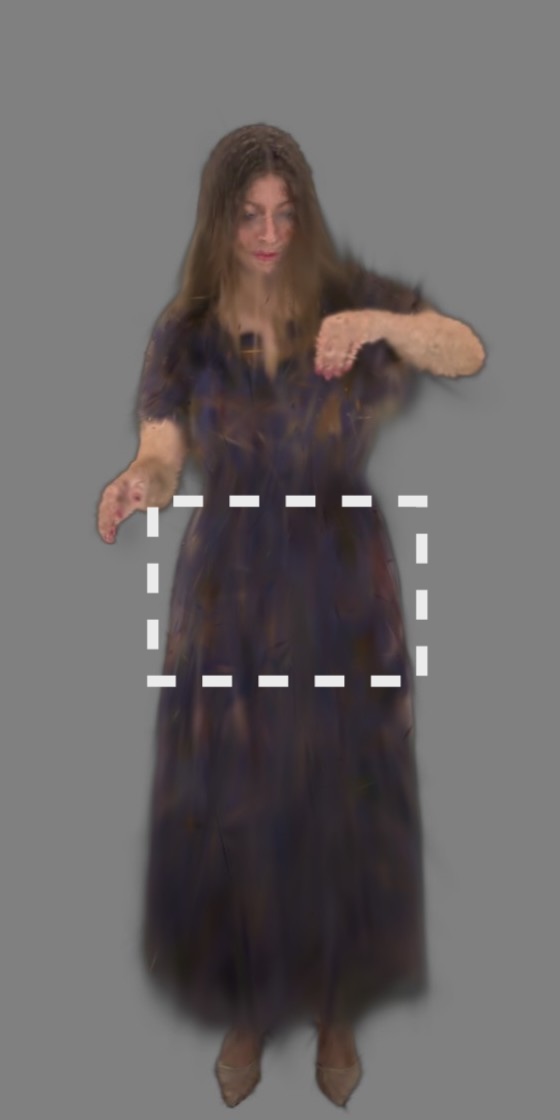} &
    \includegraphics[width=\imgw]{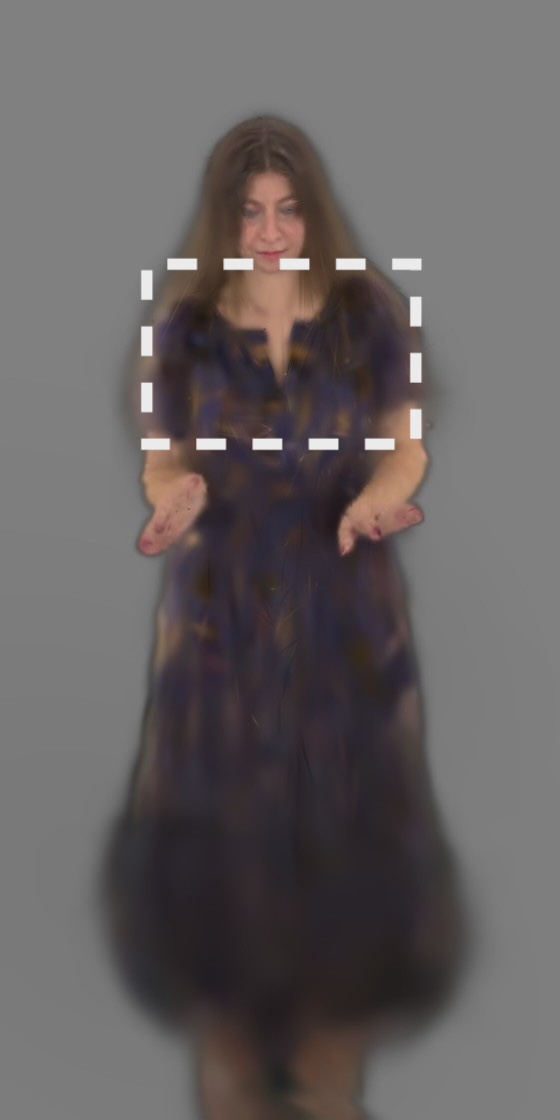} &
    \includegraphics[width=\imgw]{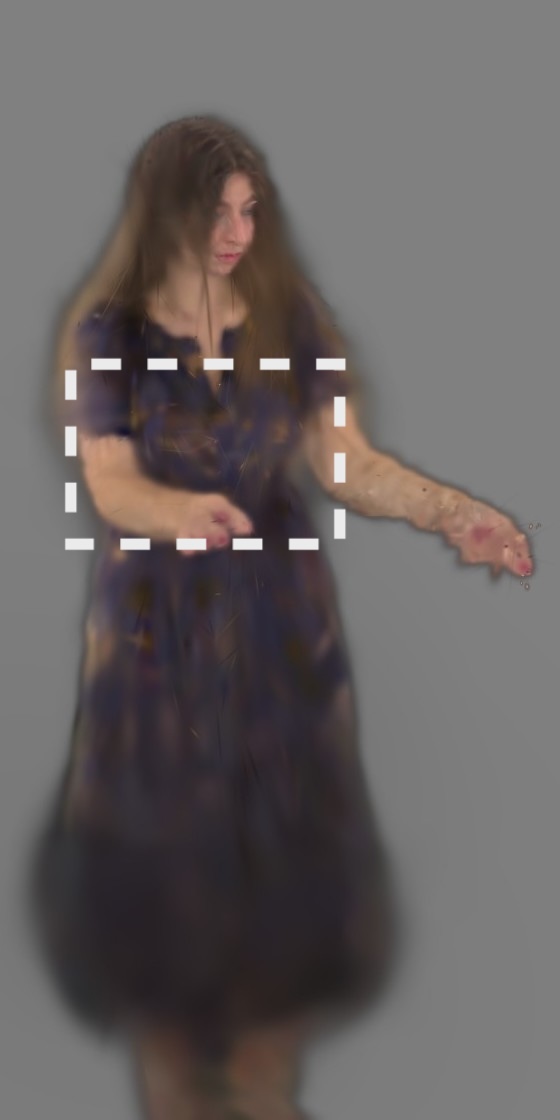} &
    \includegraphics[width=\imgw]{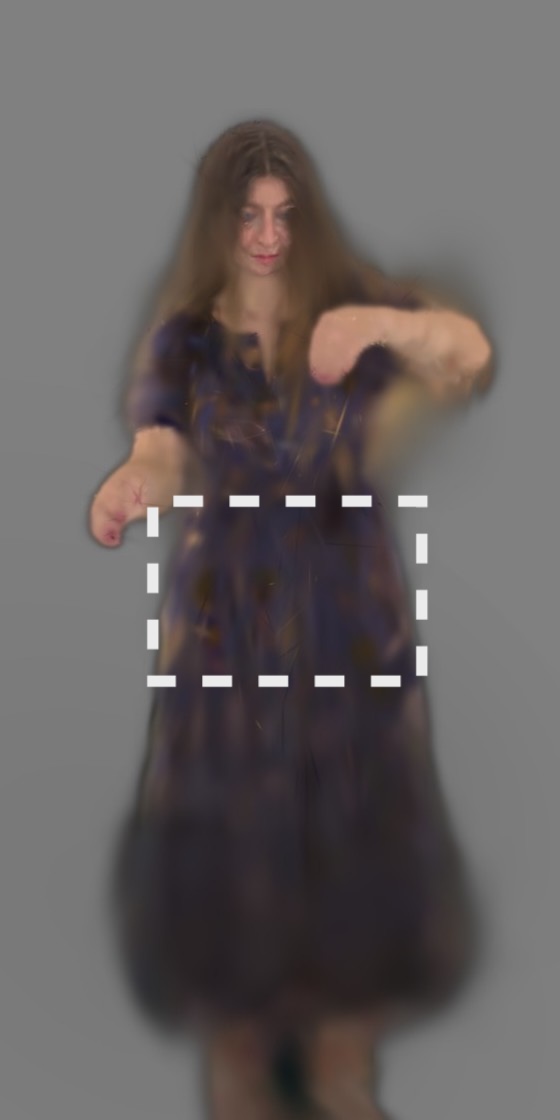} &
    \includegraphics[width=\imgw]{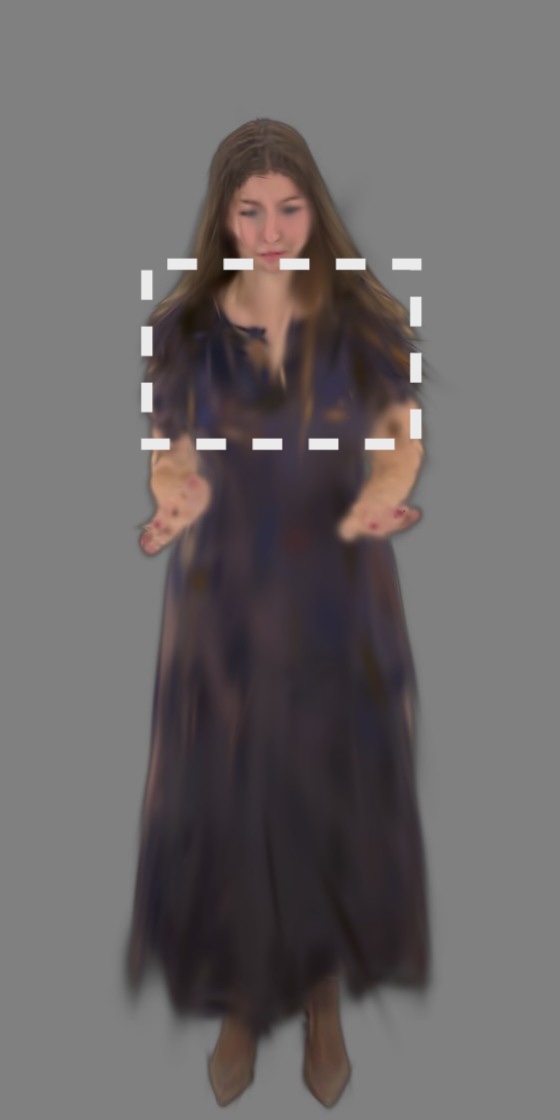} &
    \includegraphics[width=\imgw]{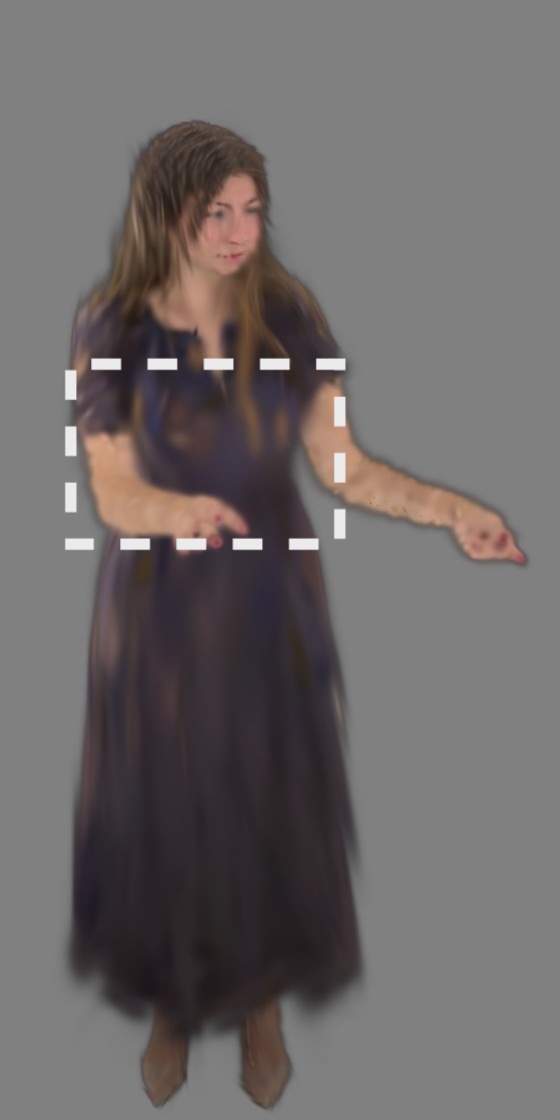} &
    \includegraphics[width=\imgw]{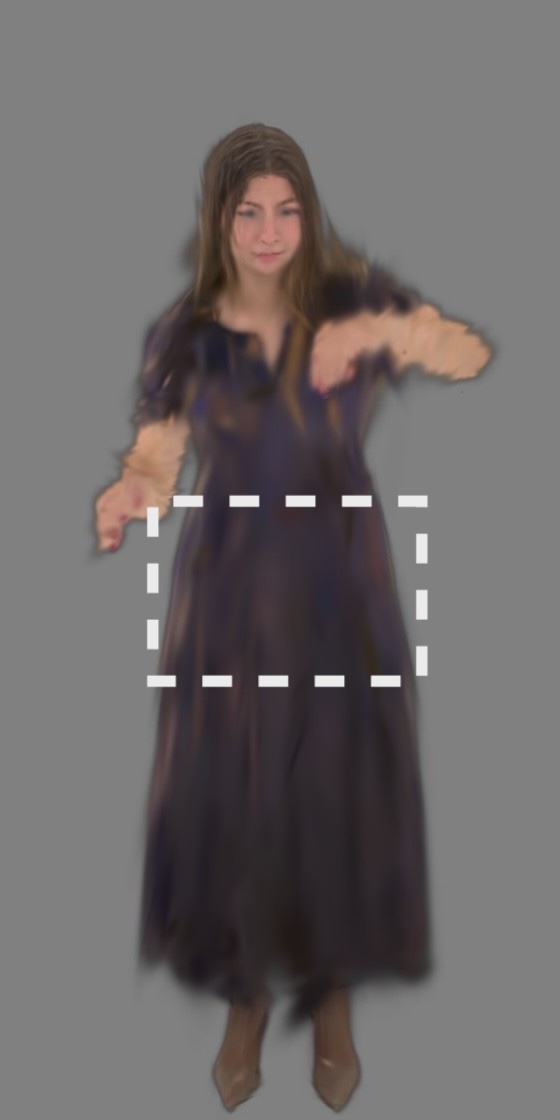} \\[-1pt]
    \includegraphics[width=\imgw]{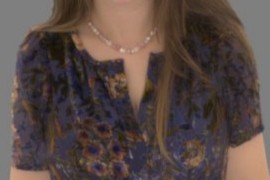} &
    \includegraphics[width=\imgw]{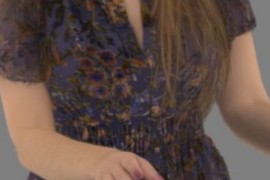} &
    \includegraphics[width=\imgw]{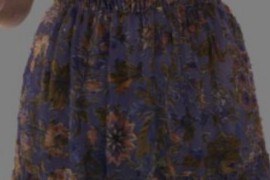} &
    \includegraphics[width=\imgw]{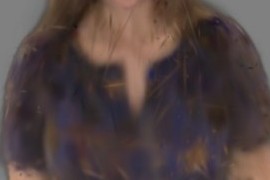} &
    \includegraphics[width=\imgw]{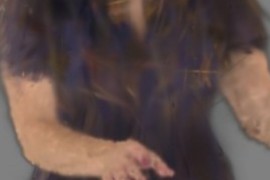} &
    \includegraphics[width=\imgw]{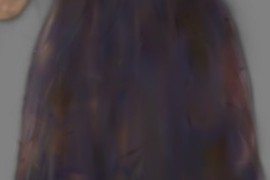} &
    \includegraphics[width=\imgw]{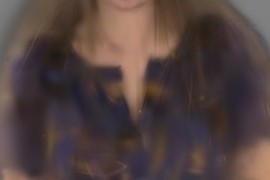} &
    \includegraphics[width=\imgw]{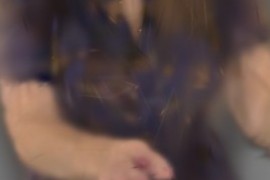} &
    \includegraphics[width=\imgw]{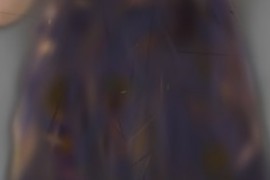} &
    \includegraphics[width=\imgw]{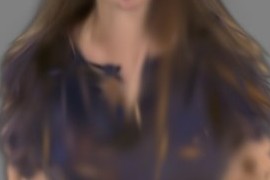} &
    \includegraphics[width=\imgw]{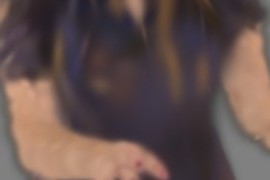} &
    \includegraphics[width=\imgw]{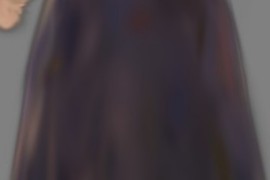} \\[2pt]
    
    \includegraphics[width=\imgw]{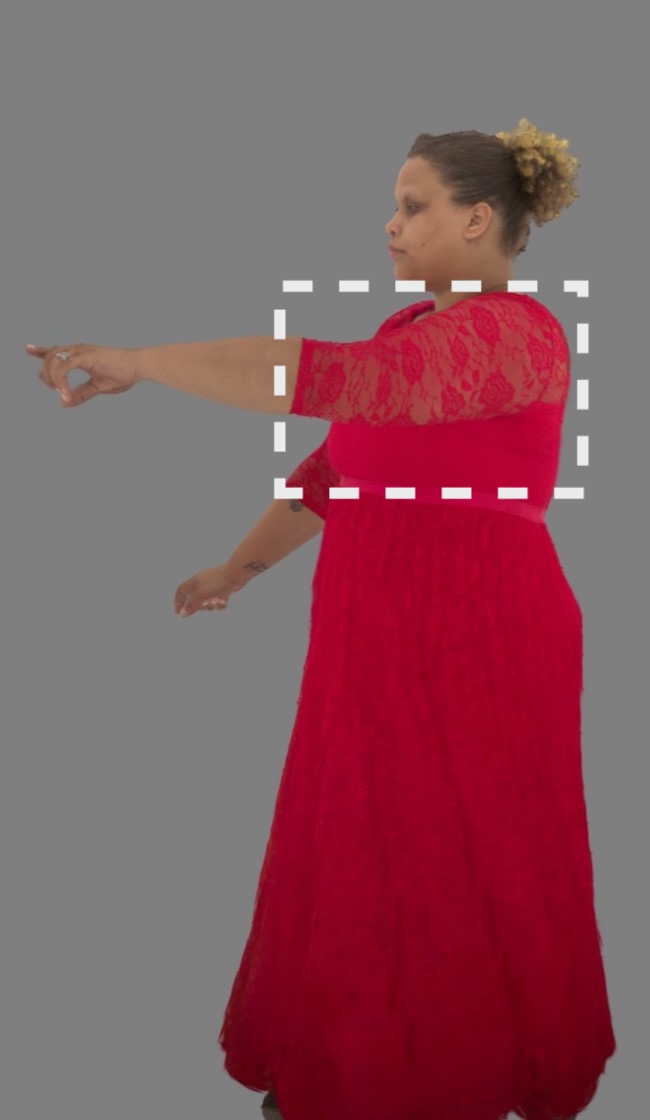} &
    \includegraphics[width=\imgw]{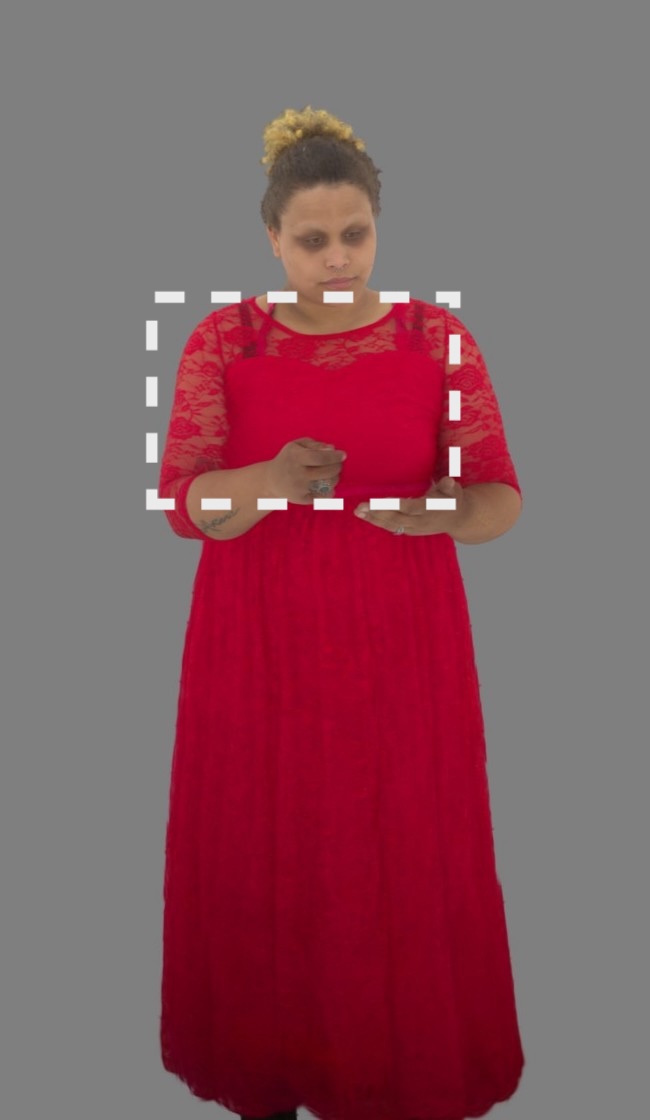} &
    \includegraphics[width=\imgw]{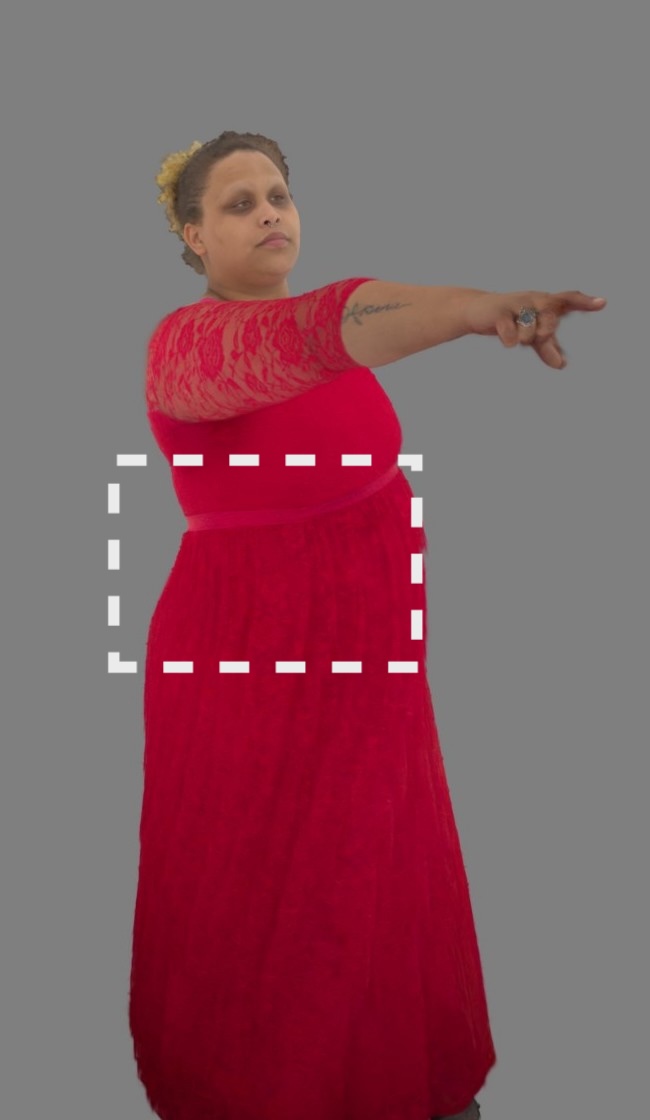} &
    \includegraphics[width=\imgw]{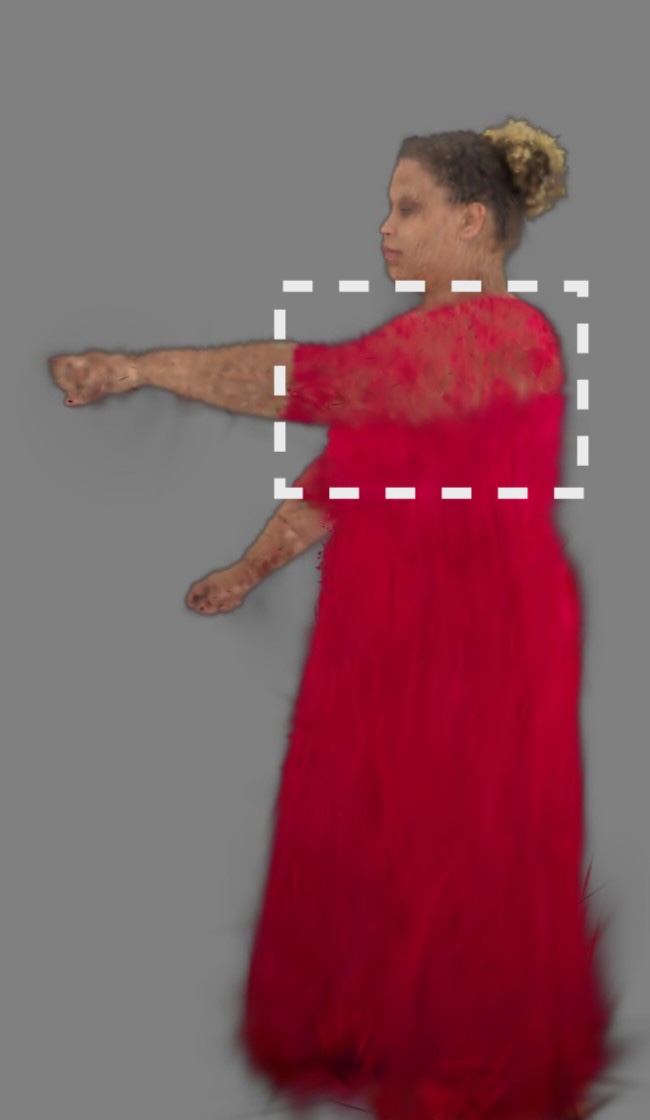} &
    \includegraphics[width=\imgw]{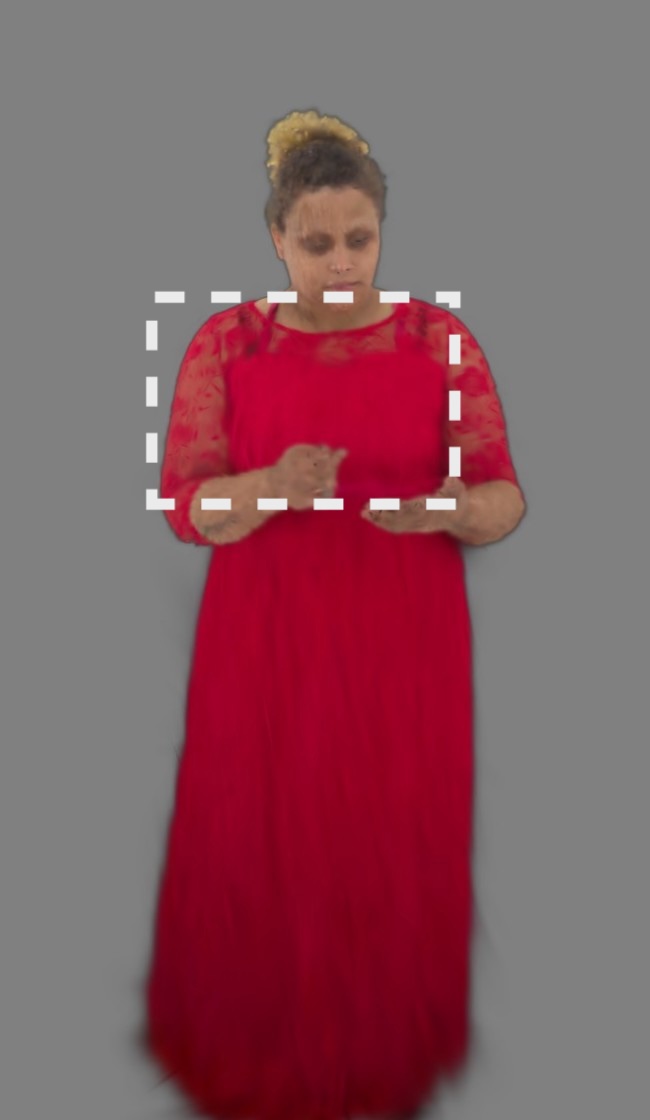} &
    \includegraphics[width=\imgw]{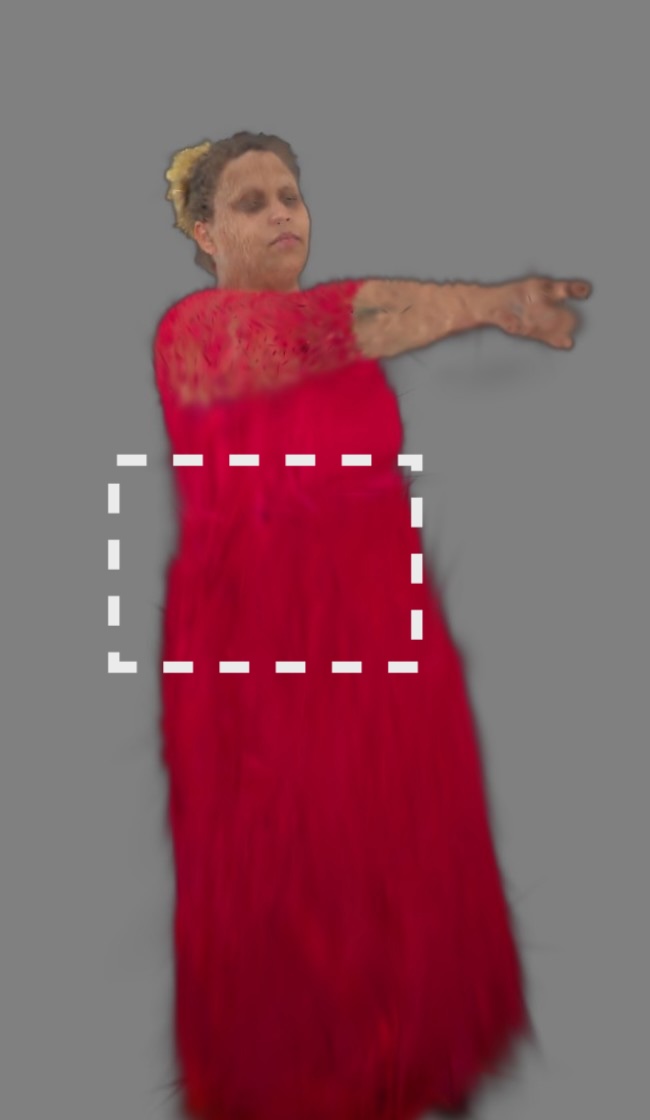} &
    \includegraphics[width=\imgw]{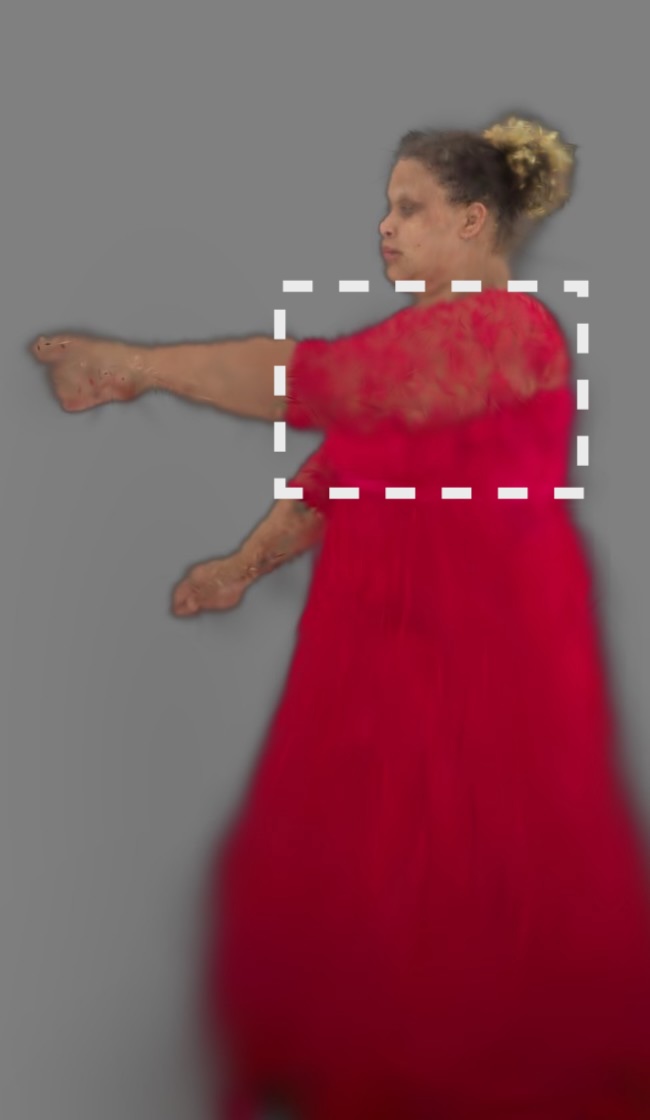} &
    \includegraphics[width=\imgw]{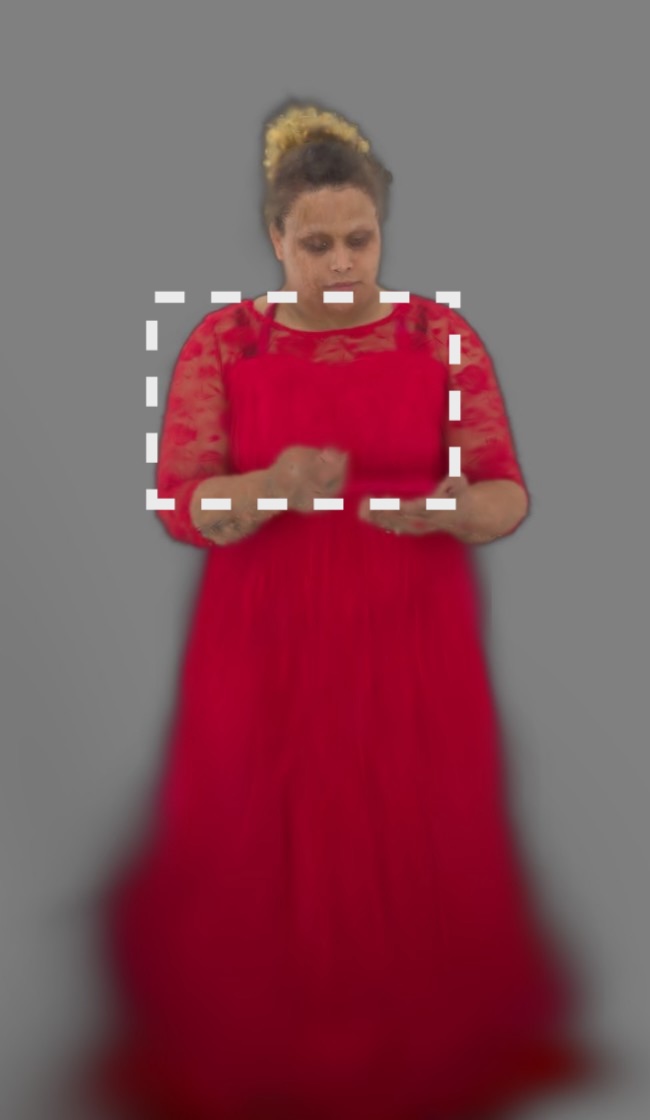} &
    \includegraphics[width=\imgw]{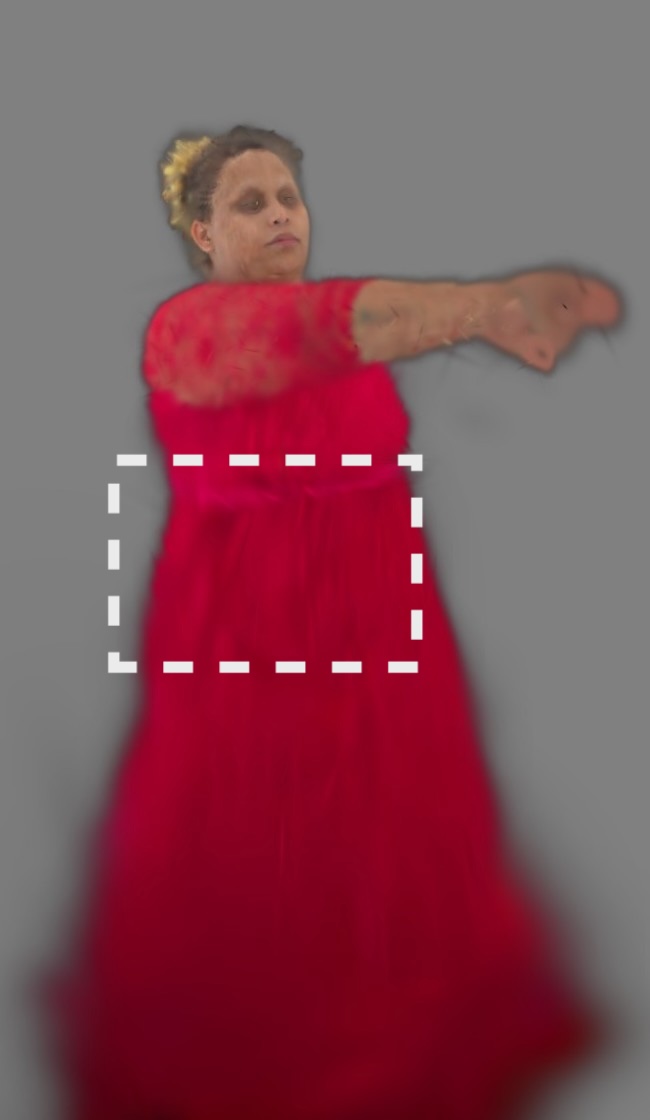} &
    \includegraphics[width=\imgw]{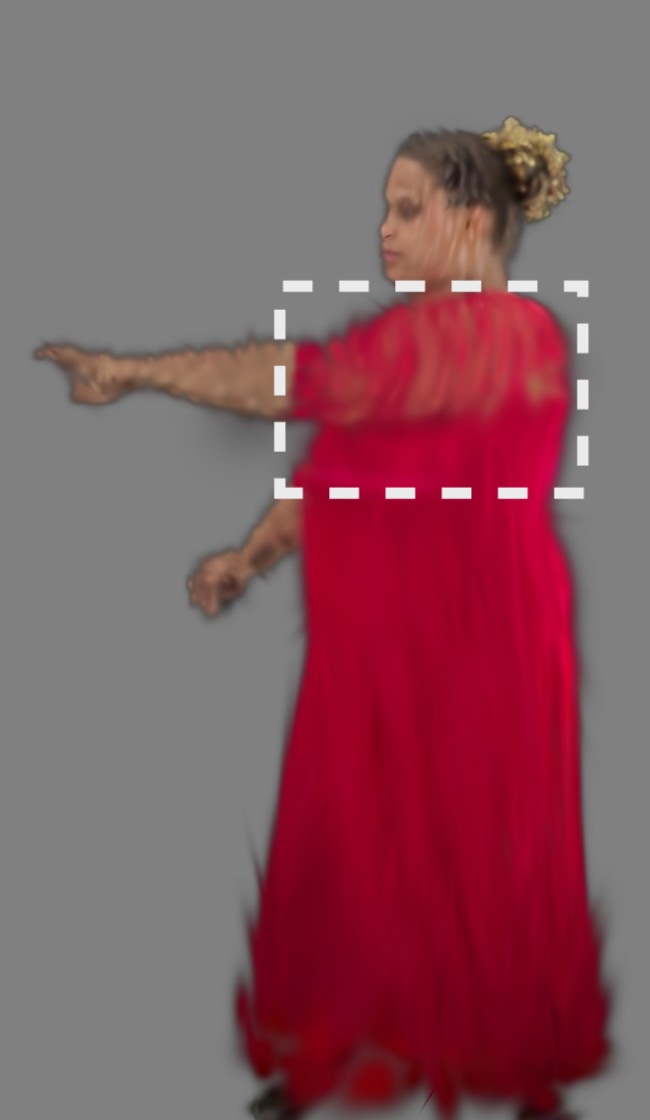} &
    \includegraphics[width=\imgw]{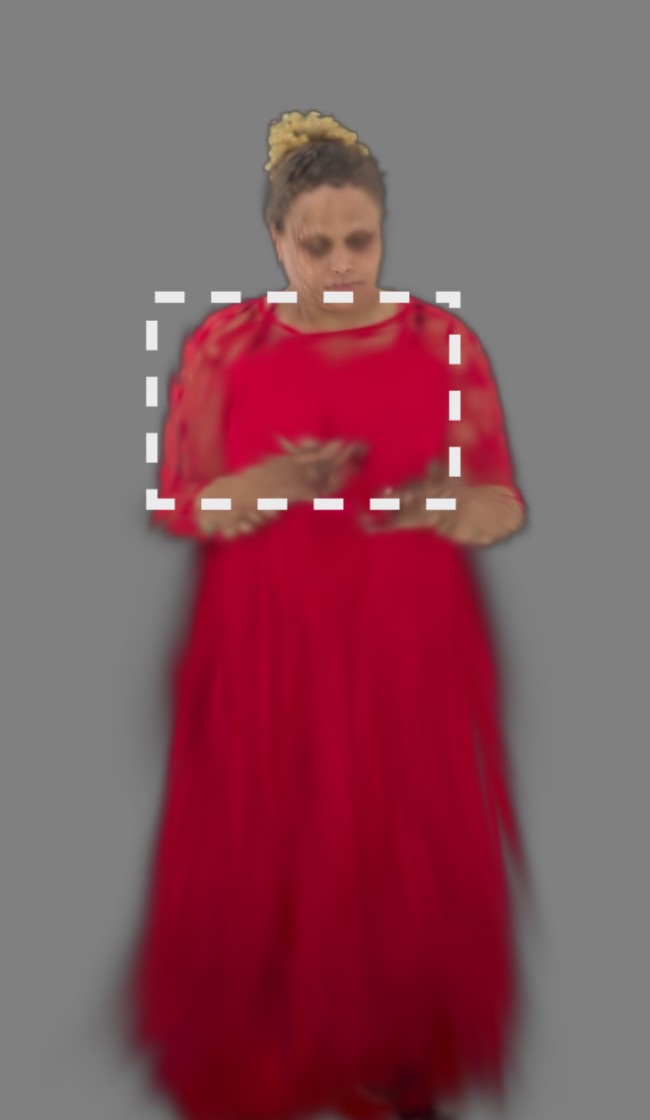} &
    \includegraphics[width=\imgw]{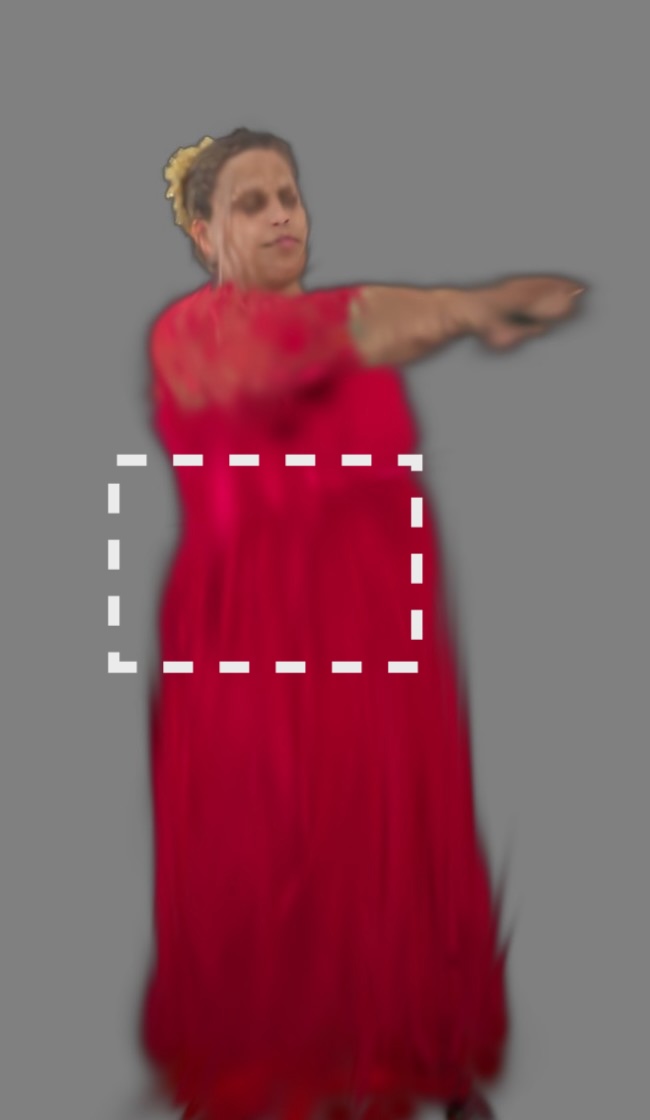} \\[-1pt]
    \includegraphics[width=\imgw]{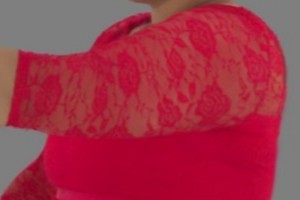} &
    \includegraphics[width=\imgw]{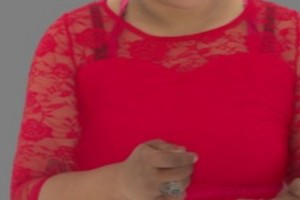} &
    \includegraphics[width=\imgw]{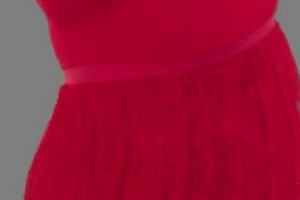} &
    \includegraphics[width=\imgw]{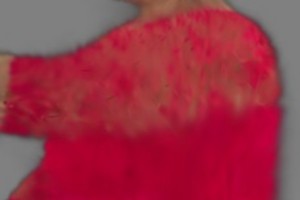} &
    \includegraphics[width=\imgw]{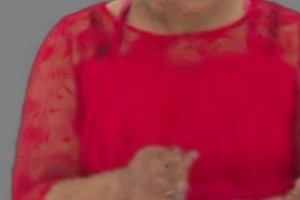} &
    \includegraphics[width=\imgw]{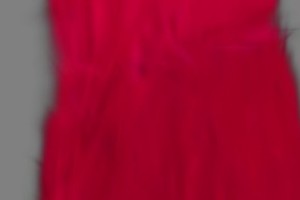} &
    \includegraphics[width=\imgw]{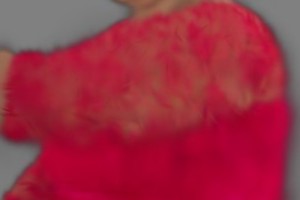} &
    \includegraphics[width=\imgw]{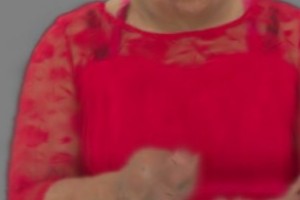} &
    \includegraphics[width=\imgw]{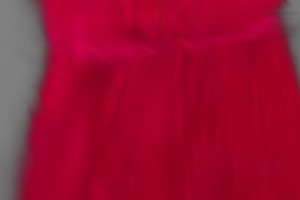} &
    \includegraphics[width=\imgw]{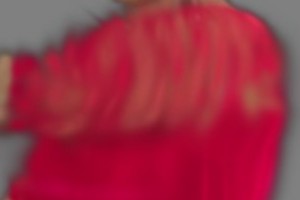} &
    \includegraphics[width=\imgw]{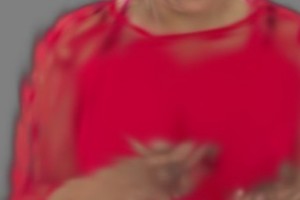} &
    \includegraphics[width=\imgw]{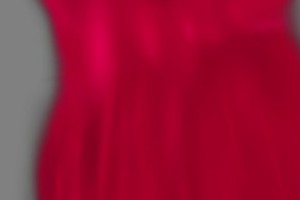} \\[2pt]
    
    \includegraphics[width=\imgw]{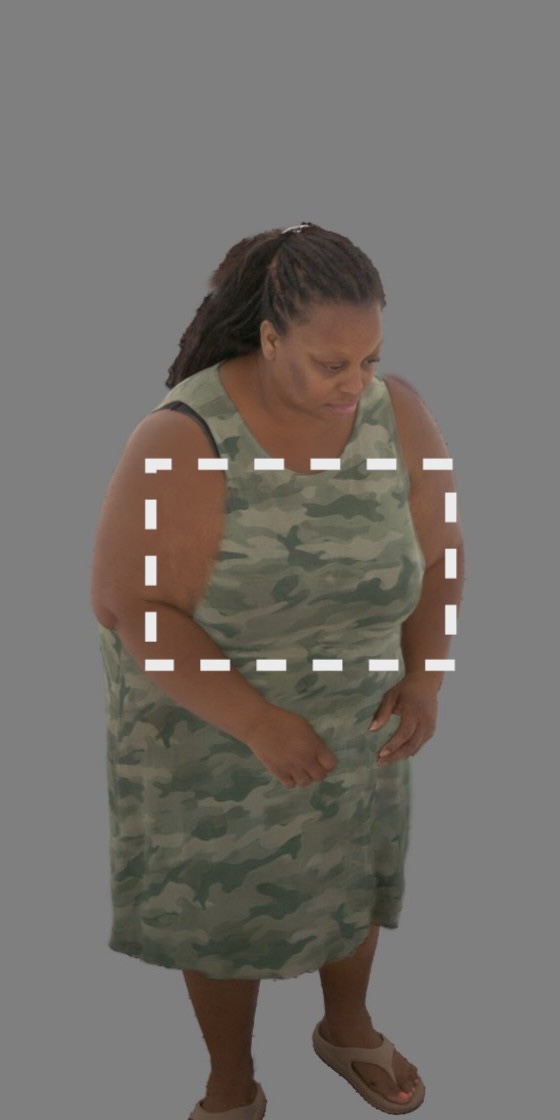} &
    \includegraphics[width=\imgw]{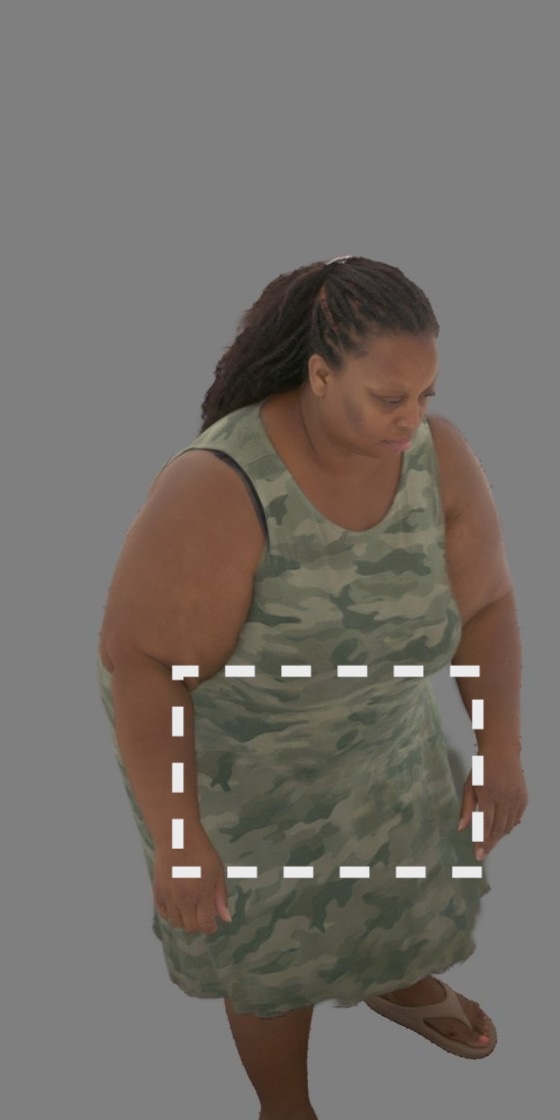} &
    \includegraphics[width=\imgw]{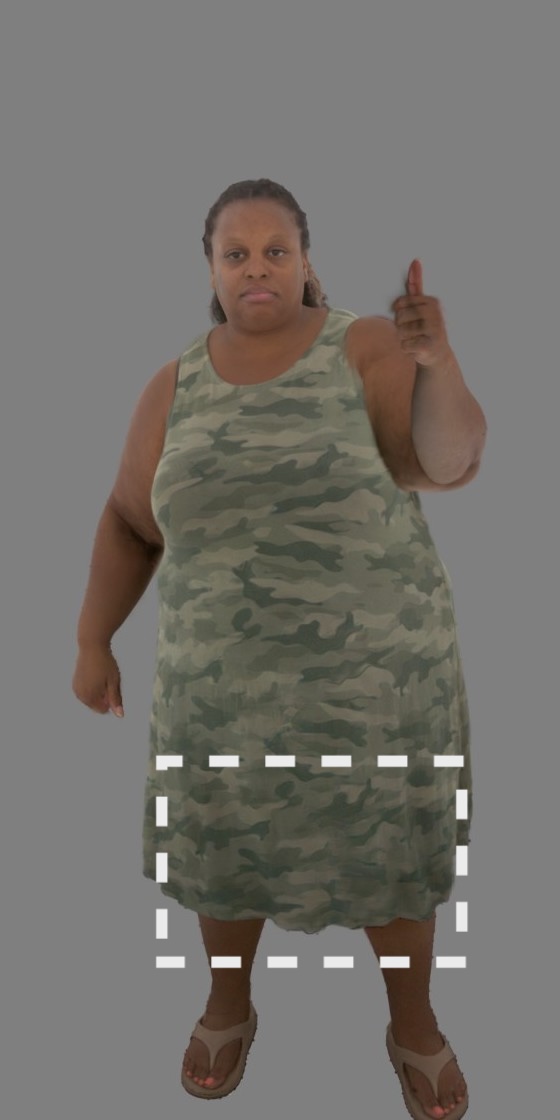} &
    \includegraphics[width=\imgw]{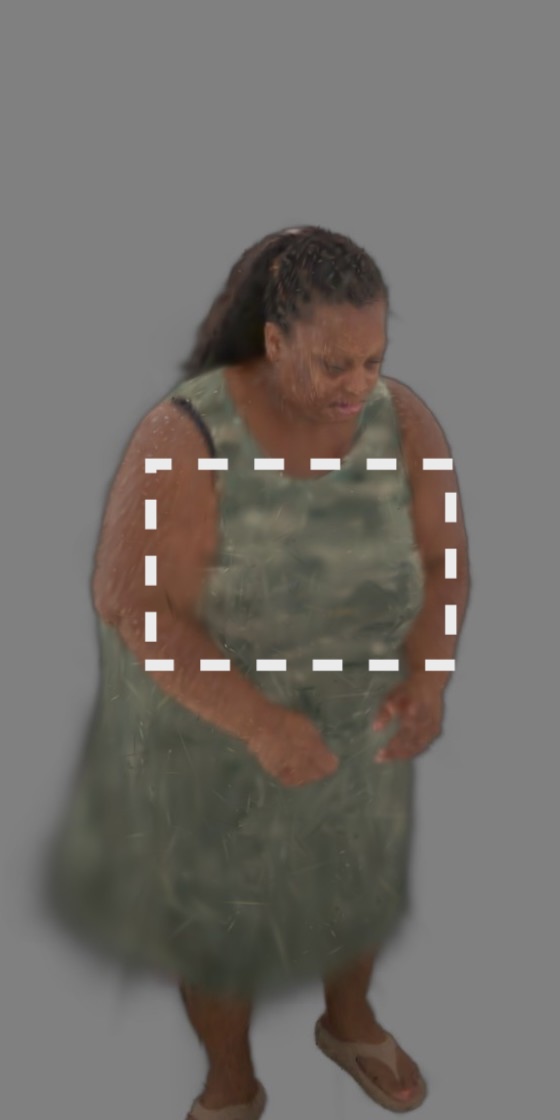} &
    \includegraphics[width=\imgw]{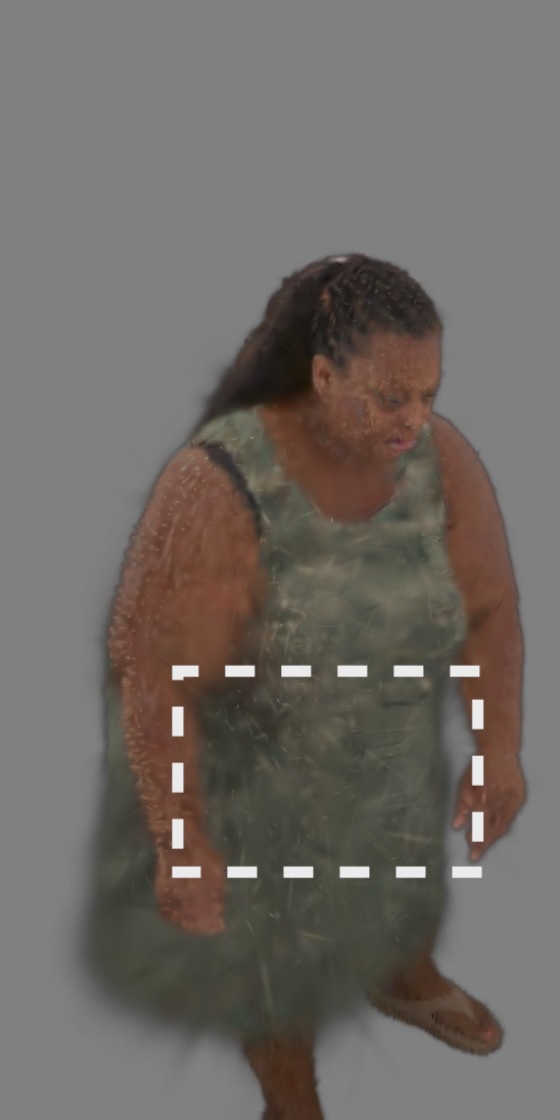} &
    \includegraphics[width=\imgw]{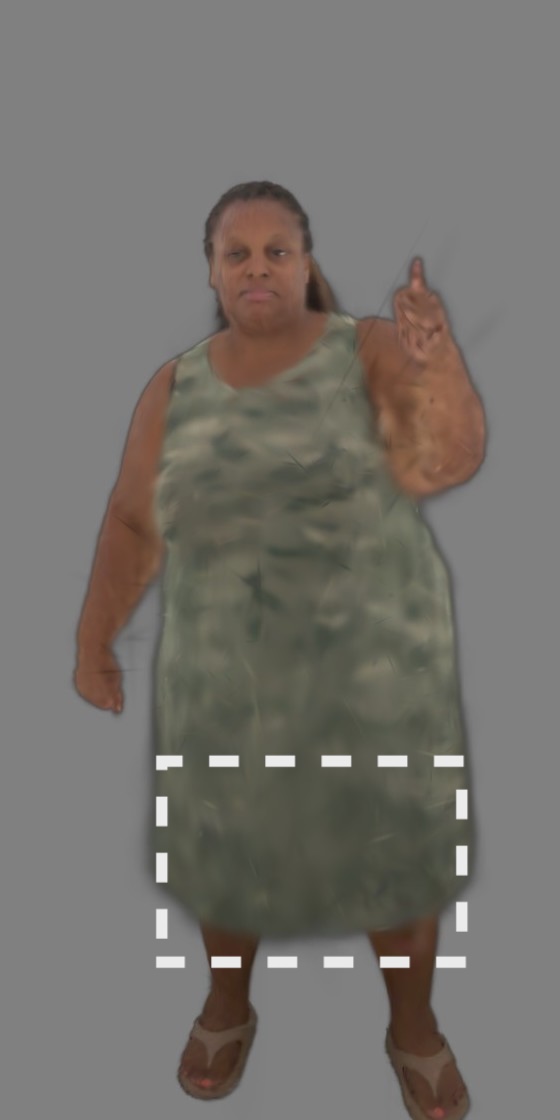} &
    \includegraphics[width=\imgw]{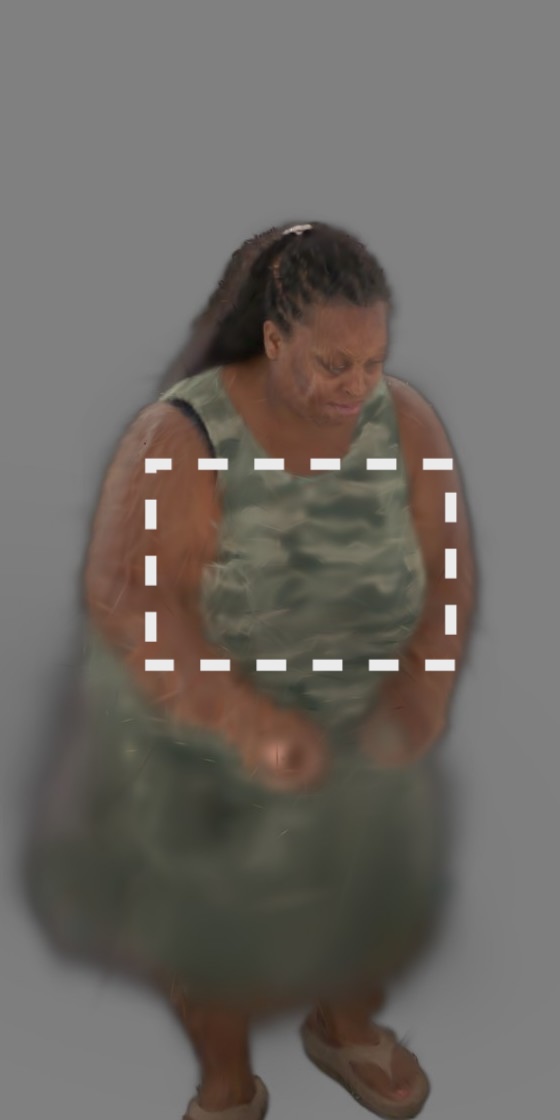} &
    \includegraphics[width=\imgw]{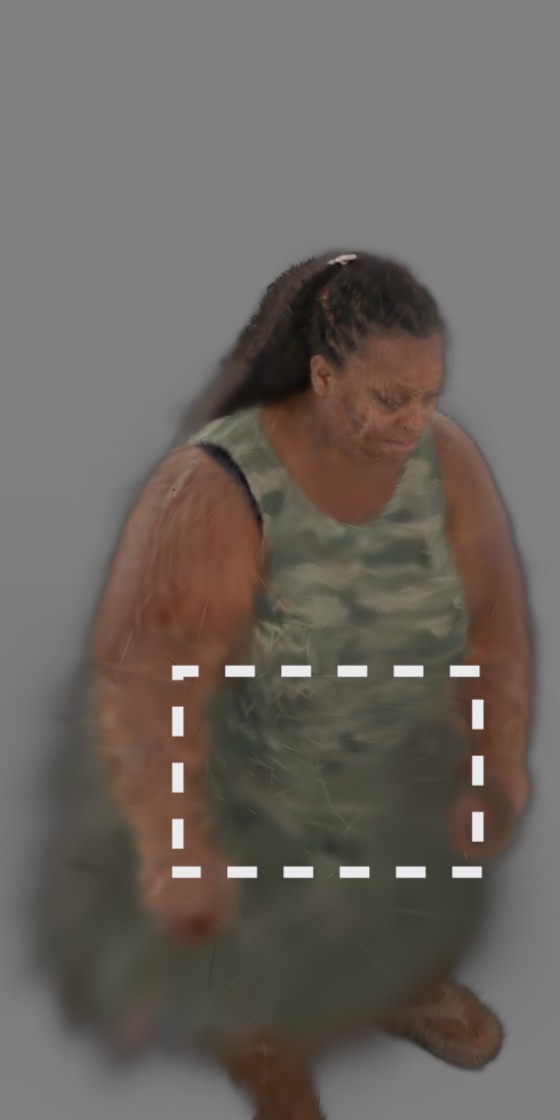} &
    \includegraphics[width=\imgw]{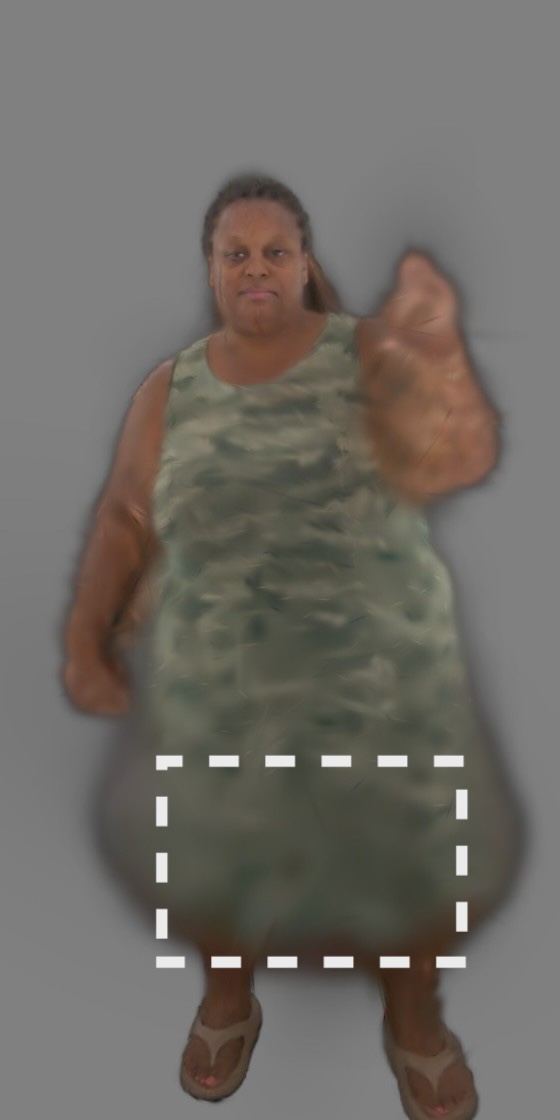} &
    \includegraphics[width=\imgw]{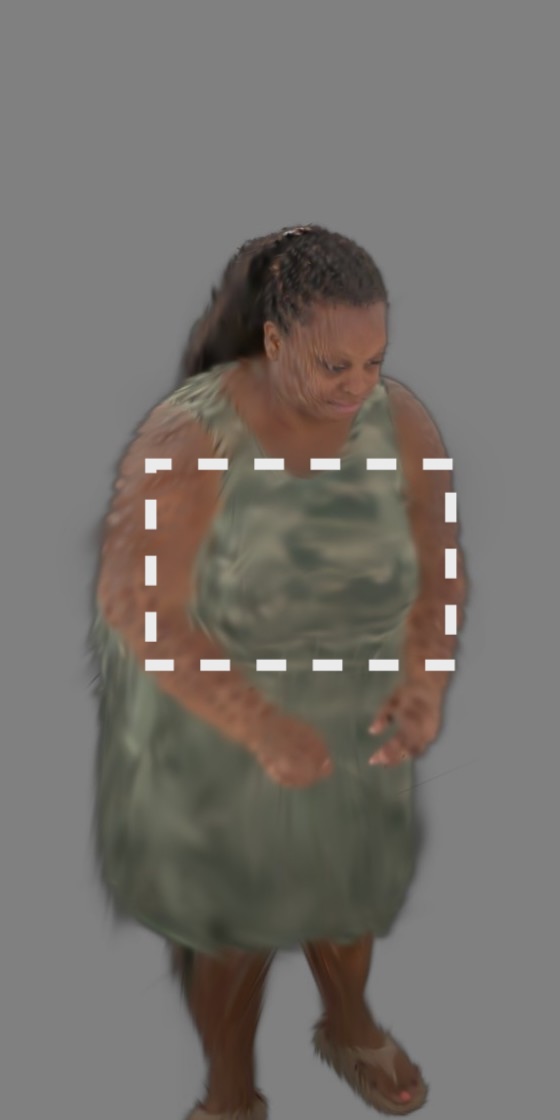} &
    \includegraphics[width=\imgw]{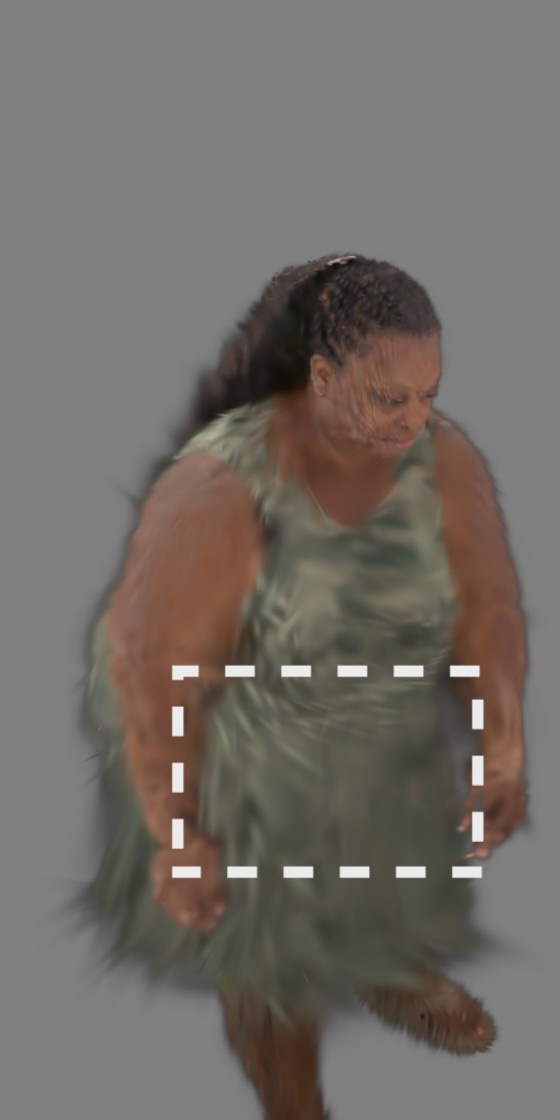} &
    \includegraphics[width=\imgw]{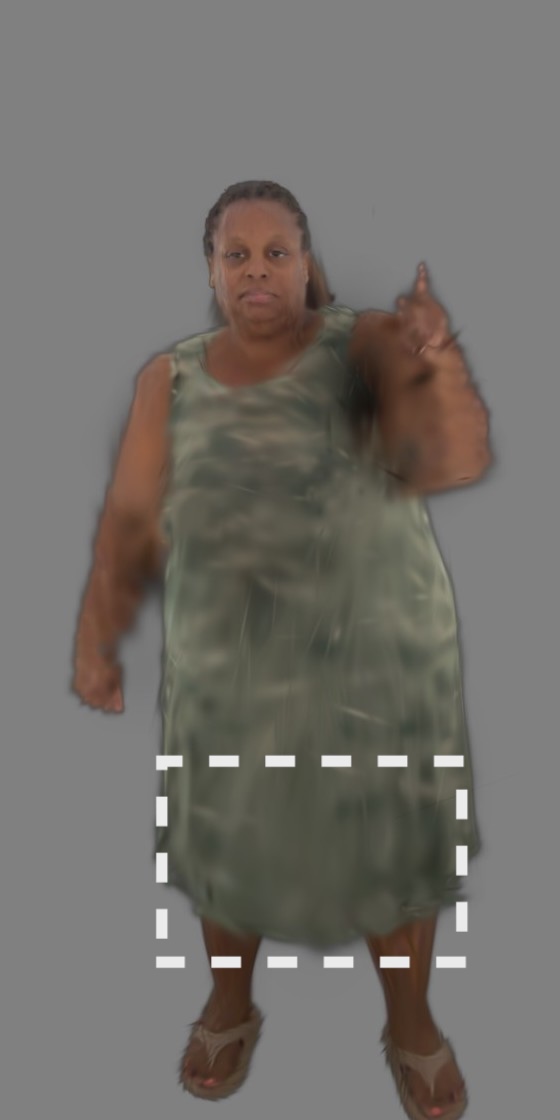} \\[-1pt]
    \includegraphics[width=\imgw]{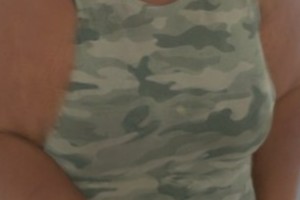} &
    \includegraphics[width=\imgw]{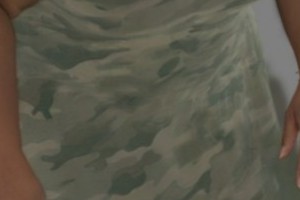} &
    \includegraphics[width=\imgw]{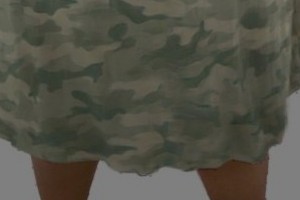} &
    \includegraphics[width=\imgw]{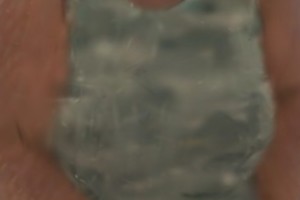} &
    \includegraphics[width=\imgw]{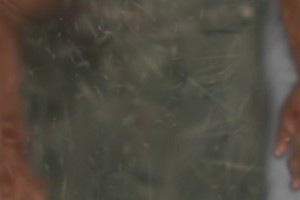} &
    \includegraphics[width=\imgw]{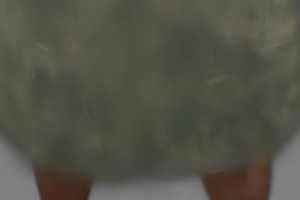} &
    \includegraphics[width=\imgw]{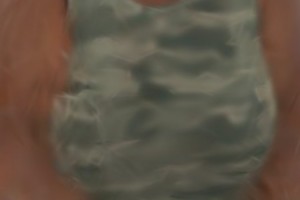} &
    \includegraphics[width=\imgw]{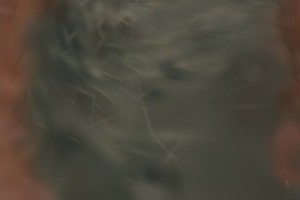} &
    \includegraphics[width=\imgw]{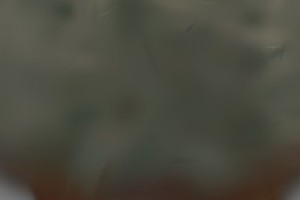} &
    \includegraphics[width=\imgw]{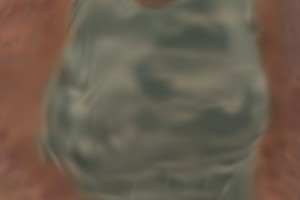} &
    \includegraphics[width=\imgw]{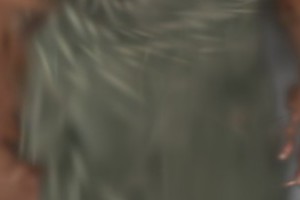} &
    \includegraphics[width=\imgw]{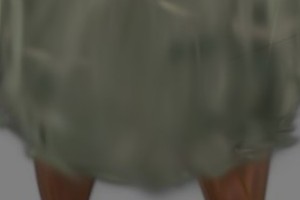} \\[0pt]
    
    \multicolumn{3}{c}{Ours} &
    \multicolumn{3}{c}{ToMiE~\cite{zhan2024tomiemodulargrowthenhanced}} &
    \multicolumn{3}{c}{Seq-Avatar~\cite{xu2025seqavatar}} &
    \multicolumn{3}{c}{$R^3$-Avatar~\cite{Zhan2025R3AvatarRA}} \\
    \end{tabular}
    \caption{\textbf{Self-reenactment results on held-out novel pose sequences.} Our method generates plausible loose-clothing motion while maintaining high rendering quality, preserving fine garment patterns and realistic wrinkles. In contrast, baseline methods tend to lose high-frequency details in high-motion regions and introduce blur or artifacts near garment boundaries.}
    \label{fig:novel_pose}
\end{figure*}

\begin{figure*}[ht]
  \newcommand{\initStateAblationWidth}{0.17\columnwidth}
  \centering
  \footnotesize
  \setlength{\tabcolsep}{2pt}
  \begin{tabular}{@{}c c@{}}
    {\setlength{\tabcolsep}{0pt}
    \begin{tabular}{@{}c*{5}{c}@{}}
      Init Latent 1 &
      \includegraphics[width=\initStateAblationWidth]{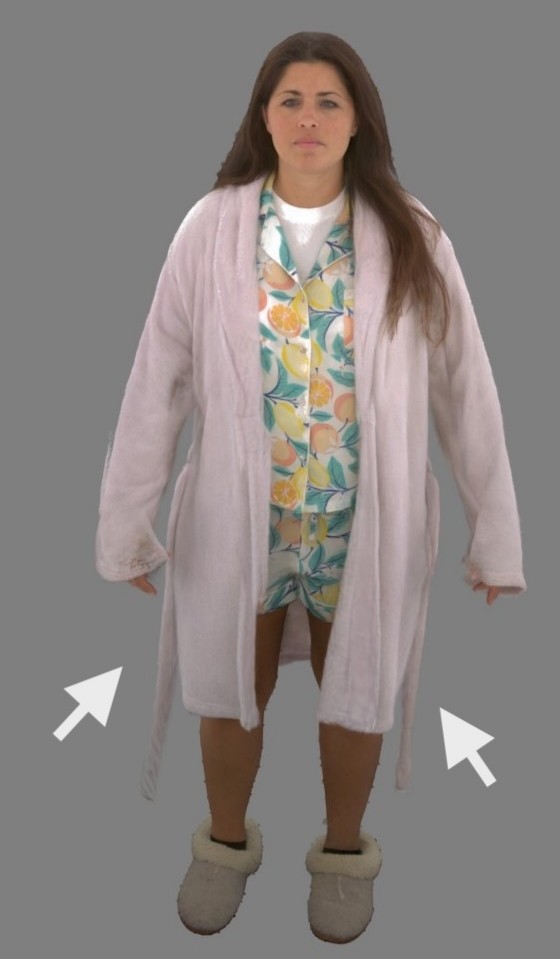} &
      \includegraphics[width=\initStateAblationWidth]{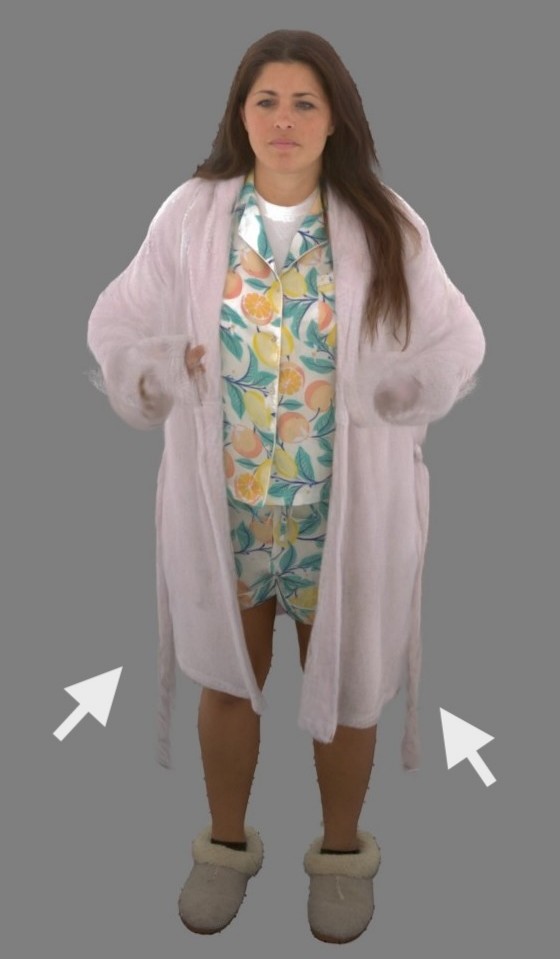} &
      \includegraphics[width=\initStateAblationWidth]{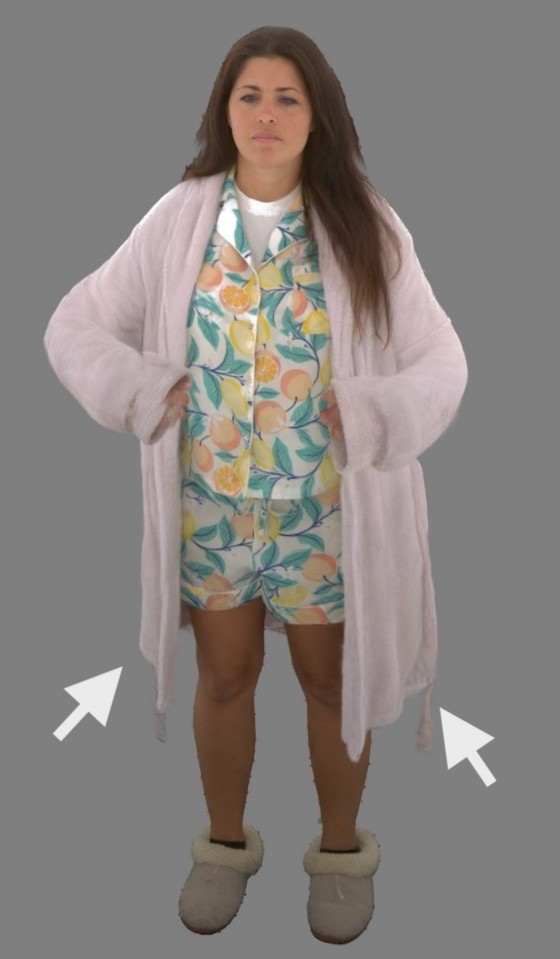} &
      \includegraphics[width=\initStateAblationWidth]{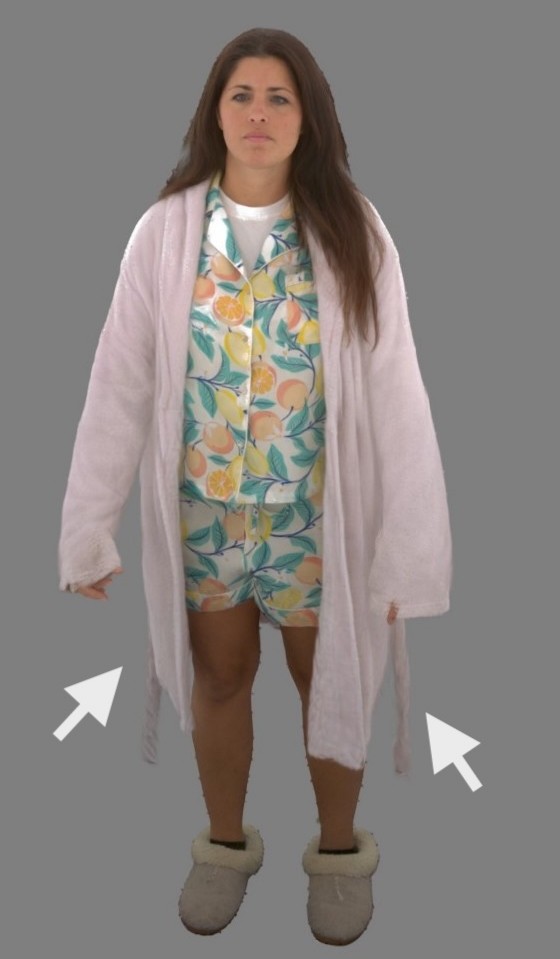} &
      \includegraphics[width=\initStateAblationWidth]{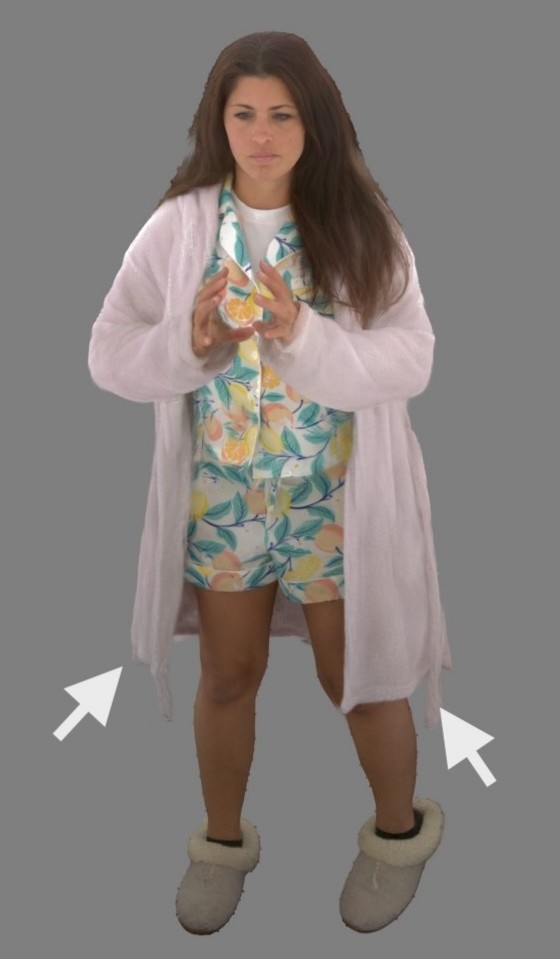} \\
      Init Latent 3 &
      \includegraphics[width=\initStateAblationWidth]{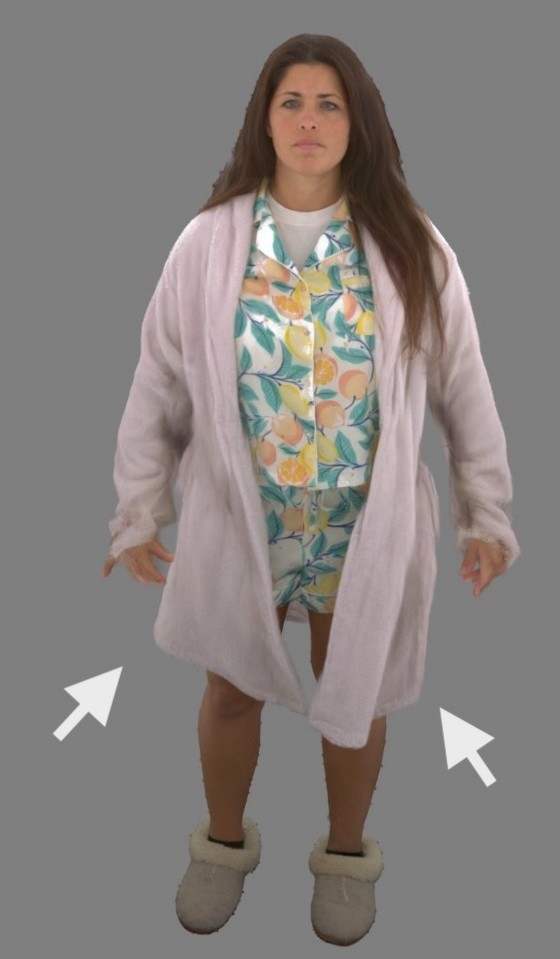} &
      \includegraphics[width=\initStateAblationWidth]{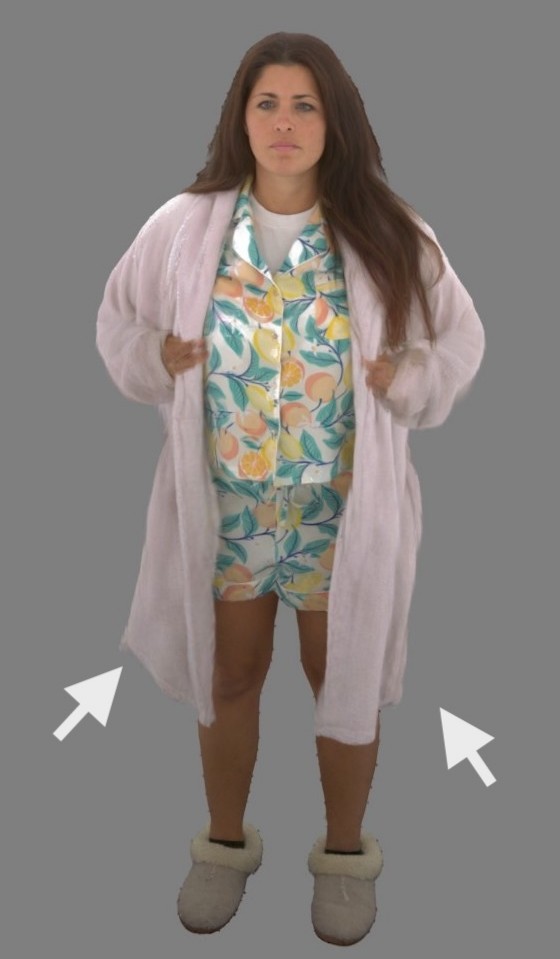} &
      \includegraphics[width=\initStateAblationWidth]{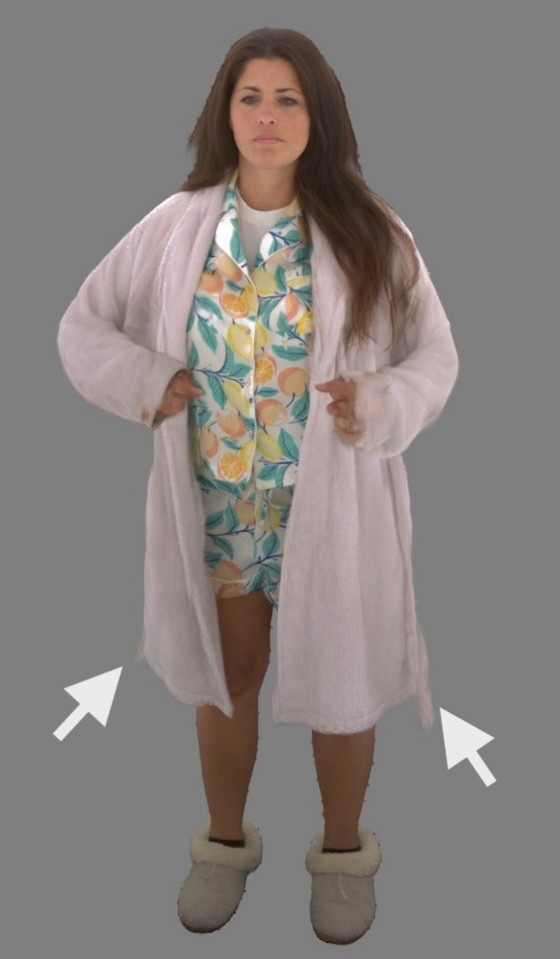} &
      \includegraphics[width=\initStateAblationWidth]{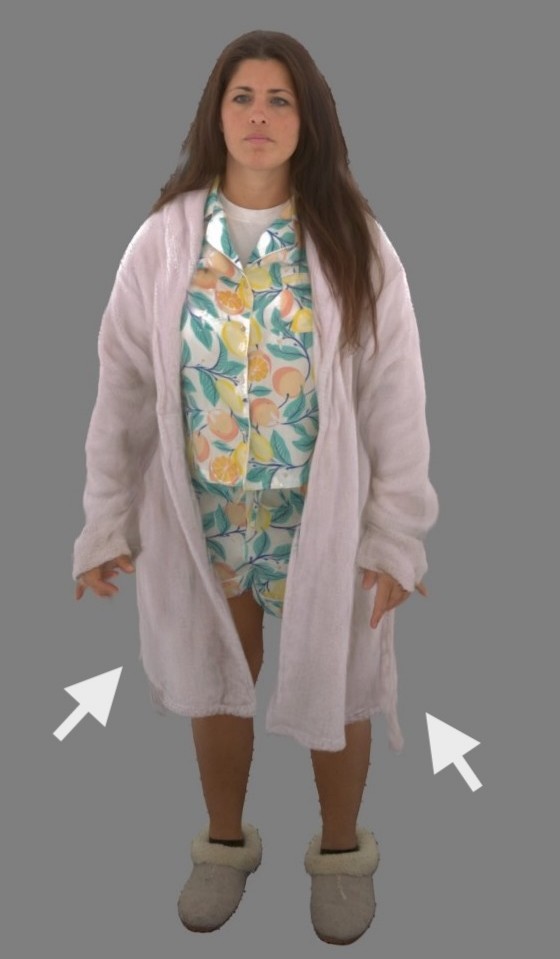} &
      \includegraphics[width=\initStateAblationWidth]{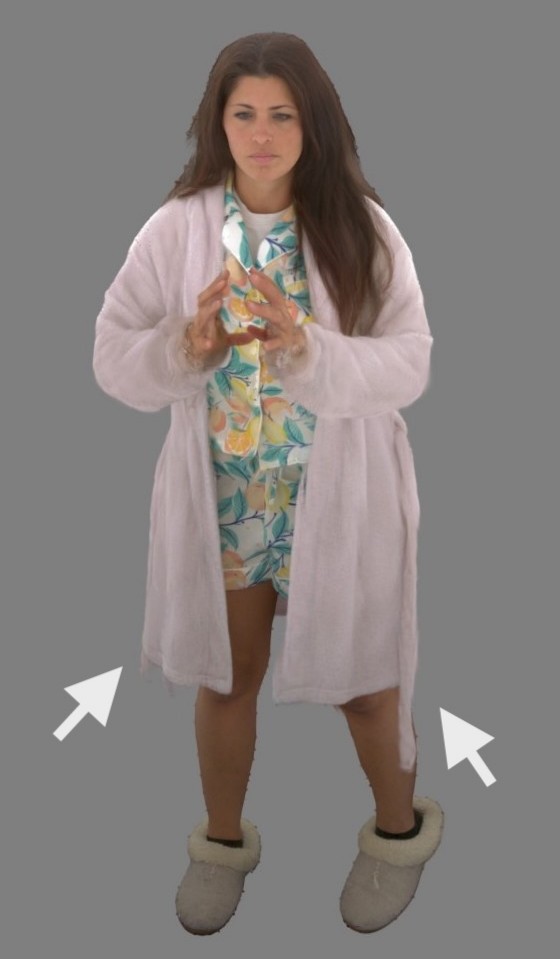} \\
    \end{tabular}} &
    {\setlength{\tabcolsep}{0pt}
    \begin{tabular}{@{}c*{5}{c}@{}}
      Init Latent 2 &
      \includegraphics[width=\initStateAblationWidth]{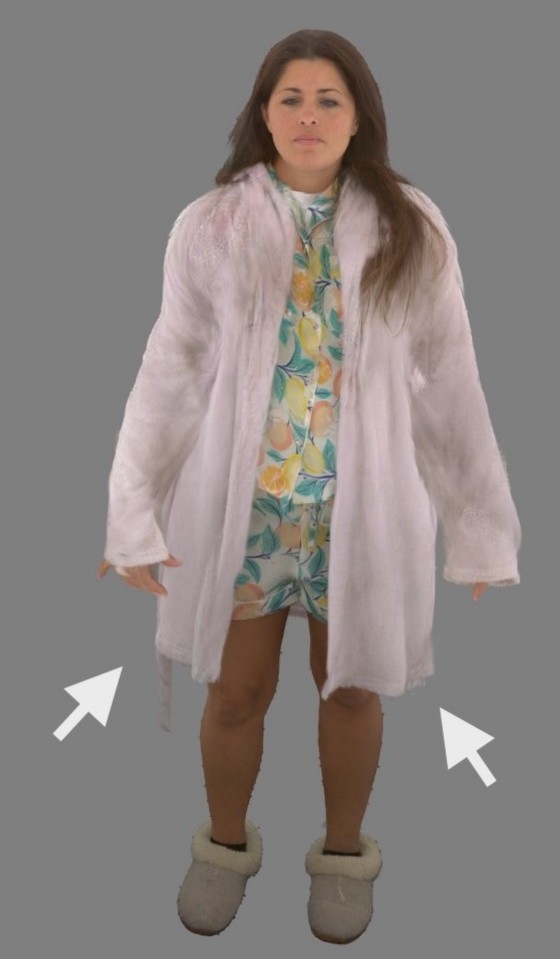} &
      \includegraphics[width=\initStateAblationWidth]{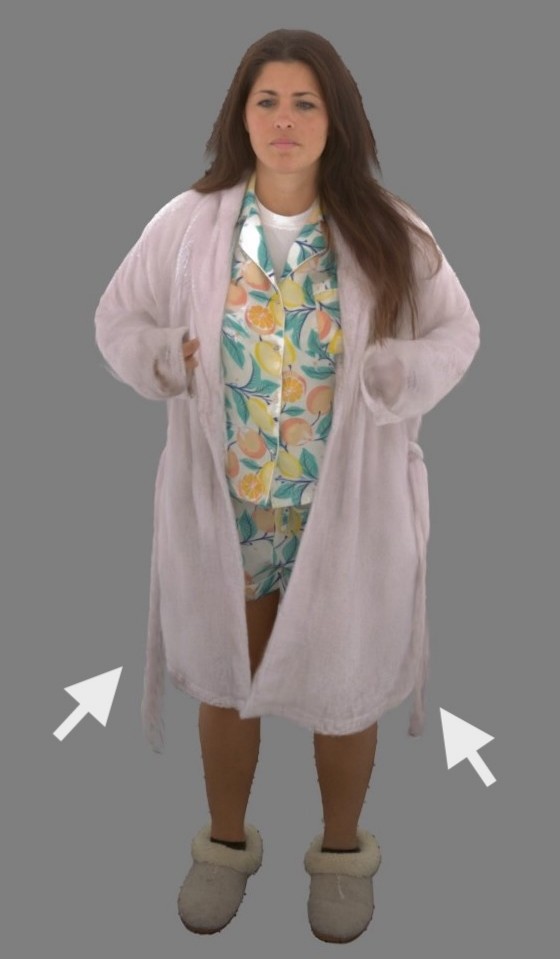} &
      \includegraphics[width=\initStateAblationWidth]{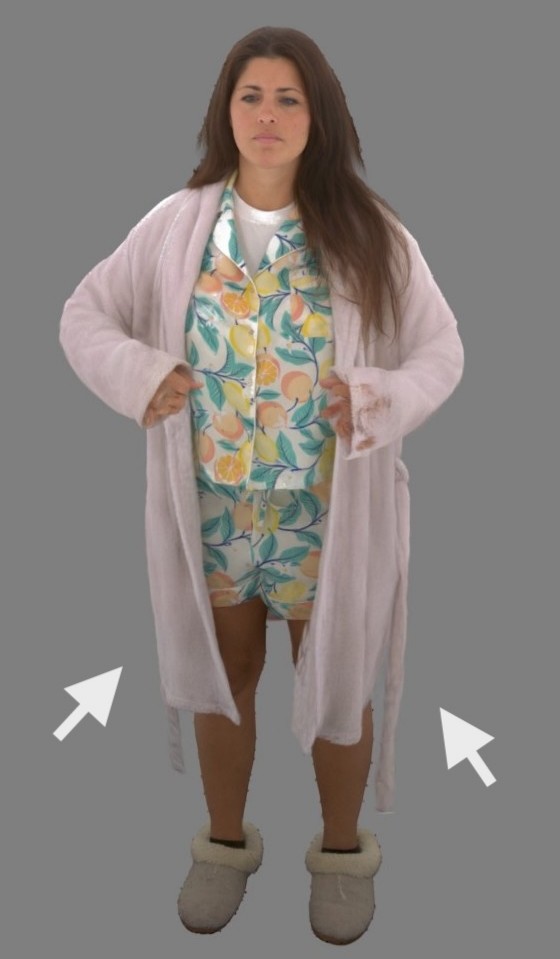} &
      \includegraphics[width=\initStateAblationWidth]{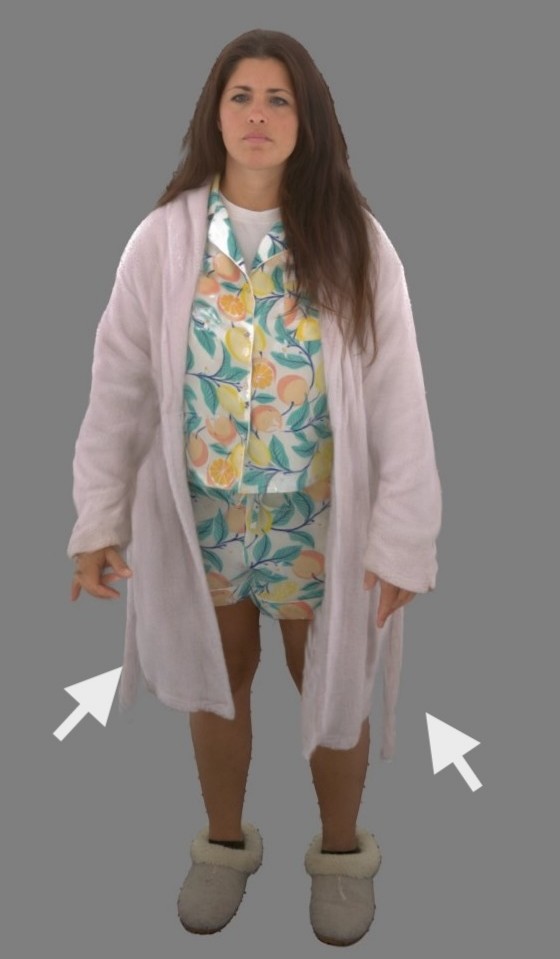} &
      \includegraphics[width=\initStateAblationWidth]{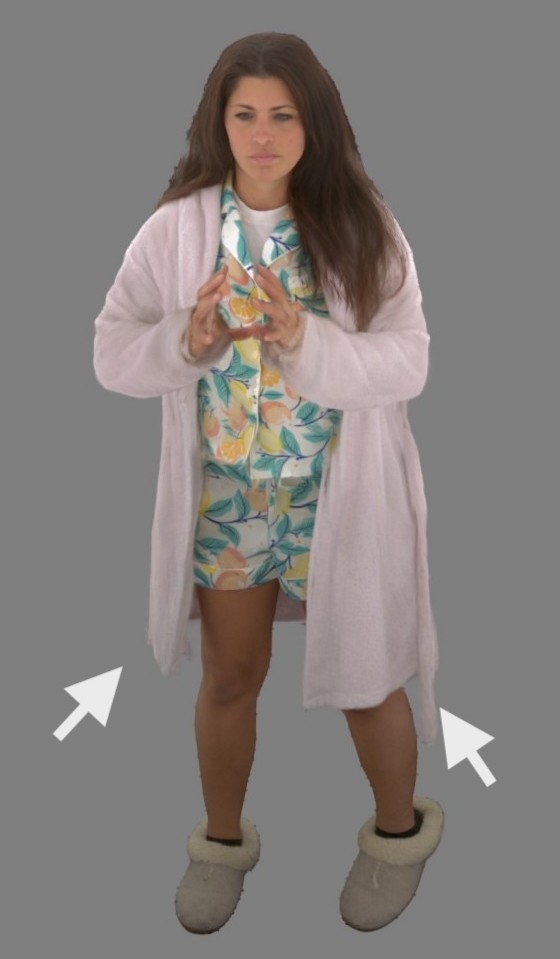} \\
      Init Latent 4 &
      \includegraphics[width=\initStateAblationWidth]{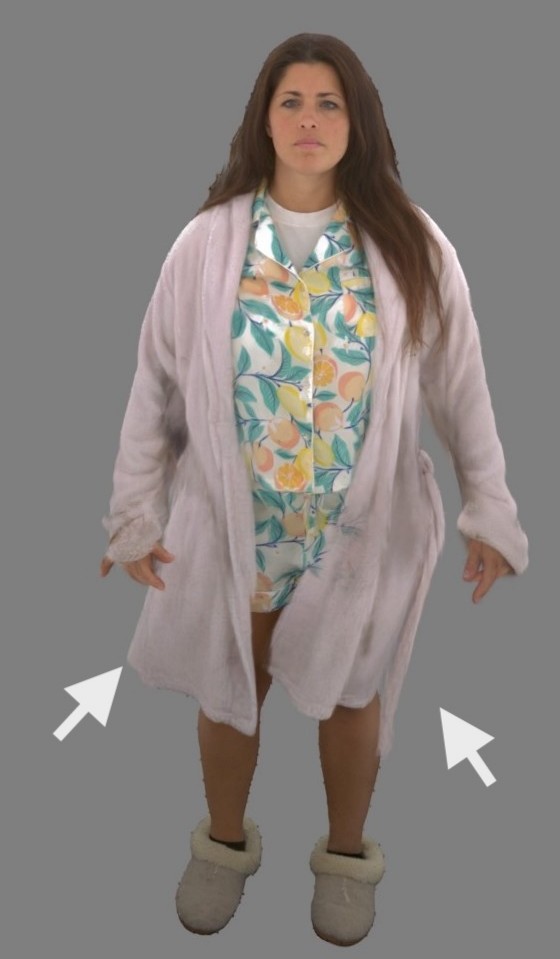} &
      \includegraphics[width=\initStateAblationWidth]{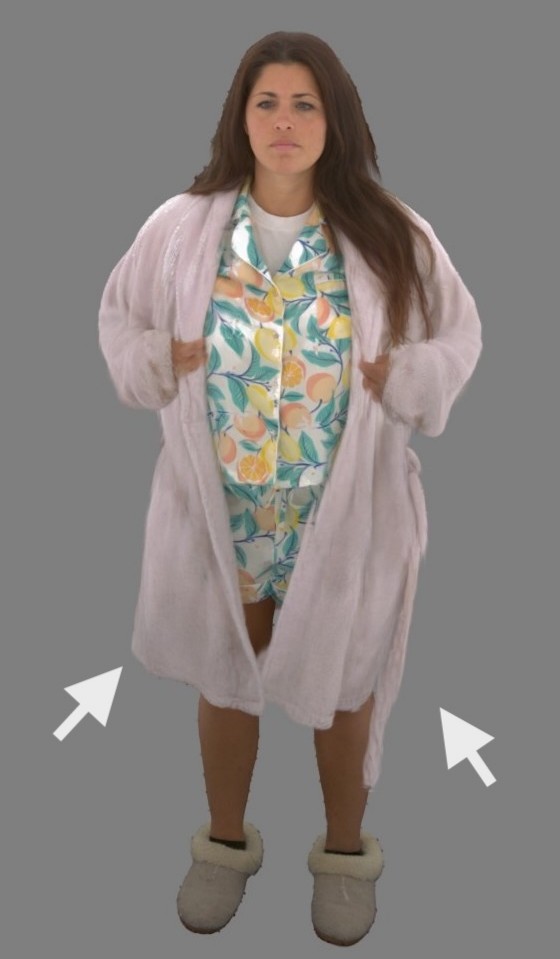} &
      \includegraphics[width=\initStateAblationWidth]{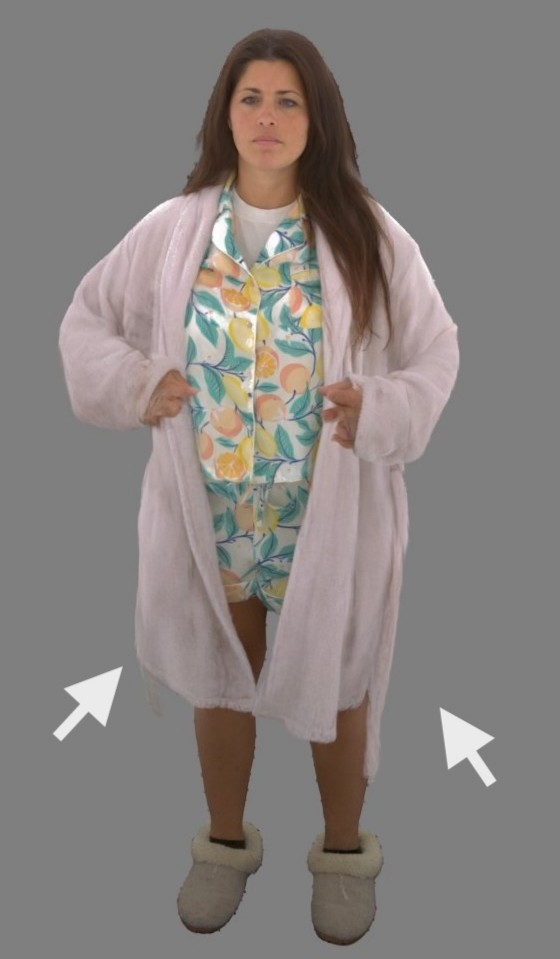} &
      \includegraphics[width=\initStateAblationWidth]{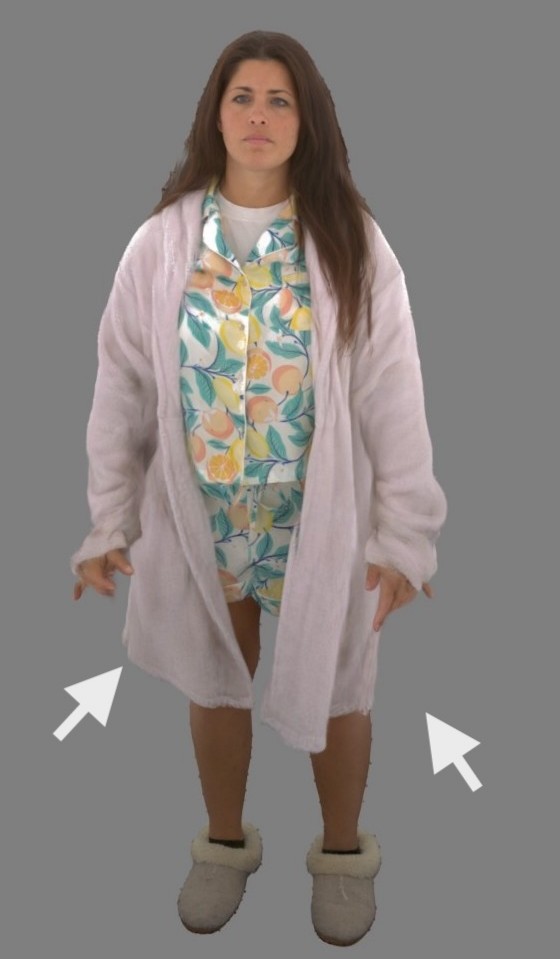} &
      \includegraphics[width=\initStateAblationWidth]{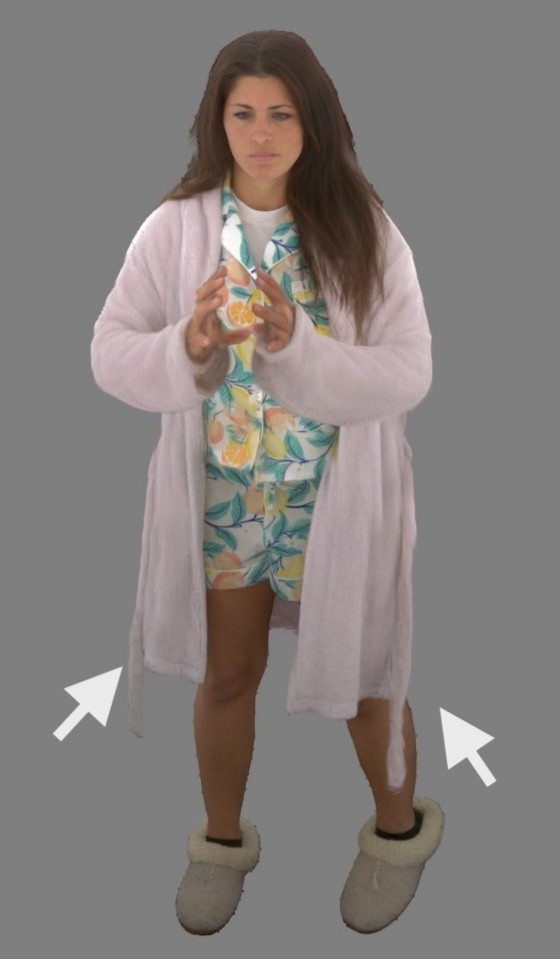} \\
    \end{tabular}}
  \end{tabular}
  \caption{\textbf{Effect of initial conditions.} Varying the initial residual latent yields distinct motion trajectories for the same driving pose sequence, producing diverse yet plausible garment dynamics.}
  \label{fig:ablation_init_state}
\end{figure*}

\section{Experiment}
\label{sec:experiment}

\boldparagraph{Baselines}
We compare our method against recent learning-based human avatar approaches that model dynamic deformations not explained by pose alone, such as those from loose clothing. We select methods with open-source implementations: ToMiE~\cite{zhan2024tomiemodulargrowthenhanced}, Seq-Avatar~\cite{xu2025seqavatar}, and $R^3$-Avatar~\cite{Zhan2025R3AvatarRA}.

\boldparagraph{Dataset} We captured nine sequences in a multi-camera light-stage setup, covering loose garments such as long dresses, knee-length dresses, and loose upper garments.
Subjects perform everyday motions: reaching, bending, stretching, casual stepping, and gesturing in conversation. The garments swing and fold across the sequence.
Each sequence contains approximately $4{,}000$ frames at $30\,\mathrm{fps}$. For each subject, we split the sequence roughly in half into two disjoint portions: one for training, and a held-out portion with more varied gestures used for novel-pose evaluation.

The capture setup comprises 512 cameras. We train on images from 50 randomly selected cameras surrounding the subject and hold out 10 additional cameras for evaluation. All videos are captured at 5328$\times$4608 and downsampled to one-quarter resolution for all experiments.

\begin{table}[t]
\centering
\caption{\textbf{Quantitative comparison on novel view synthesis} (training poses, held-out cameras), measured by average PSNR ($\uparrow$), SSIM ($\uparrow$), and LPIPS ($\downarrow$).}
\label{Tab::Res-novel-view}
\begin{tabular}{c|ccc}
\toprule
Methods & PSNR ($\uparrow$) & SSIM ($\uparrow$) & LPIPS ($\downarrow$) \\
\midrule
ToMiE~\cite{zhan2024tomiemodulargrowthenhanced} & 29.29 & 0.930 & 0.083 \\
Seq-Avatar~\cite{xu2025seqavatar} & 24.17 & 0.848 & 0.100  \\
$R^3$-Avatar~\cite{Zhan2025R3AvatarRA} & 26.33 & 0.916 & 0.092 \\
\midrule
Ours & \textbf{33.18} & \textbf{0.980} & \textbf{0.045} \\
\bottomrule
\end{tabular}%
\end{table}

\begin{table}[t]
\centering
\caption{\textbf{Quantitative comparison on self-reenactment} under novel poses in terms of average FID ($\downarrow$), and KID ($\downarrow$).}
\label{Tab::Res-Self-reenact}
\begin{tabular}{c|cc}
\toprule
Methods & FID ($\downarrow$) & KID ($\downarrow$) \\
\midrule
ToMiE~\cite{zhan2024tomiemodulargrowthenhanced} & 120.51 & 0.121 \\
Seq-Avatar~\cite{xu2025seqavatar} & 153.32 & 0.176 \\
$R^3$-Avatar~\cite{Zhan2025R3AvatarRA} & 119.01 & 0.115 \\
\midrule
Ours & \textbf{36.56} & \textbf{0.021} \\
\bottomrule
\end{tabular}%
\end{table}

\subsection{Results and Discussion}
\boldparagraph{Quantitative Comparison}
On novel-view synthesis over the training sequences, our method outperforms all baselines on every metric (Table~\ref{Tab::Res-novel-view}). We report PSNR and SSIM for pixel-level fidelity and LPIPS~\cite{Zhang2018TheUE} for perceptual similarity. The gains are consistent across the three metrics, indicating more accurate and perceptually faithful renderings on these challenging sequences.

Novel-pose self-reenactment (Table~\ref{Tab::Res-Self-reenact}) requires a different evaluation protocol. Under history-dependent dynamics, multiple plausible garment configurations can correspond to the same pose, and we do not assume access to the ground-truth garment configuration at test time. Per-frame alignment metrics such as PSNR/SSIM/LPIPS are therefore uninformative here, since the rendered garment configuration may legitimately differ from the captured one. We instead evaluate realism at the distribution level using FID~\cite{parmar2021cleanfid} and KID~\cite{Binkowski2018DemystifyingMG} between rendered novel-pose animations and the corresponding captured sequences, where lower scores indicate a closer match to the distribution of real images. Our method achieves substantially lower FID and KID than every baseline, indicating more realistic appearance under unseen poses.

\boldparagraph{Qualitative Comparison}
The metric gains carry over to visual quality. In the novel-view comparisons of Figure~\ref{fig:novel_view}, our renderings remain sharp where the baselines oversmooth regions undergoing large deformation: folds in the yellow dress (row 1), embroidery around the upper chest (row 2), and the dotted and grid patterns on the garments (row 3) are all preserved by our method but blurred or washed out by the others. This pattern holds across subjects and garment types, with additional video comparisons in the supplementary material.

The same advantage carries to unseen poses. On the held-out sequences (Figure~\ref{fig:novel_pose}), our method produces plausible garment motion while preserving fine patterns and realistic wrinkles, whereas the baselines lose high-frequency detail in high-motion regions and introduce blur or artifacts near garment boundaries. The temporal quality of the resulting clothing motion is most apparent in the accompanying videos.
	
Figure~\ref{fig:cross_grid} illustrates the behavior of our dynamic motion model by transferring a single source pose sequence to multiple target identities. Our latent dynamics model produce realistic clothing motion on each target, including swinging and folding of dresses and the white robe, while identity-specific garment details such as embroidery and patterns remain intact. For each target subject, the same unseen driving pose sequence produces garment-appropriate motion, indicating that the subject-specific dynamics model handles poses beyond its training inputs.

\boldparagraph{Perceptual User Study}
To test whether viewers perceive our clothing dynamics as more realistic, we run a pairwise two-alternative forced choice (2AFC) study. Participants view pairs of novel-pose animation videos and select the one with more realistic clothing motion. Each pair shows the same pose sequence and subject with randomized left/right placement. We recruit 28 participants and collect responses over 36 video pairs spanning all subjects and garment types.

Participants preferred our method over ToMiE~\cite{zhan2024tomiemodulargrowthenhanced} (99.2\%), Seq-Avatar~\cite{xu2025seqavatar} (98.4\%), and $R^3$-Avatar~\cite{Zhan2025R3AvatarRA} (98.8\%), all significant at $p < 0.001$ under a binomial test. The preference held across both more active (97.5\%) and calmer (99.5\%) portions of the test set, indicating that the gain is not tied to a particular motion regime.

To isolate the motion contribution from rendering quality, we run a second comparison against an ablated variant that replaces the dynamics model with the fixed reference latent $\mathbf{z}_{\text{ref}}$, keeping the decoder and renderer identical. Even with rendering held constant, participants preferred the full dynamics model in 69.3\% of trials ($p < 0.001$), confirming that the perceptual gain comes from the learned latent dynamics, not rendering alone.

\subsection{Controlling Motion Dynamics}

The force decomposition of our latent dynamics model exposes controls over the resulting motion. We can scale the forces $\mathbf{F}_\text{spring}$, $\mathbf{F}_\text{damping}$, and $\mathbf{F}_\text{pose}$ from Section~\ref{sec:method_dynamics}, and we can vary the initial residual latent. Static images are limiting for motion analysis, so we refer readers to the supplementary video for clearer differences.

\begin{figure}[ht]
  \newcommand{\springAblationWidth}{0.17\columnwidth}
  \centering
  \footnotesize
  \setlength{\tabcolsep}{0pt}
  \begin{tabular}{@{}c*{5}{c}@{}}
    $\mathbf{F}_\text{spring} \times 0.5$ &
    \includegraphics[width=\springAblationWidth]{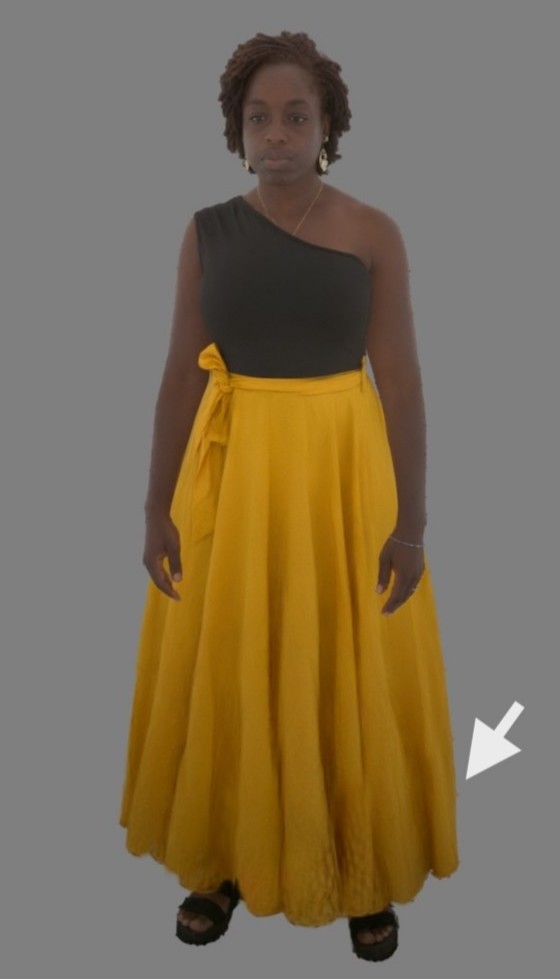} &
    \includegraphics[width=\springAblationWidth]{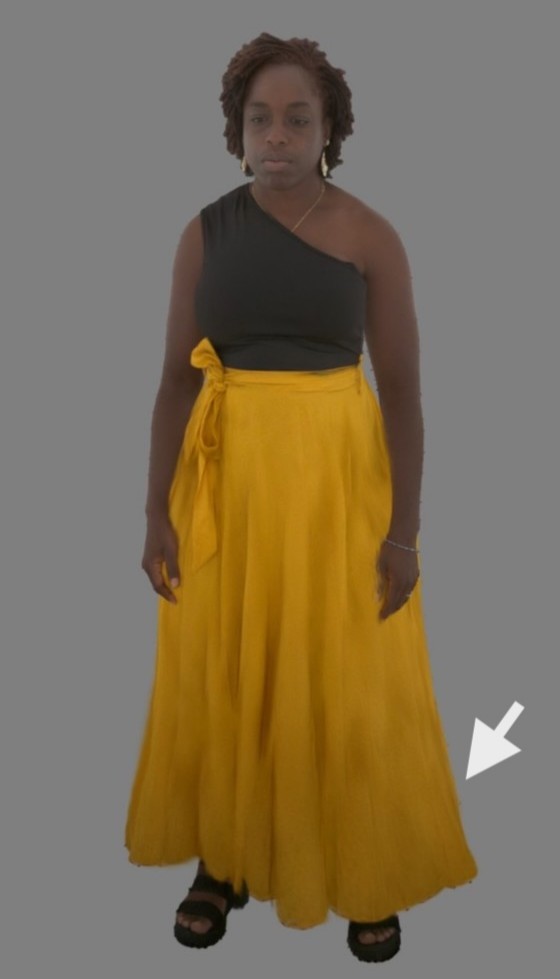} &
    \includegraphics[width=\springAblationWidth]{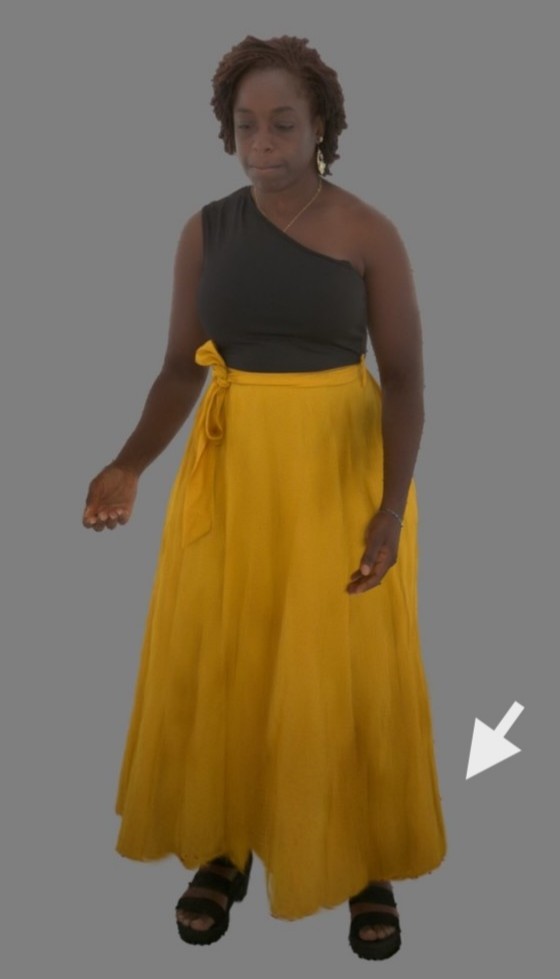} &
    \includegraphics[width=\springAblationWidth]{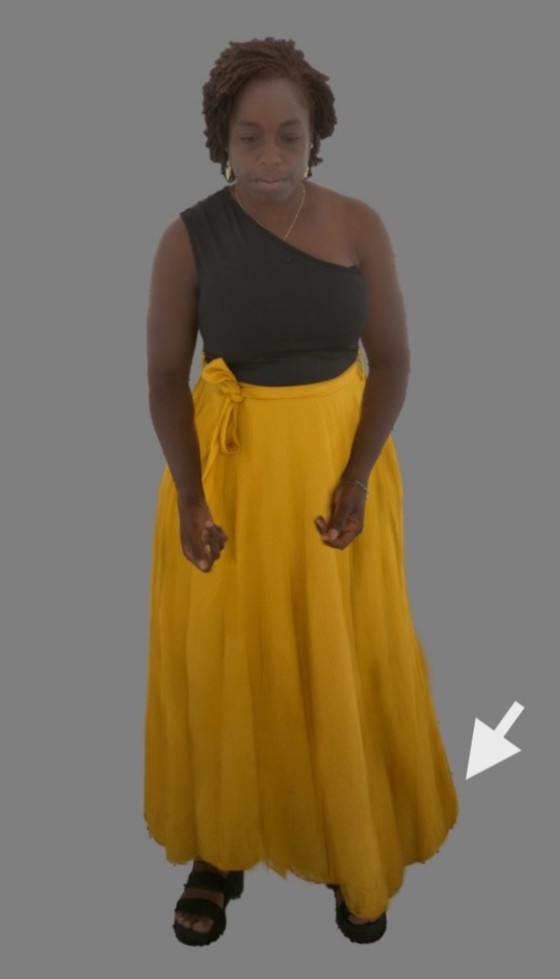} &
    \includegraphics[width=\springAblationWidth]{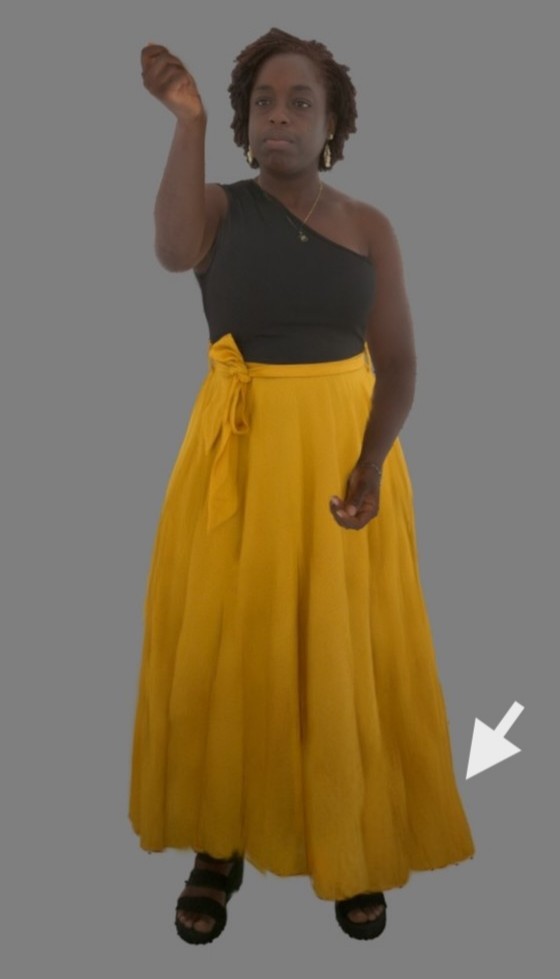} \\
    $\mathbf{F}_\text{spring} \times 1$ &
    \includegraphics[width=\springAblationWidth]{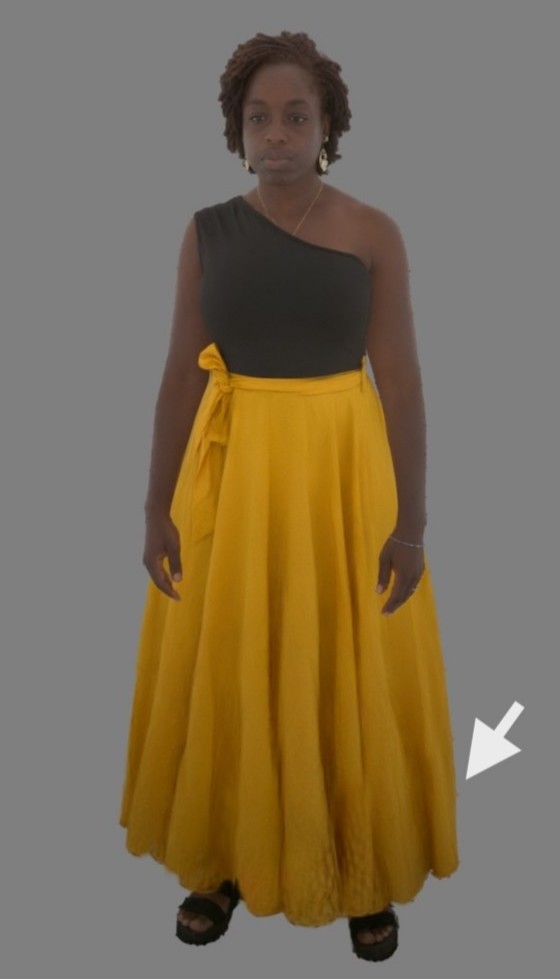} &
    \includegraphics[width=\springAblationWidth]{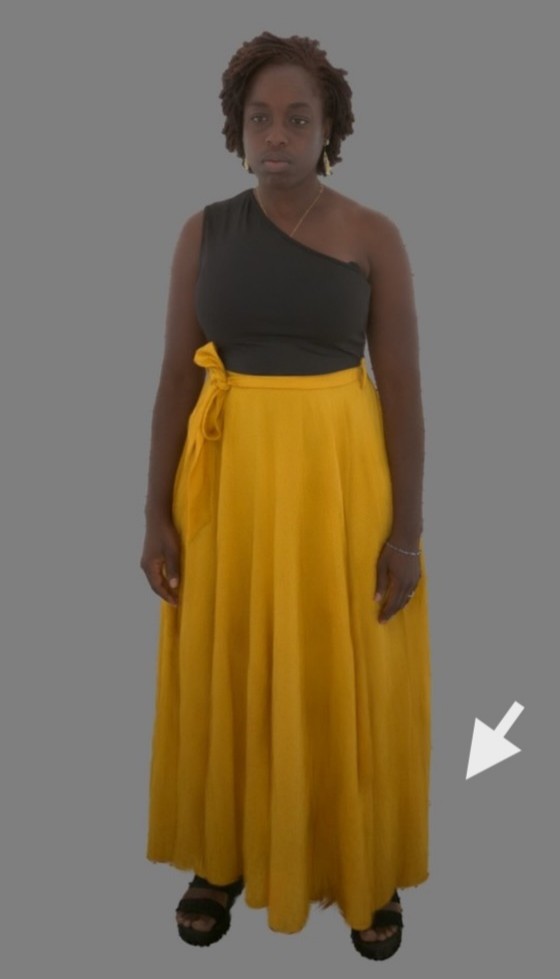} &
    \includegraphics[width=\springAblationWidth]{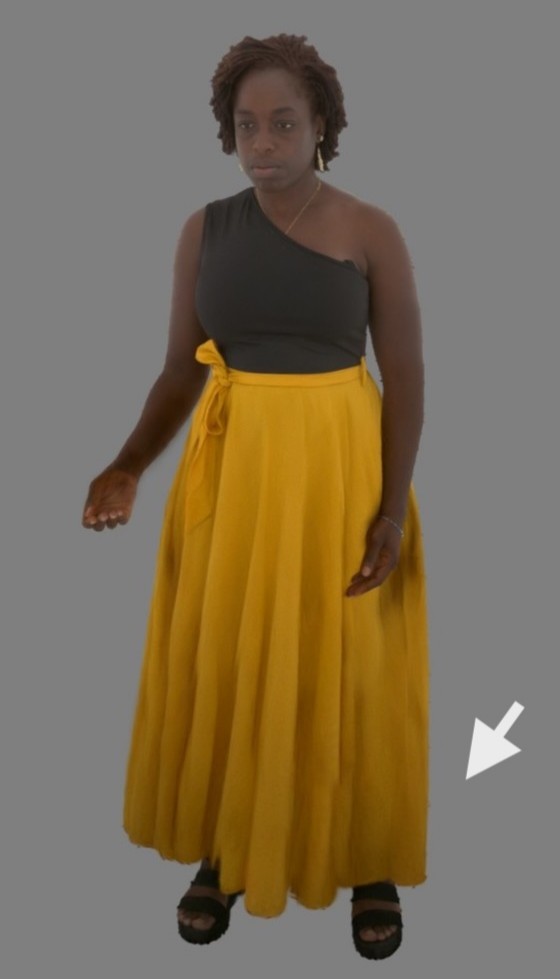} &
    \includegraphics[width=\springAblationWidth]{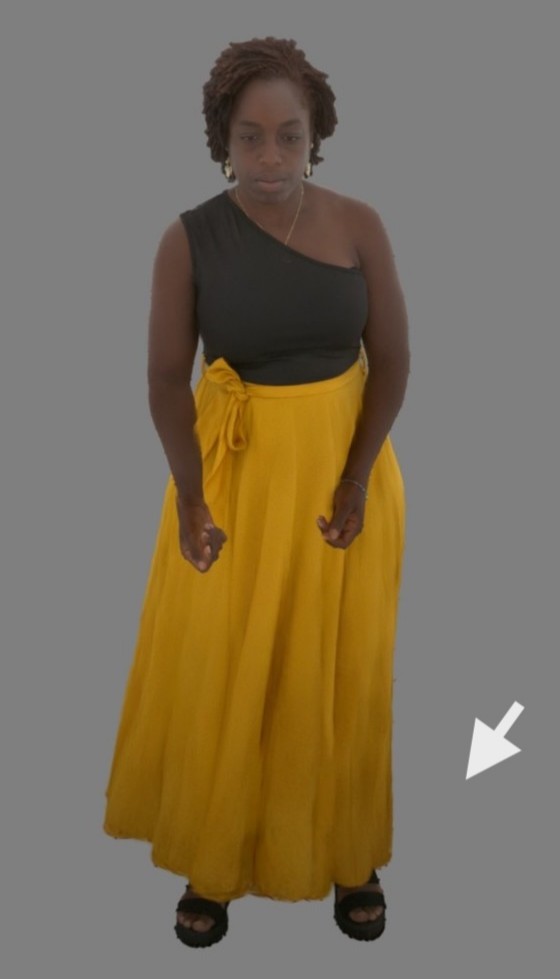} &
    \includegraphics[width=\springAblationWidth]{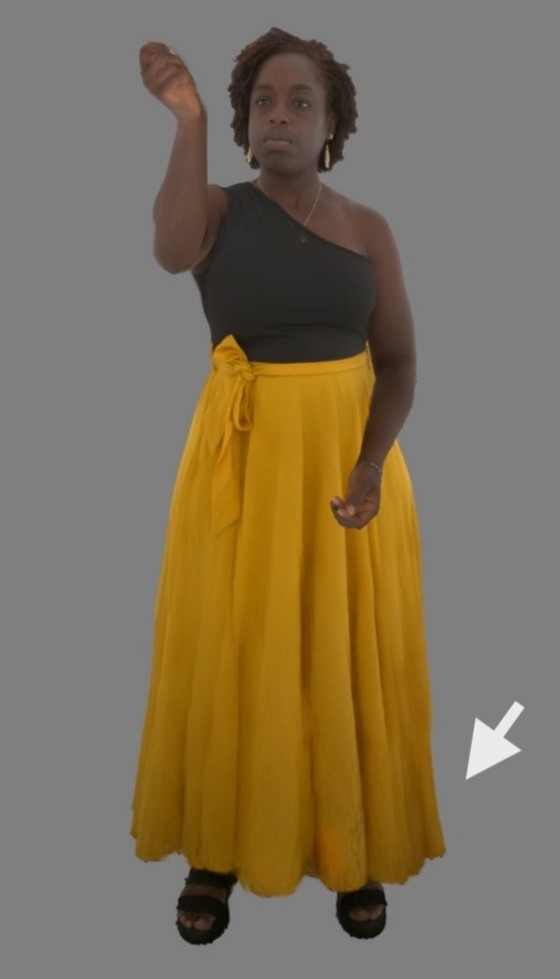} \\
    $\mathbf{F}_\text{spring} \times 2$ &
    \includegraphics[width=\springAblationWidth]{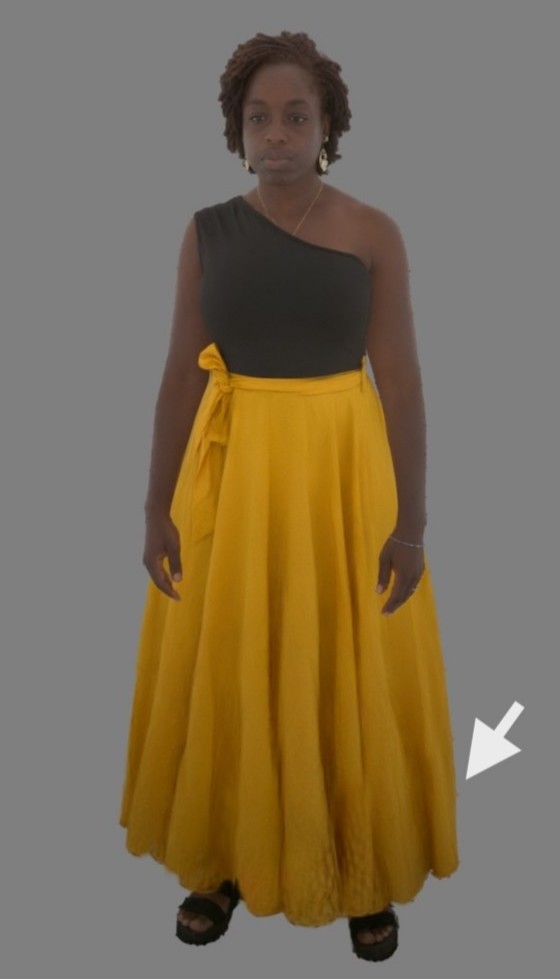} &
    \includegraphics[width=\springAblationWidth]{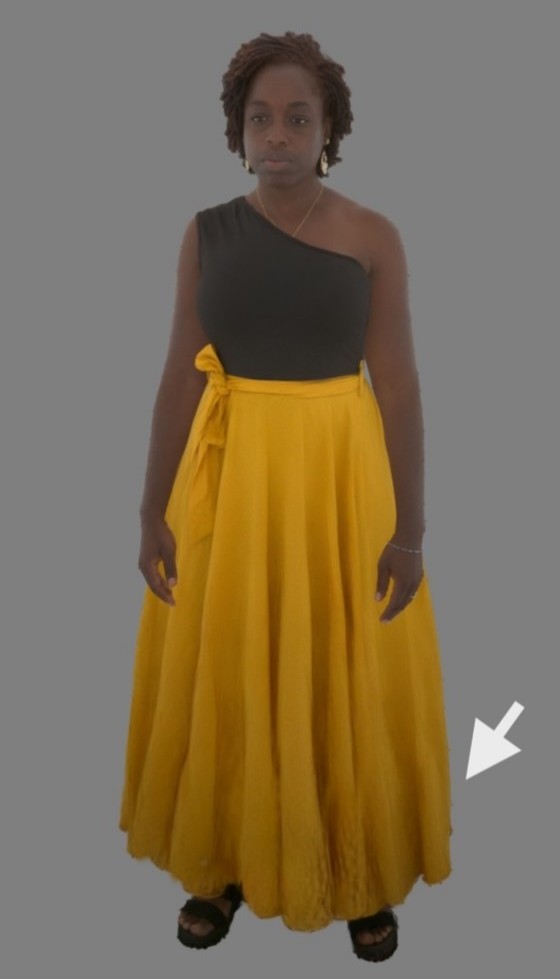} &
    \includegraphics[width=\springAblationWidth]{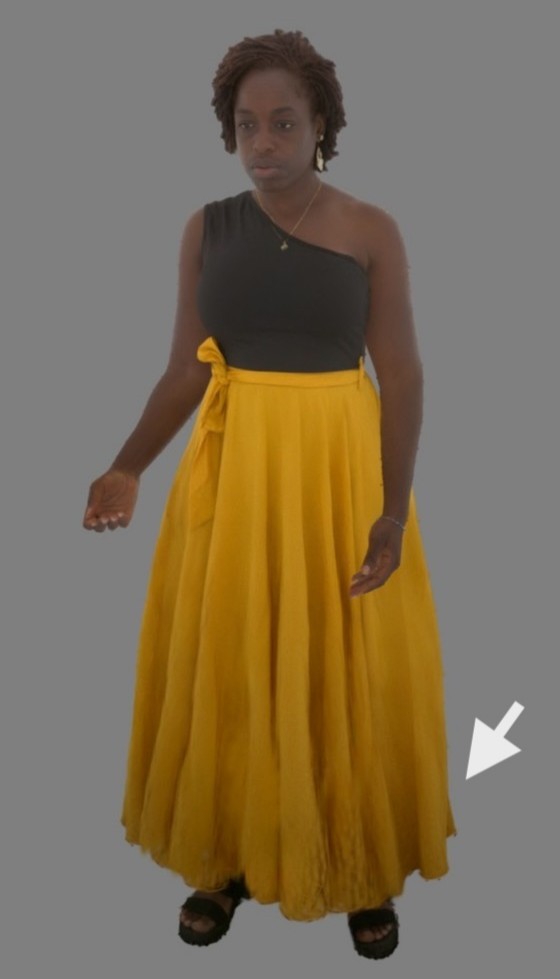} &
    \includegraphics[width=\springAblationWidth]{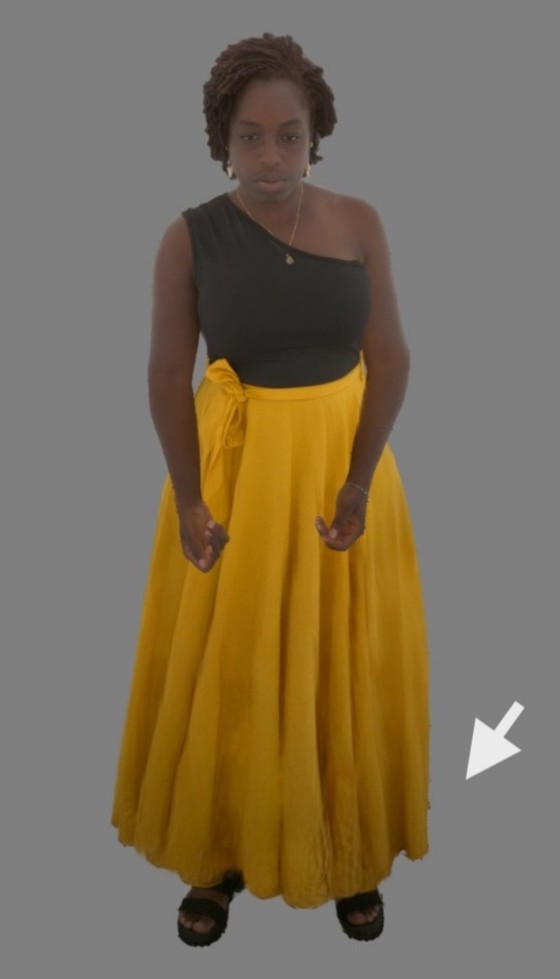} &
    \includegraphics[width=\springAblationWidth]{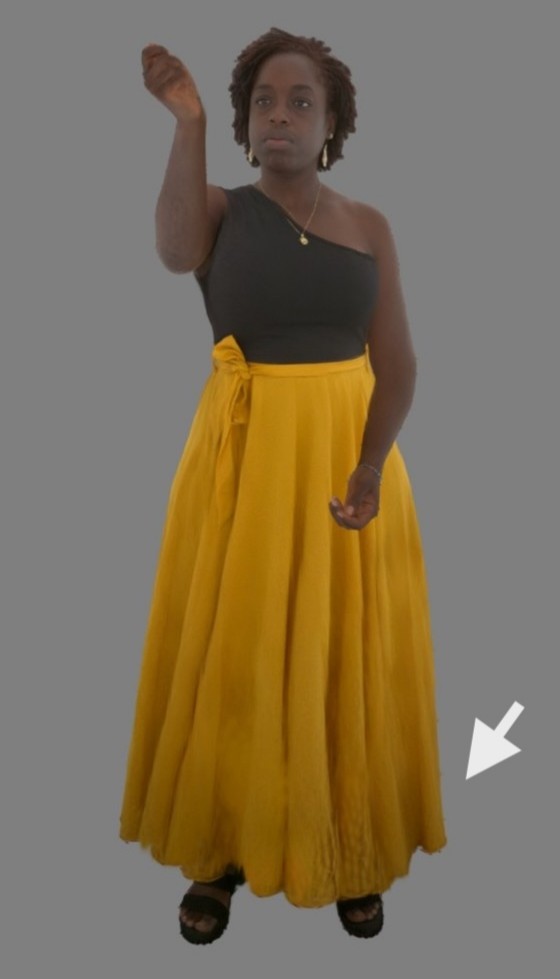} \\
  \end{tabular}
  \caption{\textbf{Effect of spring force on clothing motion dynamics.} Increasing the spring force pulls the cloth more strongly toward its rest state, resulting in stiffer motion, while decreasing it allows larger drapes and smoother swings, as visible in the dress folds.}
  \label{fig:ablation_spring}
\end{figure}

\boldparagraph{Scaling Spring Force} The spring force pulls the cloth back toward its rest state. Increasing it produces stiffer motion that resists stretching and bending; decreasing it allows larger drapes and smoother swings, visible as deeper folds in the dress in Figure~\ref{fig:ablation_spring}.

\begin{figure}[ht]
  \newcommand{\dampingAblationWidth}{0.16\columnwidth}
  \centering
  \footnotesize
  \setlength{\tabcolsep}{0pt}
  \begin{tabular}{@{}c*{5}{c}@{}}
    $\mathbf{F}_\text{damping} \times 0.5$ &
    \includegraphics[width=\dampingAblationWidth]{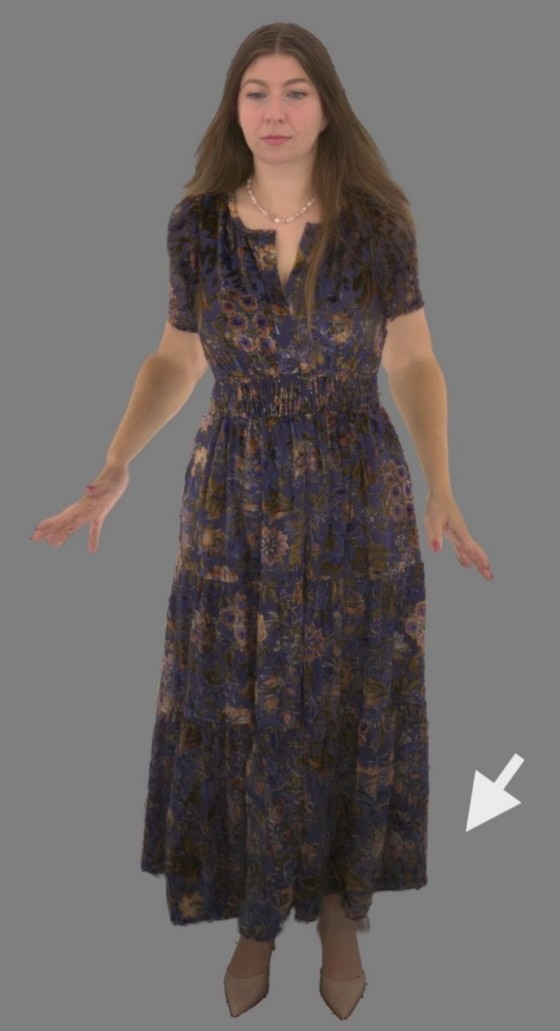} &
    \includegraphics[width=\dampingAblationWidth]{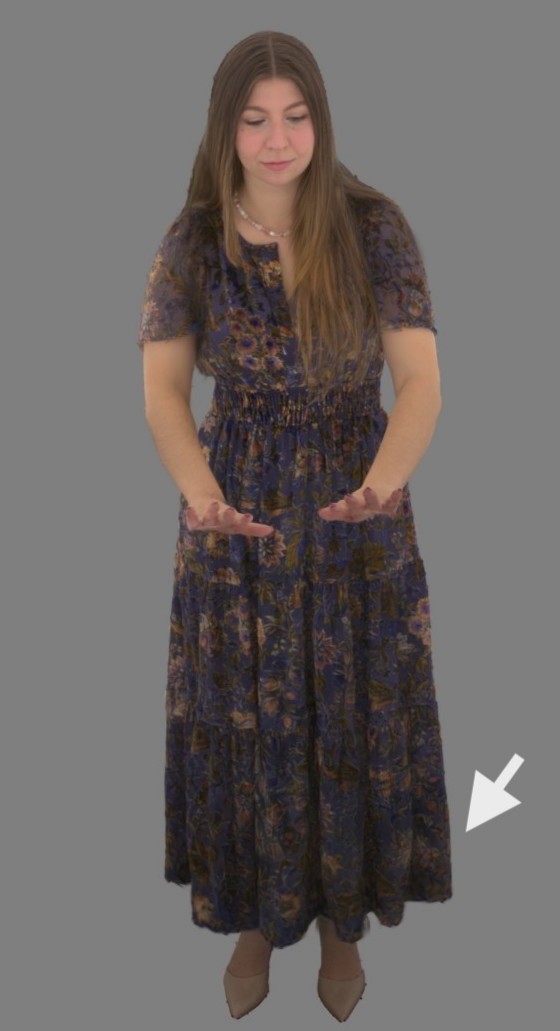} &
    \includegraphics[width=\dampingAblationWidth]{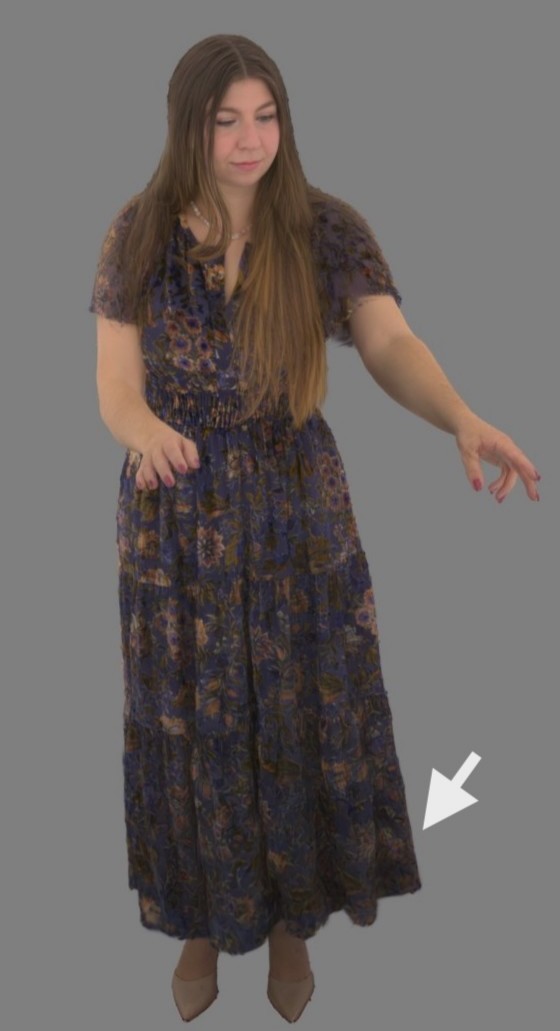} &
    \includegraphics[width=\dampingAblationWidth]{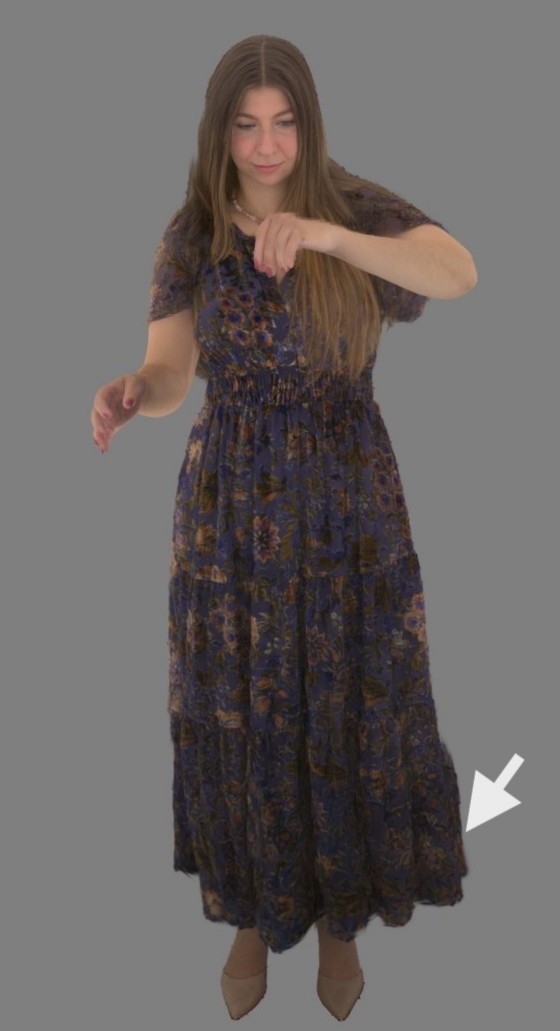} &
    \includegraphics[width=\dampingAblationWidth]{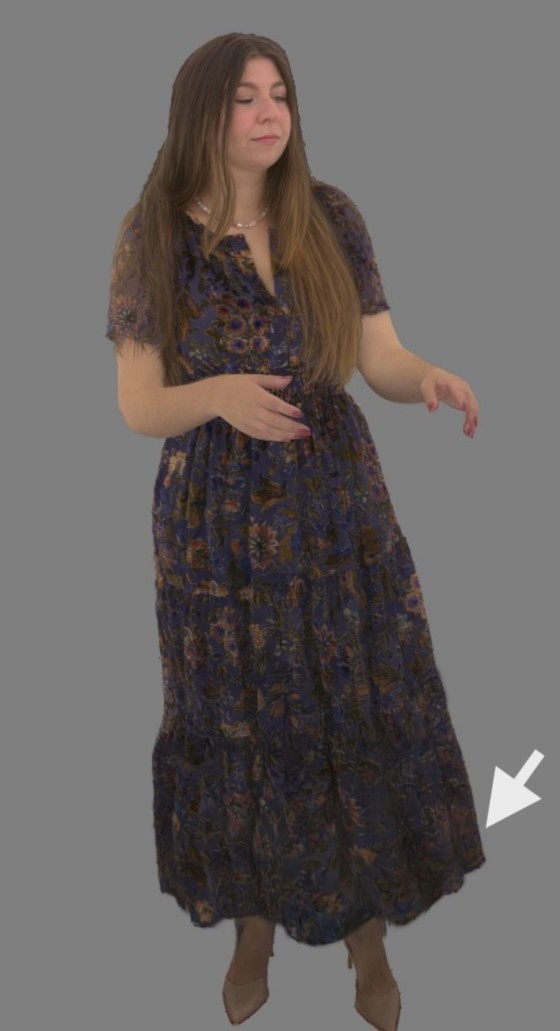} \\
    $\mathbf{F}_\text{damping} \times 1$ &
    \includegraphics[width=\dampingAblationWidth]{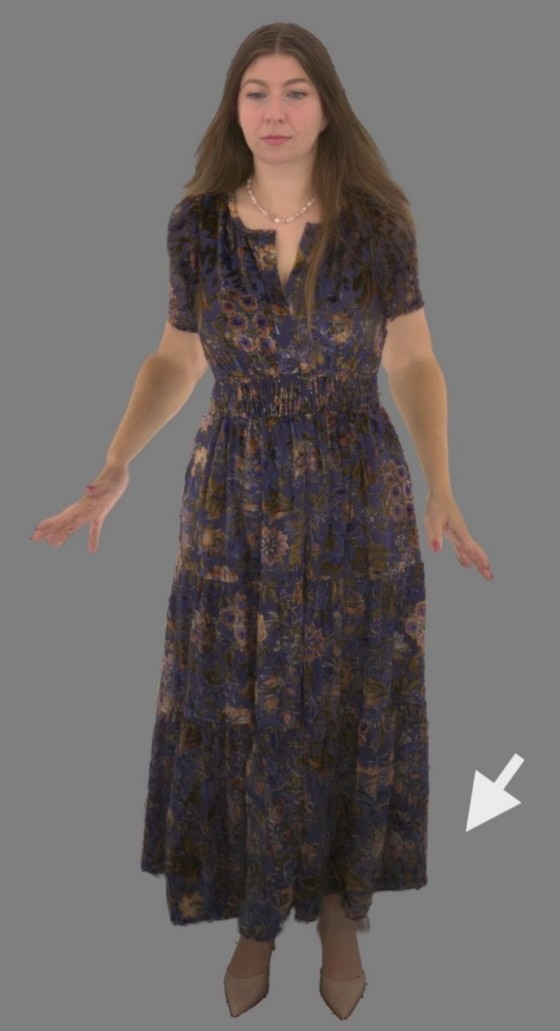} &
    \includegraphics[width=\dampingAblationWidth]{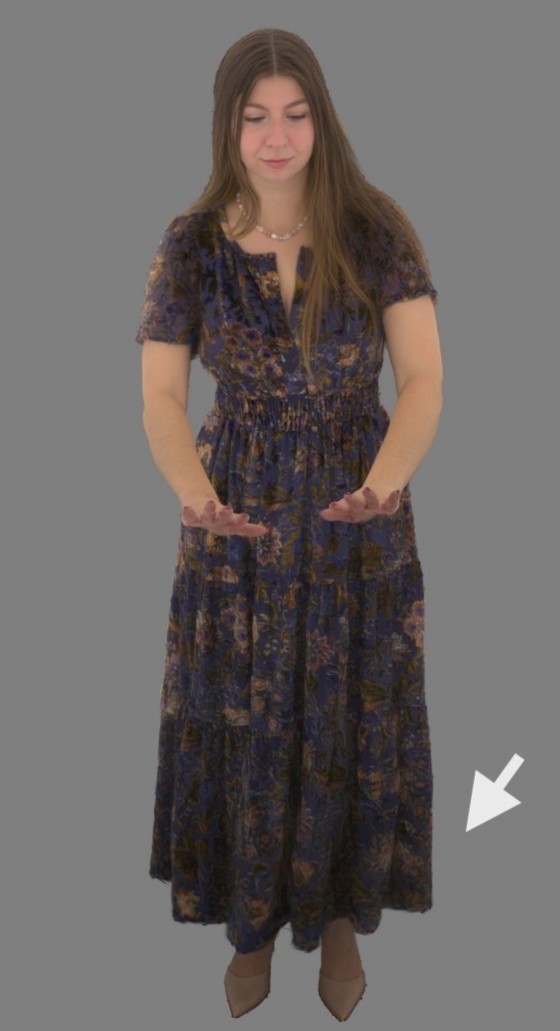} &
    \includegraphics[width=\dampingAblationWidth]{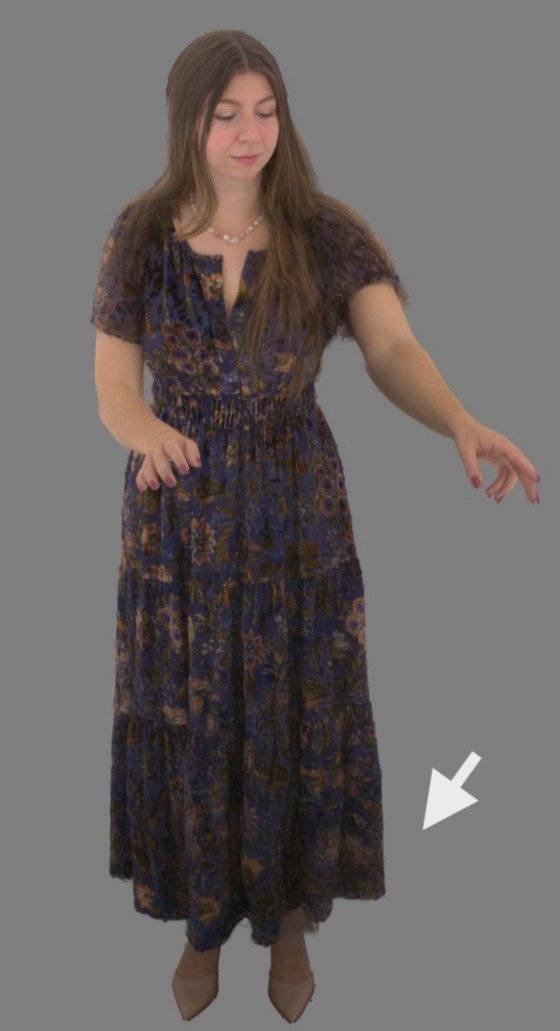} &
    \includegraphics[width=\dampingAblationWidth]{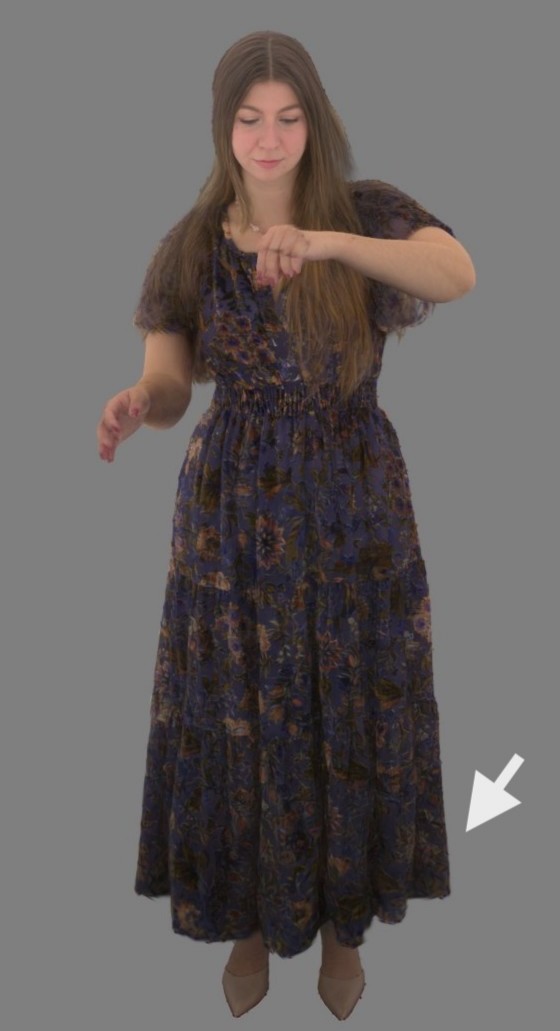} &
    \includegraphics[width=\dampingAblationWidth]{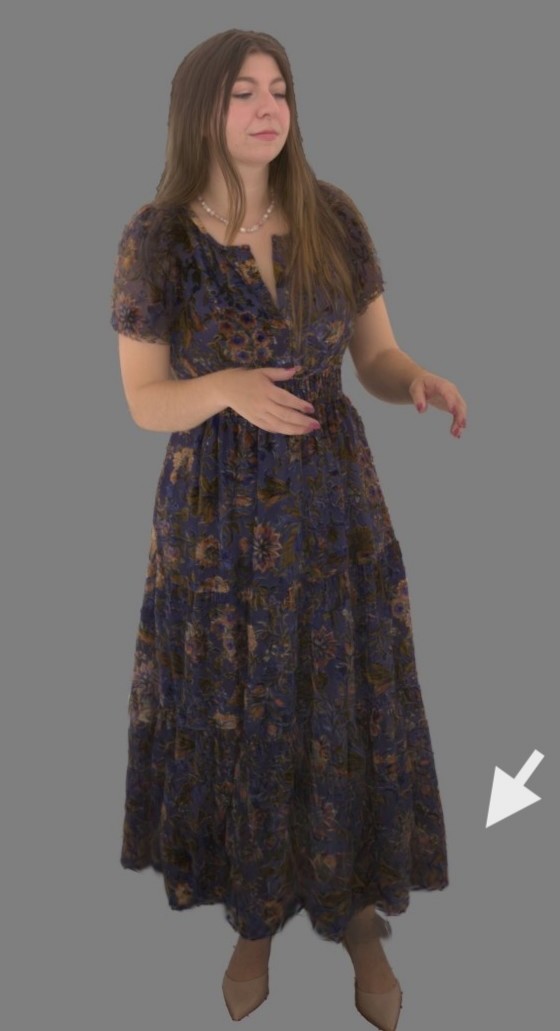} \\
    $\mathbf{F}_\text{damping} \times 2$ &
    \includegraphics[width=\dampingAblationWidth]{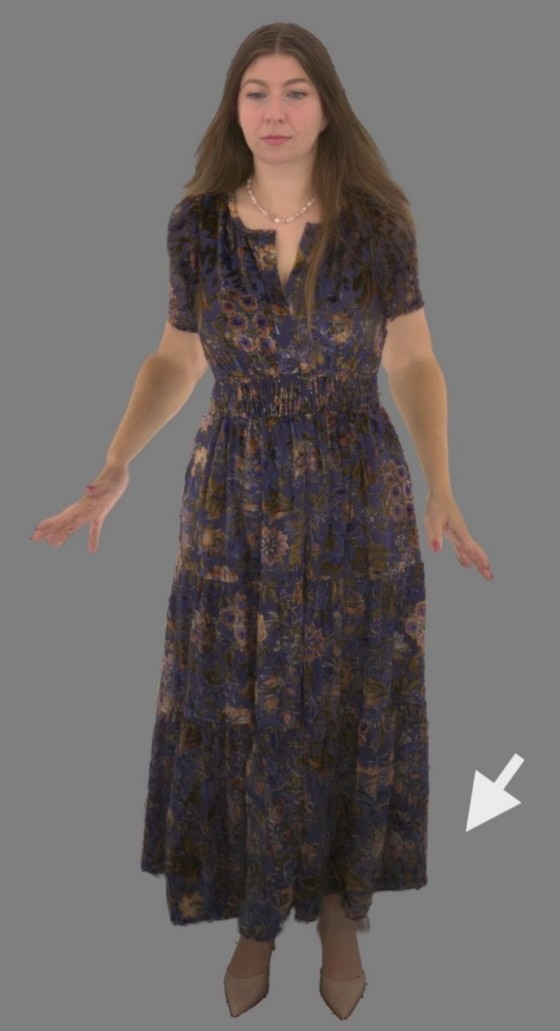} &
    \includegraphics[width=\dampingAblationWidth]{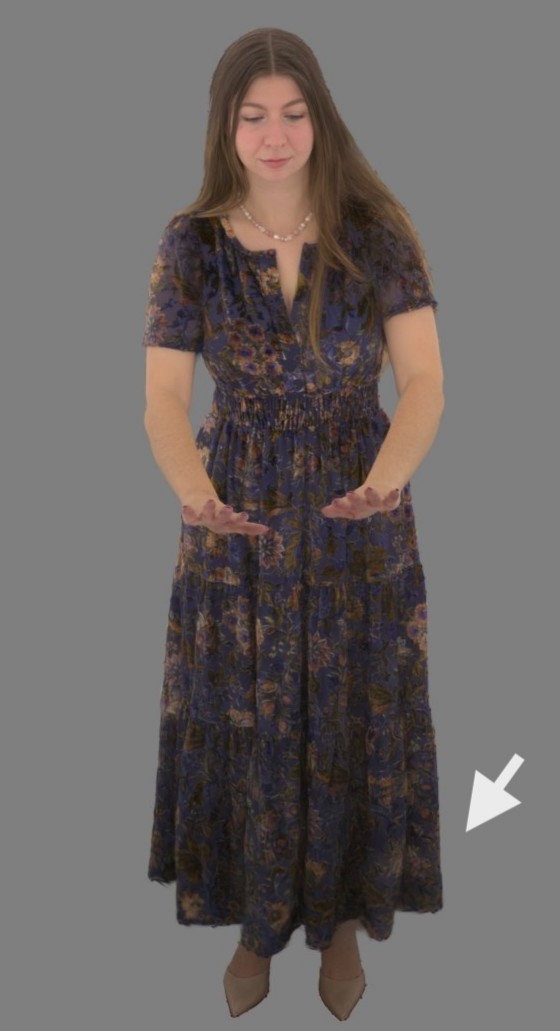} &
    \includegraphics[width=\dampingAblationWidth]{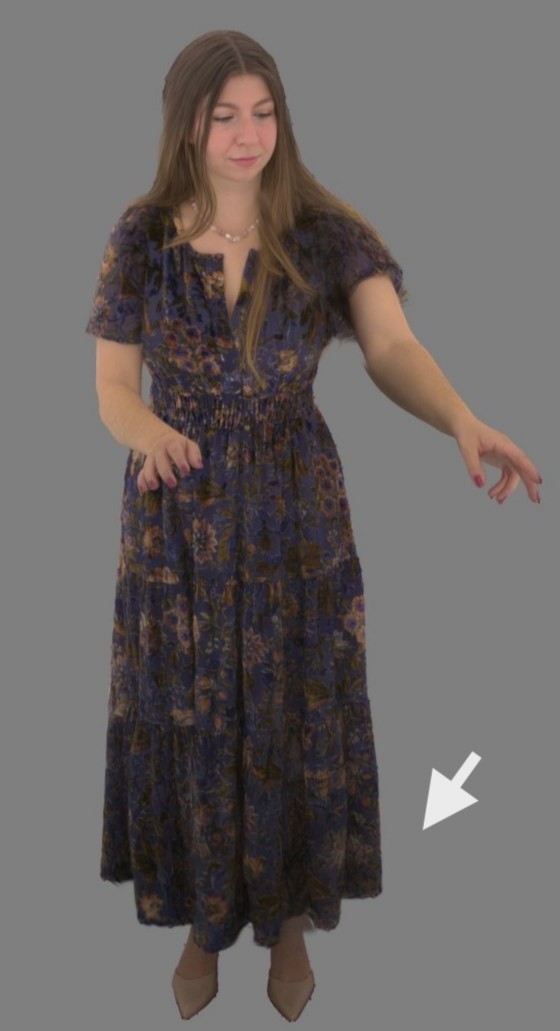} &
    \includegraphics[width=\dampingAblationWidth]{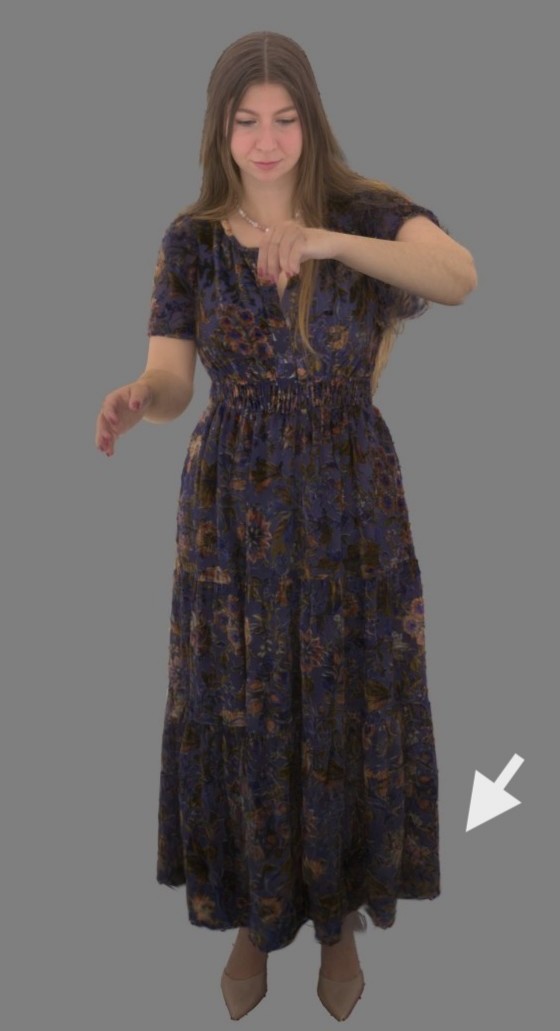} &
    \includegraphics[width=\dampingAblationWidth]{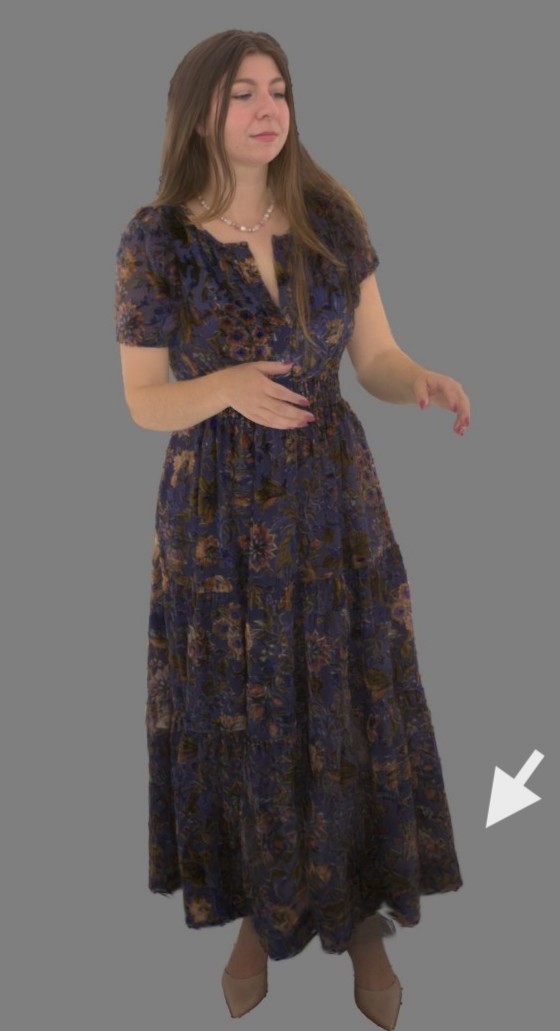} \\
  \end{tabular}
  \caption{\textbf{Effect of damping force on clothing motion dynamics.}Increasing damping dissipates kinetic energy more aggressively, leading to faster settling, while decreasing damping preserves oscillations and produces longer-lasting, more lively motion.}
  \label{fig:ablation_damping}
\end{figure}

\boldparagraph{Scaling Damping Force} The damping force dissipates kinetic energy. Higher damping attenuates velocity more aggressively and settles the cloth faster; lower damping preserves oscillations, so the motion lingers and appears livelier in Figure~\ref{fig:ablation_damping}.

\begin{figure}[ht]
  \newcommand{\poseAblationWidth}{0.17\columnwidth}
  \centering
  \footnotesize
  \setlength{\tabcolsep}{0pt}
  \begin{tabular}{@{}c*{5}{c}@{}}
    $\mathbf{F}_\text{pose} \times 0.5$ &
    \includegraphics[width=\poseAblationWidth]{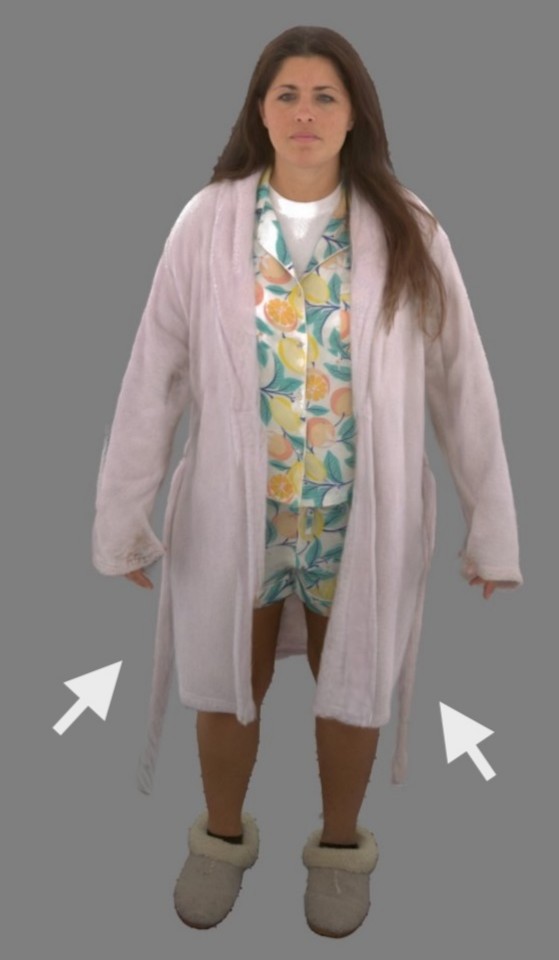} &
    \includegraphics[width=\poseAblationWidth]{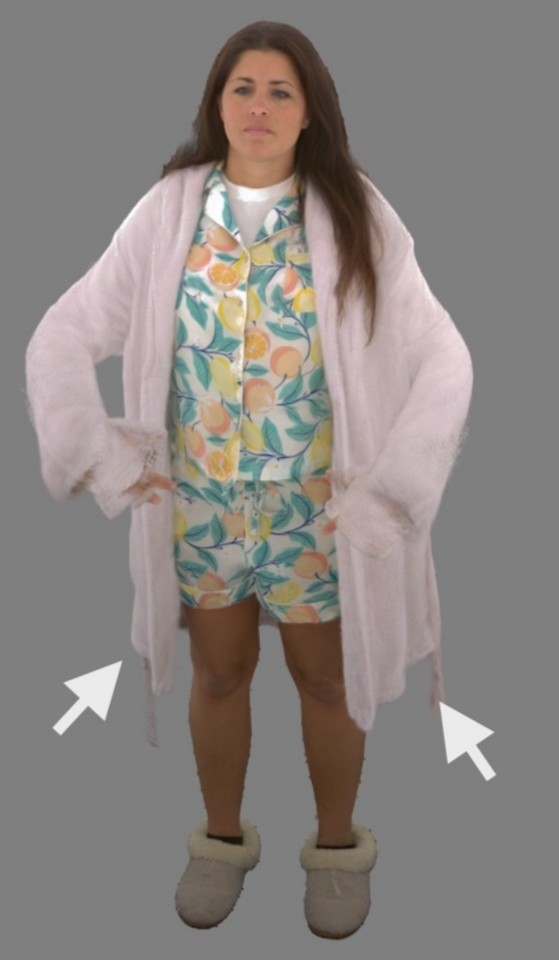} &
    \includegraphics[width=\poseAblationWidth]{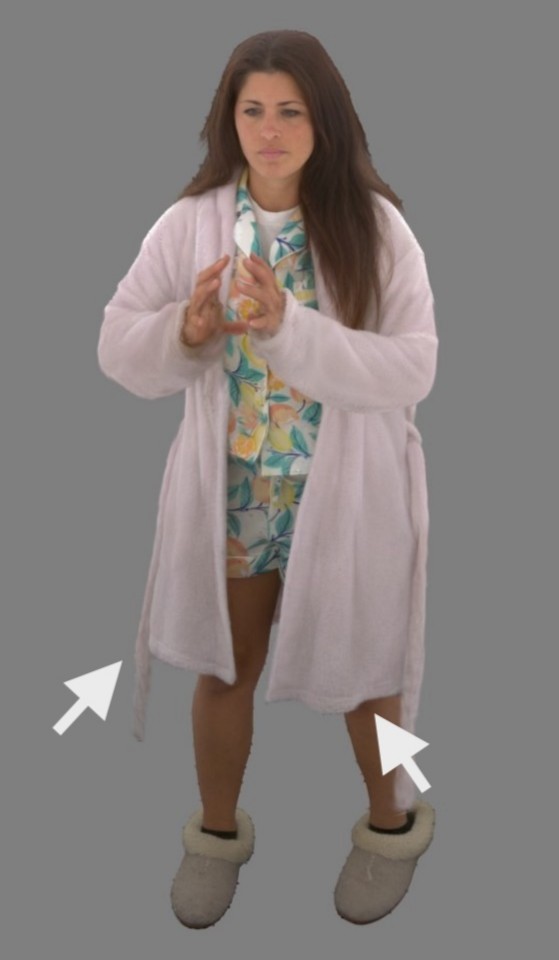} &
    \includegraphics[width=\poseAblationWidth]{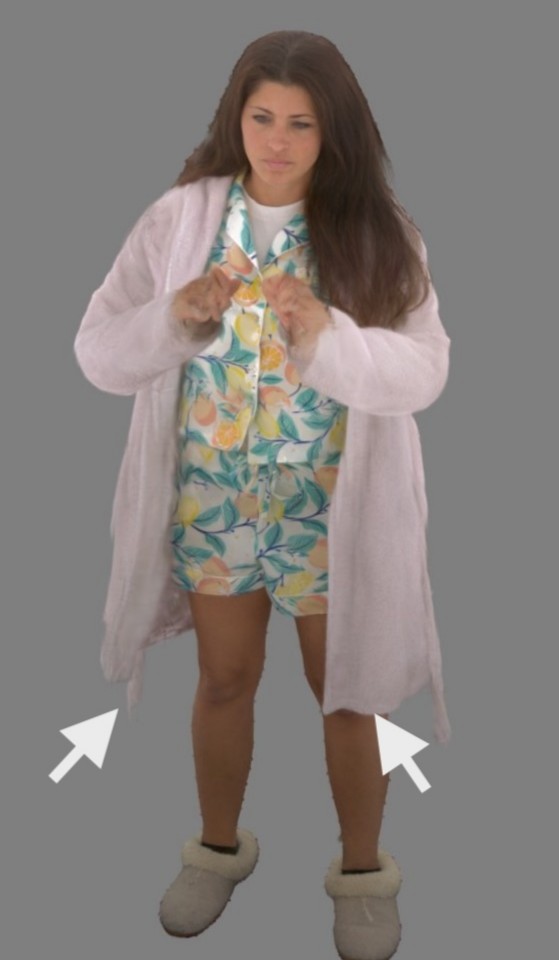} &
    \includegraphics[width=\poseAblationWidth]{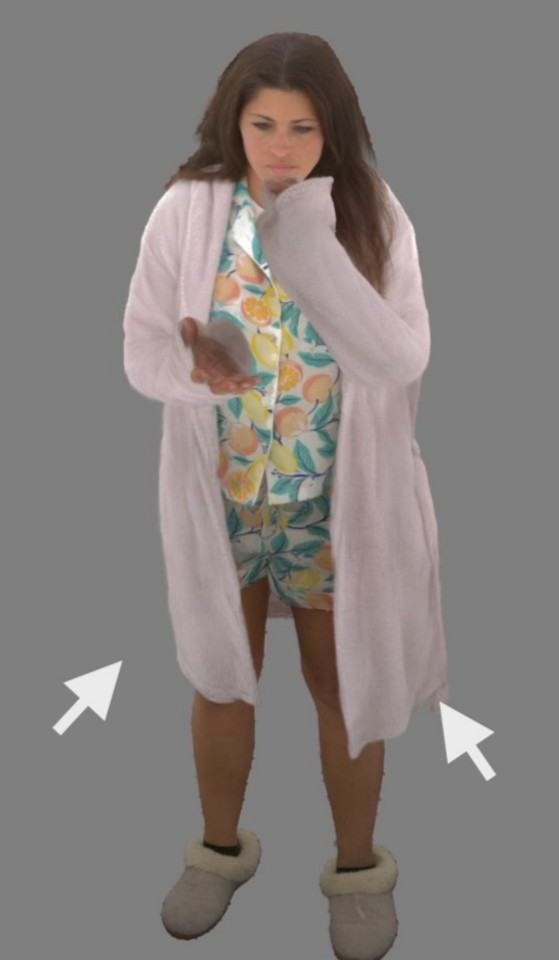} \\
    $\mathbf{F}_\text{pose} \times 1$ &
    \includegraphics[width=\poseAblationWidth]{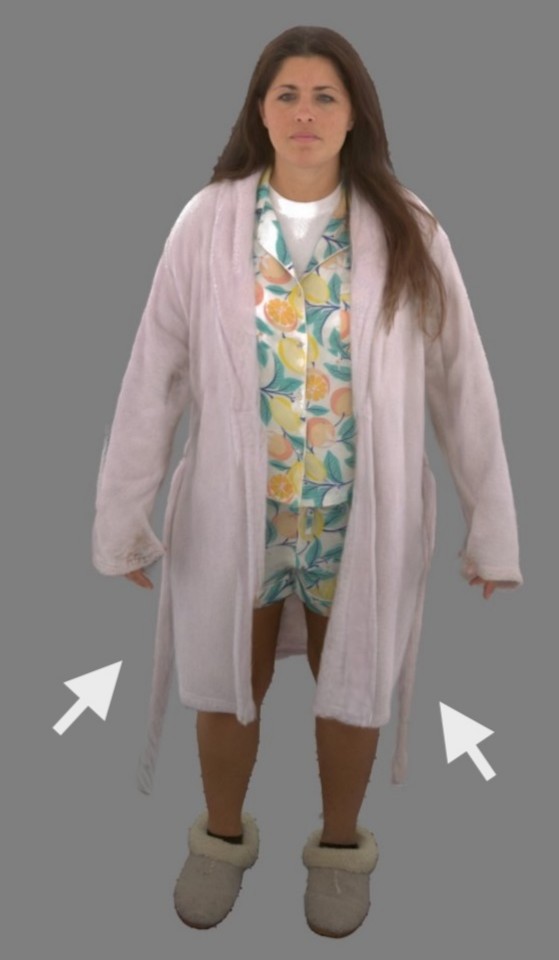} &
    \includegraphics[width=\poseAblationWidth]{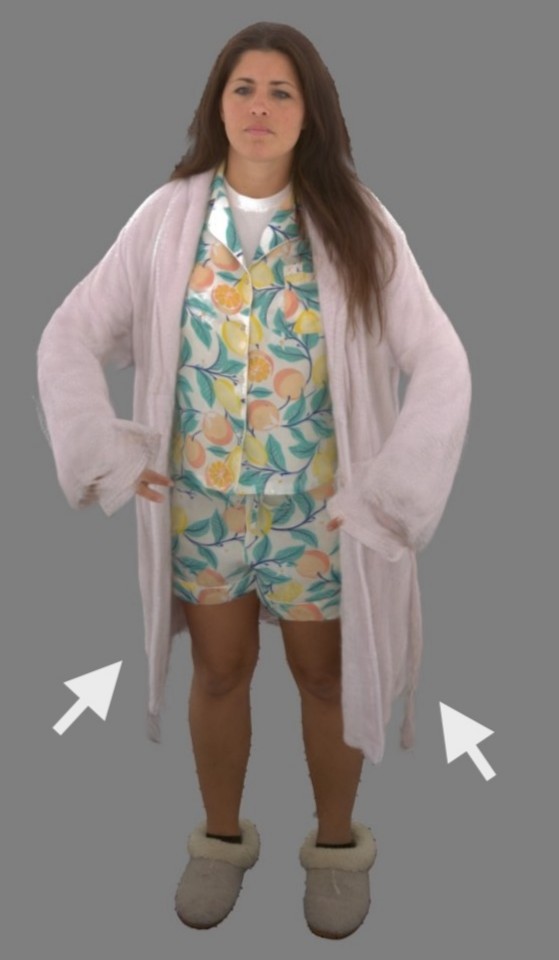} &
    \includegraphics[width=\poseAblationWidth]{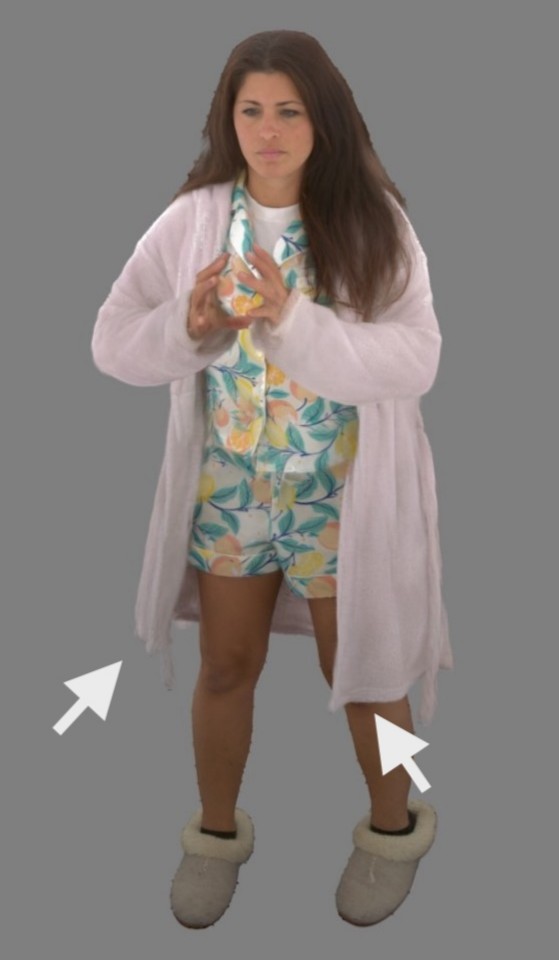} &
    \includegraphics[width=\poseAblationWidth]{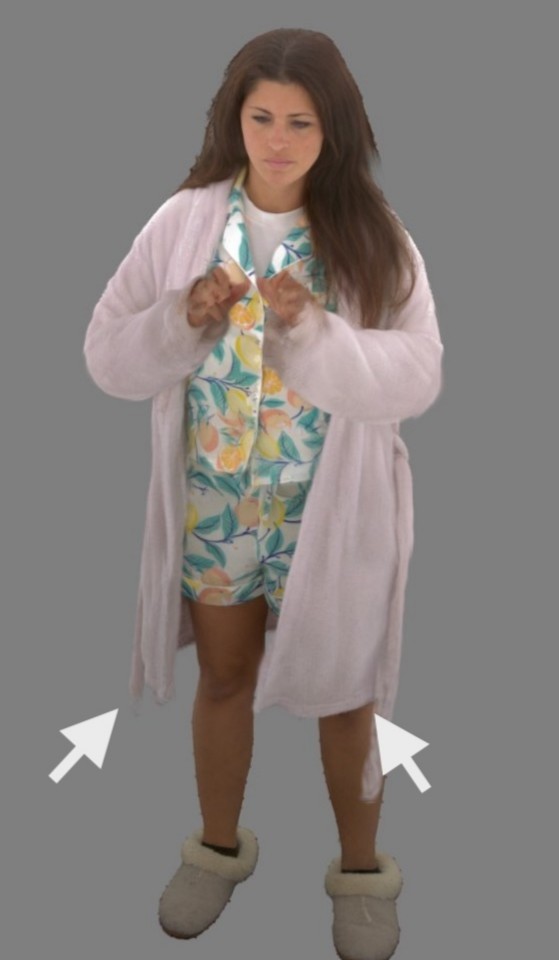} &
    \includegraphics[width=\poseAblationWidth]{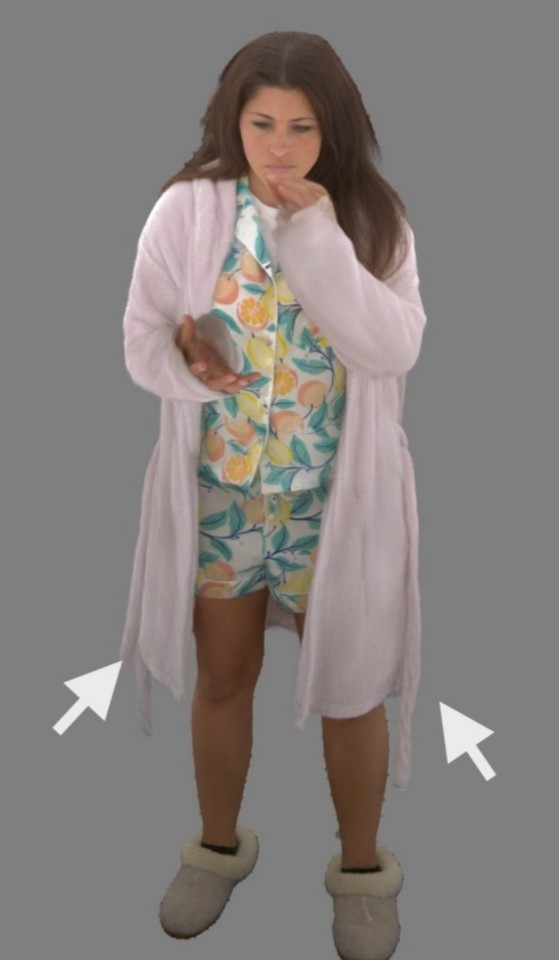} \\
    $\mathbf{F}_\text{pose} \times 2$ &
    \includegraphics[width=\poseAblationWidth]{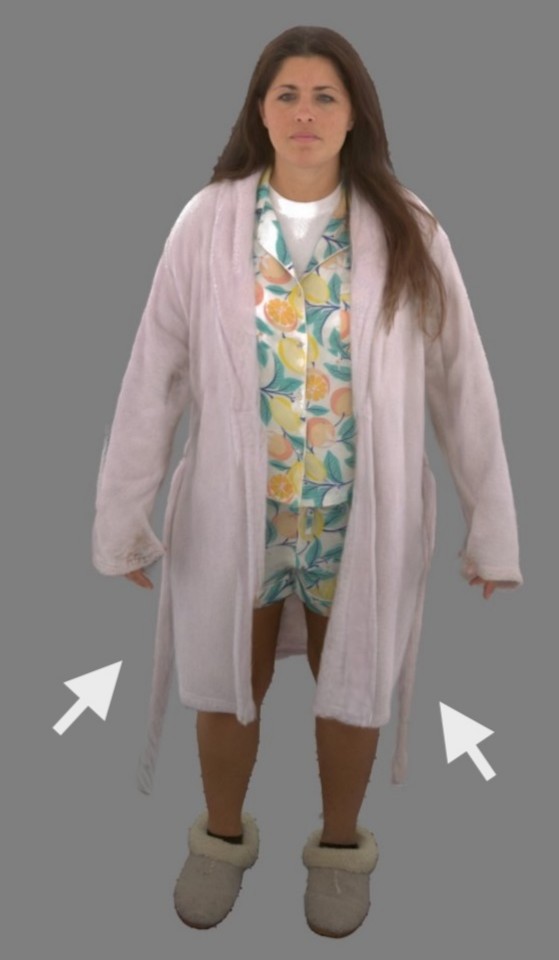} &
    \includegraphics[width=\poseAblationWidth]{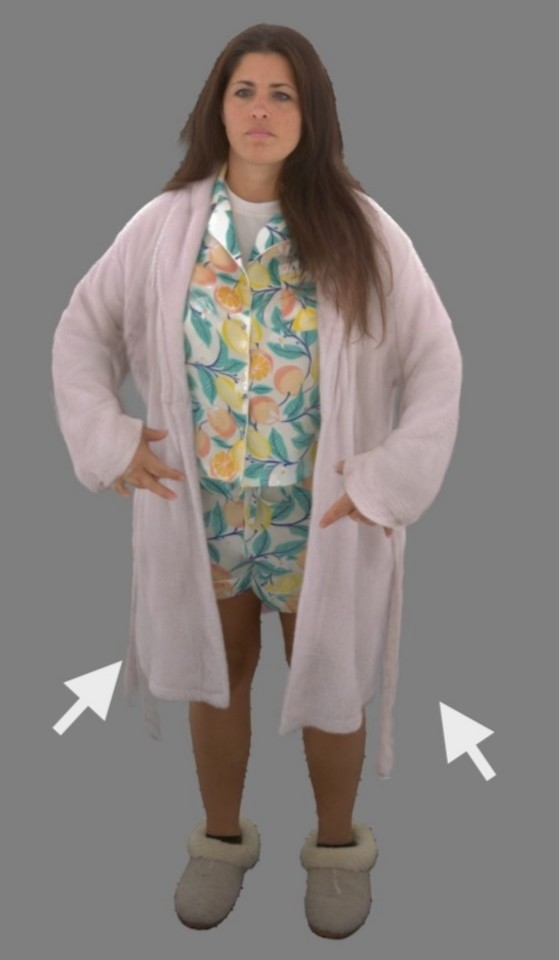} &
    \includegraphics[width=\poseAblationWidth]{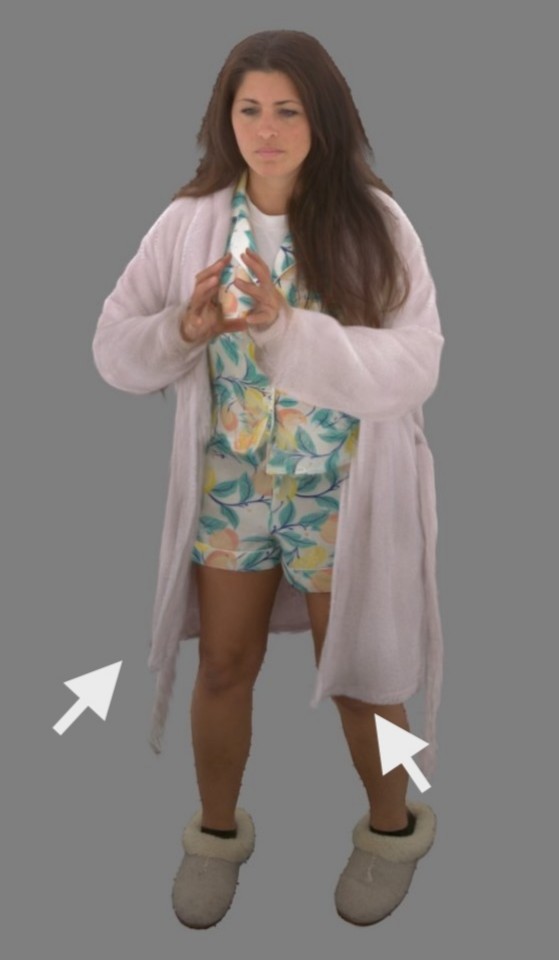} &
    \includegraphics[width=\poseAblationWidth]{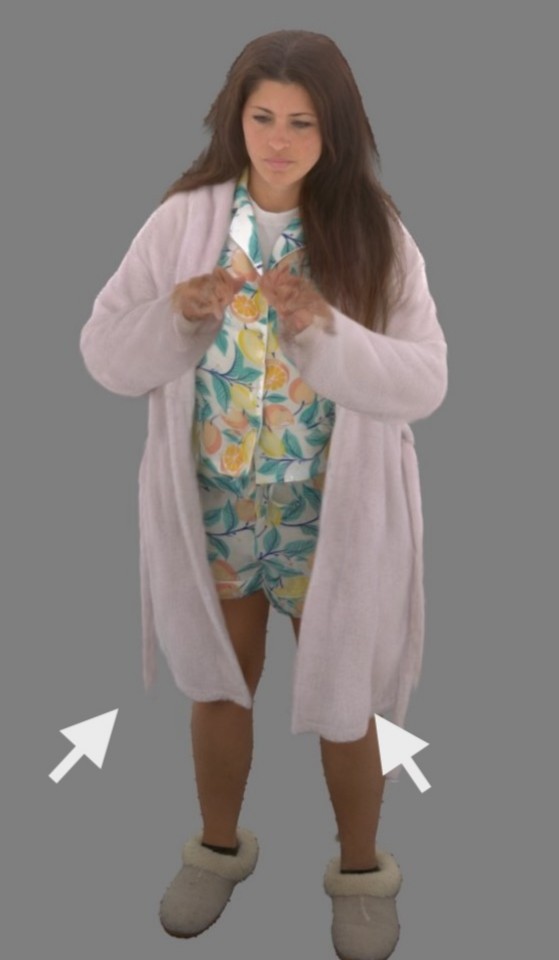} &
    \includegraphics[width=\poseAblationWidth]{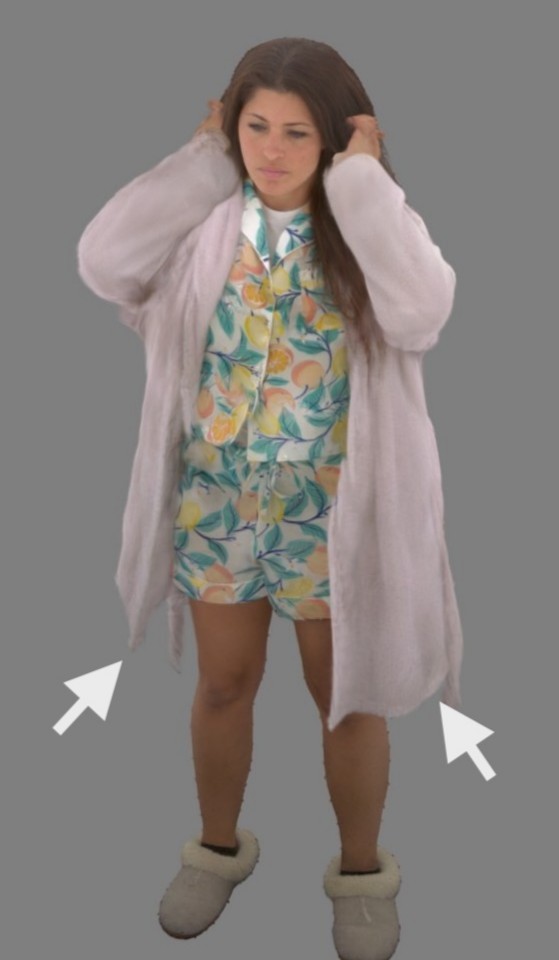} \\
  \end{tabular}
  \caption{\textbf{Effect of pose force on clothing motion dynamics.} Higher pose force increases the responsiveness of the garment to body pose changes, leading to tighter coupling and faster reaction.}
  \label{fig:ablation_pose}
\end{figure}

\boldparagraph{Scaling Pose Force} The pose force couples cloth dynamics to the driving body pose. Increasing it tightens this coupling, so the cloth tracks pose changes faster, evident in the tighter, snappier motion of the white robe in Figure~\ref{fig:ablation_pose}.

\boldparagraph{Different Initial Conditions} Holding the pose sequence fixed and varying the initial residual latent yields distinct yet plausible garment trajectories (Figure~\ref{fig:ablation_init_state}). The same body motion can therefore drive a family of clothing rollouts, and the dynamics model remains stable across these starting states.

\section{Discussion and Conclusion}

\boldparagraph{Limitations}
Our method has a few limitations. First, we use a non-relightable appearance model, so illumination is baked into the learned appearance. As a result, the residual latent can absorb lighting variations and produce inconsistent shadows under novel-pose driving; enabling the full relightable appearance model of our base avatar~\cite{Wang2025RelightableFG} would alleviate this.
Second, the residual latent is not explicitly disentangled from the driving signal and may still absorb pose-dependent cues; an explicit disentanglement loss may help~\cite{Bagautdinov2021DrivingsignalAF}.
Third, like other data-driven avatars that learn deformation directly from capture without explicit garment templates or contact priors, our model does not enforce hard constraints such as non-penetration. This is most visible in cross-driving, where transferring a fixed pose sequence to subjects with larger body shapes occasionally produces body--clothing interpenetration or self-intersection.
Finally, the dynamics model is learned per subject, so generalizing across identities requires retraining; extending it to a multi-subject or universal setting is left for future work.

\boldparagraph{Conclusion}
Prior data-driven avatars fix, regress, or retrieve the auxiliary latent, reducing it to a near-static pose-to-deformation map. We argue instead that this latent is more usefully treated as a quantity with its own \emph{temporal dynamics}, and that explicitly modeling how it evolves is what unlocks history-dependent, temporally coherent loose-clothing motion. We realize this with a transformer-based decoder conditioned on driving signals and a \emph{dynamics residual latent}. A spring-damper latent update then rolls the residual latent forward at inference from pose history and prior latent state. Quantitative metrics and a perceptual user study confirm more natural and diverse full-body animations under everyday motions with loose clothing. The force decomposition also exposes controls such as stiffness and damping. Because the dynamics model is orthogonal to the decoder and adds negligible cost, we expect it to transfer to other backbones and extend to a broader range of dynamic deformations.


\begin{acks}
We thank Igor Santesteban for early codebase support, Carsten Stoll and
Jinlong Yang for help with the body rig, and Nicholas Dahm for
infrastructure support. 
\end{acks}

\bibliographystyle{ACM-Reference-Format}
\bibliography{sample-base}

\clearpage

\appendix
\clearpage
\setcounter{page}{1}

\appendix
\onecolumn
\setcounter{figure}{0}
\setcounter{table}{0}
\renewcommand{\thefigure}{S\arabic{figure}}
\renewcommand{\thetable}{S\arabic{table}}

\begin{center}
{\Large\bfseries Supplementary Material}\\[0.25em]
{\large for ``Latent Dynamics for Full Body Avatar Animation''}
\end{center}
\vspace{0.75em}

\boldparagraph{Overview}
This supplementary document provides detailed implementation specifics and additional analyses to complement the main paper. It includes: (i) \hyperref[app:impl_details]{Implementation Details}, which describe the encoder (\hyperref[app:pose_encoder]{Sec.~\ref*{app:pose_encoder}}), the residual-latent-augmented transformer decoder and its key design choices (\hyperref[app:decoder_details]{Sec.~\ref*{app:decoder_details}}), and details for our latent dynamics model, including feature construction, and training strategy (\hyperref[app:dynamics_details]{Sec.~\ref*{app:dynamics_details}}); (ii) \hyperref[app:ablation]{Ablation Study}, which isolates the contributions of the transformer-based decoder (\hyperref[app:ablation_decoder]{Sec.~\ref*{app:ablation_decoder}}) and our latent dynamics design (\hyperref[app:ablation_dynamics]{Sec.~\ref*{app:ablation_dynamics}});
(iii) \hyperref[app:baseline_quality]{Additional Discussion on Baseline Quality (Sec.~\ref*{app:baseline_quality})}, analyzing baseline behavior under a reduced-data training setting; and (iv) \hyperref[app:limitations]{Failure Cases (Sec.~\ref*{app:limitations})}, providing a concrete failure example for the contact-related limitation discussed in the main paper.

\section{Implementation Details}
\label{app:impl_details}

\subsection{Encoder}
\label{app:pose_encoder}
As described in Sec.~3.1 of the main paper, we use a lightweight encoder to map the driving signal $\mathbf{P}_t$ to a driving latent $\mathbf{p}_t \in \mathbb{R}^{288}$. The encoder is implemented as a 3-layer MLP: three linear layers with a hidden dimension of 256 and LeakyReLU activations. We adopt the Momentum Human Rig (MHR)~\cite{MHR:2025} parameterization for $\mathbf{P}_t$, and decompose it as $\mathbf{P}_t = (\mathbf{K}_b, \mathbf{K}_f)$, where $\mathbf{K}_b$ comprises body keypoints and $\mathbf{K}_f$ comprises face keypoints. Each is transformed by the inverse of its respective root (body root for $\mathbf{K}_b$, face root for $\mathbf{K}_f$) before being passed to $\mathcal{E}$. The latent dynamics model of Sec.~3.3 consumes body-keypoint history only; face keypoints enter the pipeline solely through the encoder $\mathcal{E}$ and the decoder, since garment motion is physically driven by body motion.

\subsection{Pose+Clothing-Conditioned Decoder}
\label{app:decoder_details}

\boldparagraph{Architecture hyperparameters}
Table~\ref{tab:decoder_hparams} reports the key architectural choices used in our transformer-based decoder (Sec.~3 in the main paper).

\begin{table}[h]
    \centering
    \setlength{\tabcolsep}{6pt}
    \begin{tabular}{l c}
        \toprule
        \textbf{Component} & \textbf{Value} \\
        \midrule
        Input tokens $N_{in}$ & 4096 \\
        Model width $d$ & 1024 \\
        Latent array size $N_{latent}$ & 512 \\
        Transformer blocks $L$ & 12 \\
        Output queries $N_{out}$ & 4096 \\
        Output feature map $H{\times}W$ & $64{\times}64$ \\
        Num upsampling blocks & 3 \\
        Final UV resolution & $512{\times}512$ \\
        Driving latent dim $d_p$ & 288 \\
        Residual latent dim $d_z$ & 128 \\
        \bottomrule
    \end{tabular}
    \caption{\textbf{Decoder hyperparameters.} The transformer follows a Perceiver-IO-style latent-array design; dense UV outputs are produced by reshaping $N_{out}$ output tokens into a $H{\times}W$ feature map and decoding with convolutional upsampling blocks.}
    \label{tab:decoder_hparams}
\end{table}

\begin{figure}[t]
    \centering
    \includegraphics[width=0.8\textwidth]{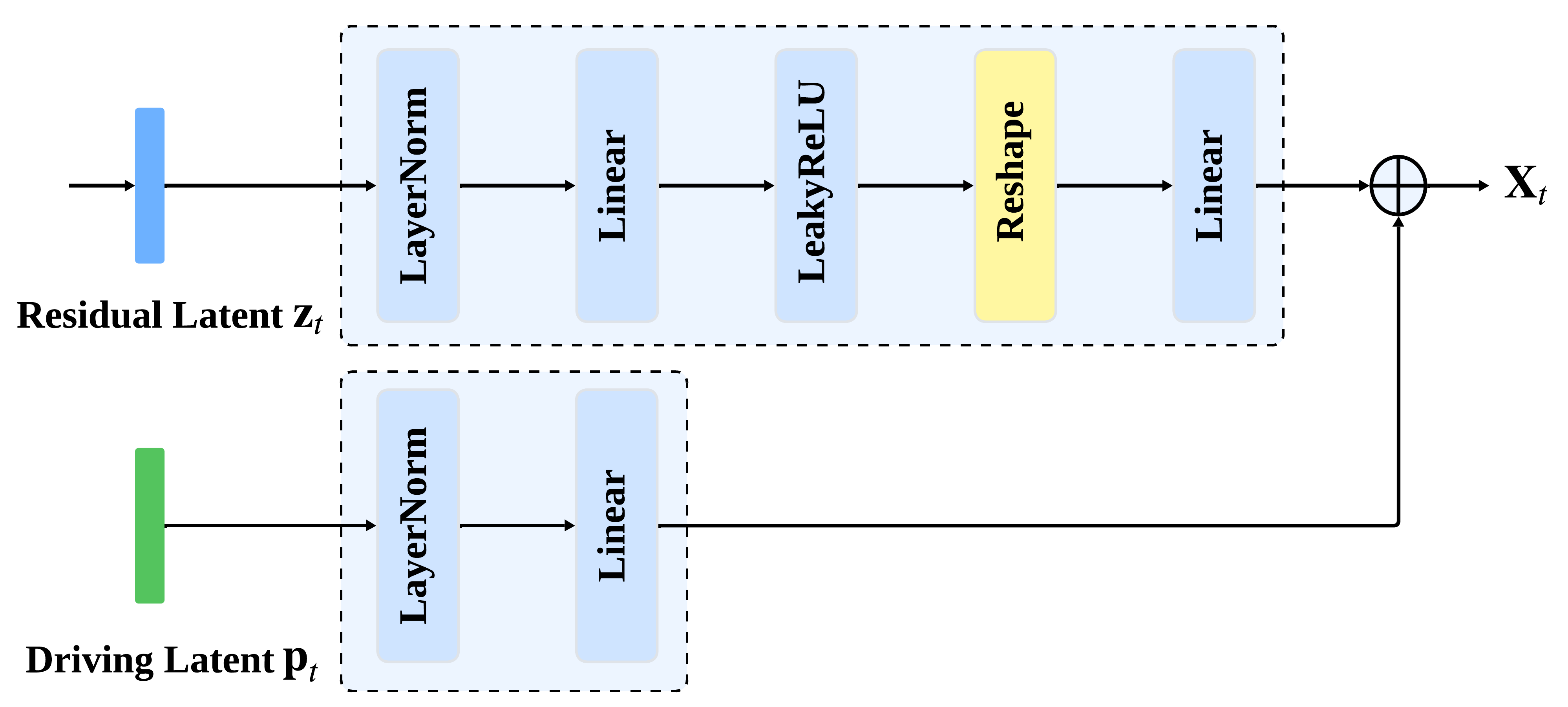}
    \caption{\textbf{Input-token construction.} We expand the residual latent $\mathbf{z}_t$ into per-token features and add a broadcast driving embedding $\mathbf{p}_t$ to form the input token array $\mathbf{X}_t$ for the transformer. Each linear block denotes a learnable linear map; the per-token feature dimension is brought to the model width $d$ before fusion.}
    \label{fig:token_construction}
\end{figure}

\boldparagraph{Input-token construction from driving and residual latents}
Let $\mathbf{p}_t$ denote the driving embedding and $\mathbf{z}_t\in\mathbb{R}^{128}$ denote the standardized residual latent at time $t$ (Sec.~3 in the main paper). We form $N_{in}$ input tokens by first expanding the two conditioning signals into token-aligned features and then mapping to width $d$ (Fig.~\ref{fig:token_construction}):
\begin{itemize}
    \item \textbf{Residual Latent branch.} We apply layer normalization~\cite{ba2016layernormalization} to $\mathbf{z}_t$ and map it with a linear layer from $128$ to $N_{in}\!\times\!128$ and reshape to a tensor in $\mathbb{R}^{N_{in}\times 128}$. A second linear layer maps the per-token feature dimension $128 \rightarrow d$, yielding residual tokens $\mathbf{X}^{(z)}_t\in\mathbb{R}^{N_{in}\times d}$.
    \item \textbf{Driving Latent branch.} We apply layer normalization~\cite{ba2016layernormalization} to $\mathbf{p}_t$ and map it with a linear layer to $\mathbb{R}^{d}$ and broadcast to all $N_{in}$ tokens, yielding $\mathbf{X}^{(p)}_t\in\mathbb{R}^{N_{in}\times d}$.
    \item \textbf{Token fusion.} We combine the two branches by addition, $\mathbf{X}_t=\mathbf{X}^{(z)}_t+\mathbf{X}^{(p)}_t$, and feed $\mathbf{X}_t$ to the Perceiver-IO attention stack as keys/values.
\end{itemize}

\begin{figure}[t]
    \centering
    \includegraphics[width=0.8\textwidth]{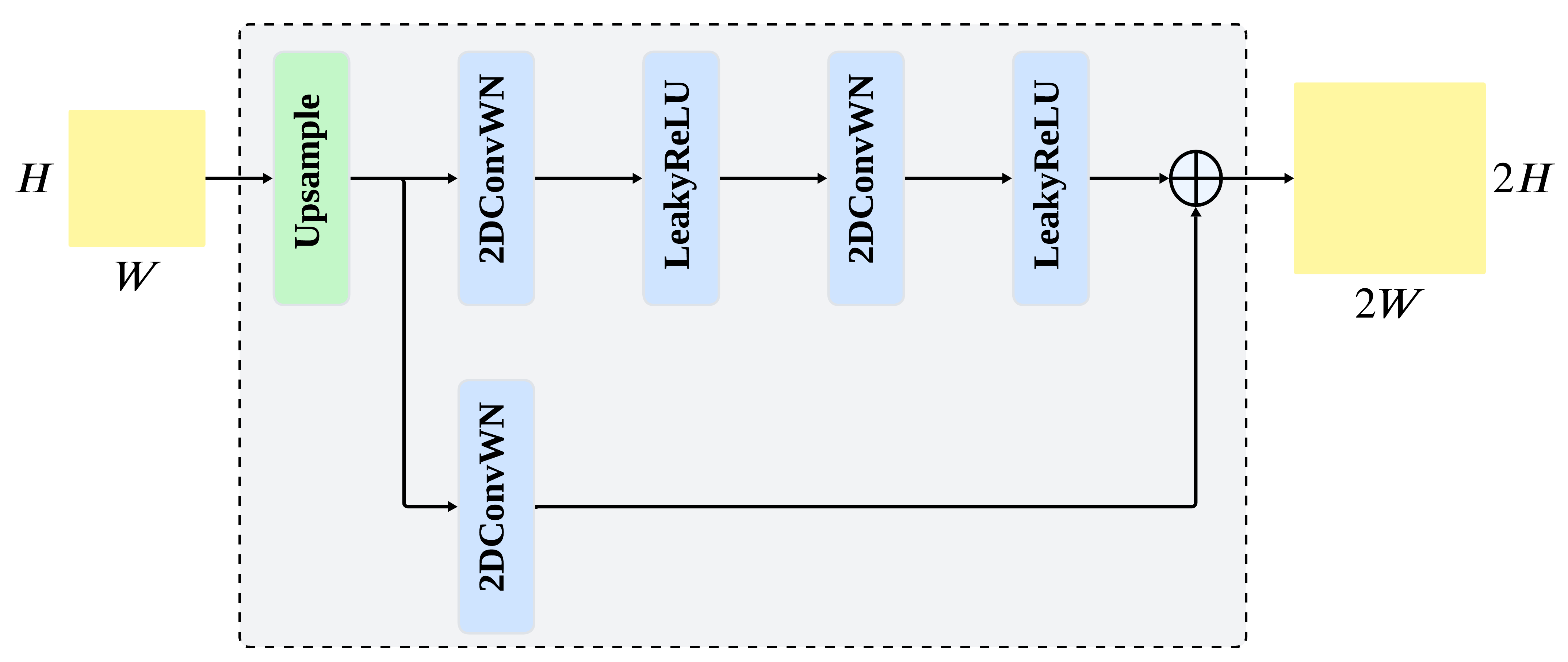}
    \caption{\textbf{Upsampling block.} Each block upsamples bilinearly, applies two 2D convolutions with weight normalization, and adds a skip connection. Chaining three blocks maps $64{\times}64$ features to $512{\times}512$ UV outputs.}
    \label{fig:upsampling_block}
\end{figure}

\boldparagraph{Convolutional upsampling blocks}
After the final cross-attention, we reshape the $N_{out}$ output tokens into a feature map in $\mathbb{R}^{64\times 64\times d}$ and decode dense UV maps using three upsampling blocks (Fig.~\ref{fig:upsampling_block}). Each block performs bilinear upsampling by a factor of 2, followed by two convolutions, and adds a residual/skip connection from the upsampled features to the block output.

\boldparagraph{Avatar training objective and schedule}
We train the avatar (Gaussian parameters + decoder) with multi-view supervision using a weighted sum of photometric reconstruction ($\ell_1$), perceptual loss (LPIPS), and a foreground mask loss computed from the rendered alpha matte:
\begin{align}
    \mathcal{L}_{avatar} = \lambda_{1}\mathcal{L}_{1} + \lambda_{p}\mathcal{L}_{LPIPS} + \lambda_{m}\mathcal{L}_{mask}.
\end{align}
Unless otherwise noted, we use $\lambda_{1}{=}1$, $\lambda_{p}{=}0.1$, and $\lambda_{m}{=}0.5$, and train for 150k iterations with a global batch size of 64 using the AdamW optimizer~\cite{Loshchilov2017DecoupledWD} with learning rate $1{\times}10^{-4}$.

\subsection{Latent Dynamics Model Details}
\label{app:dynamics_details}

\boldparagraph{Test-time pipeline}
Our pipeline does \emph{not} require any input image at inference. Given a per-subject avatar trained as in Sec.~3 of the main paper, animation from a novel driving signal sequence $\{\mathbf{P}_t\}_{t=1}^{T}$ proceeds as follows:
\begin{enumerate}
    \item Initialize $\mathbf{z}_0 = \mathbf{z}_{\text{ref}}$ and $\mathbf{v}_0 = \mathbf{0}$, where $\mathbf{z}_{\text{ref}}$ is fixed at training time as the standardized SAPIENS$+$PCA latent of the first training frame for that subject (Sec.~3.2 of the main paper).
    \item At each step $t$, compute the pose descriptor $\mathbf{f}^{\text{pose}}_t$ from a short body-keypoint history and the previous-state feature $\mathbf{f}^{\text{state}}_{t-1}=[\mathbf{z}_{t-1};\,\mathbf{v}_{t-1};\,\|\mathbf{v}_{t-1}\|_2]$, predict the force parameters $\mathbf{g}_t,\boldsymbol{\kappa}_t,\mathbf{c}_t,\mathbf{m}_t$, and roll out Eq.~(10) of the main paper to obtain $(\mathbf{z}_t,\mathbf{v}_t)$.
    \item Feed $\mathbf{z}_t$ together with the encoded driving latent $\mathbf{p}_t$ to the transformer-based decoder for rendering.
\end{enumerate}
The image-based extraction of $\mathbf{z}_t$ via SAPIENS$+$PCA is used \emph{only at training time} to supervise the avatar/decoder and to provide targets for the dynamics model. During training, alignment between the pose used to compute $\mathbf{f}^{\text{pose}}_t$ and the frontal image used to extract $\mathbf{z}_t$ is enforced by drawing both from the same synchronized capture frame; at test time, no such alignment is needed and only the input pose sequence drives the model.

\boldparagraph{Rotational kinematics}
We compute per-joint rotational signals from joint rotations $\{\mathbf{R}^{(j)}_t\}_{j=1}^{J}\in SO(3)$ using the $SO(3)$ log map (axis-angle vectors). In particular, we use pose-to-reference rotations
$\boldsymbol{\xi}^{(j)}_t=\log_{SO(3)}\!\big(\mathbf{R}^{(j)\top}_{ref}\mathbf{R}^{(j)}_{t}\big)$,
inter-frame angular velocity
$\boldsymbol{\omega}^{(j)}_t=\tfrac{1}{\Delta t}\log_{SO(3)}\!\big(\mathbf{R}^{(j)\top}_{t-1}\mathbf{R}^{(j)}_{t}\big)$,
and finite differences for angular acceleration and jerk:
\begin{align}
    \boldsymbol{\alpha}^{(j)}_t &= \tfrac{1}{\Delta t}\big(\boldsymbol{\omega}^{(j)}_{t}-\boldsymbol{\omega}^{(j)}_{t-1}\big), \\
    \boldsymbol{\eta}^{(j)}_t &= \tfrac{1}{\Delta t}\big(\boldsymbol{\alpha}^{(j)}_{t}-\boldsymbol{\alpha}^{(j)}_{t-1}\big).
\end{align}

\boldparagraph{Pose features $\mathbf{f}^{pose}_t$}
We construct a compact pose descriptor by grouping joints into 14 anatomical groups (left/right upper leg, left/right lower leg, left/right foot, left/right upper arm, left/right lower arm, left/right wrist, core, and head). For each group $g$, we compute six scalars: four mean magnitudes,
$\mathbb{E}_{j\in g}\!\left[\|\boldsymbol{\xi}^{(j)}_t\|_2\right]$,
$\mathbb{E}_{j\in g}\!\left[\|\boldsymbol{\omega}^{(j)}_t\|_2\right]$,
$\mathbb{E}_{j\in g}\!\left[\|\boldsymbol{\alpha}^{(j)}_t\|_2\right]$,
$\mathbb{E}_{j\in g}\!\left[\|\boldsymbol{\eta}^{(j)}_t\|_2\right]$,
and two ``signed'' statistics capturing coherent group motion,
$\left\|\mathbb{E}_{j\in g}\!\left[\boldsymbol{\xi}^{(j)}_t\right]\right\|_2$ and
$\left\|\mathbb{E}_{j\in g}\!\left[\boldsymbol{\omega}^{(j)}_t\right]\right\|_2$.
Additionally, for each of the six bilateral pairs (left/right upper leg, left/right lower leg, left/right foot, left/right upper arm, left/right lower arm, and left/right wrist), we include two symmetry features: the differences of angular-velocity and angular-acceleration magnitudes between the left and right groups. Concatenating all features yields $\mathbf{f}^{pose}_t\in\mathbb{R}^{d_p}$ with $d_p=6\times 14 + 2\times 6=96$.

\boldparagraph{Previous-state features $\mathbf{f}^{state}_{t-1}$}
In the standardized latent space, we summarize the previous step with
$\mathbf{f}^{state}_{t-1}=[\mathbf{z}_{t-1};\,\mathbf{v}_{t-1};\,\|\mathbf{v}_{t-1}\|_2]\in\mathbb{R}^{d_s}$, where $d_s=2d_z+1$ (thus $d_s=257$ for $d_z=128$).

\boldparagraph{Network design for force parameters}
At each time step, a lightweight network predicts the parameters in Eq.~(\ref{eq:latent_dynamics}) of the main paper:
$\mathbf{g}_t,\boldsymbol{\kappa}_t,\mathbf{c}_t,\mathbf{m}_t$.
We implement this predictor as 4 MLPs (one per parameter group) operating on the concatenated input $[\mathbf{f}^{pose}_t;\mathbf{f}^{state}_{t-1}]$. Each network uses 4 fully connected layers with hidden width 256 and GELU activations, followed by a linear projection to $\mathbb{R}^{d_z}$. We enforce positivity of $\boldsymbol{\kappa}_t,\mathbf{c}_t,\mathbf{m}_t$ with a softplus nonlinearity.

\boldparagraph{User-controllable force scaling}
The force decomposition of Eq.~(\ref{eq:latent_dynamics}) in the main paper admits a per-component scalar gain that exposes the user controls used in Sec.~4.2 of the main paper. Concretely, we replace the three forces in the acceleration with scaled versions
\begin{align}
    \mathbf{F}^{\alpha}_{\text{pose},t}    &= \alpha_{\text{pose}}\, \mathbf{g}_t, \notag\\
    \mathbf{F}^{\alpha}_{\text{damping},t} &= \alpha_{\text{damp}}\, \mathbf{c}_t \odot \mathbf{v}_t, \notag\\
    \mathbf{F}^{\alpha}_{\text{spring},t}  &= \alpha_{\text{spring}}\, \boldsymbol{\kappa}_t \odot (\mathbf{z}_t - \mathbf{z}_{\text{ref}}),
\end{align}
where $\alpha_{\text{pose}},\alpha_{\text{damp}},\alpha_{\text{spring}}\in\mathbb{R}_{\ge 0}$ are user-supplied scalar gains, and form the acceleration as
\begin{align}
    \mathbf{a}_t = \big(\mathbf{F}^{\alpha}_{\text{pose},t} - \mathbf{F}^{\alpha}_{\text{damping},t} - \mathbf{F}^{\alpha}_{\text{spring},t}\big) \oslash \mathbf{m}_t.
\end{align}
Each $\alpha$ acts as a global gain on its corresponding force component while leaving the per-dimension network outputs $\mathbf{g}_t,\mathbf{c}_t,\boldsymbol{\kappa}_t$ unchanged. The default setting $\alpha_{\text{pose}}=\alpha_{\text{damp}}=\alpha_{\text{spring}}=1$ recovers the un-modified update of Eq.~(\ref{eq:latent_dynamics}); the figures in Sec.~4.2 of the main paper sweep one component at a time over $\{0.5, 1, 2\}$ while holding the other two at $1$.

\boldparagraph{Dynamics training strategy}
We train the latent dynamics model on extracted latent trajectories with an MSE loss in the standardized latent space, using multi-step rollouts with curriculum teacher forcing. We use the Adam optimizer~\cite{Kingma2014AdamAM} with learning rate $5{\times}10^{-5}$. Concretely, we train for 1500 epochs with batch size 256 on a single H100 GPU; during training we gradually (i) increase the rollout length from 4 to 50 steps and (ii) decrease the teacher-forcing probability from 0.9 to 0.02.

\subsection{Runtime and Computational Cost}
\label{app:runtime}

Table~\ref{tab:runtime} reports a per-frame inference breakdown of our pipeline measured on a single NVIDIA H100 GPU. The latent dynamics model adds only $2.1$\,ms per frame, about $3.1\%$ of the per-frame budget, so its contribution is negligible relative to the decoder and Gaussian splatting. The full unoptimized pipeline runs at approximately $15$\,FPS, while the dynamics model alone runs at $\geq\!450$\,FPS standalone. The dynamics model is therefore orthogonal to the decoder: adding our test-time evolution to any real-time decoder preserves real-time behavior.

For training time, the avatar (Gaussian parameters together with the transformer-based decoder) takes $\approx 2.5$ days on $8{\times}$H100. The latent dynamics model is trained in parallel to the avatar/decoder, on the extracted latent trajectories, and takes $\approx 1$ hour on a single H100.

\begin{table}[h]
    \centering
    \setlength{\tabcolsep}{6pt}
    \begin{tabular}{l c c}
        \toprule
        \textbf{Component} & \textbf{Time (ms)} & \textbf{Share} \\
        \midrule
        Encoder $\mathcal{E}$ & 1.8 & 2.6\% \\
        Latent dynamics model & 2.1 & 3.1\% \\
        Transformer decoder $+$ rendering & 64.1 & 94.3\% \\
        \midrule
        Total per frame & 68.0 & 100\% \\
        \bottomrule
    \end{tabular}
    \caption{\textbf{Inference cost per frame} on a single NVIDIA H100. Our latent dynamics model contributes about $3\%$ of total inference time.}
    \label{tab:runtime}
\end{table}

\section{Ablation Study}
\label{app:ablation}
\subsection{Decoder Architecture}
\label{app:ablation_decoder}

To further study the effectiveness of our transformer-based decoder architecture, we replace it with a convolution-based architecture similar to the decoder design in~\cite{Bagautdinov2021DrivingsignalAF}. Specifically, we remove the Perceiver-IO-style attention stack and directly decode dense UV feature maps from the conditioning signals using a convolutional network.

\begin{figure}[t]
    \centering
    \small
    \setlength{\tabcolsep}{0pt}
    \newcommand{\rowlab}[1]{\begin{tabular}[c]{@{}l@{}}#1\end{tabular}}
    \newcommand{\imgw}{0.133\linewidth}
    \newcommand{\labw}{0.18\linewidth}
    \begin{tabular}{@{}p{\labw}@{}*{6}{c}@{}}
        \raisebox{10ex}{\rowlab{Convolution-based}} &
        \includegraphics[width=\imgw]{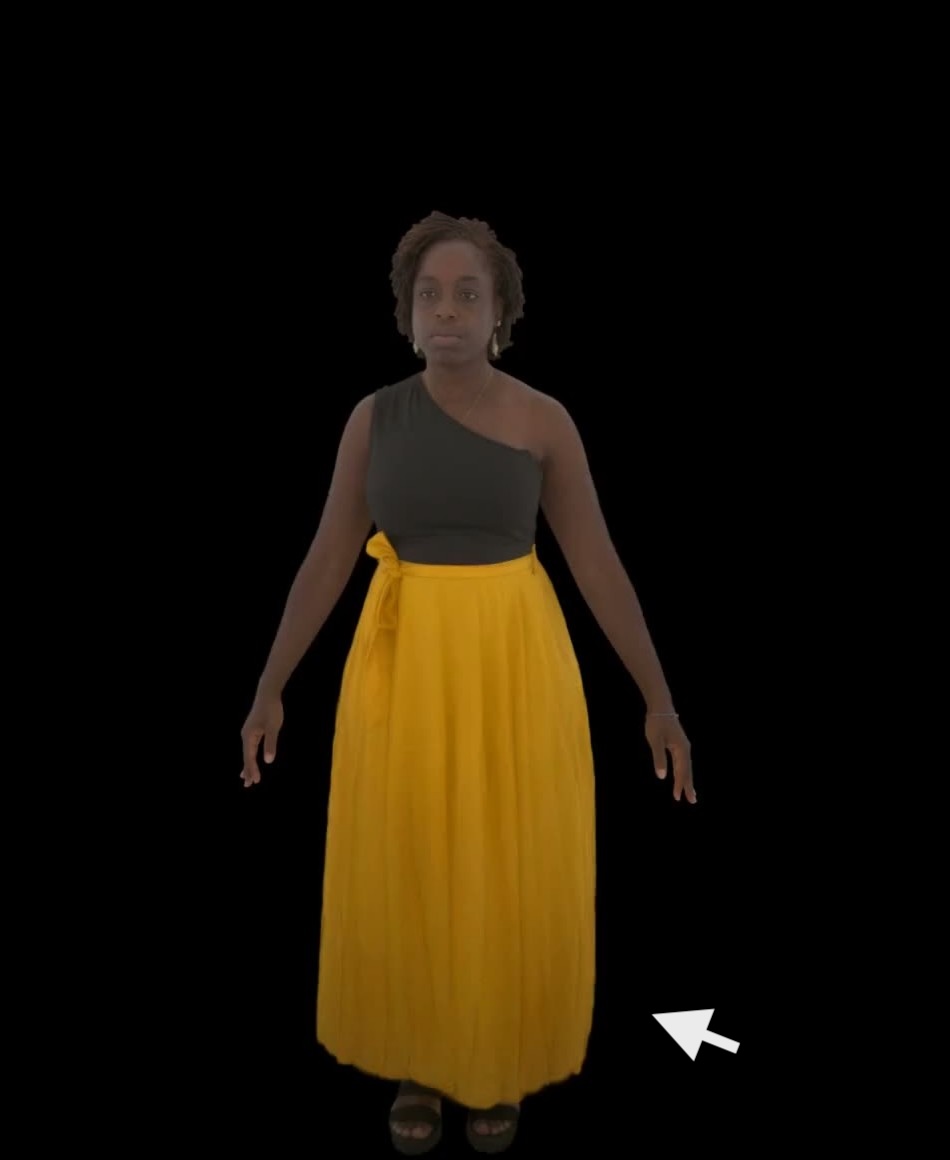} &
        \includegraphics[width=\imgw]{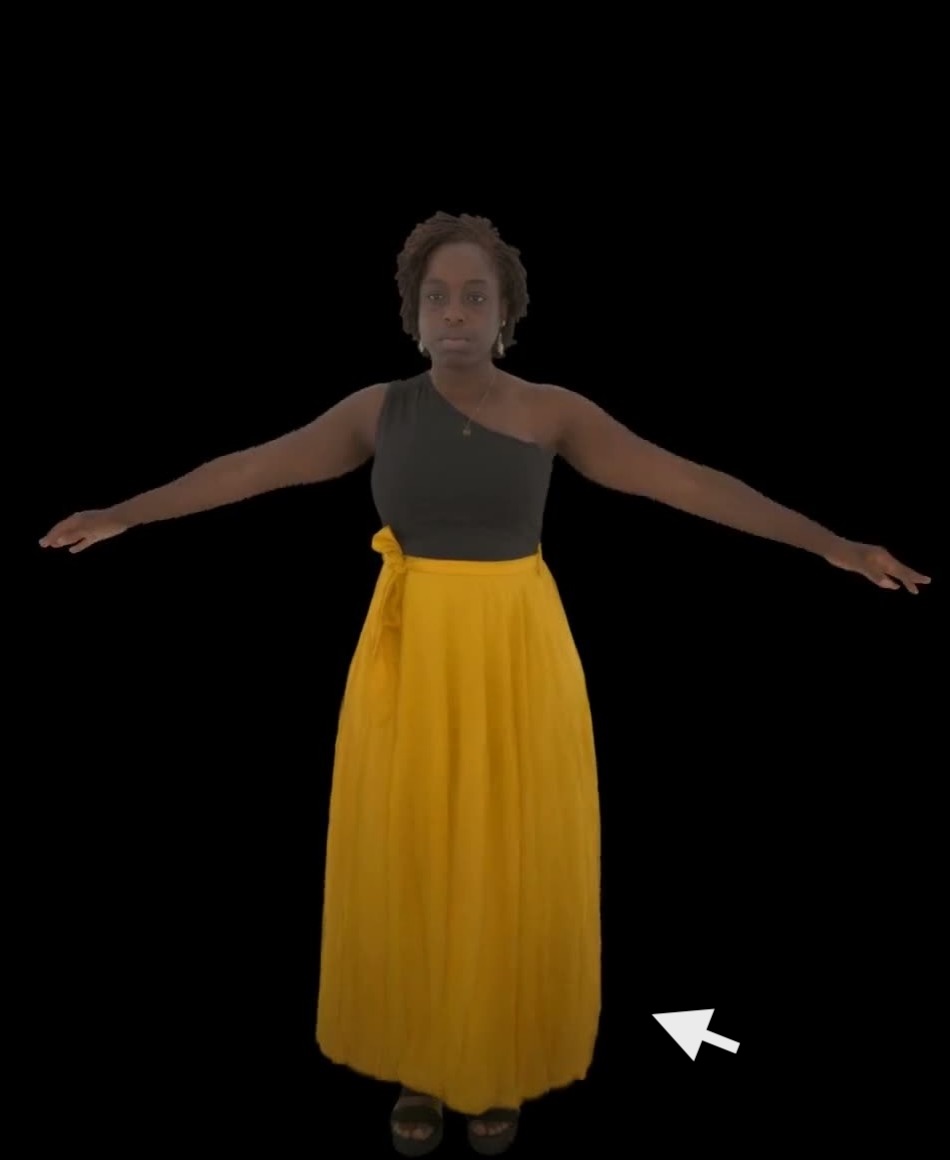} &
        \includegraphics[width=\imgw]{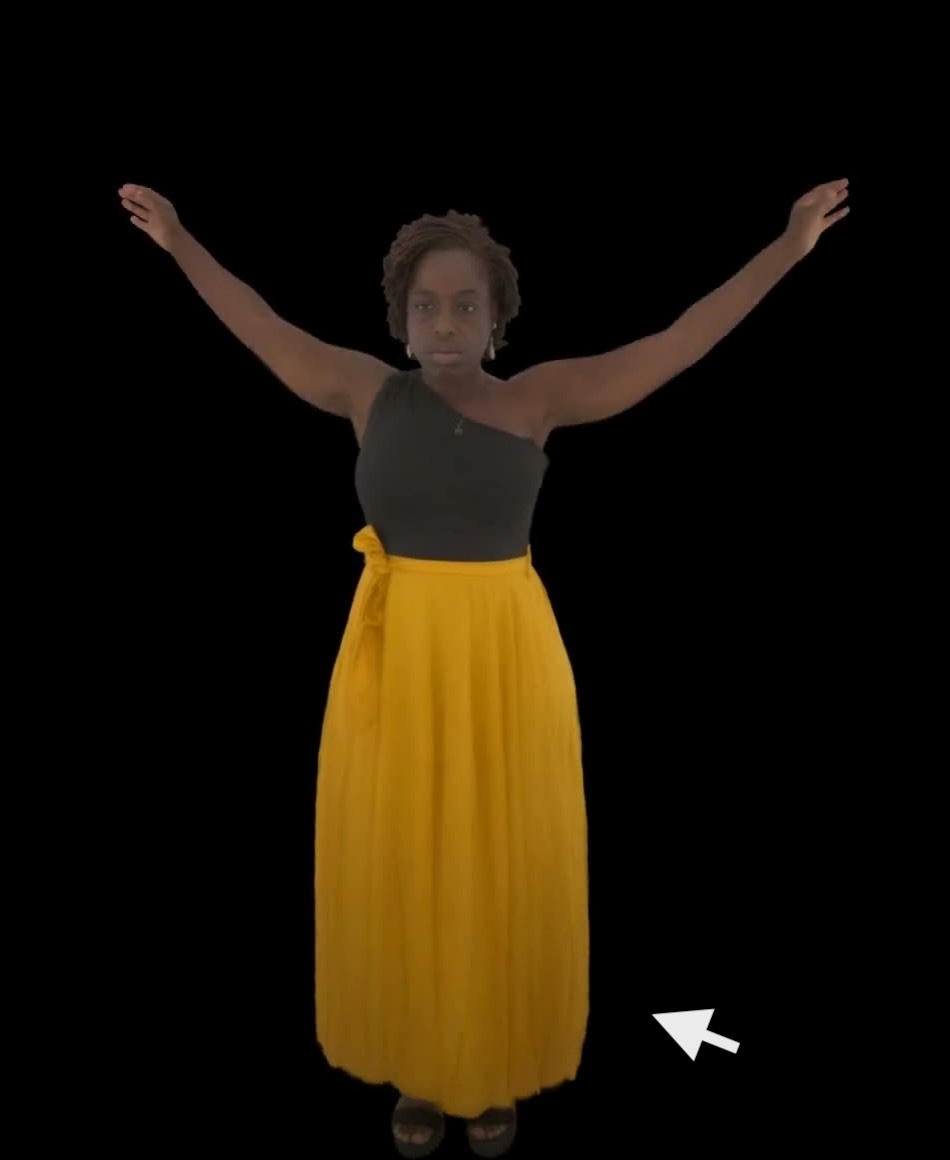} &
        \includegraphics[width=\imgw]{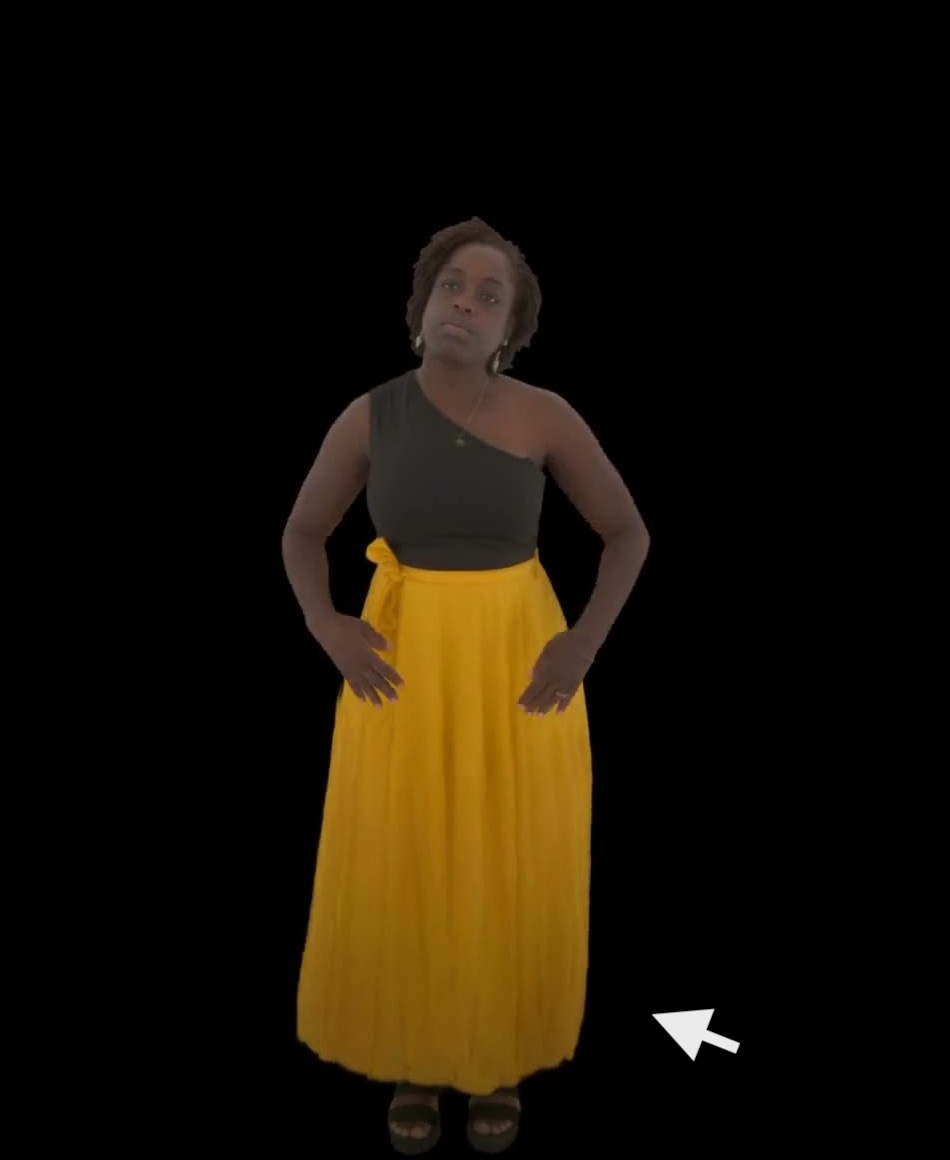} &
        \includegraphics[width=\imgw]{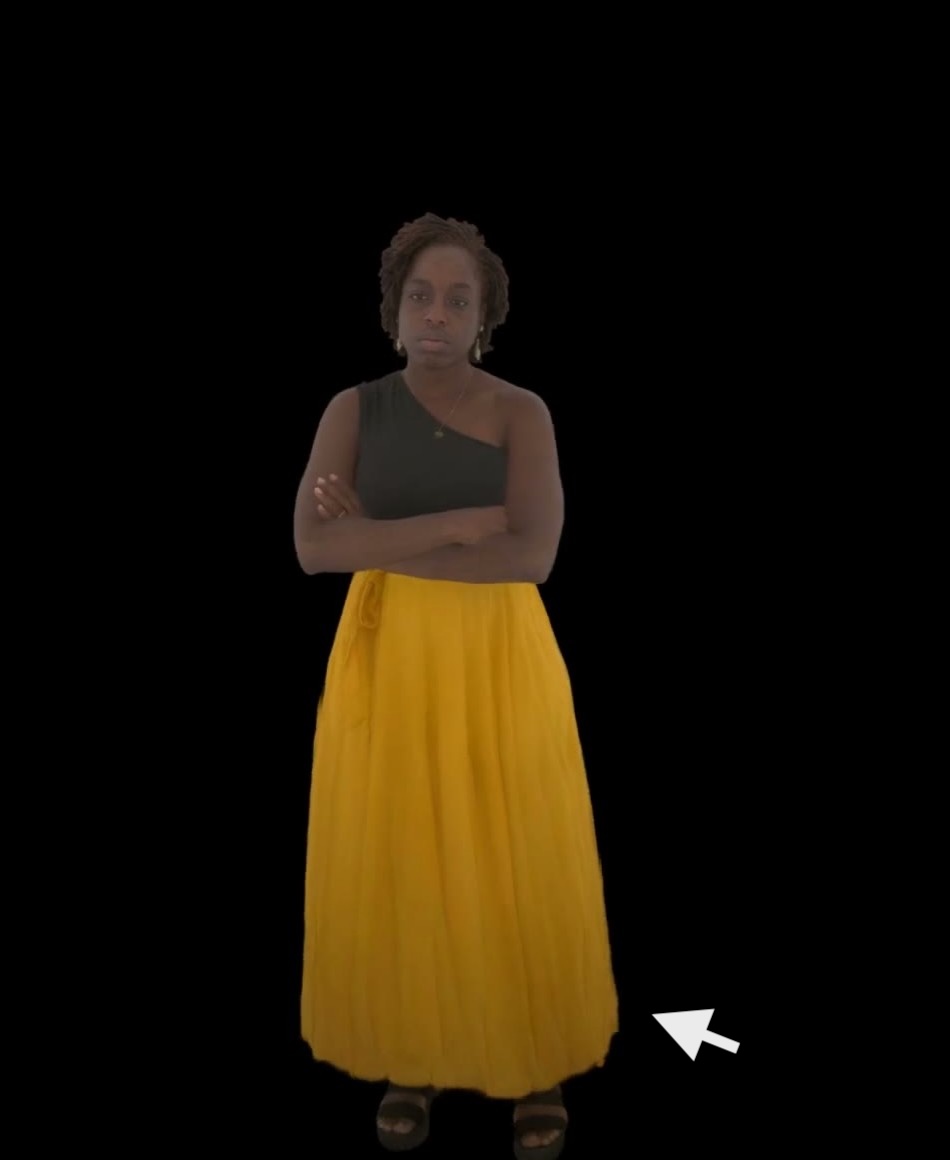} &
        \includegraphics[width=\imgw]{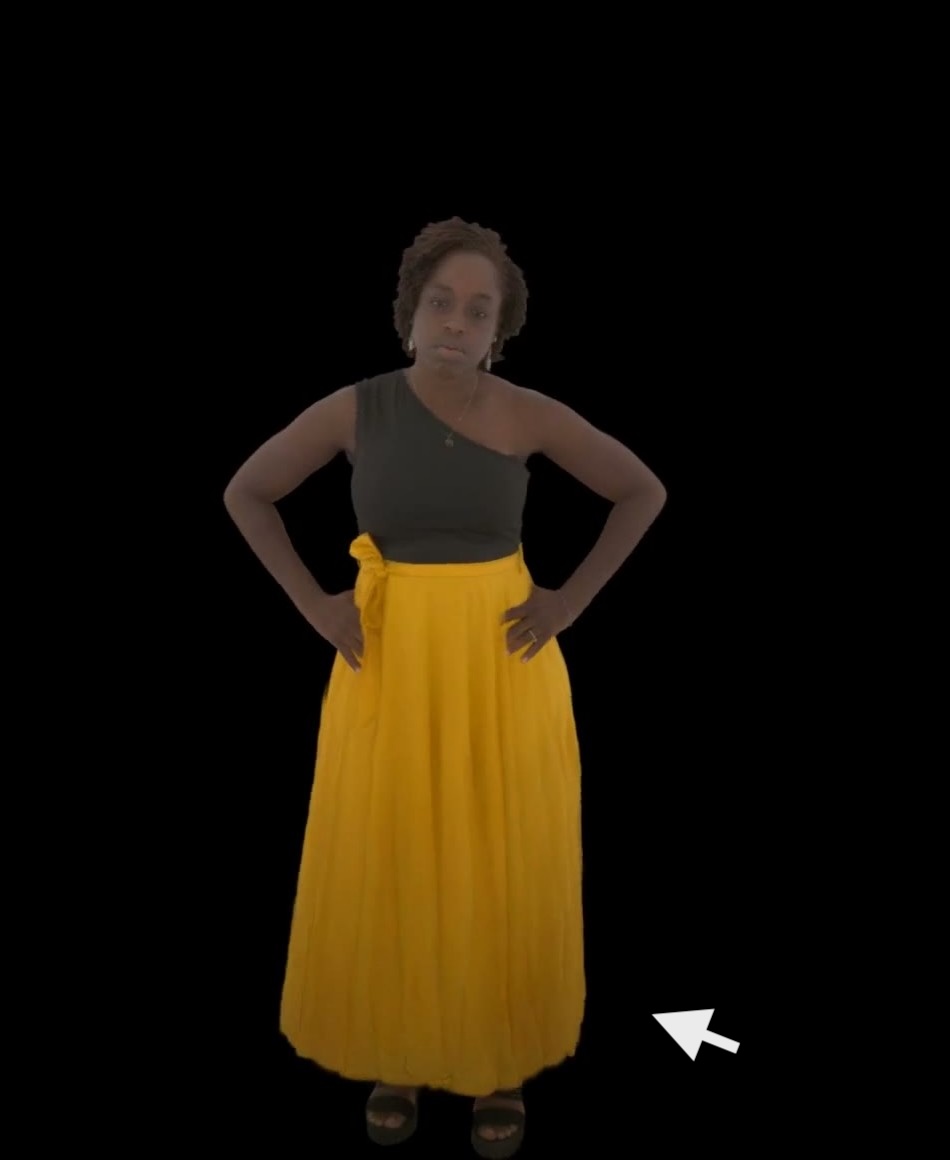} \\
        \raisebox{10ex}{\rowlab{Transformer-based\\(Ours)}} &
        \includegraphics[width=\imgw]{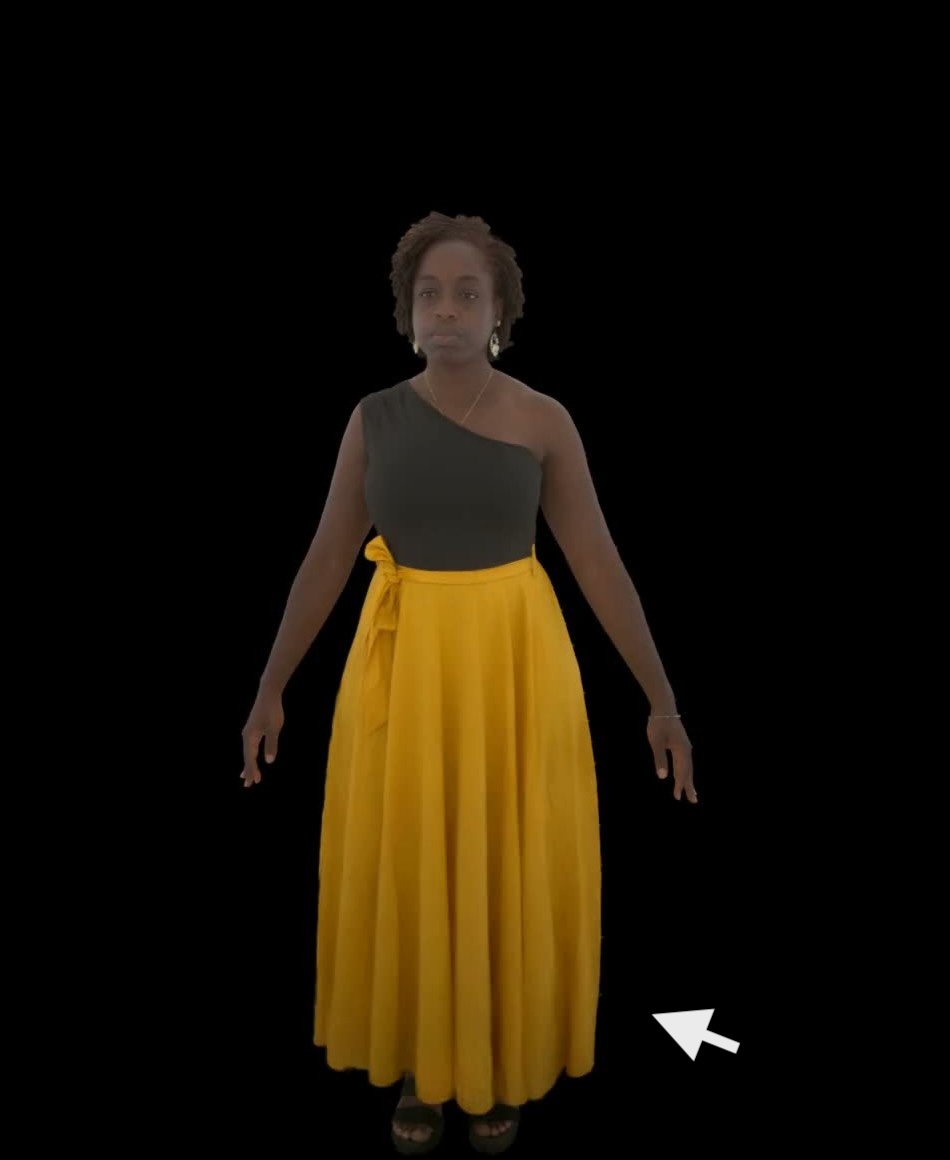} &
        \includegraphics[width=\imgw]{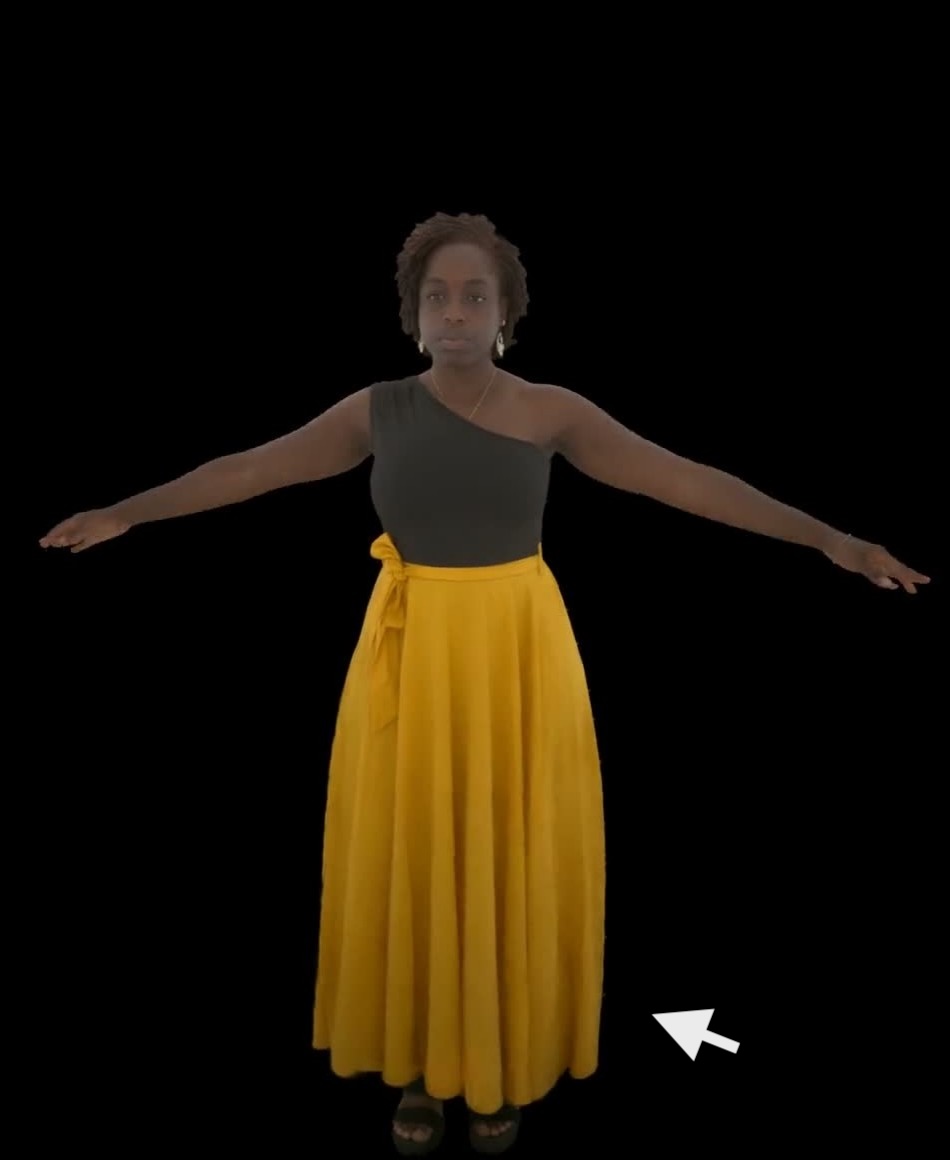} &
        \includegraphics[width=\imgw]{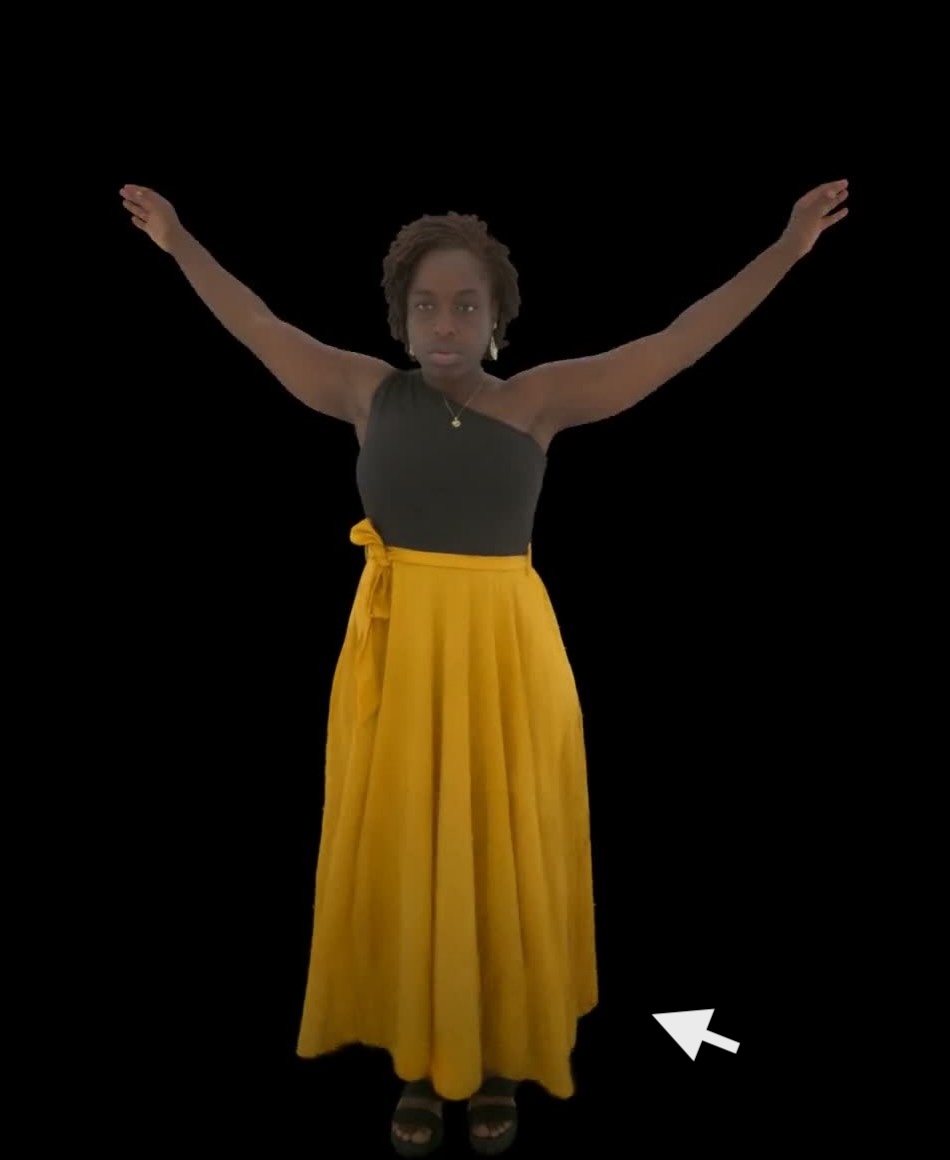} &
        \includegraphics[width=\imgw]{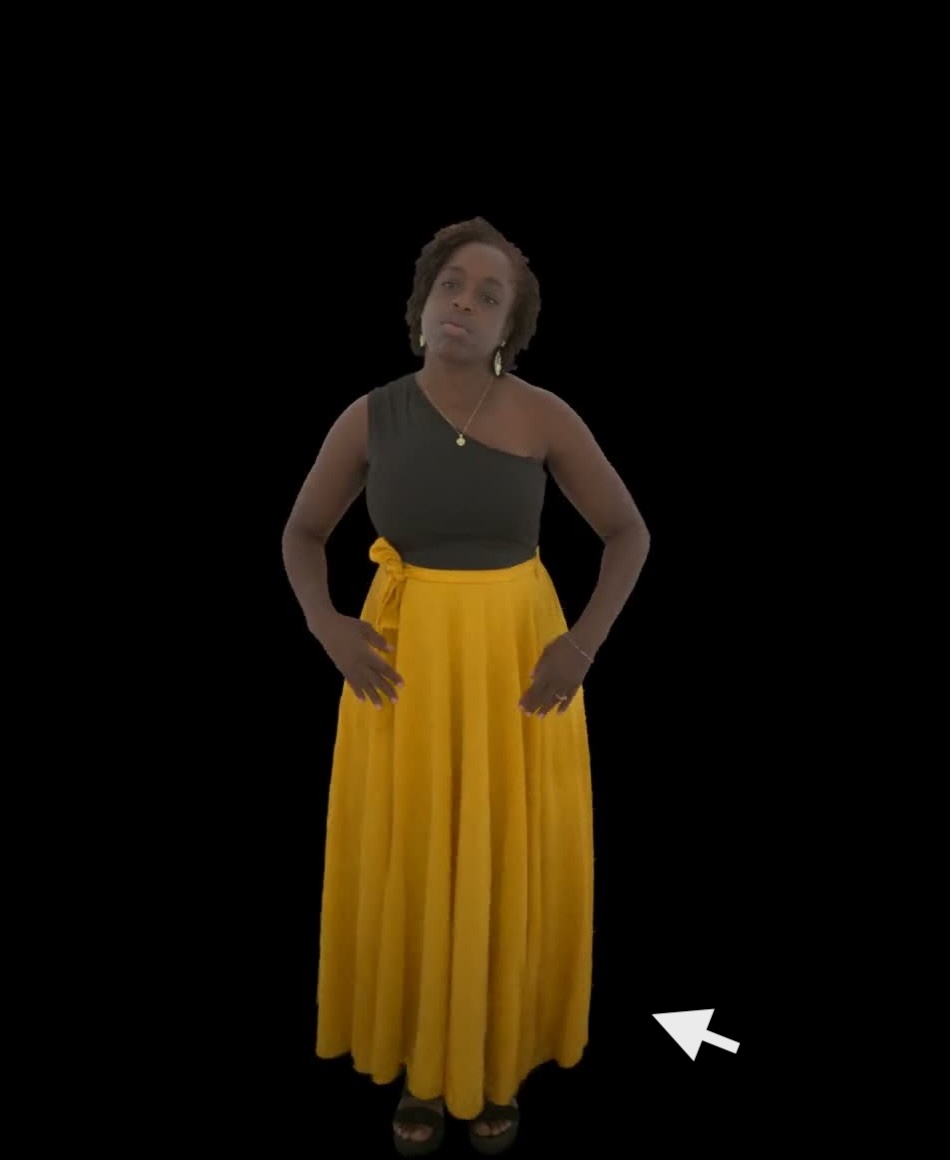} &
        \includegraphics[width=\imgw]{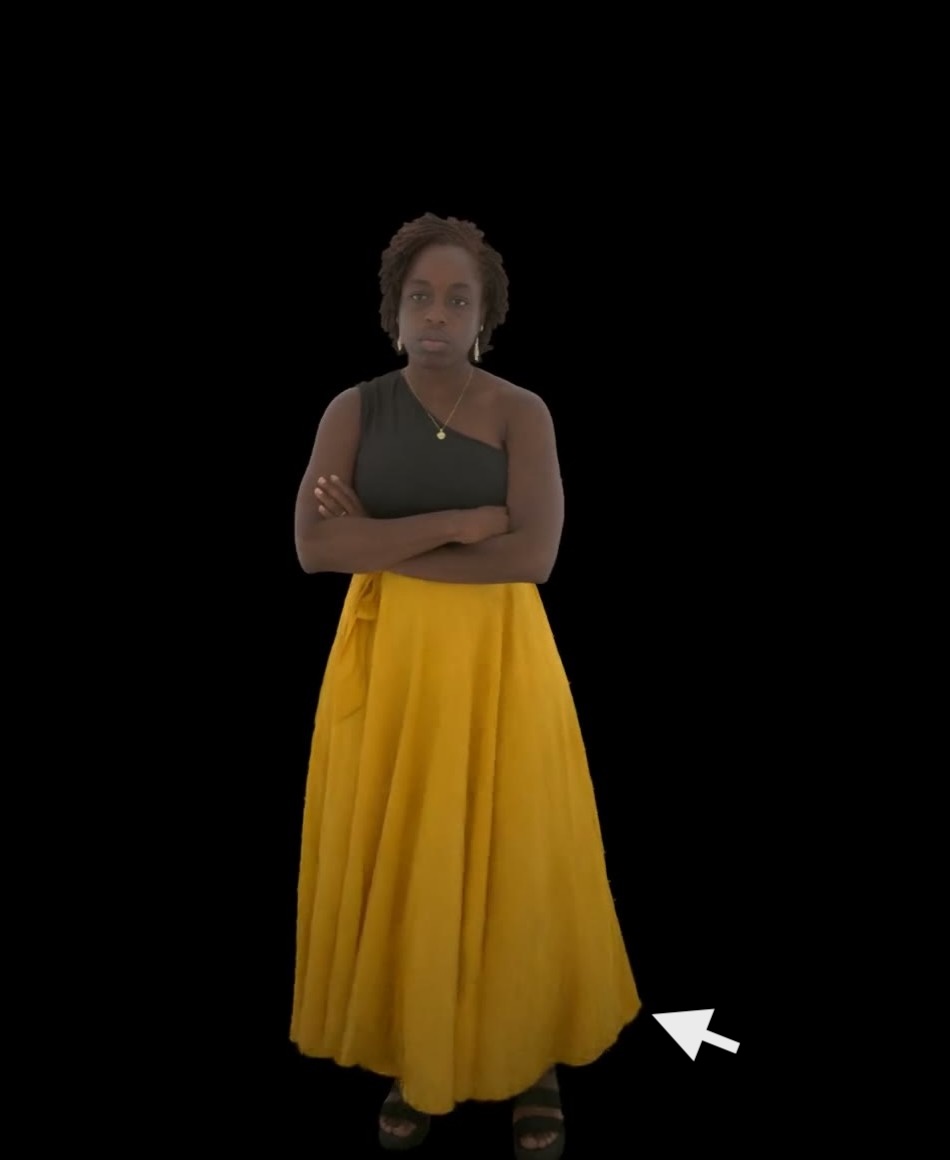} &
        \includegraphics[width=\imgw]{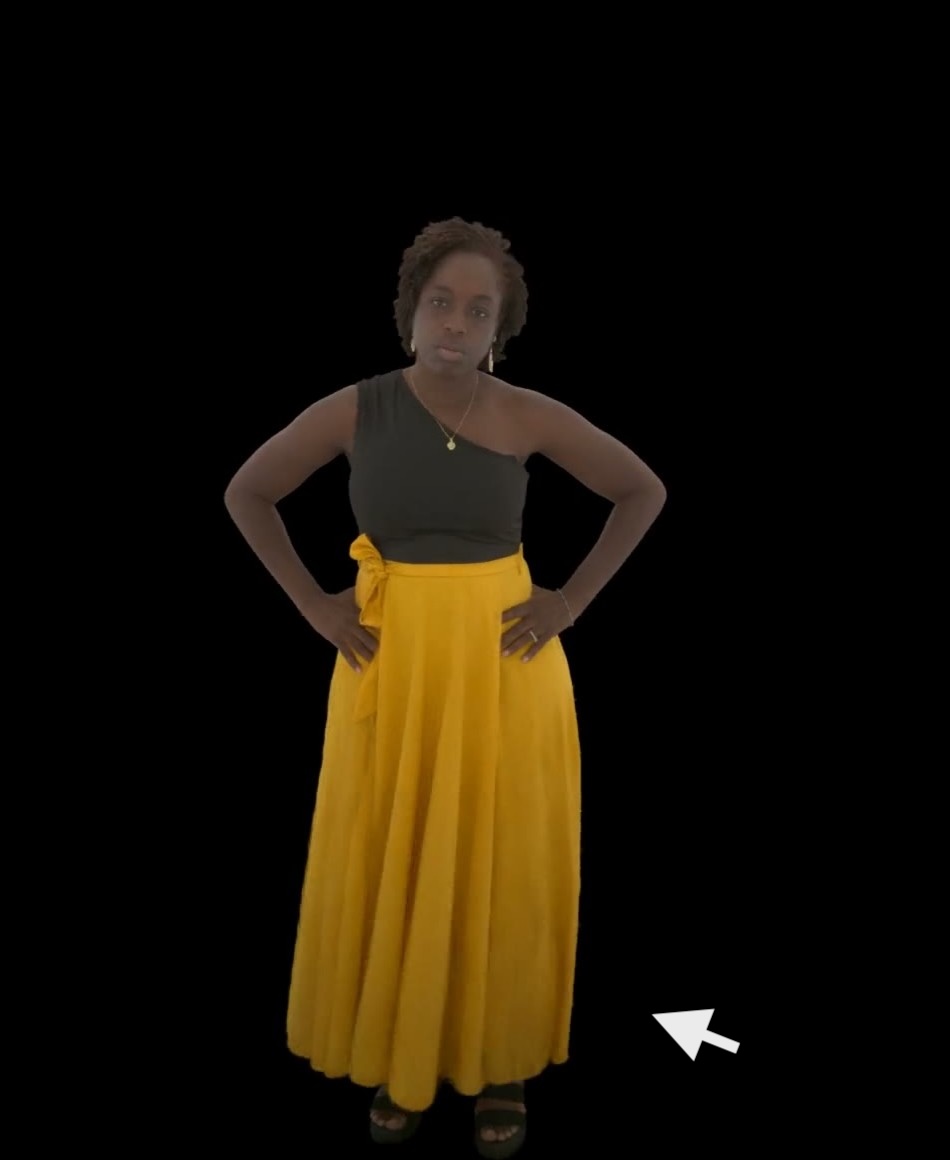} \\
        \raisebox{10ex}{\rowlab{Reference Image}} &
        \includegraphics[width=\imgw]{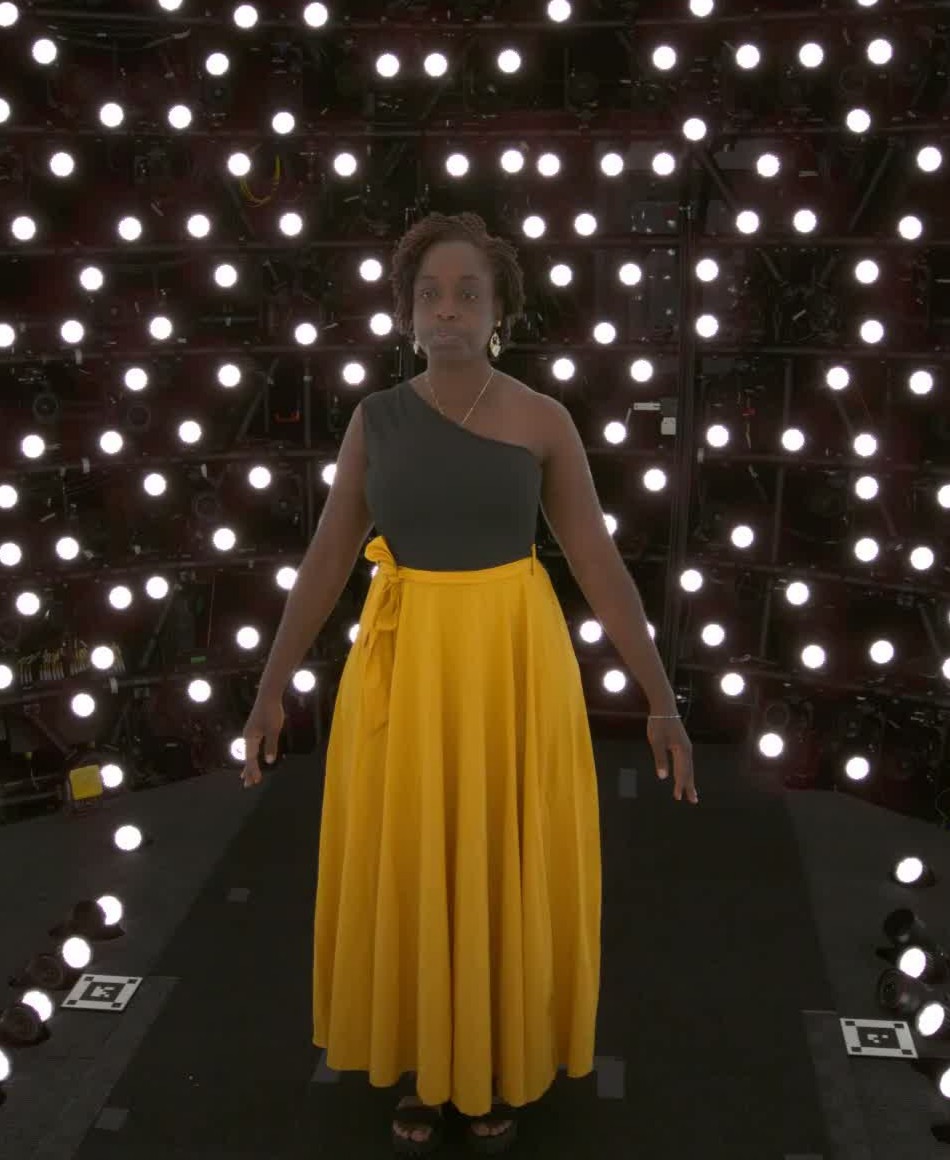} &
        \includegraphics[width=\imgw]{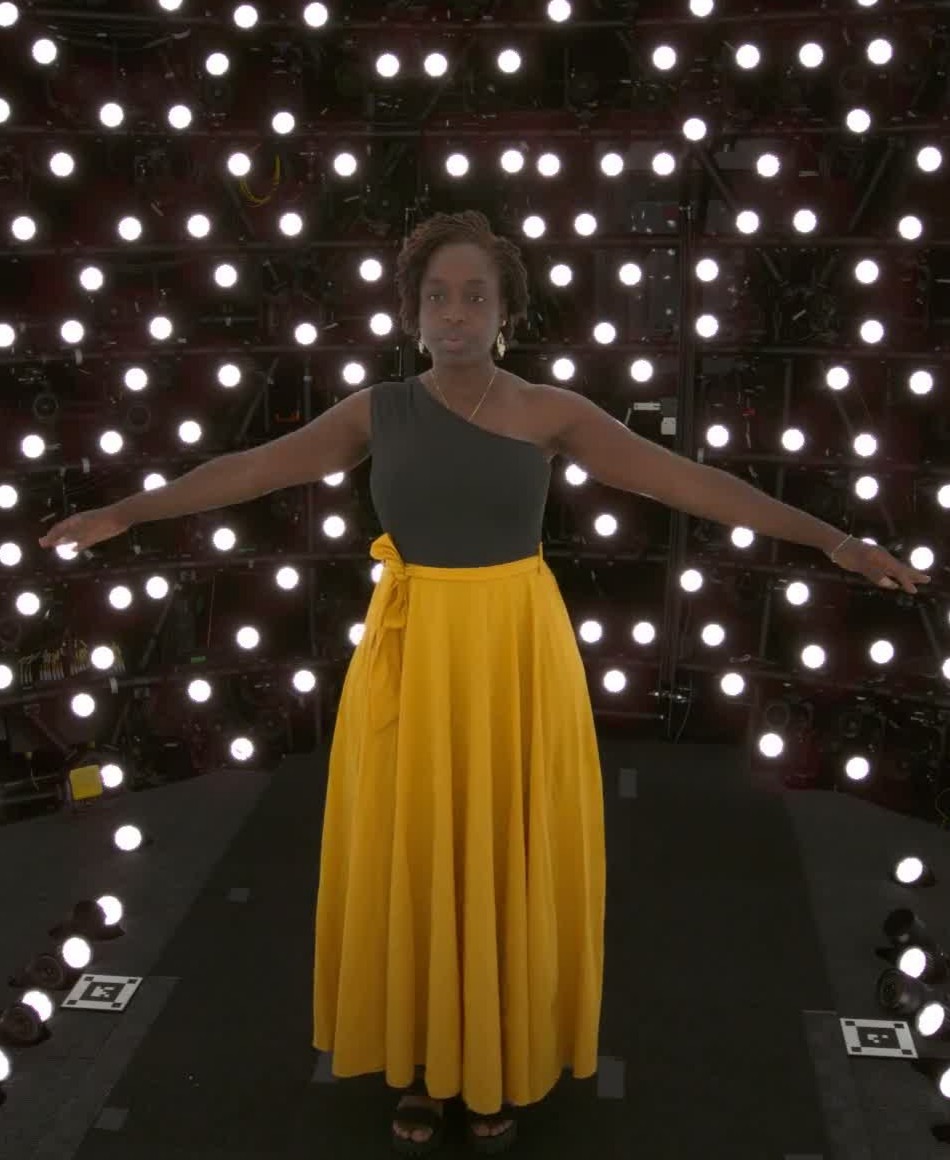} &
        \includegraphics[width=\imgw]{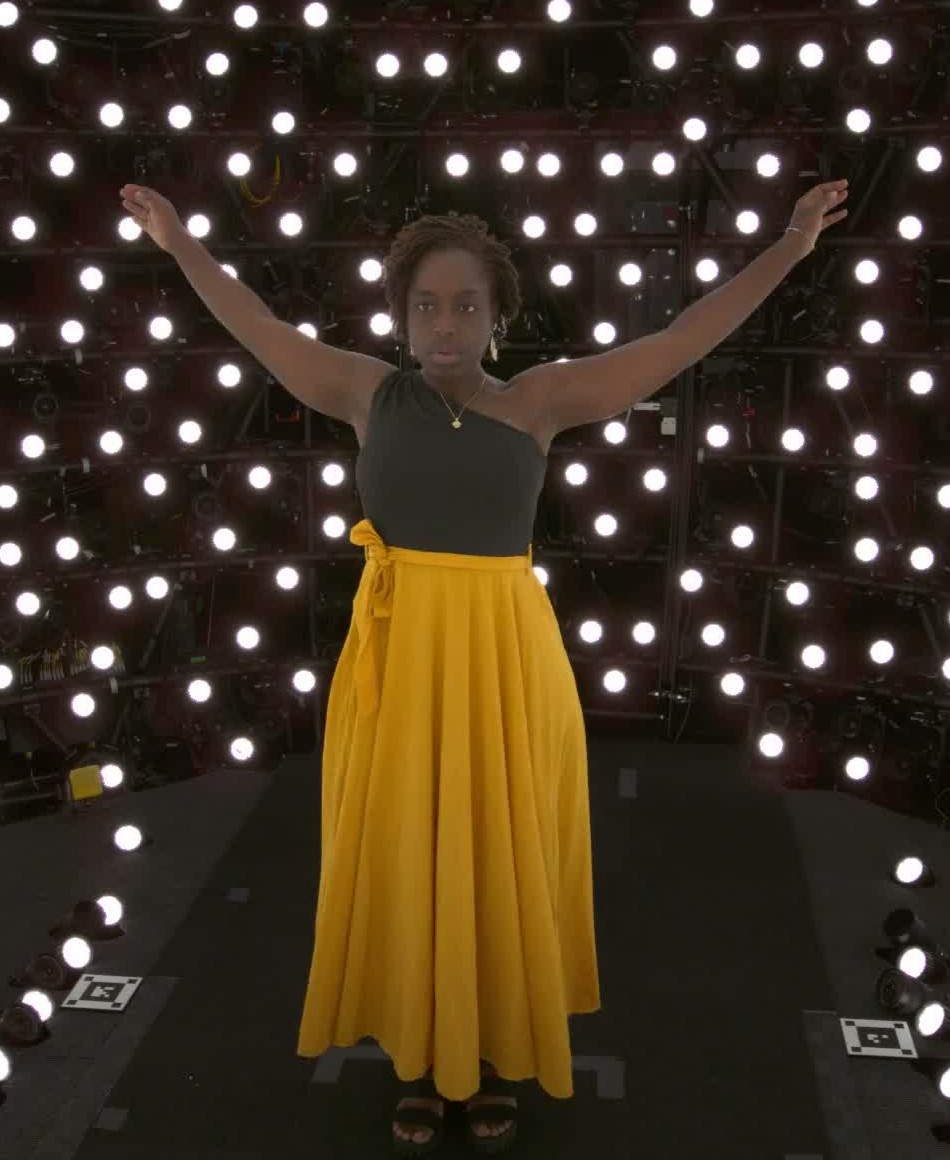} &
        \includegraphics[width=\imgw]{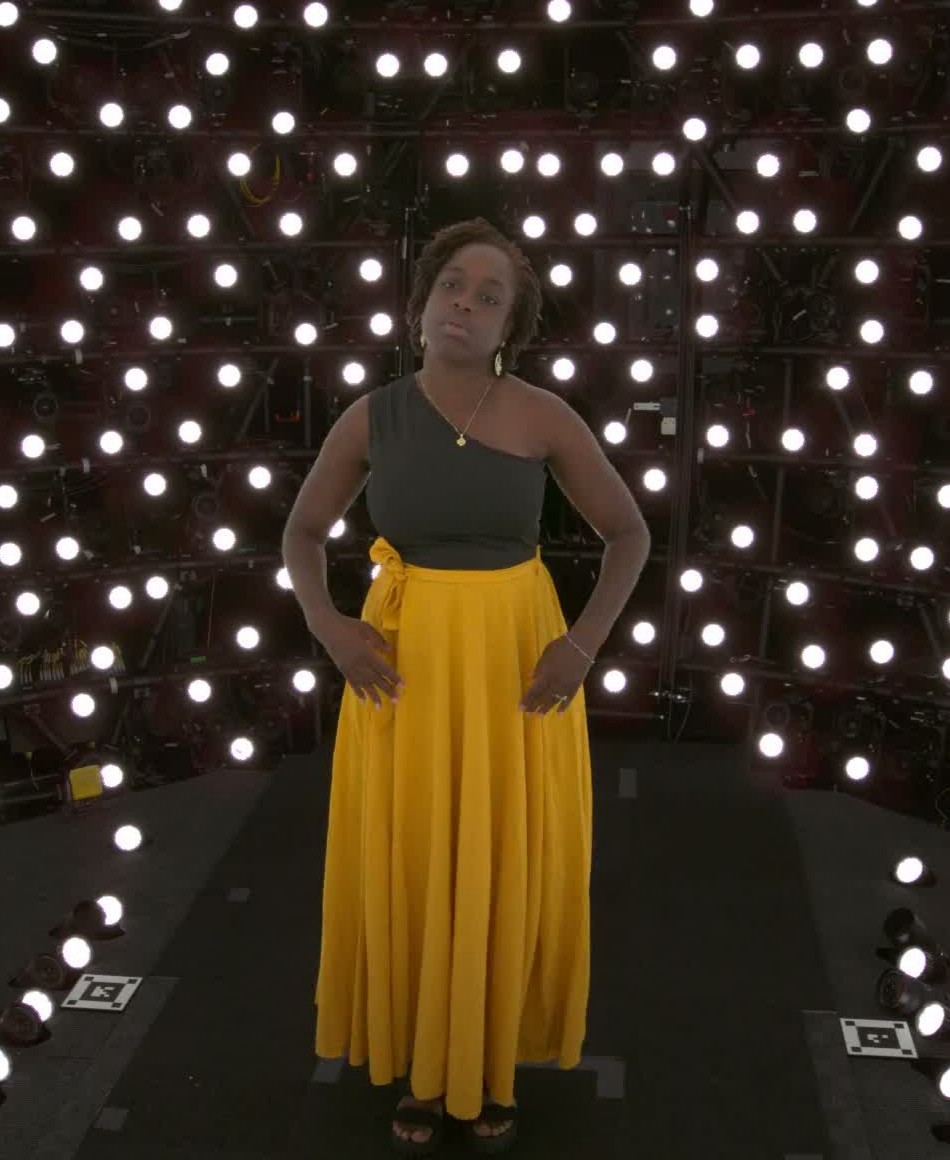} &
        \includegraphics[width=\imgw]{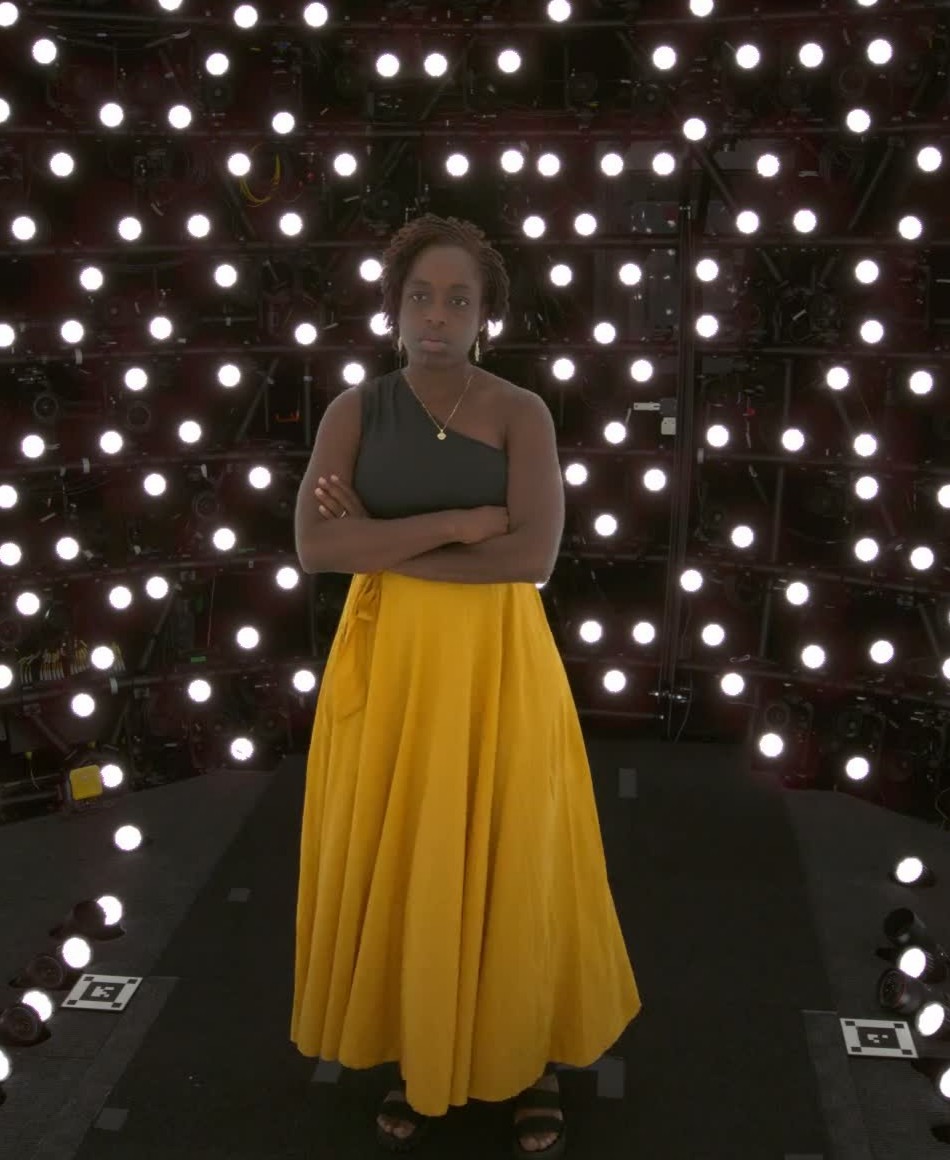} &
        \includegraphics[width=\imgw]{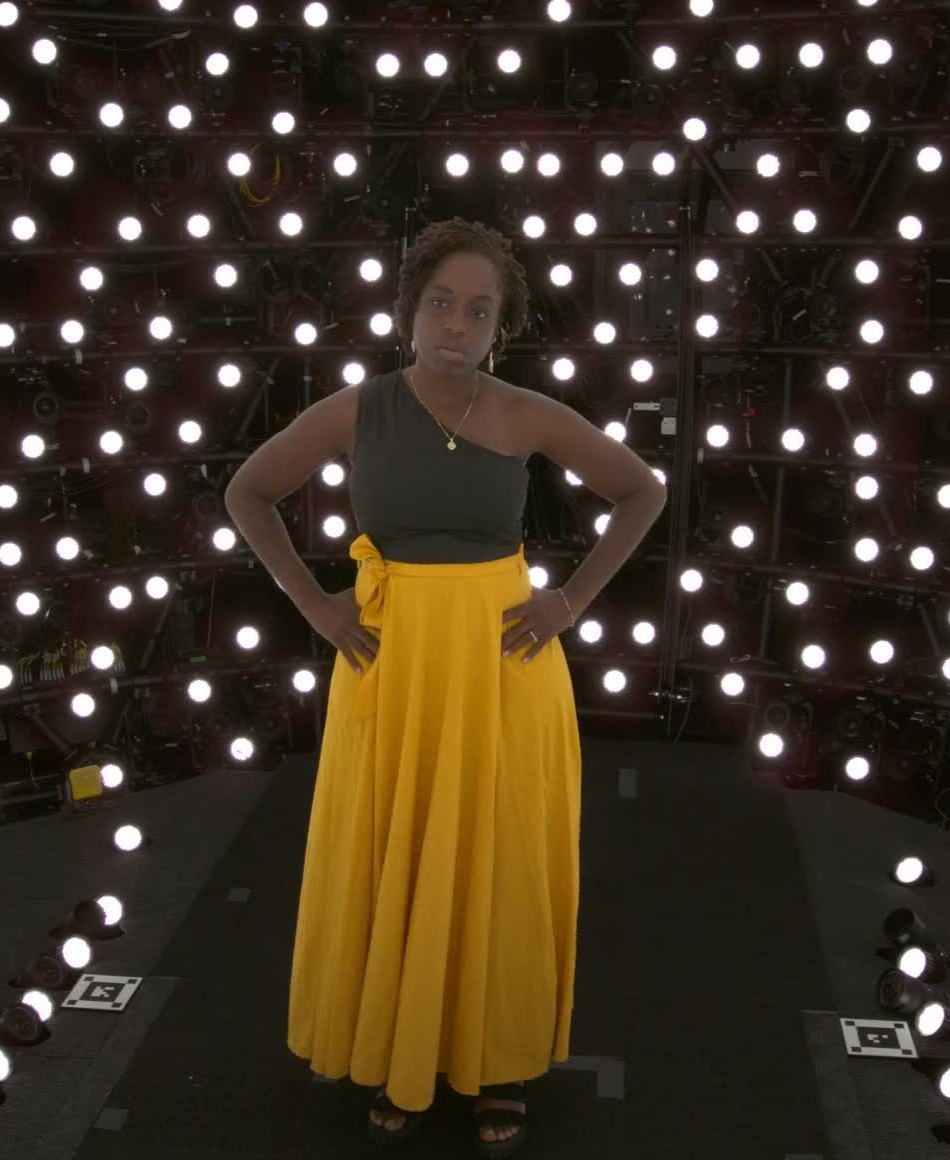} \\
    \end{tabular}
    \caption{\textbf{Decoder architecture ablation.} Compared to a convolution-based UV decoder~\cite{Bagautdinov2021DrivingsignalAF}, our transformer-based decoder reconstructs finer wrinkles and folds with richer details. See the supplementary video for a clearer temporal comparison.}
    \label{fig:ablation_decoder_arch}
\end{figure}

Figure~\ref{fig:ablation_decoder_arch} shows qualitative novel-view renderings on training sequences. Our transformer-based decoder better reconstructs fine wrinkles and folds and preserves higher-frequency garment details, while the convolutional alternative tends to produce smoother and less detailed cloth appearance. The difference is more evident in motion; please refer to the supplementary video for side-by-side comparisons.

\subsection{Latent Dynamics Model Design}
\label{app:ablation_dynamics}

Our latent dynamics model predicts latent acceleration while incorporating a spring-damper system (Sec.~3.3 in the main paper). To study the impact of this design, we compare against three alternatives:
\begin{itemize}
    \item \textbf{(i) Direct latent prediction.} Predict current step latent $\mathbf{z}_t$ directly.
    \item \textbf{(ii) Velocity prediction.} Predict latent velocity $\mathbf{v}_t$ (instead of acceleration) and update the state accordingly.
    \item \textbf{(iii) Acceleration without spring-damper.} Predict acceleration $\mathbf{a}_t$ but remove the spring-damper components from the update.
\end{itemize}

\begin{figure}[t]
    \centering
    \small
    \setlength{\tabcolsep}{0pt}
    \newcommand{\rowlab}[1]{\begin{tabular}[c]{@{}l@{}}#1\end{tabular}}
    \newcommand{\imgw}{\dimexpr(\linewidth-0.14\linewidth-24\tabcolsep)/11\relax}
    \begin{tabular}{@{}p{0.14\linewidth}@{}*{11}{c}@{}}
        \raisebox{10.0ex}{\rowlab{(i) Predict $\mathbf{z}_t$}} &
        \includegraphics[width=\imgw]{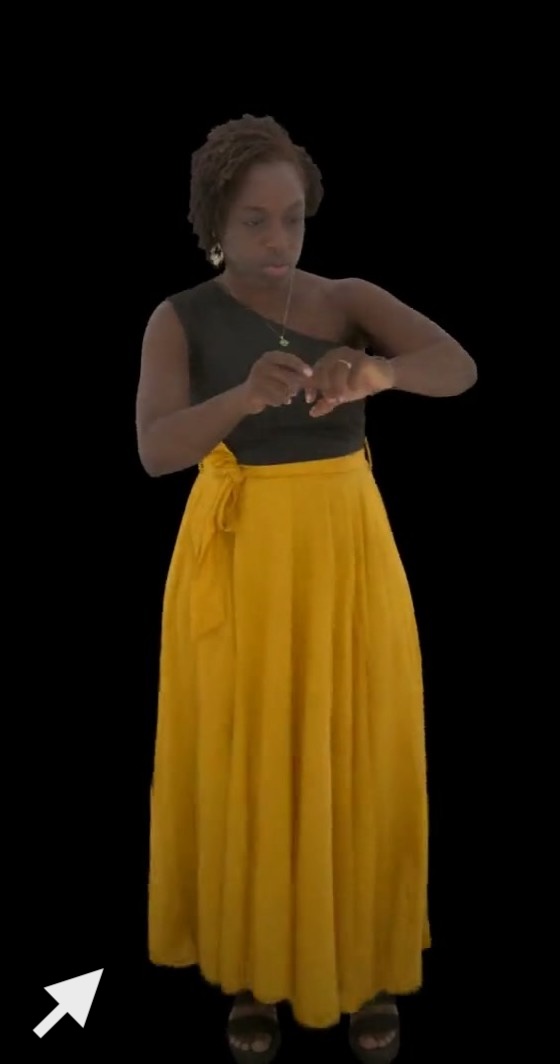} &
        \includegraphics[width=\imgw]{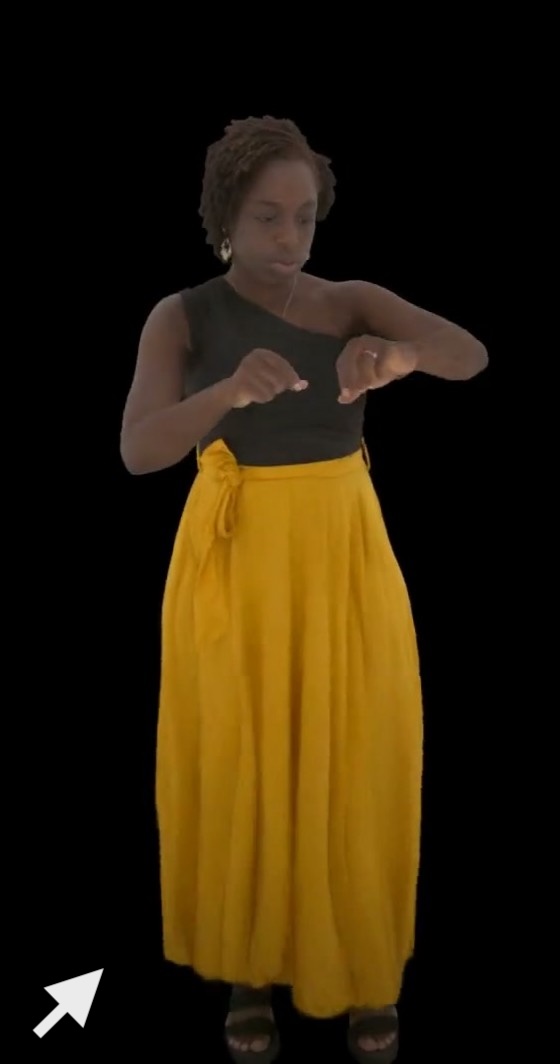} &
        \includegraphics[width=\imgw]{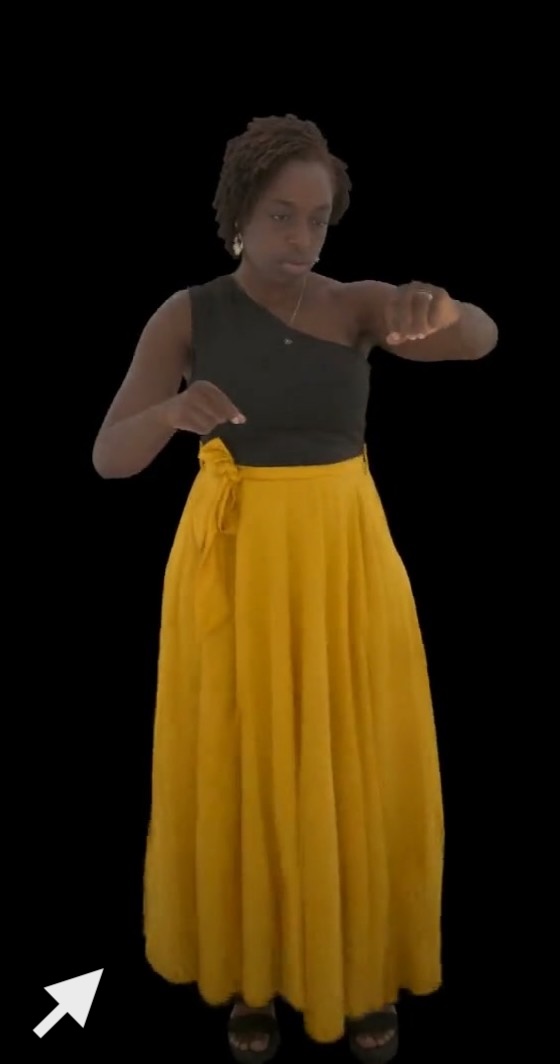} &
        \includegraphics[width=\imgw]{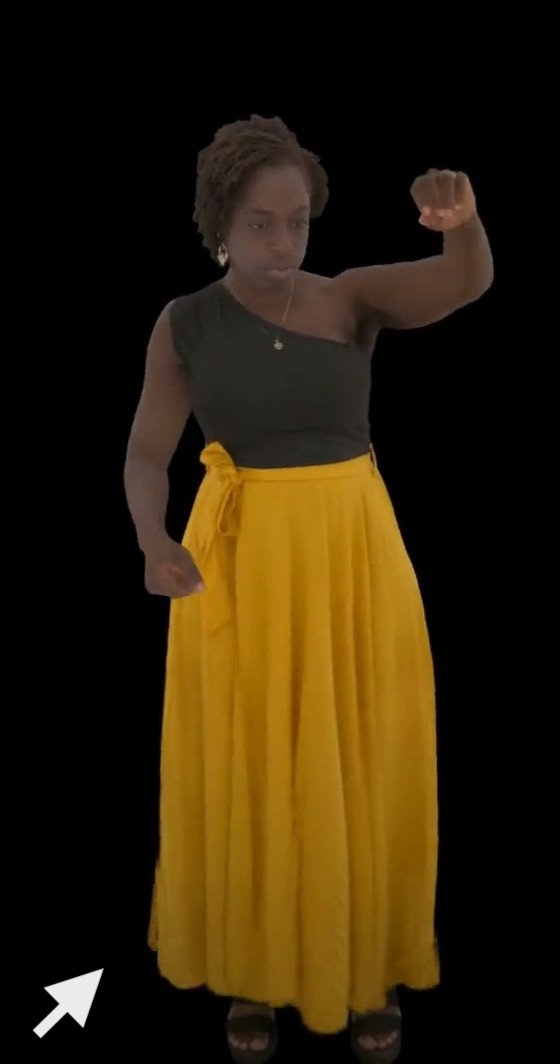} &
        \includegraphics[width=\imgw]{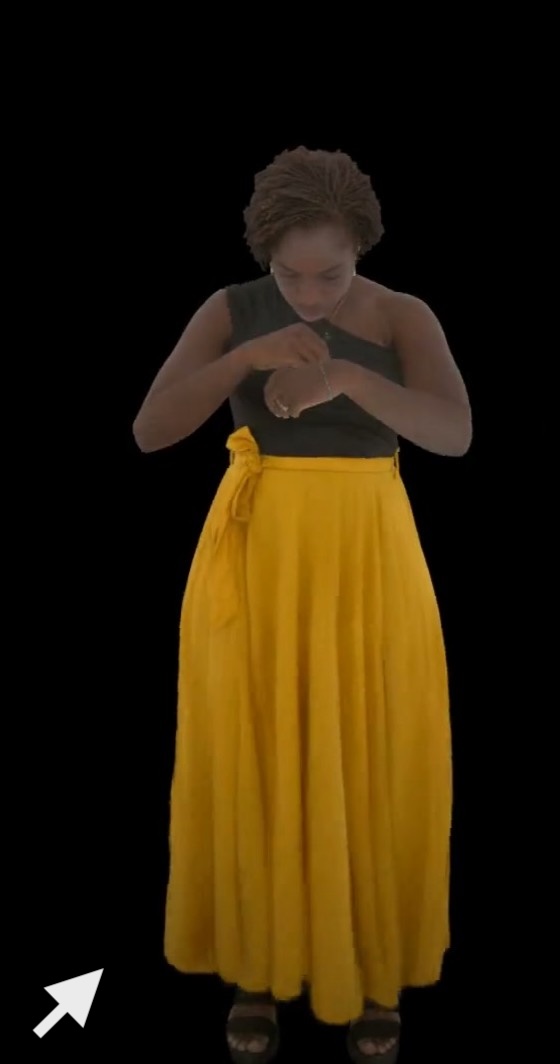} &
        \includegraphics[width=\imgw]{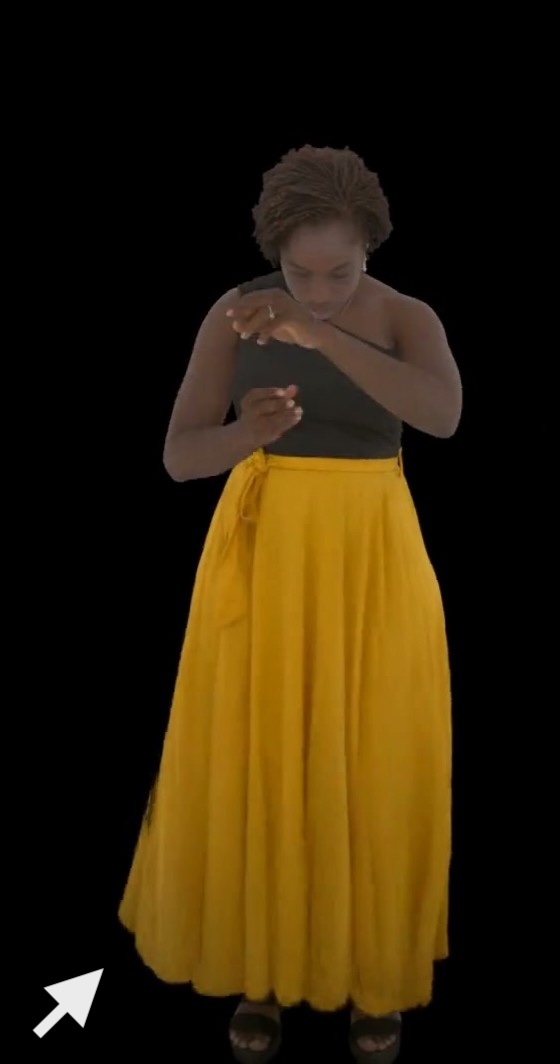} &
        \includegraphics[width=\imgw]{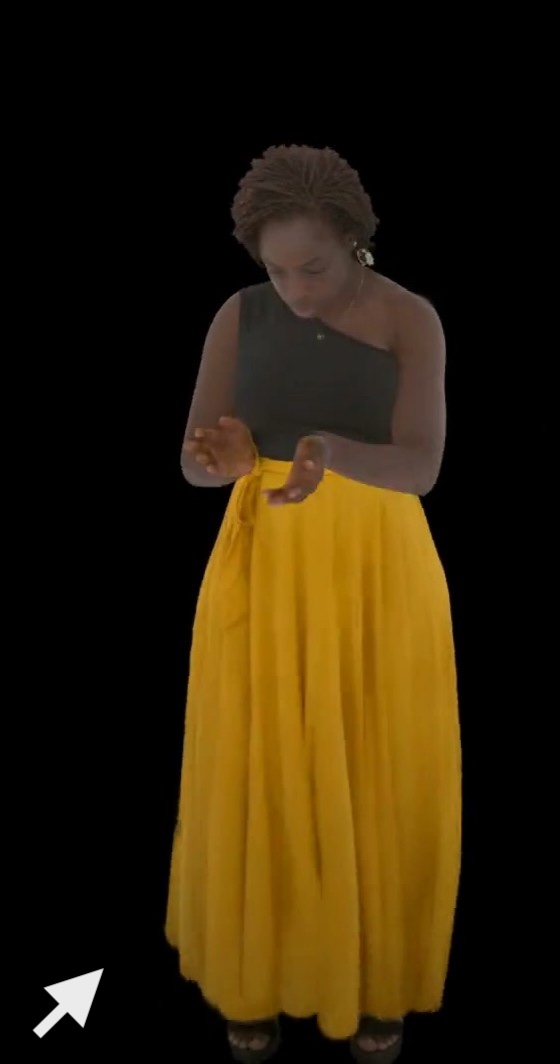} &
        \includegraphics[width=\imgw]{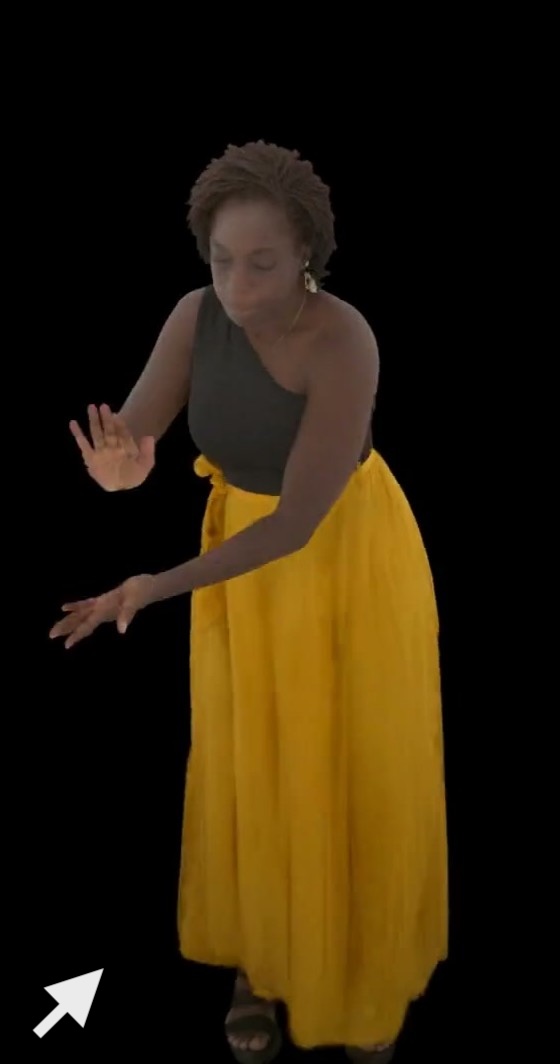} &
        \includegraphics[width=\imgw]{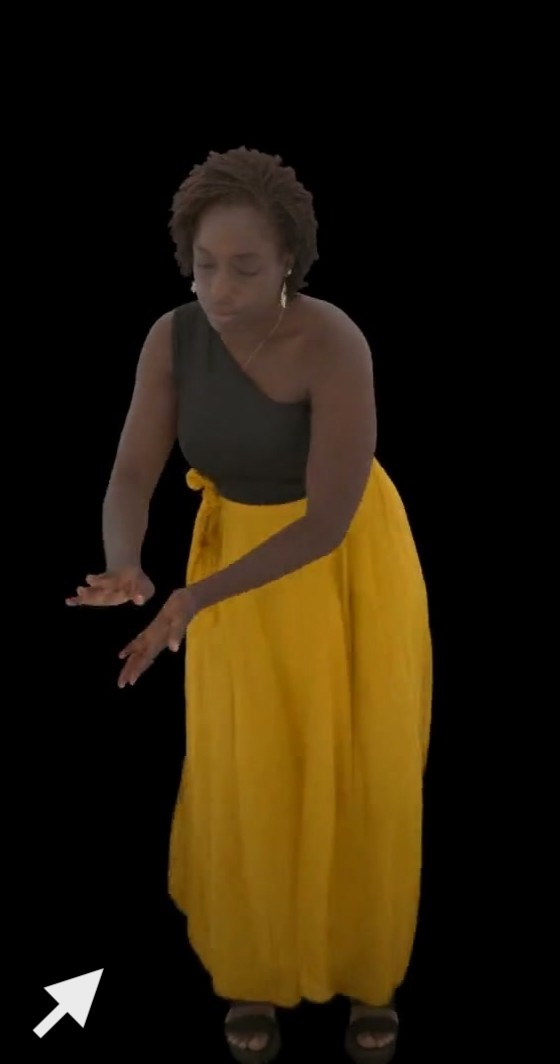} &
        \includegraphics[width=\imgw]{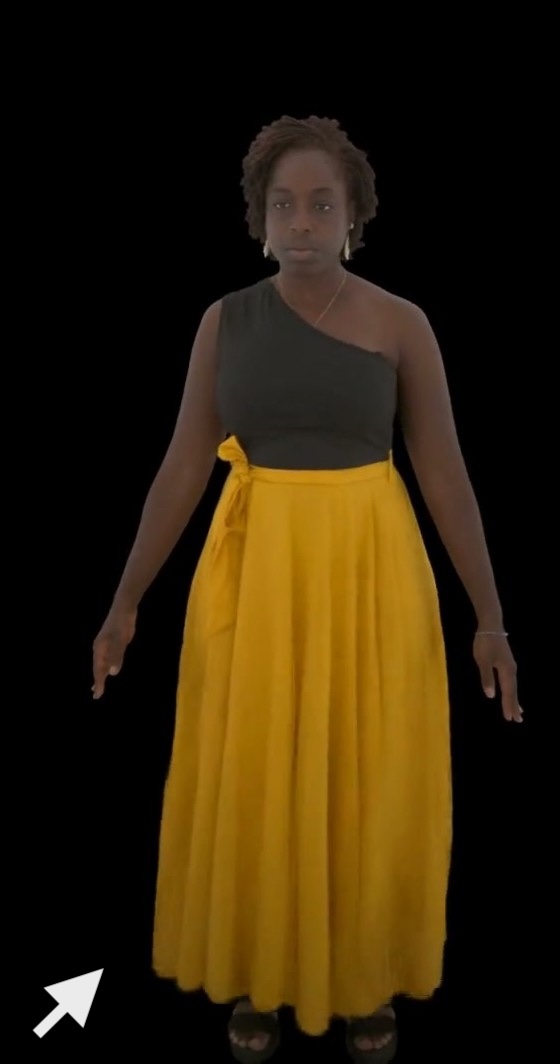} &
        \includegraphics[width=\imgw]{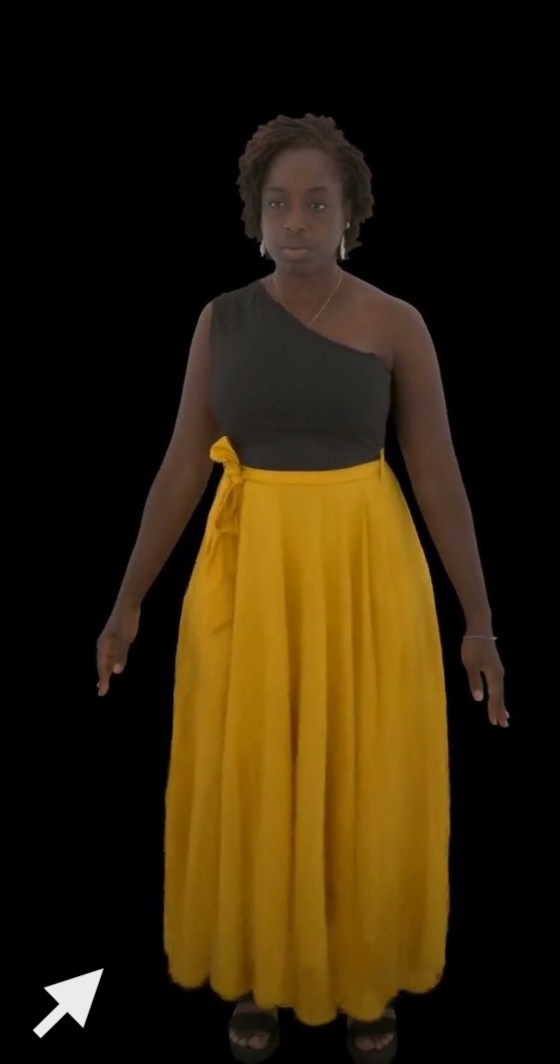} \\
        \raisebox{10.0ex}{\rowlab{(ii) Predict $\mathbf{v}_t$}} &
        \includegraphics[width=\imgw]{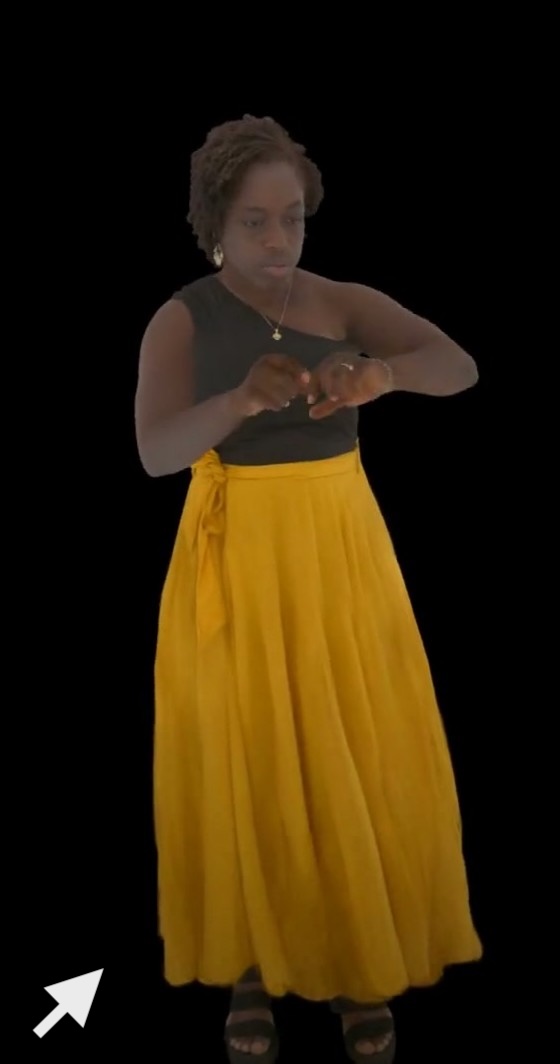} &
        \includegraphics[width=\imgw]{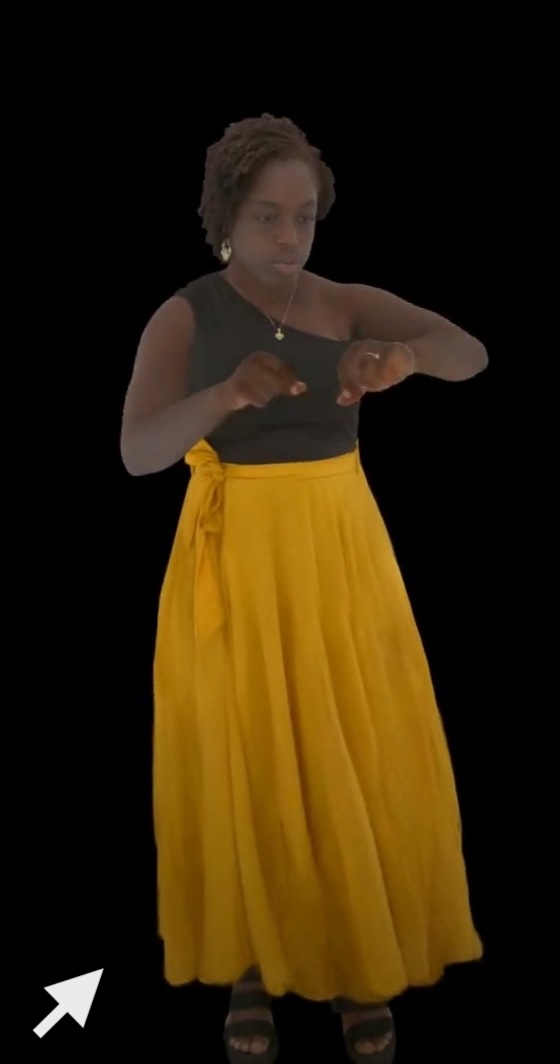} &
        \includegraphics[width=\imgw]{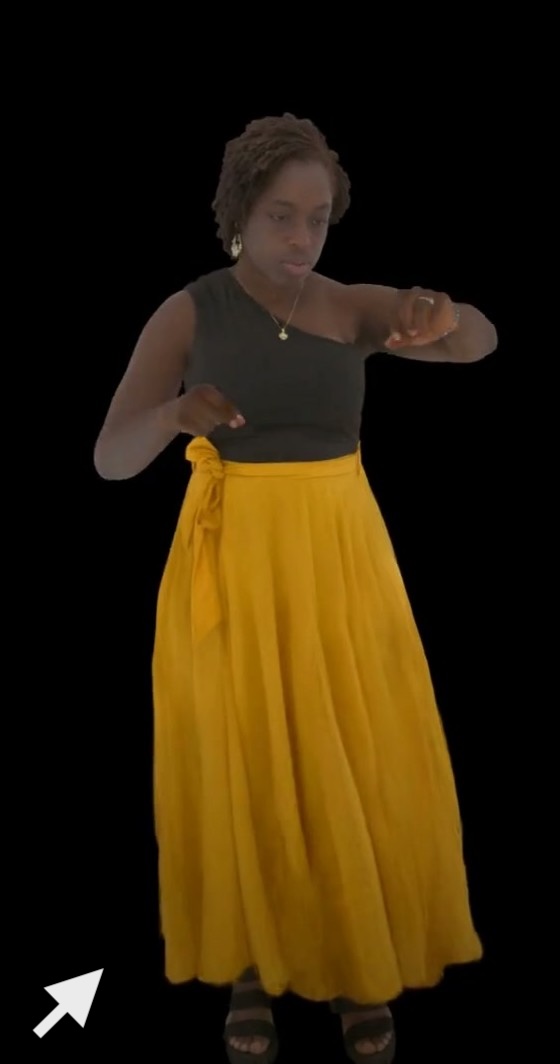} &
        \includegraphics[width=\imgw]{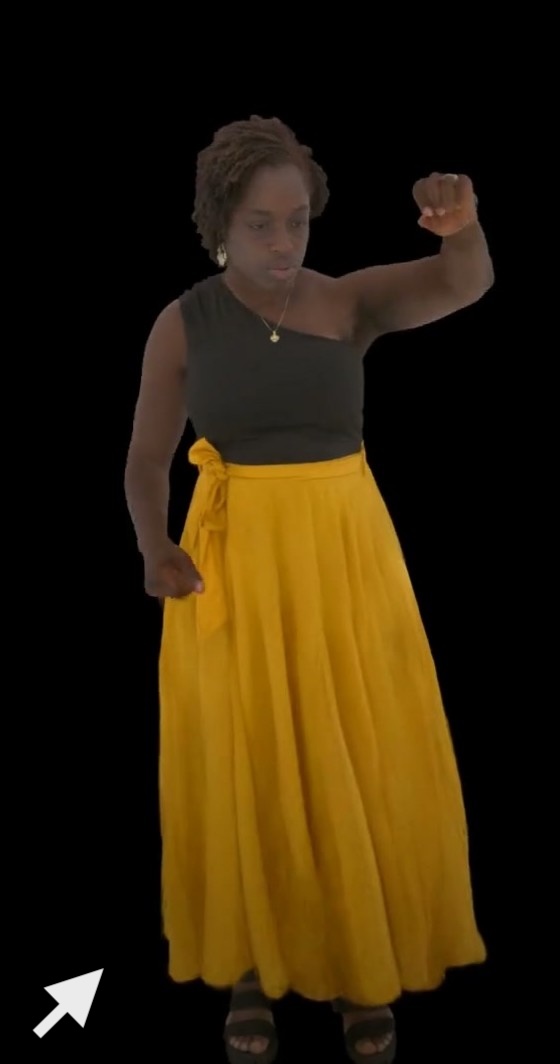} &
        \includegraphics[width=\imgw]{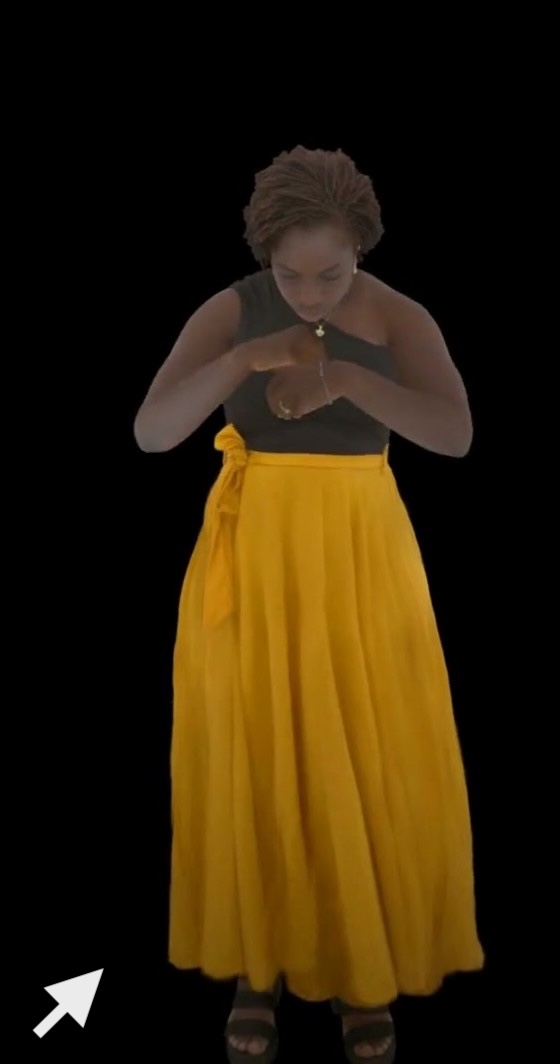} &
        \includegraphics[width=\imgw]{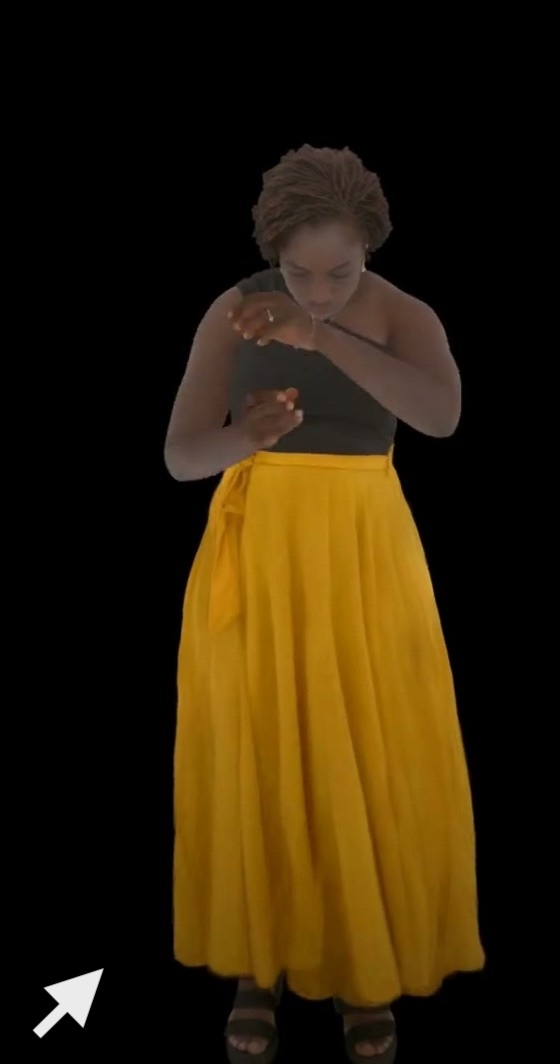} &
        \includegraphics[width=\imgw]{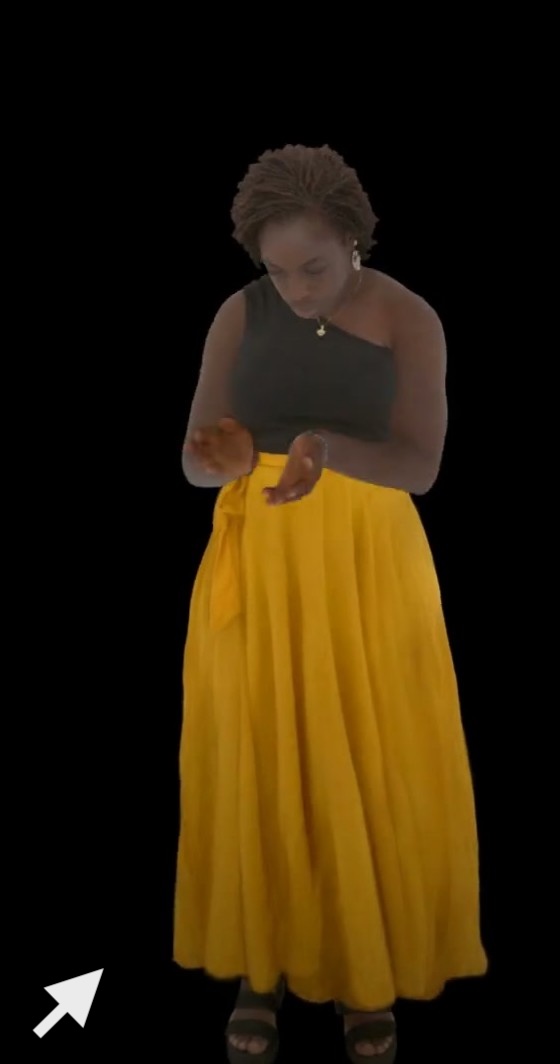} &
        \includegraphics[width=\imgw]{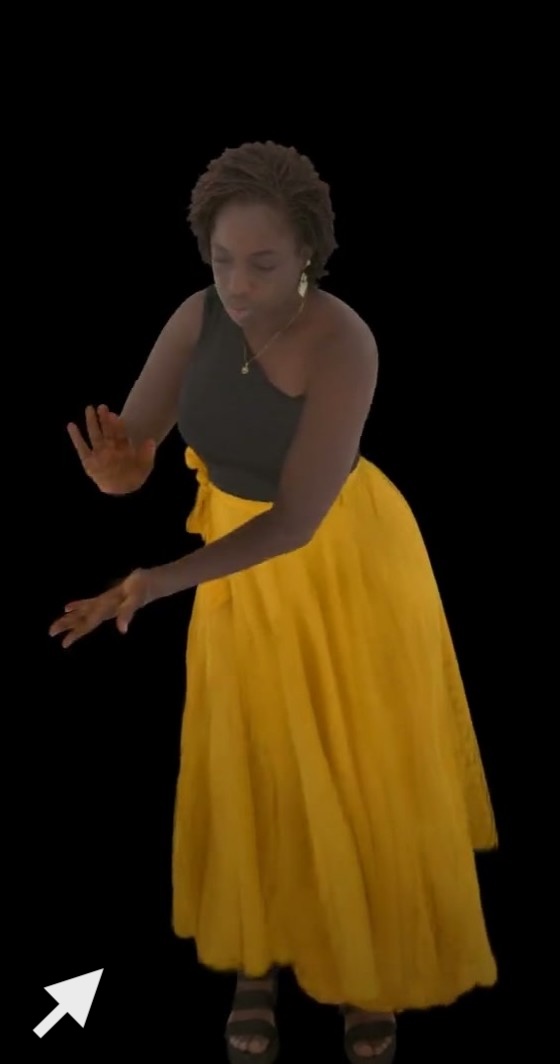} &
        \includegraphics[width=\imgw]{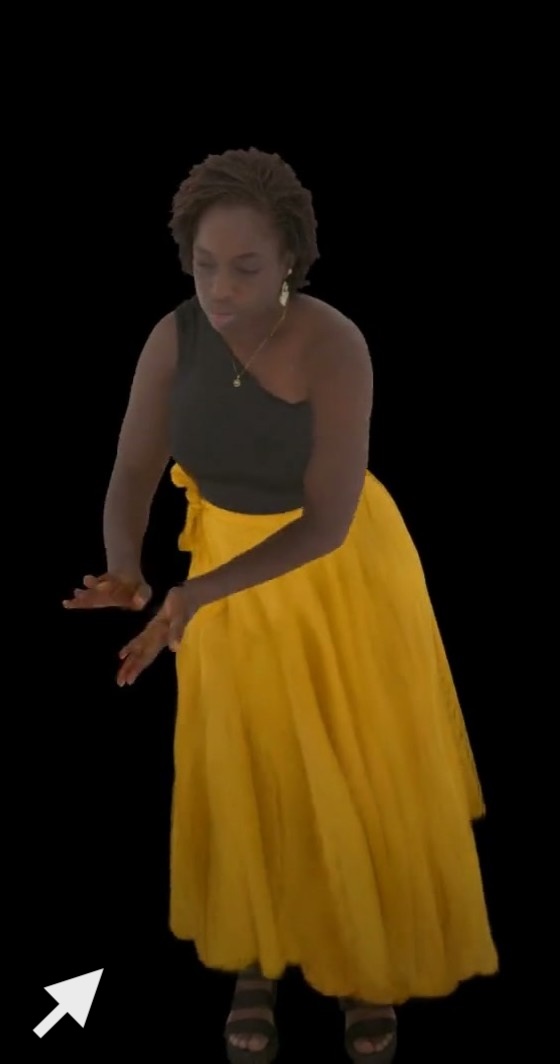} &
        \includegraphics[width=\imgw]{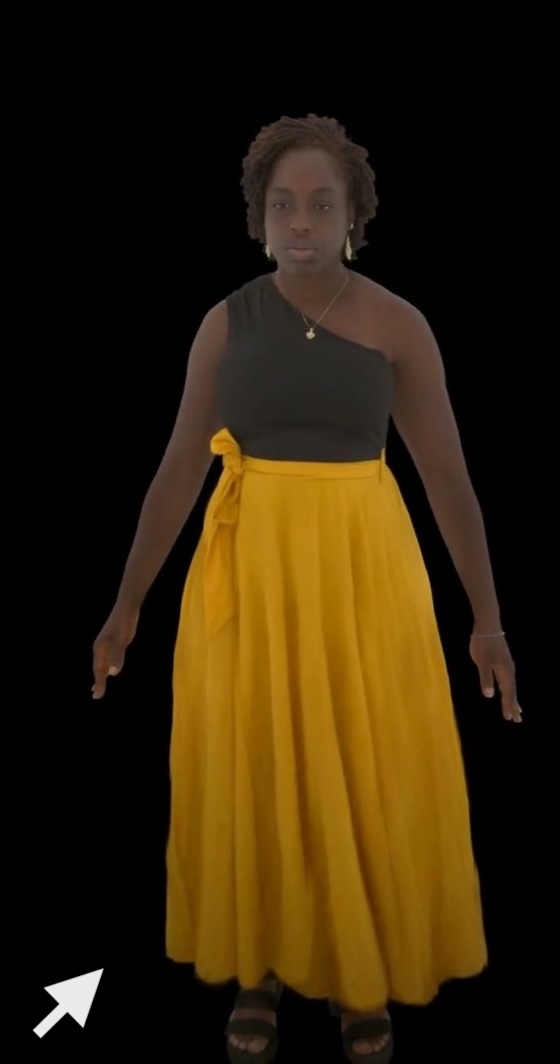} &
        \includegraphics[width=\imgw]{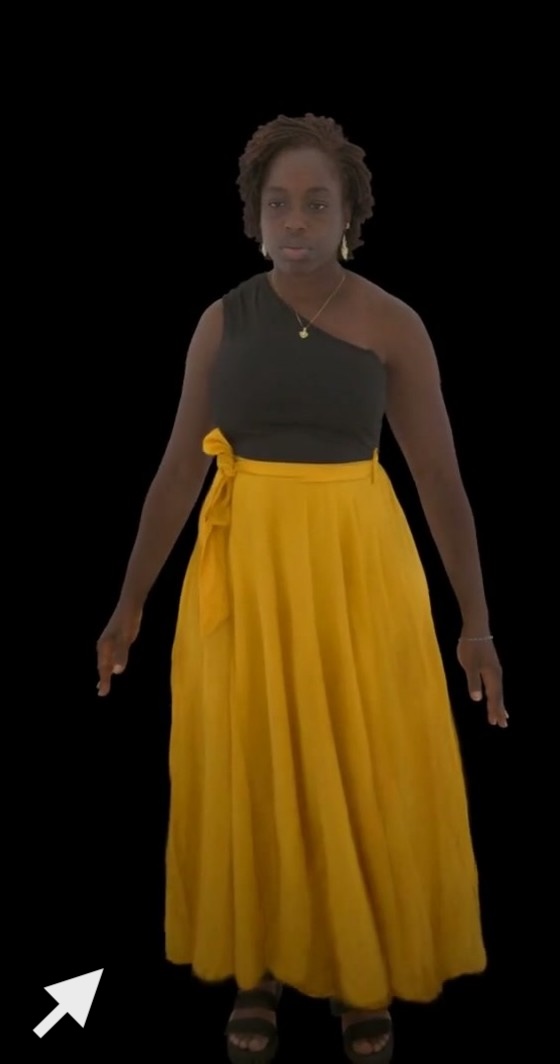} \\
        \raisebox{10.0ex}{\rowlab{(iii) Predict $\mathbf{a}_t$\\ w/o\\ Spring-damper}} &
        \includegraphics[width=\imgw]{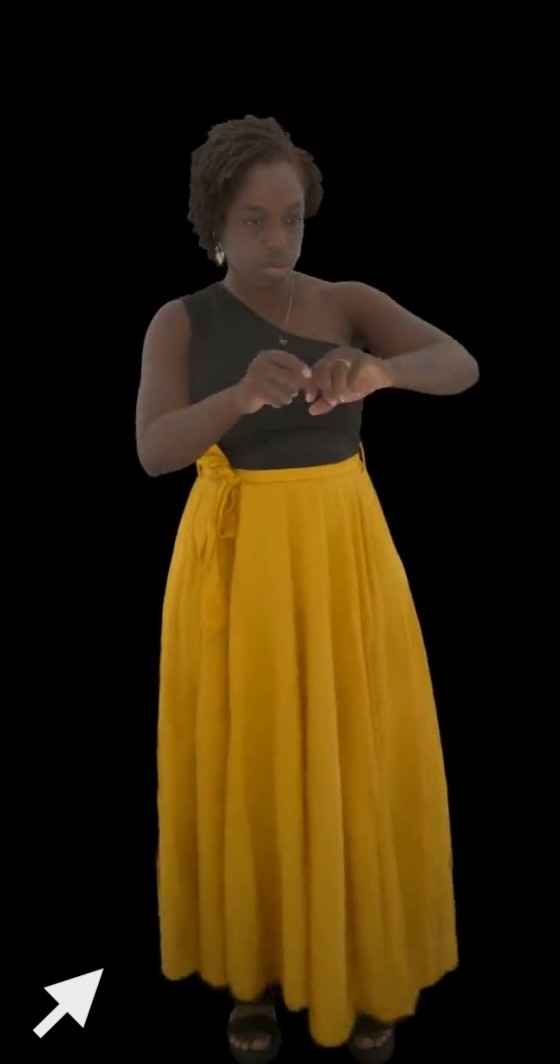} &
        \includegraphics[width=\imgw]{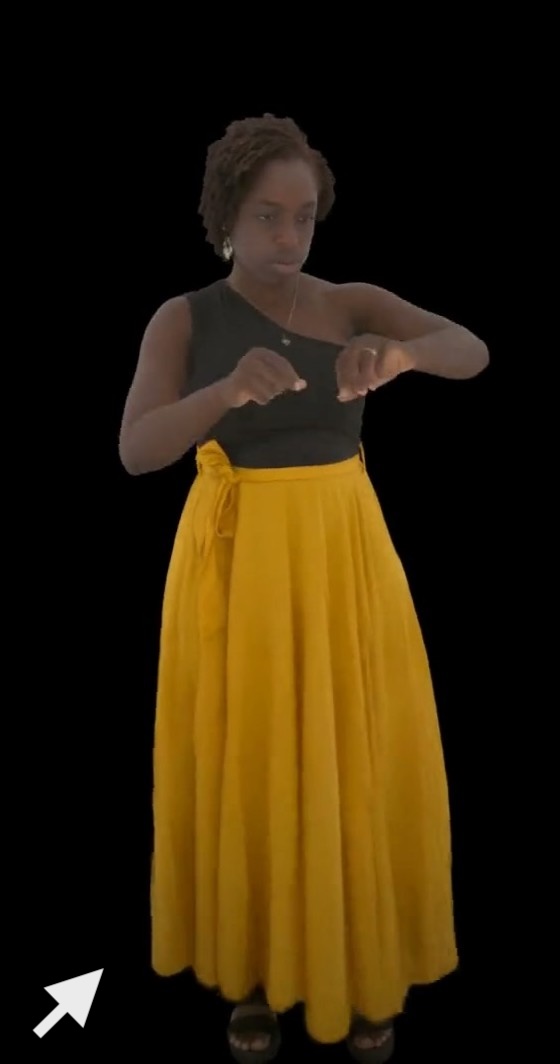} &
        \includegraphics[width=\imgw]{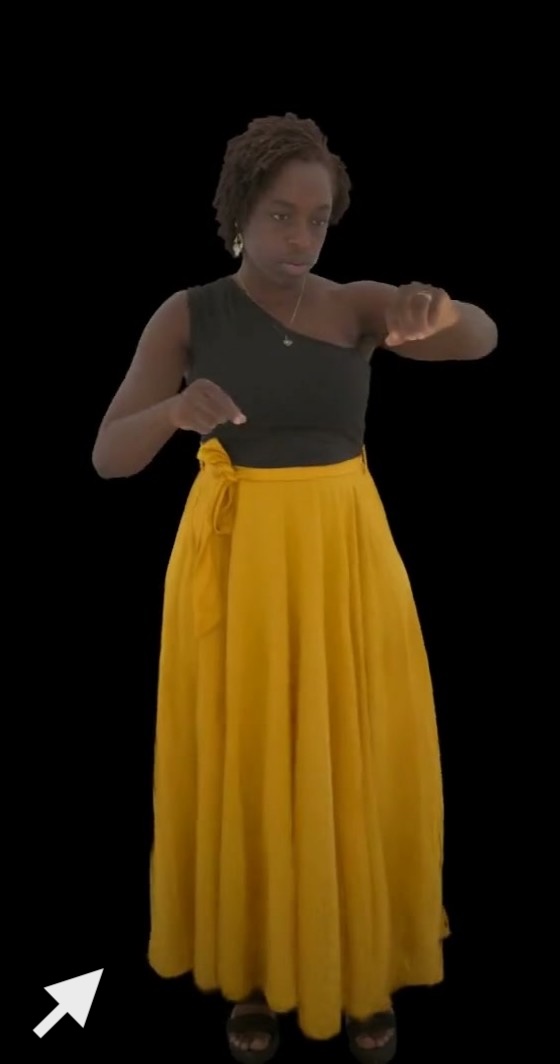} &
        \includegraphics[width=\imgw]{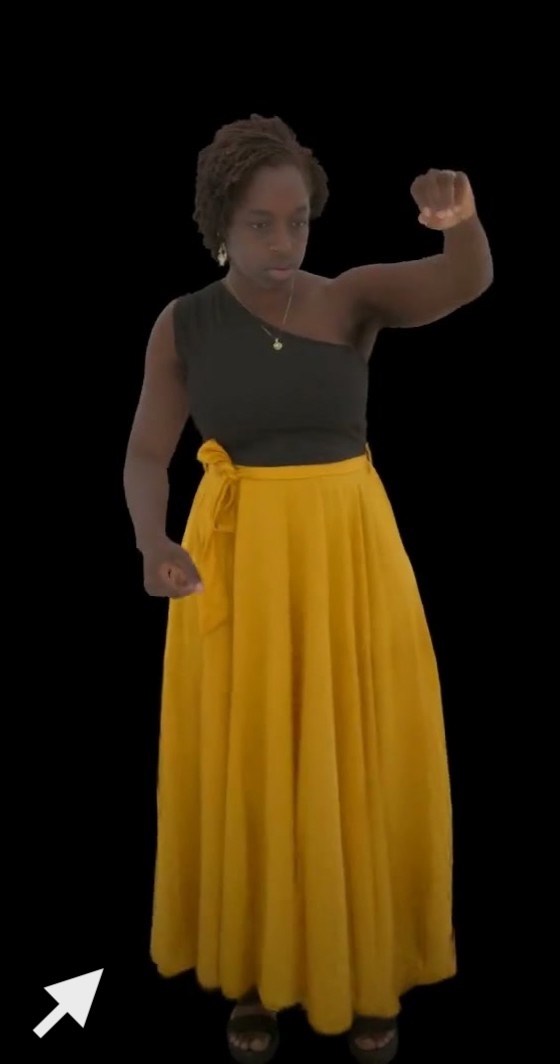} &
        \includegraphics[width=\imgw]{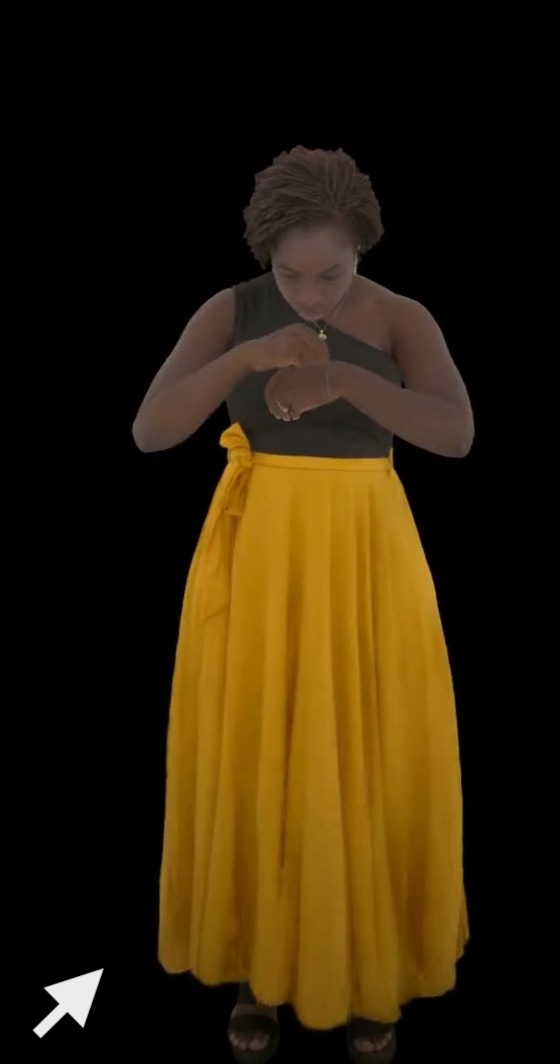} &
        \includegraphics[width=\imgw]{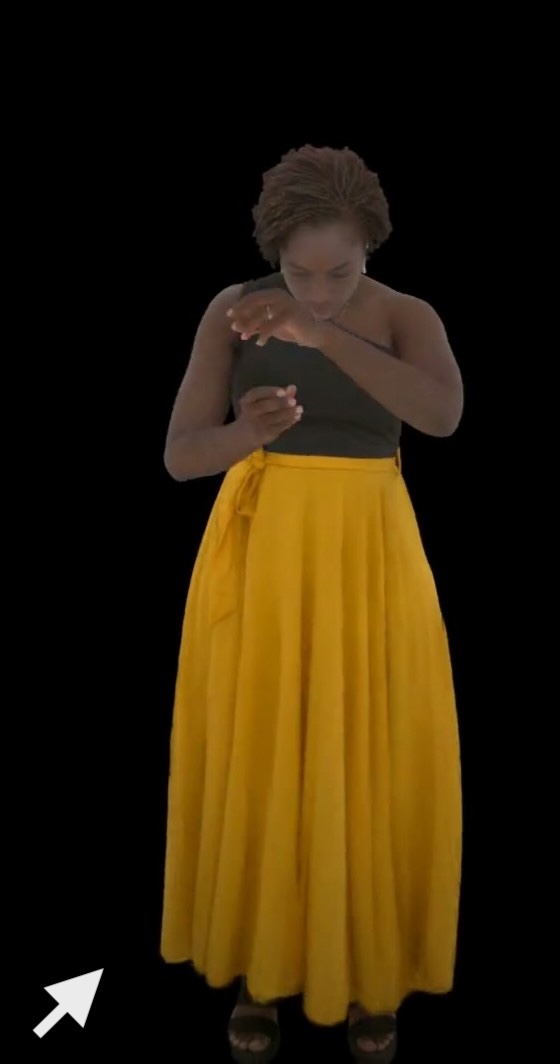} &
        \includegraphics[width=\imgw]{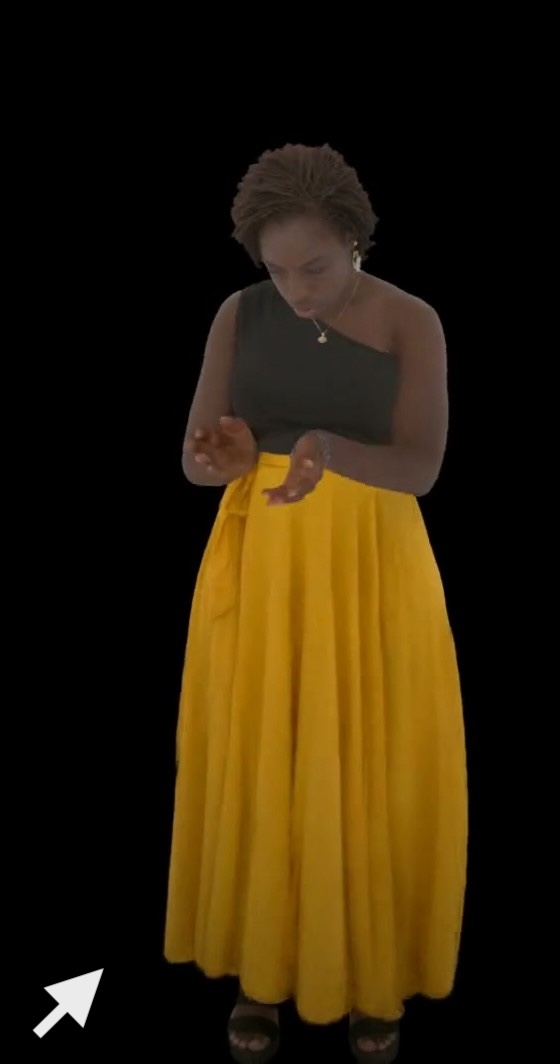} &
        \includegraphics[width=\imgw]{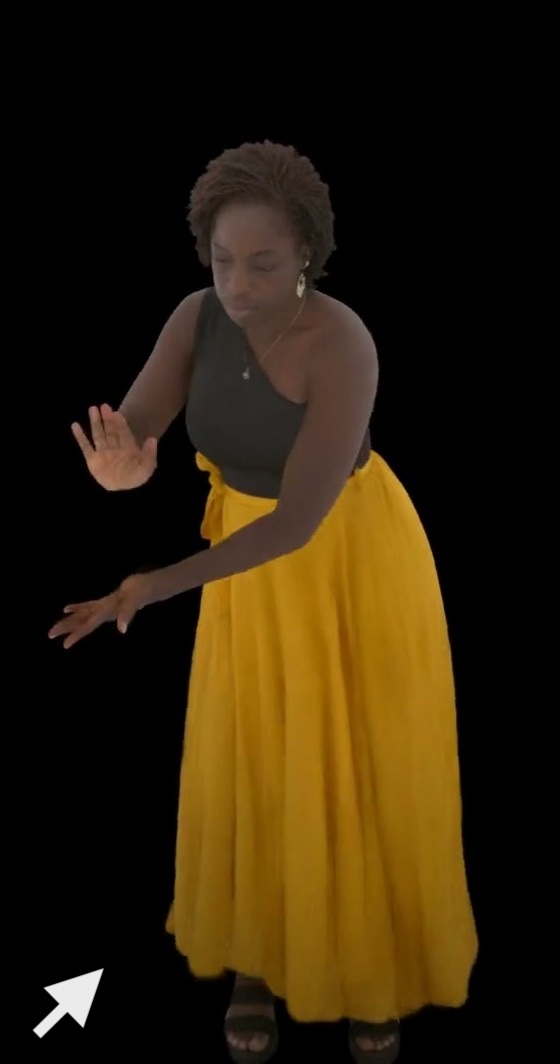} &
        \includegraphics[width=\imgw]{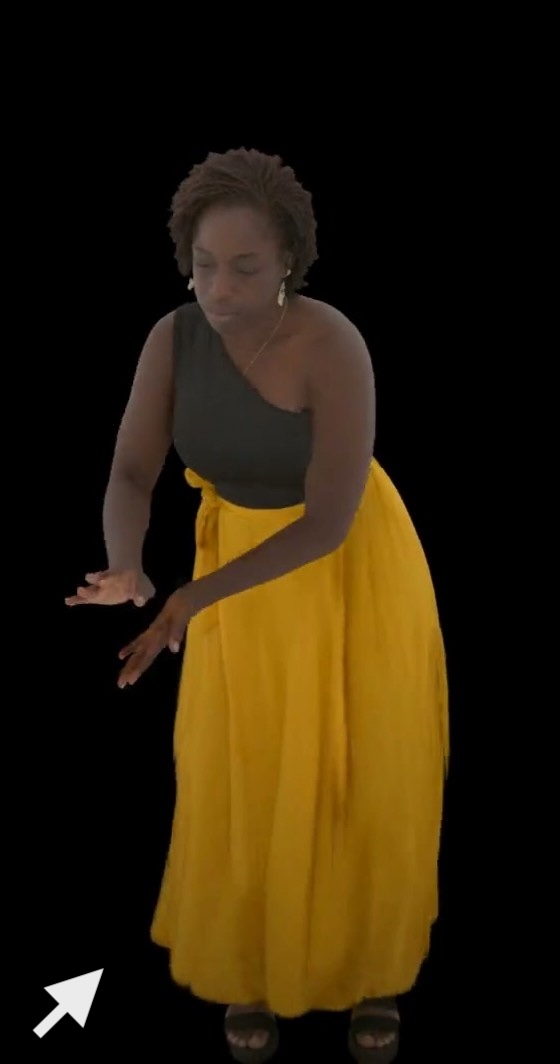} &
        \includegraphics[width=\imgw]{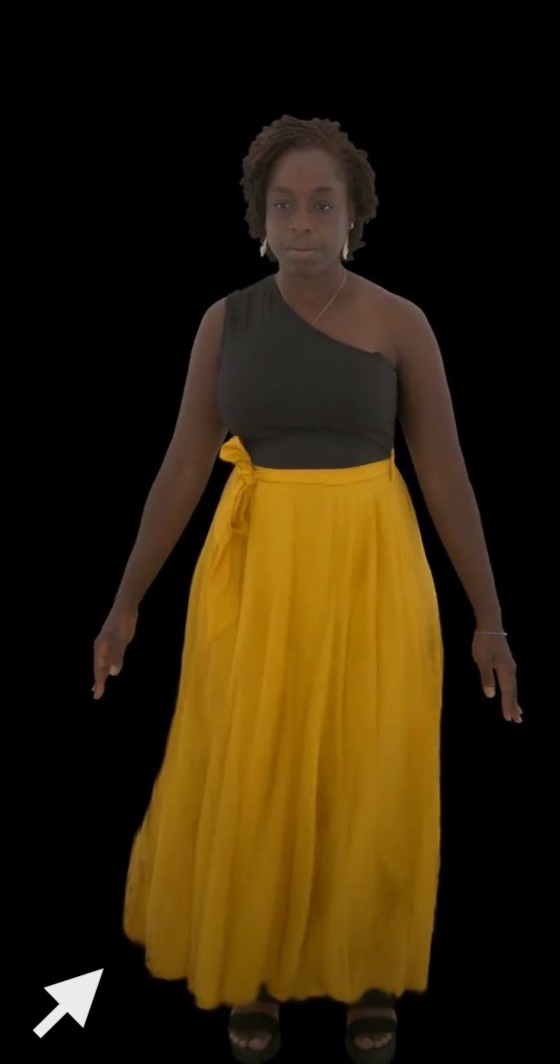} &
        \includegraphics[width=\imgw]{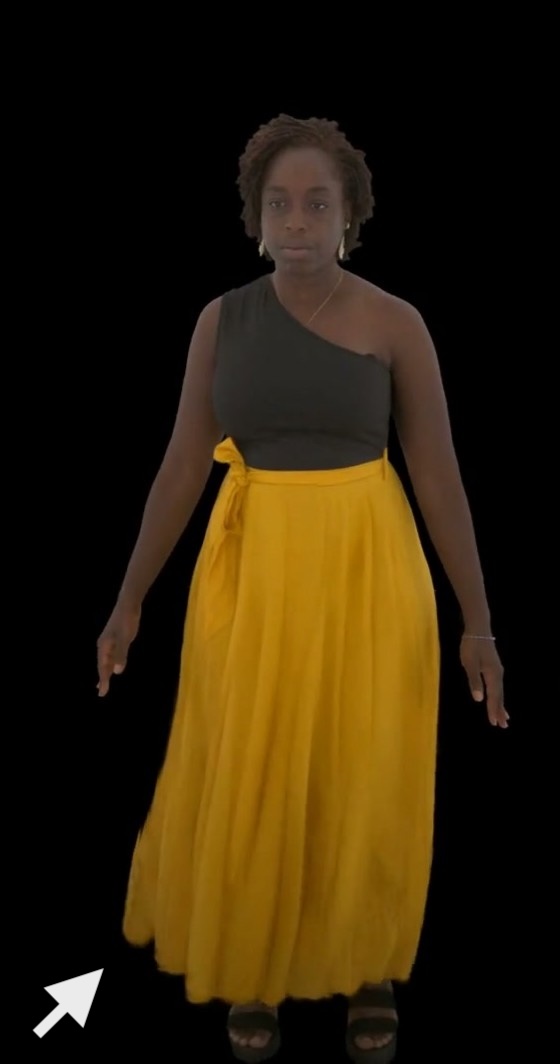} \\
        \raisebox{10.0ex}{\rowlab{Full Model}} &
        \includegraphics[width=\imgw]{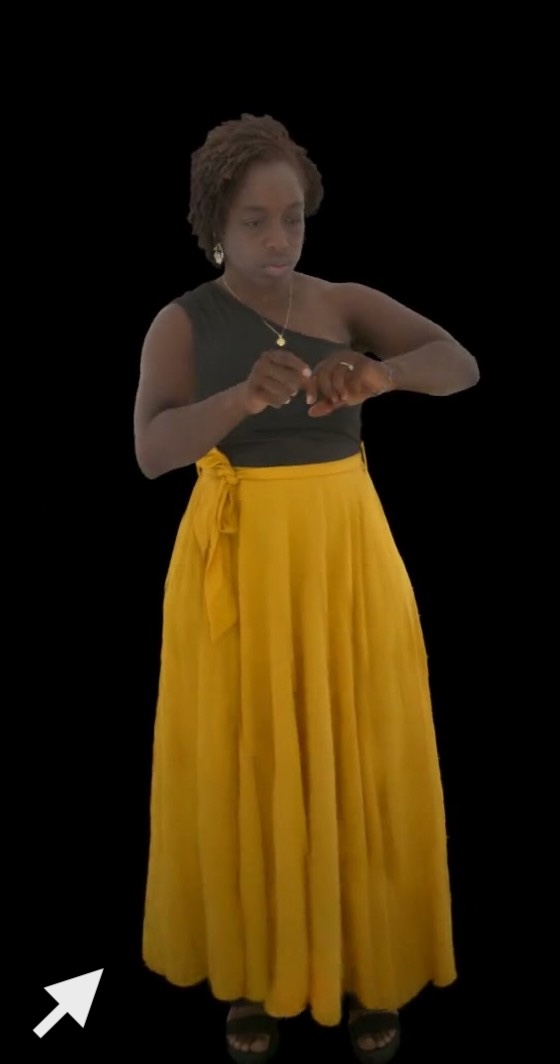} &
        \includegraphics[width=\imgw]{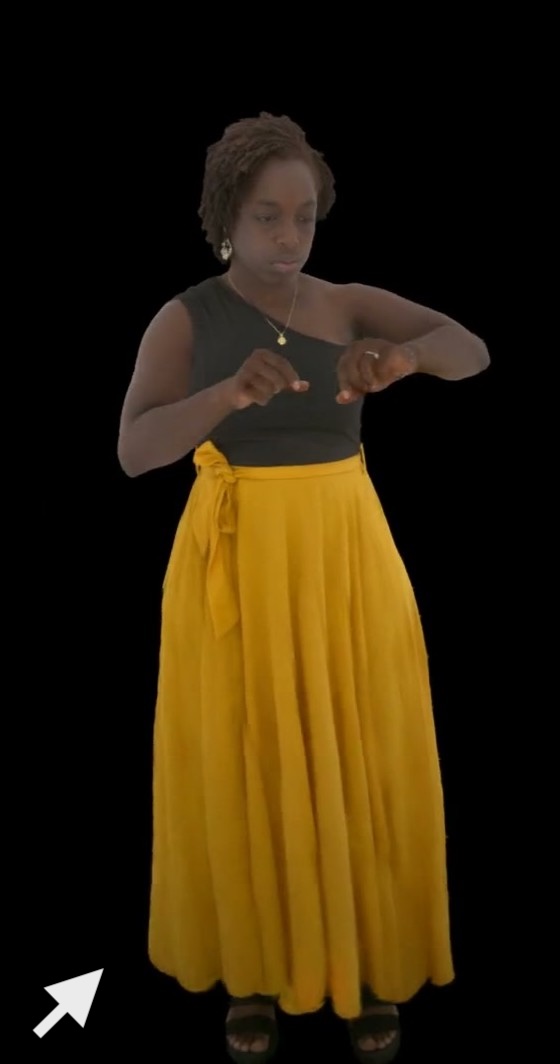} &
        \includegraphics[width=\imgw]{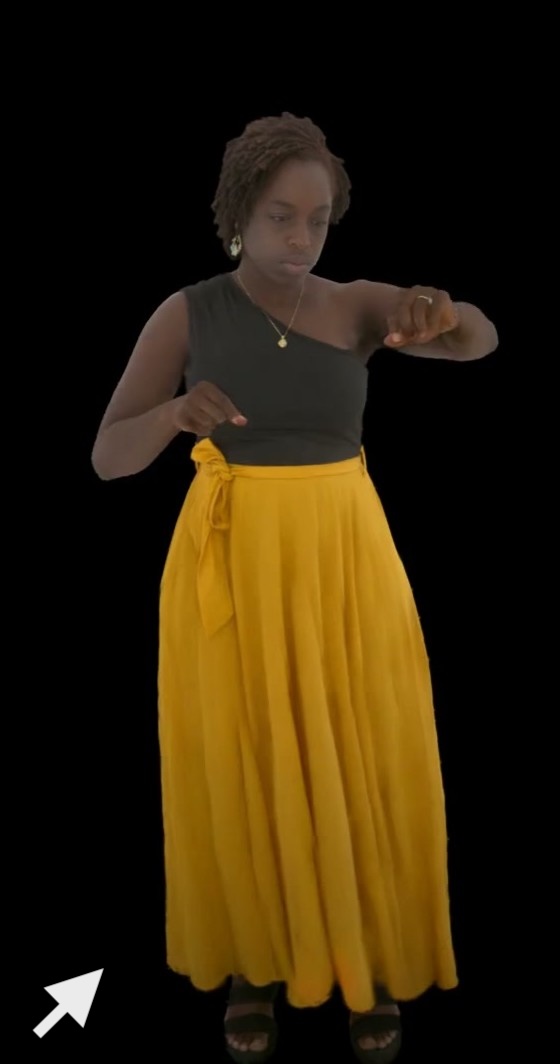} &
        \includegraphics[width=\imgw]{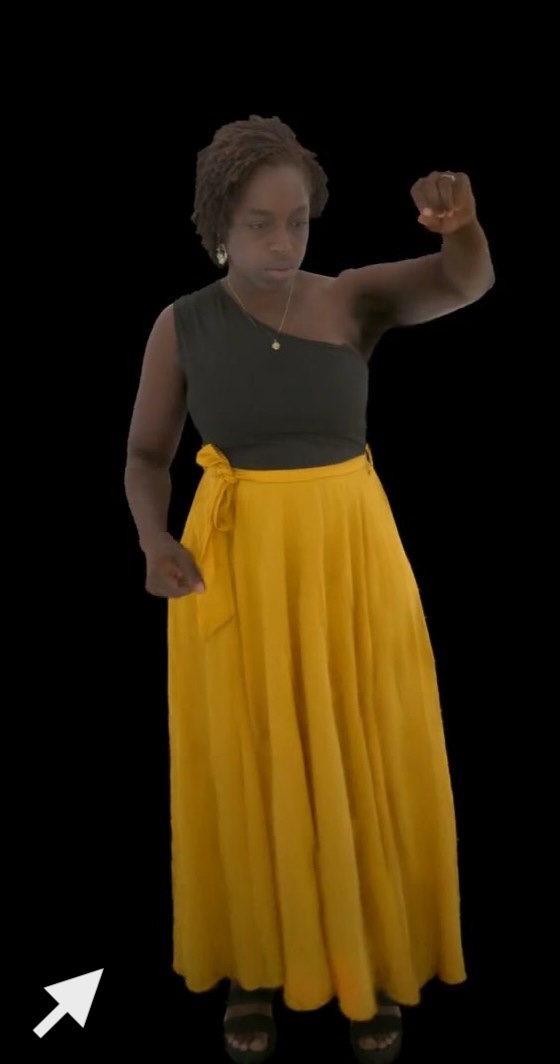} &
        \includegraphics[width=\imgw]{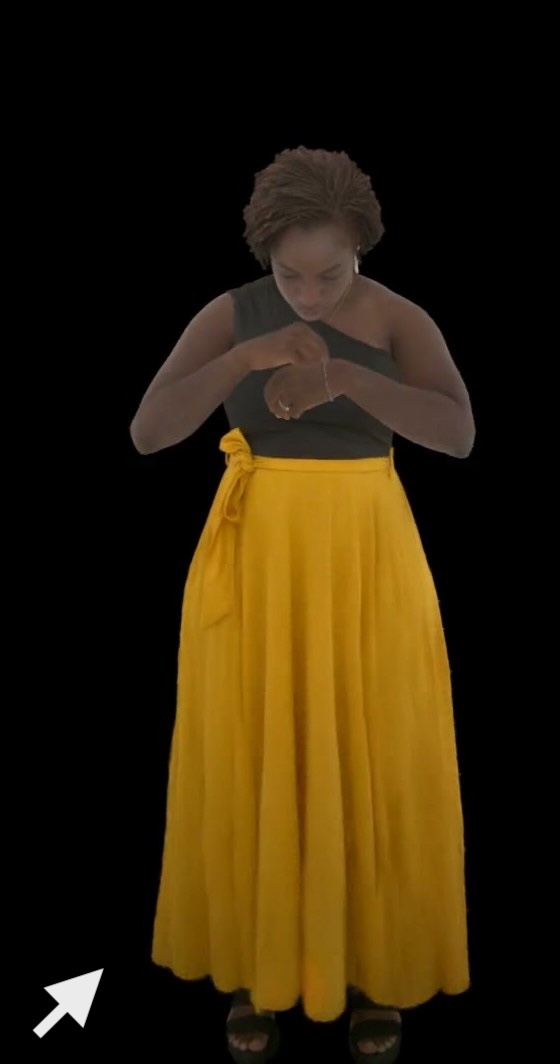} &
        \includegraphics[width=\imgw]{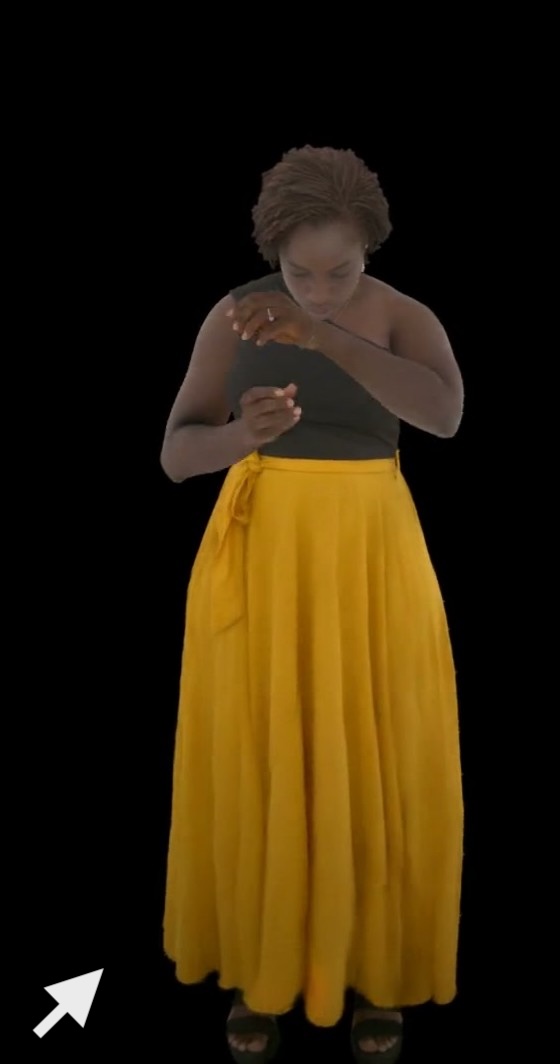} &
        \includegraphics[width=\imgw]{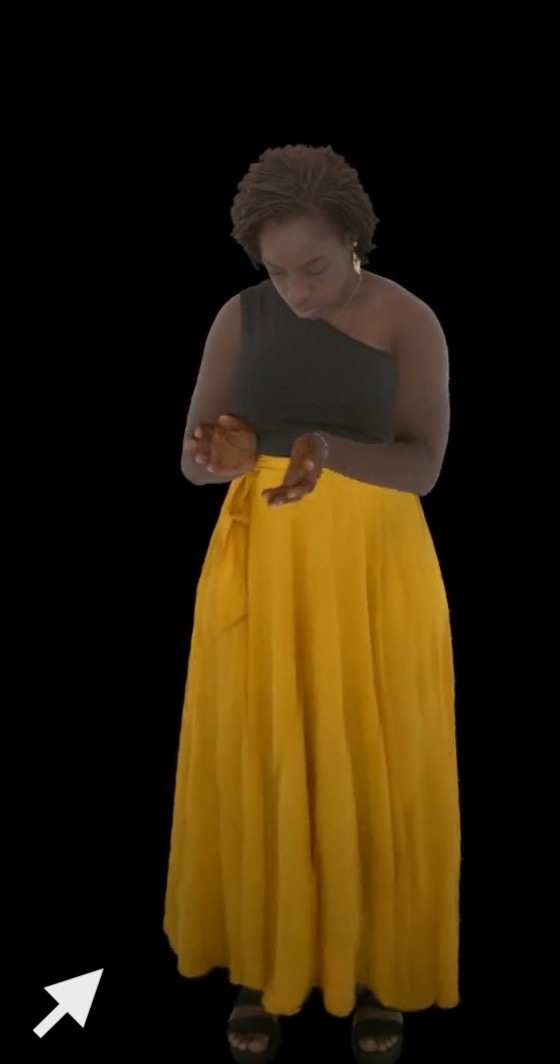} &
        \includegraphics[width=\imgw]{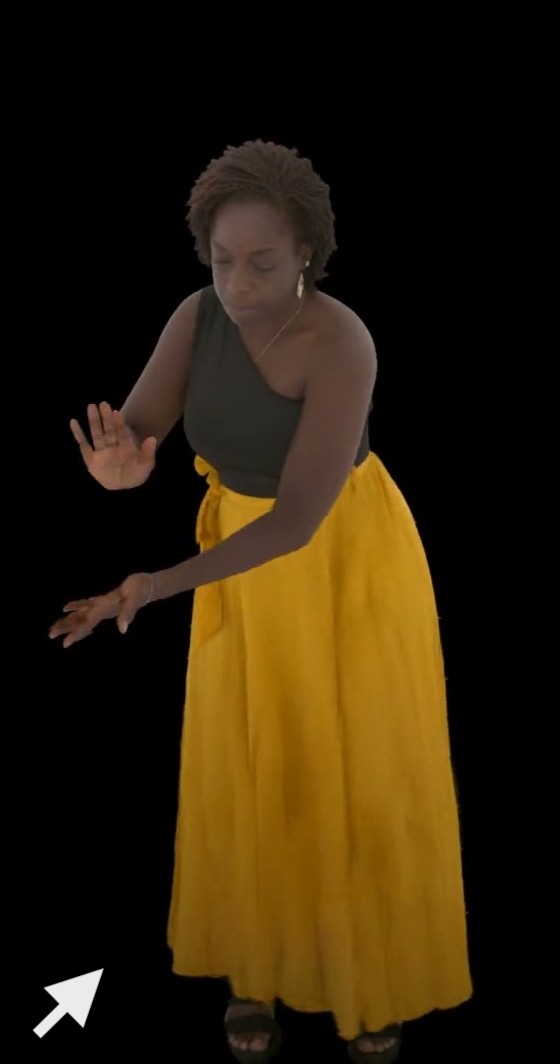} &
        \includegraphics[width=\imgw]{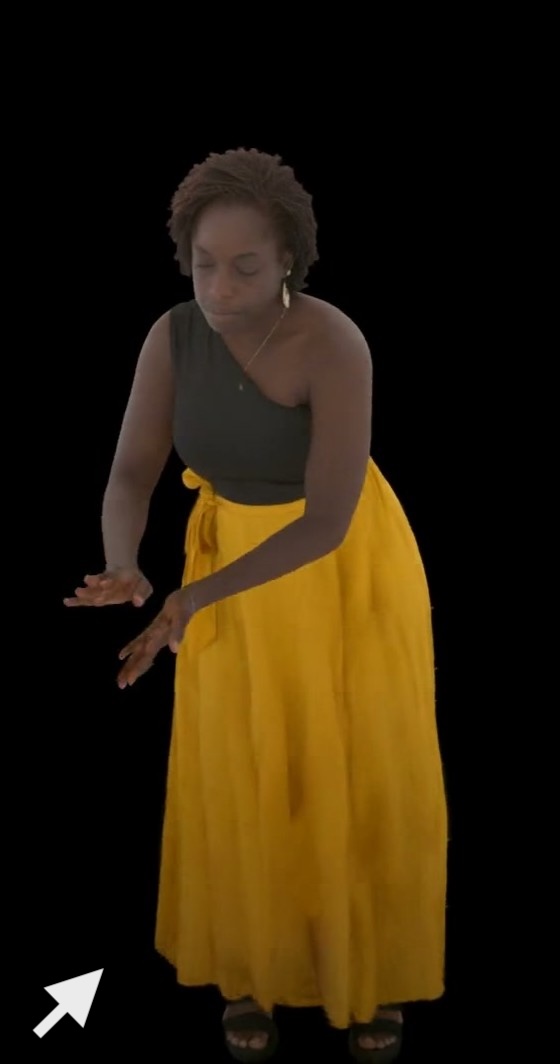} &
        \includegraphics[width=\imgw]{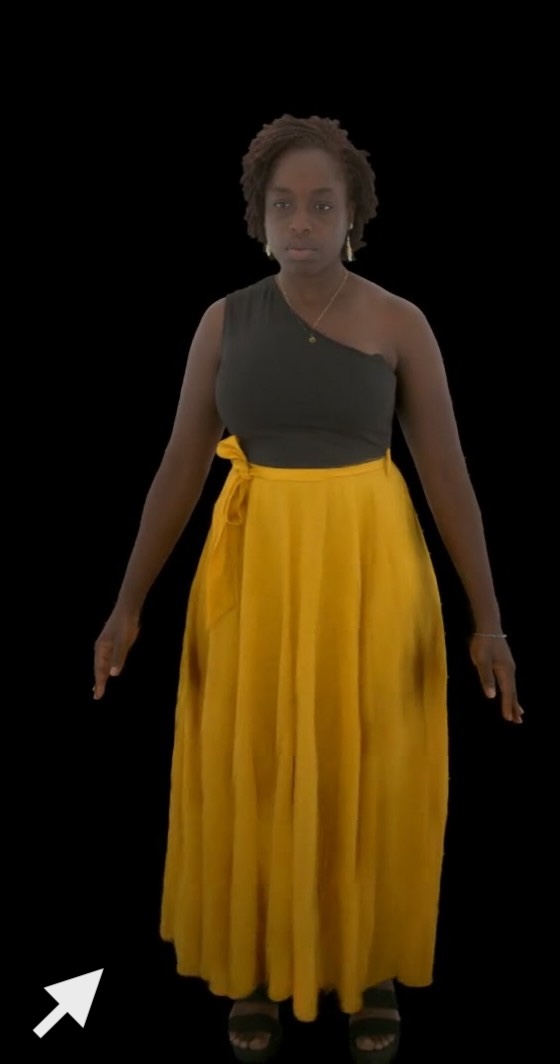} &
        \includegraphics[width=\imgw]{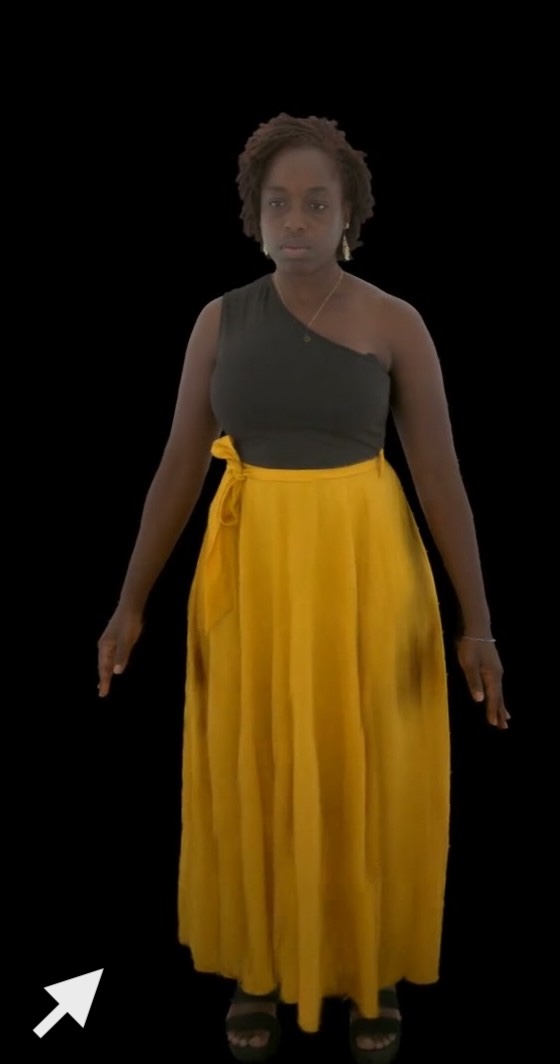} \\
    \end{tabular}
    \caption{\textbf{Latent dynamics model design ablation.} Compared to our full model with acceleration-based spring-damper dynamics (bottom), (i) direct latent $\mathbf{z}_t$ prediction jitters, (ii) velocity $\mathbf{v}_t$ prediction is overly stiff, and (iii) acceleration $\mathbf{a}_t$ without spring-damper formulation does not properly return to rest after motion stops. The supplementary video shows the temporal effects more clearly.}
    \label{fig:ablation_dynamics_design}
\end{figure}

While the differences are much more obvious in motion (see the supplementary video for details), we summarize the observed behaviors here and provide a visual reference in Figure~\ref{fig:ablation_dynamics_design}. (i) Directly predicting $\mathbf{z}_t$ leads to noticeable jitter and lacks smooth temporal transitions. (ii) Predicting velocity $\mathbf{v}_t$ produces overly stiff cloth motion, and the garment does not flow naturally. (iii) Removing the spring-damper yields flowy motion but fails to return to a plausible rest state when the body motion stops, causing lingering deformation.

\section{Discussion On Baseline Quality}
\label{app:baseline_quality}
Despite being trained for the same number of epochs on the full sequence as our method, the baselines show a performance drop compared to their reported results on public benchmarks, including DNA-Rendering~\cite{Cheng2023DNARenderingAD}, I3D-Human~\cite{Chen2024WithinTD}, MVHumanNet~\cite{Xiong2023MVHumanNetAL}, and ZJU\_MoCap~\cite{Peng2020NeuralBI}. We attribute this discrepancy to differences in capture settings and clothing complexity. For example, the baseline results on DNA-Rendering were trained on only 100 training frames, whereas our sequences provide about 2,000 training frames. I3D-Human contains about 600 training frames at a lower $512\times512$ resolution and includes limited loose-clothing diversity (only one long dress among three subjects). MVHumanNet and ZJU\_MoCap primarily feature tight clothing and thus exhibit substantially smaller dynamic deformations.
	
\begin{figure*}[ht]
  \centering
  \small
  \setlength{\tabcolsep}{0pt}
  \newcommand{\imgw}{0.093\linewidth} 
  
  \begin{tabular}{cc@{\hskip 6pt}cc@{\hskip 6pt}cc@{\hskip 6pt}cc@{\hskip 6pt}cc}
  \includegraphics[width=\imgw]{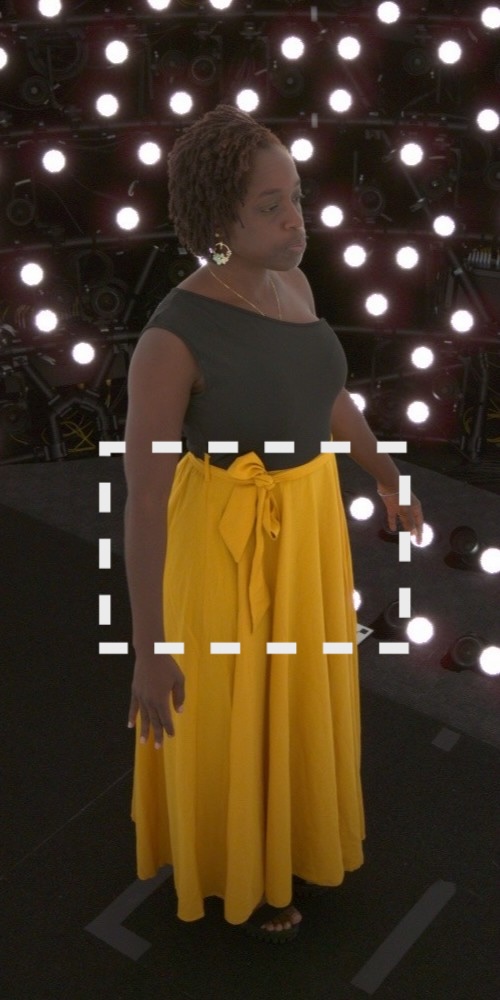} &
  \includegraphics[width=\imgw]{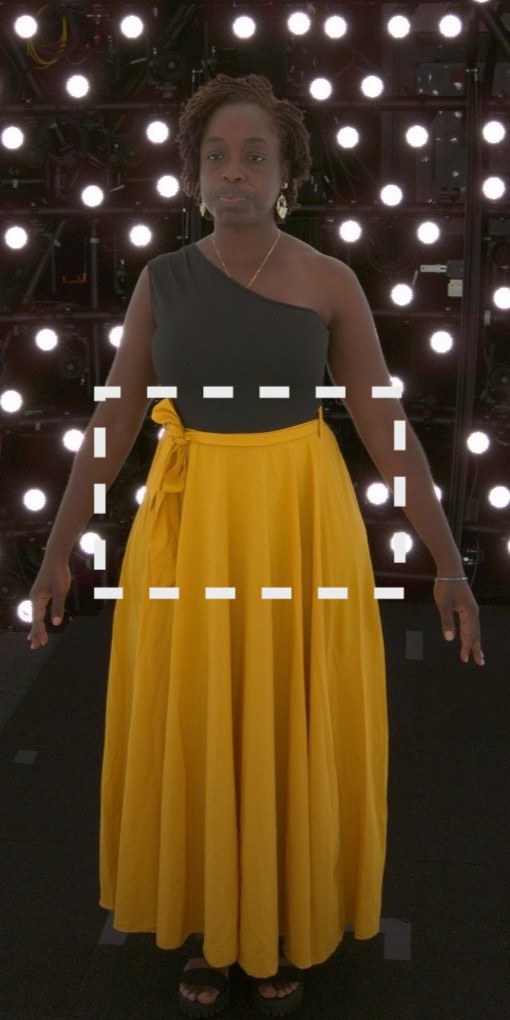} &
  \includegraphics[width=\imgw]{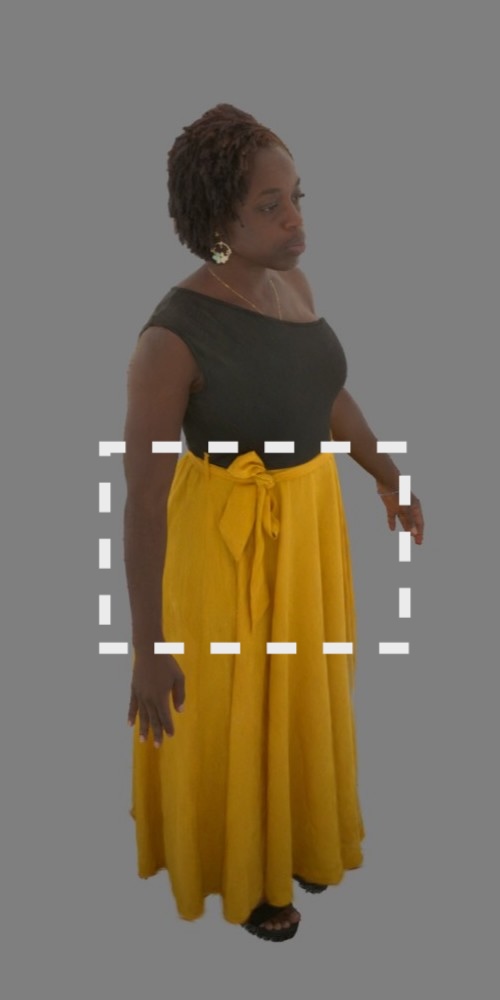} &
  \includegraphics[width=\imgw]{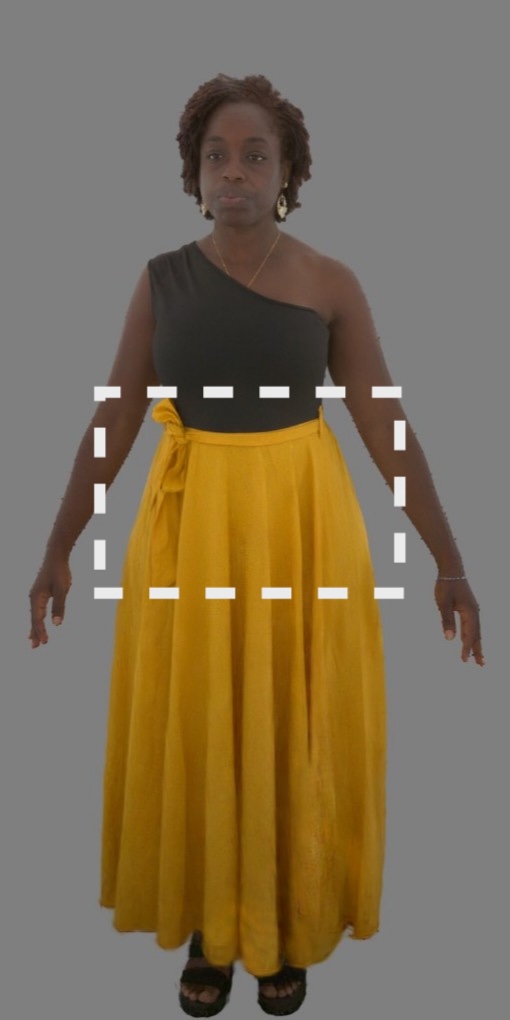} &
  \includegraphics[width=\imgw]{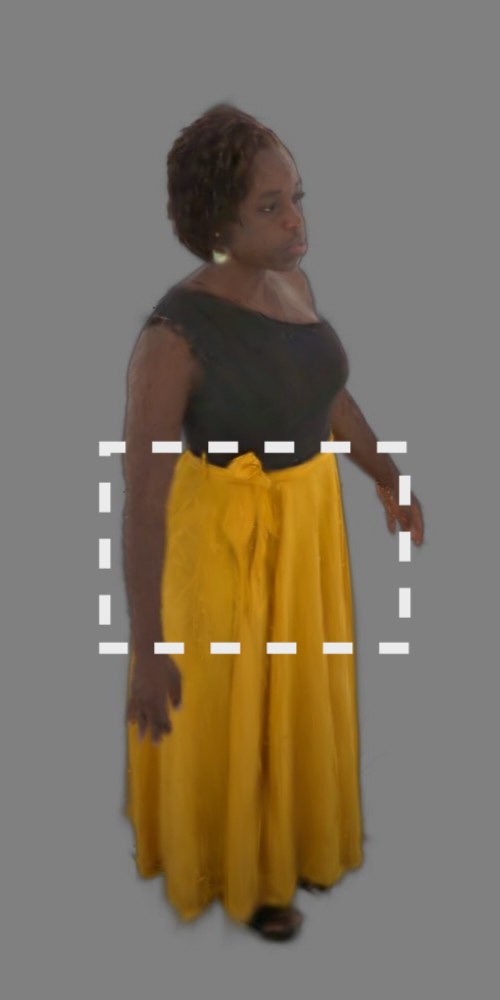} &
  \includegraphics[width=\imgw]{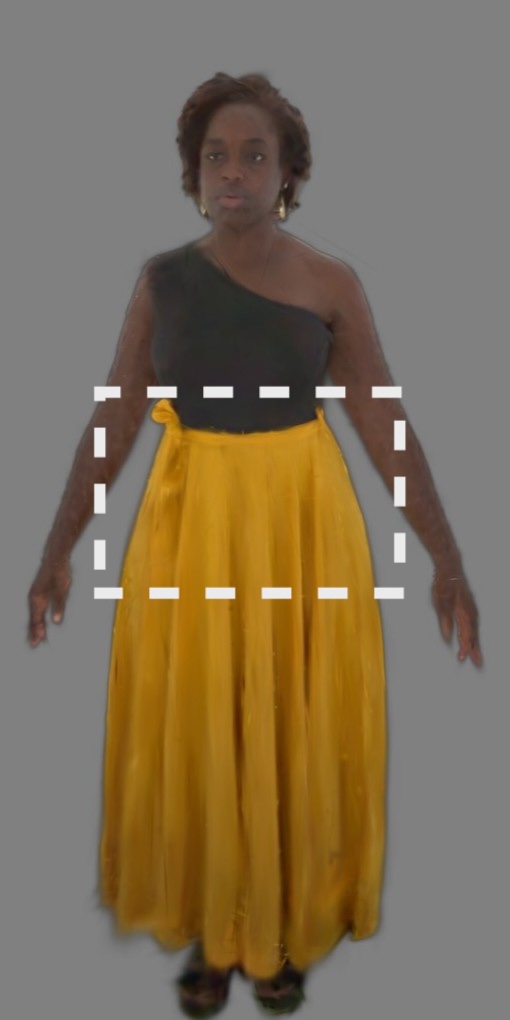} &
  \includegraphics[width=\imgw]{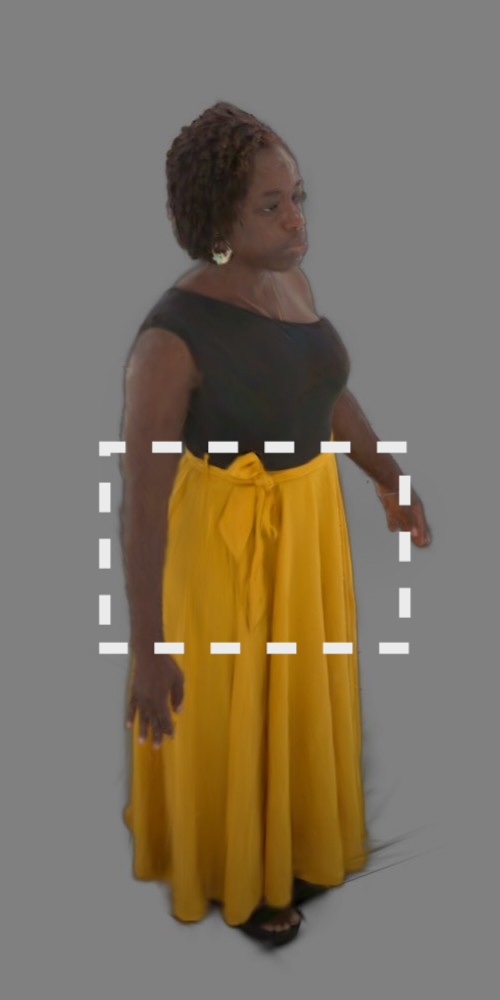} &
  \includegraphics[width=\imgw]{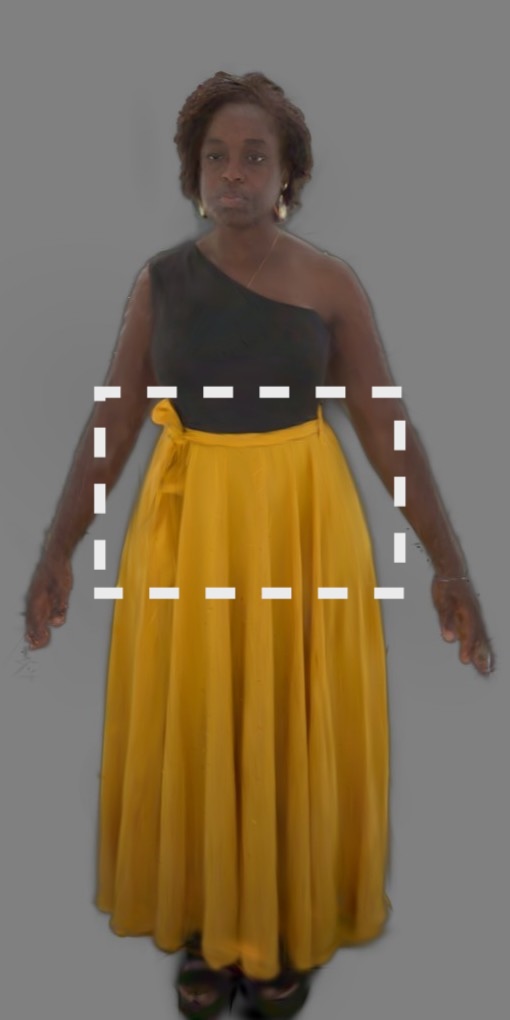} &
  \includegraphics[width=\imgw]{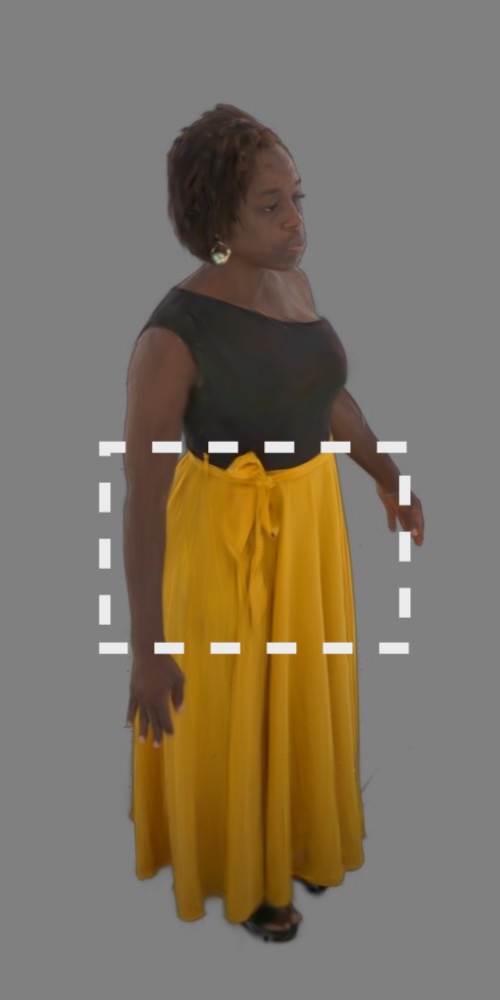} &
  \includegraphics[width=\imgw]{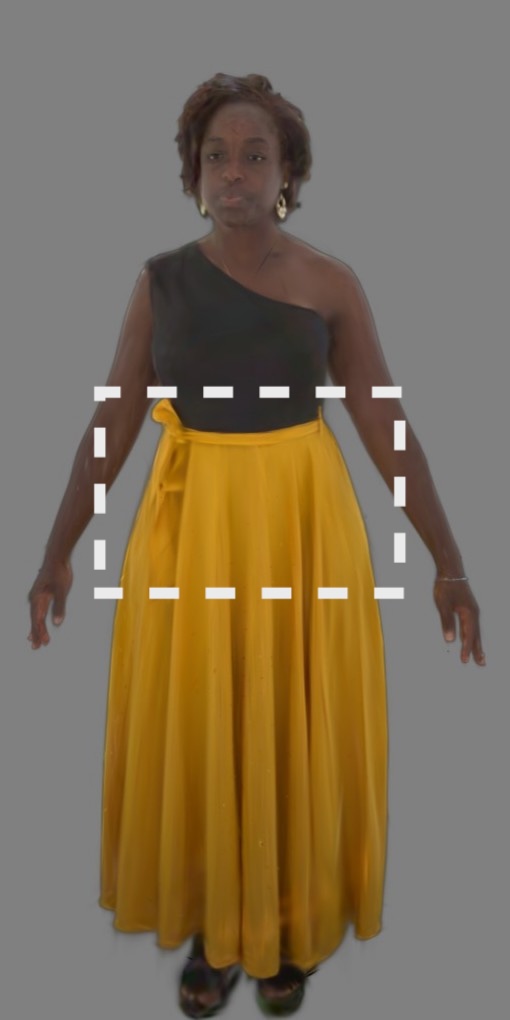} \\[-1.5pt]
  \includegraphics[width=\imgw]{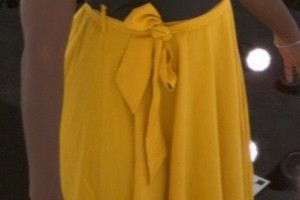} &
  \includegraphics[width=\imgw]{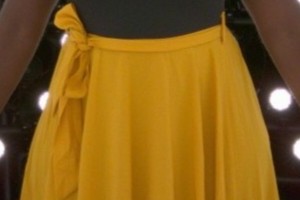} &
  \includegraphics[width=\imgw]{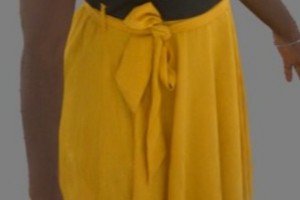} &
  \includegraphics[width=\imgw]{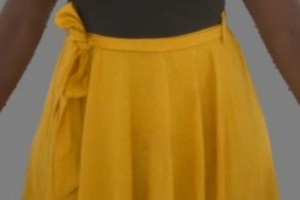} &
  \includegraphics[width=\imgw]{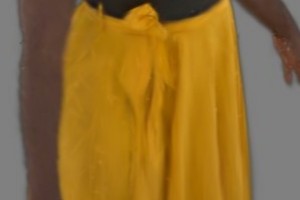} &
  \includegraphics[width=\imgw]{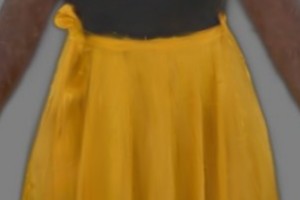} &
  \includegraphics[width=\imgw]{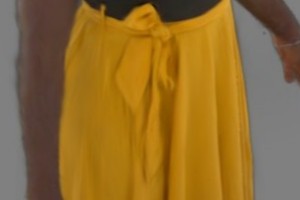} &
  \includegraphics[width=\imgw]{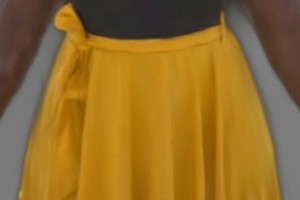} &
  \includegraphics[width=\imgw]{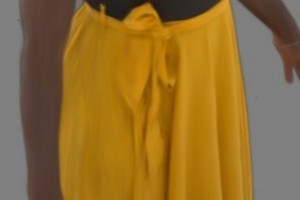} &
  \includegraphics[width=\imgw]{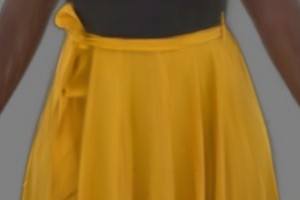} \\[2pt]
  \includegraphics[width=\imgw]{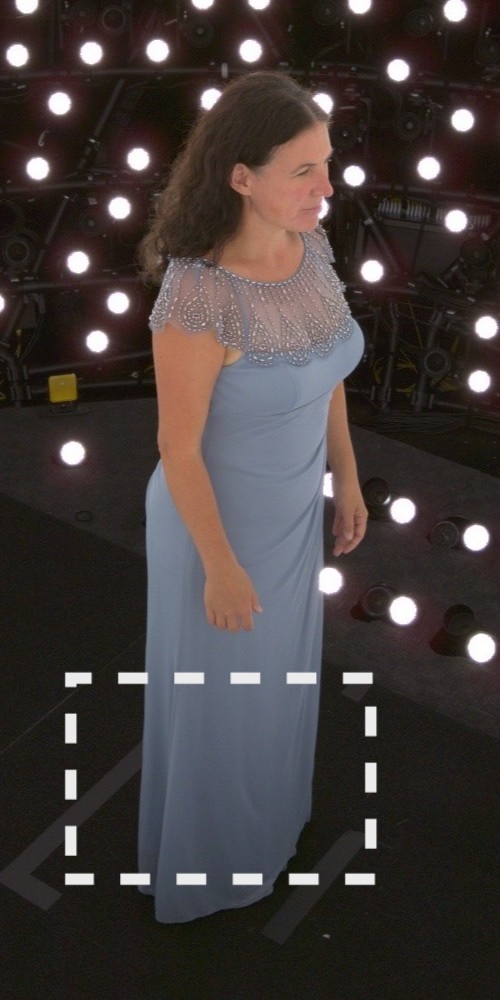} &
  \includegraphics[width=\imgw]{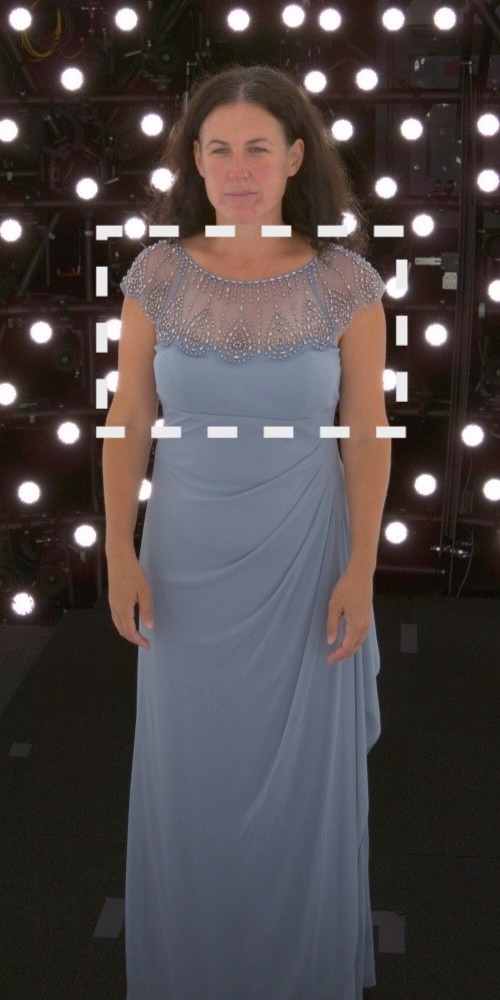} &
  \includegraphics[width=\imgw]{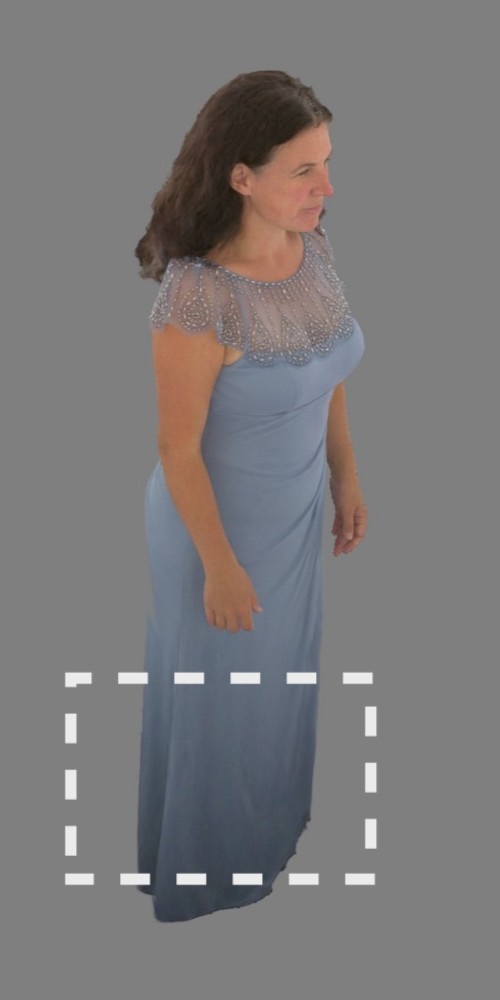} &
  \includegraphics[width=\imgw]{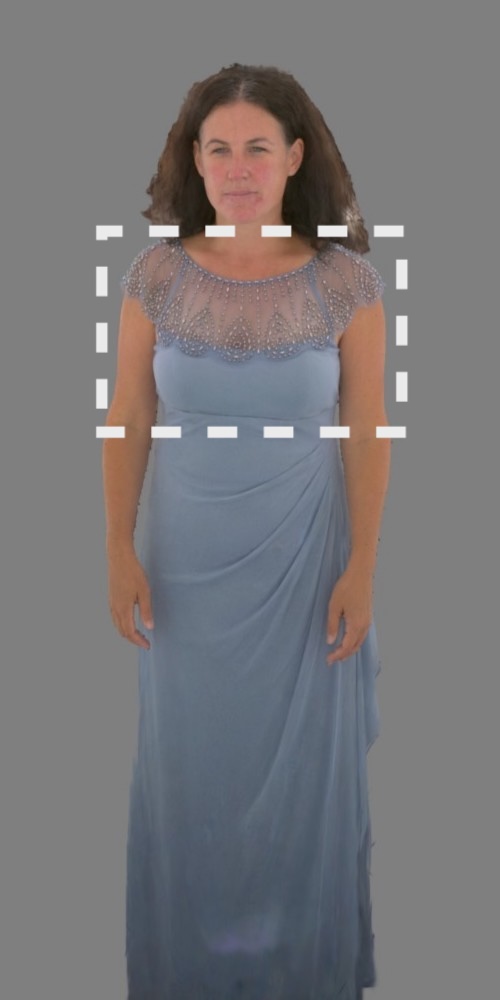} &
  \includegraphics[width=\imgw]{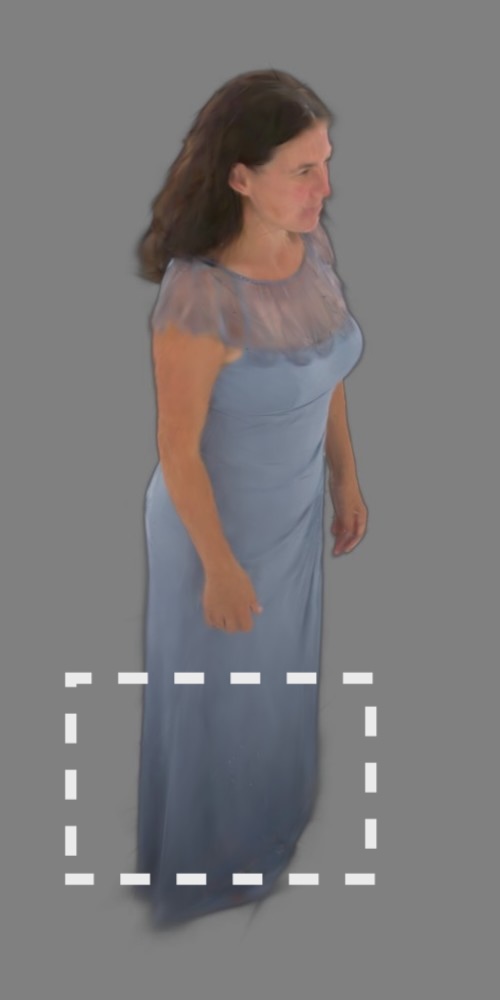} &
  \includegraphics[width=\imgw]{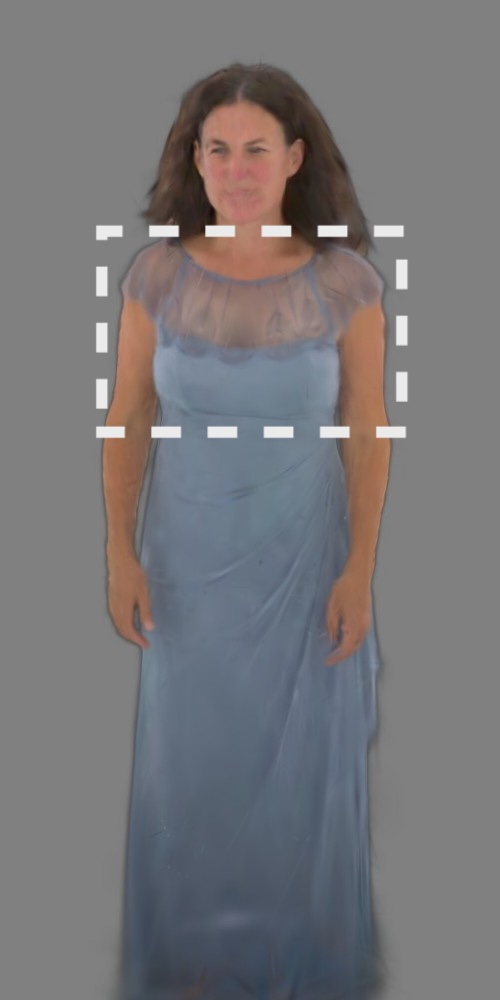} &
  \includegraphics[width=\imgw]{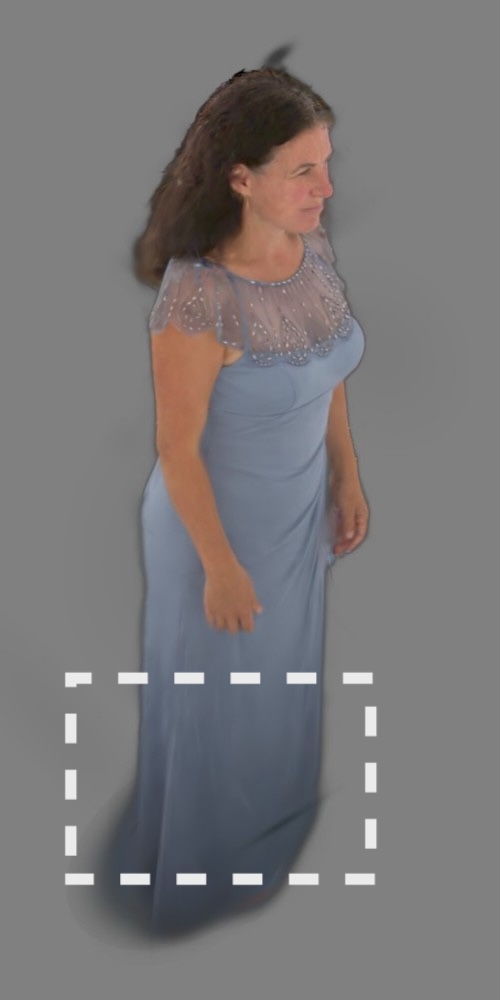} &
  \includegraphics[width=\imgw]{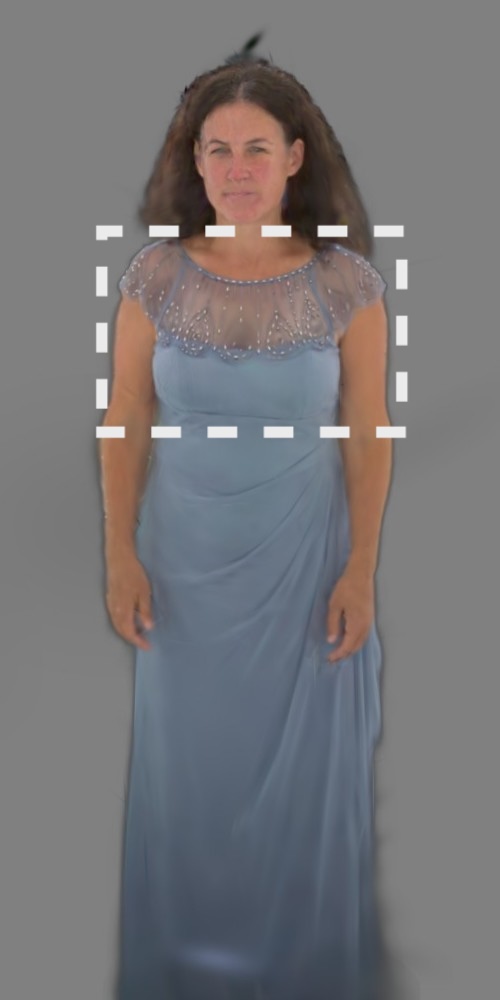} &
  \includegraphics[width=\imgw]{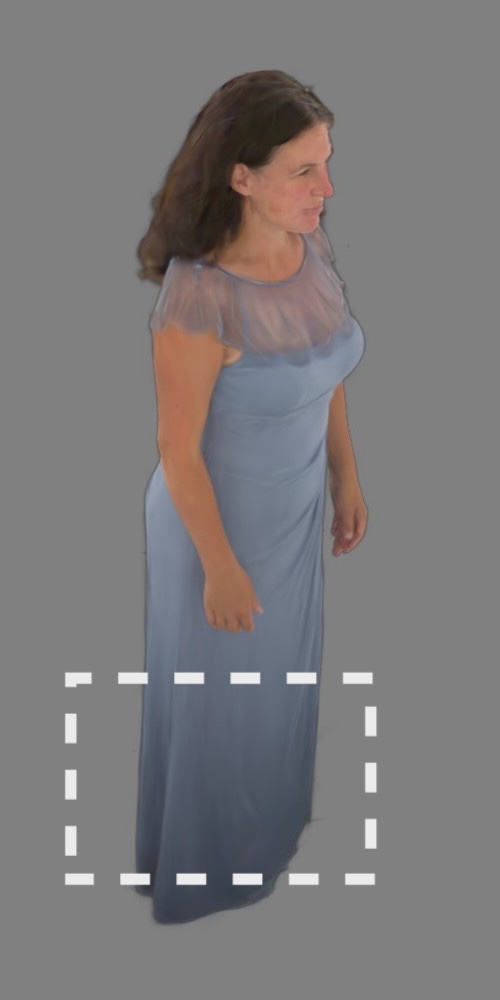} &
  \includegraphics[width=\imgw]{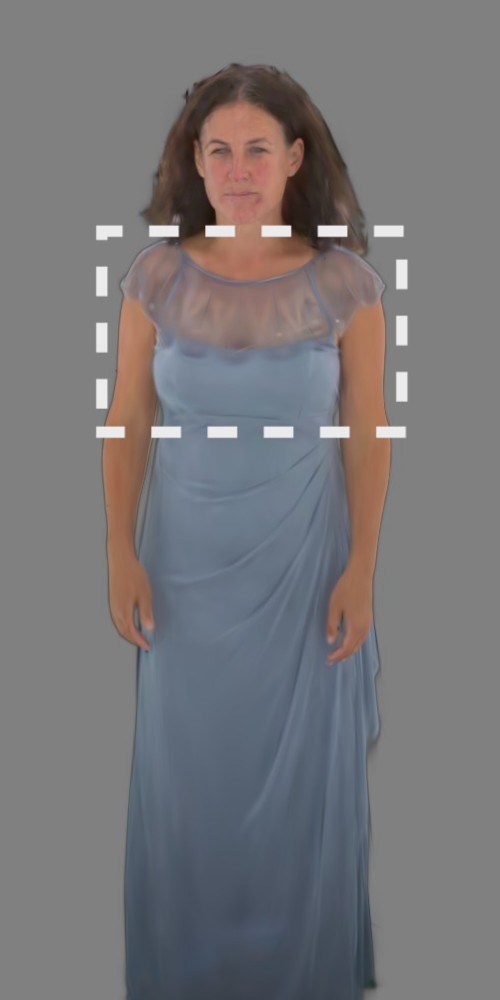} \\[-1.5pt]
  \includegraphics[width=\imgw]{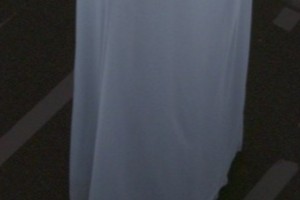} &
  \includegraphics[width=\imgw]{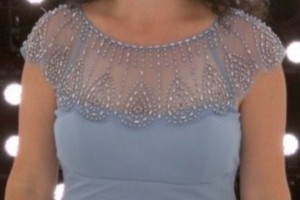} &
  \includegraphics[width=\imgw]{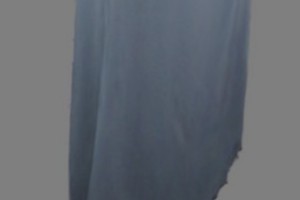} &
  \includegraphics[width=\imgw]{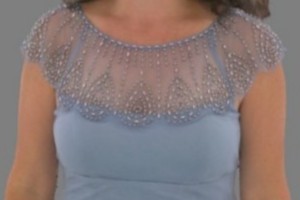} &
  \includegraphics[width=\imgw]{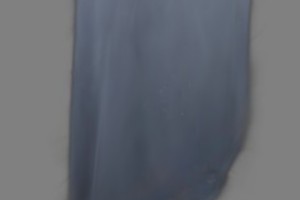} &
  \includegraphics[width=\imgw]{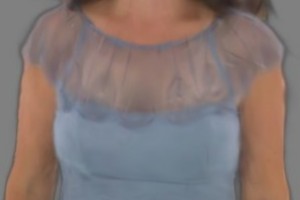} &
  \includegraphics[width=\imgw]{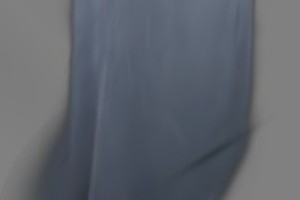} &
  \includegraphics[width=\imgw]{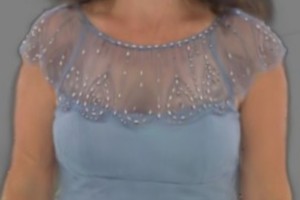} &
  \includegraphics[width=\imgw]{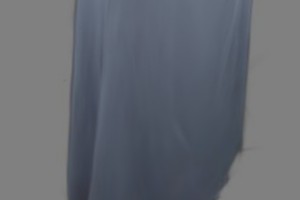} &
  \includegraphics[width=\imgw]{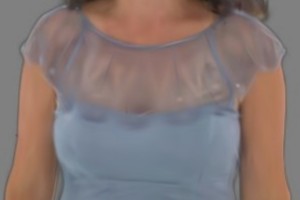} \\[2pt]
  \includegraphics[width=\imgw]{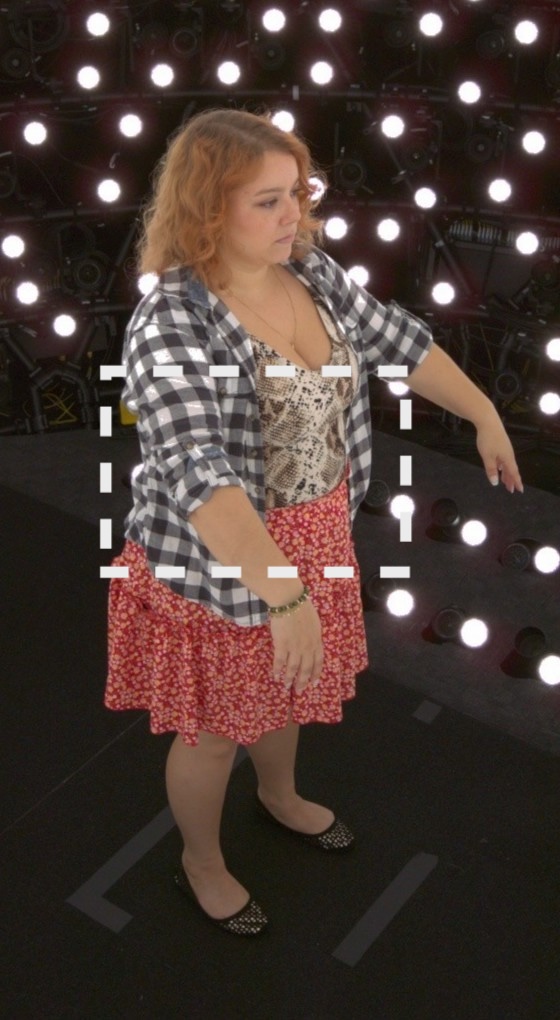} &
  \includegraphics[width=\imgw]{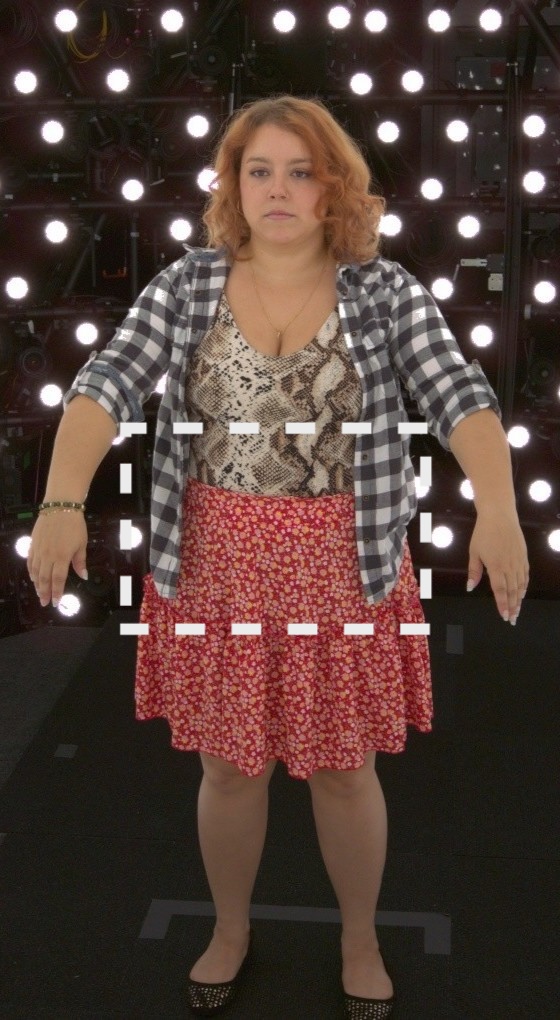} &
  \includegraphics[width=\imgw]{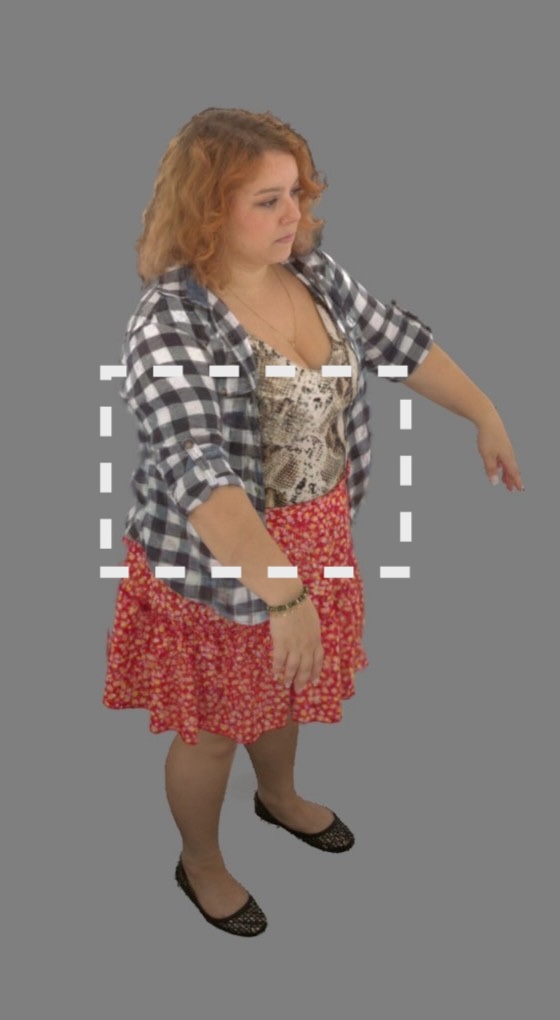} &
  \includegraphics[width=\imgw]{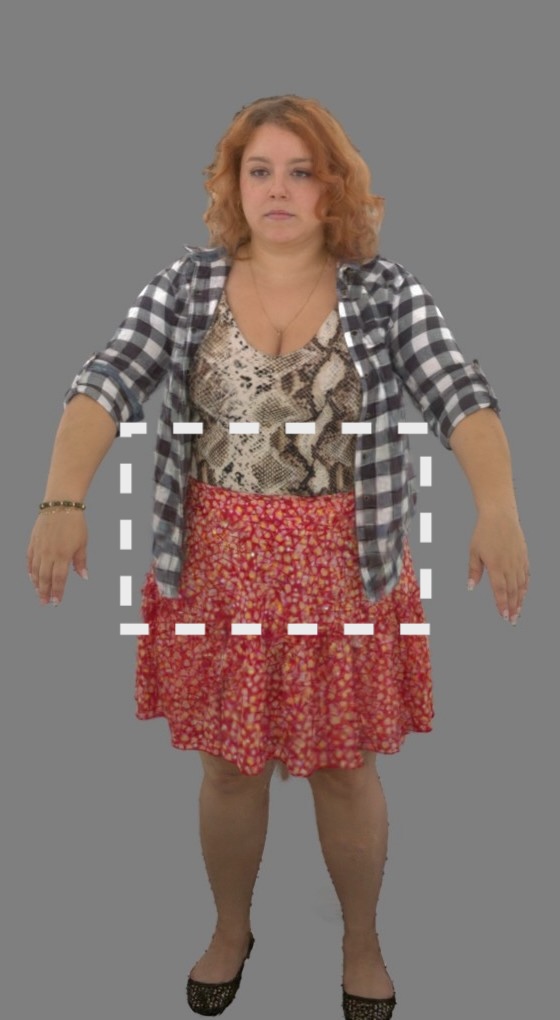} &
  \includegraphics[width=\imgw]{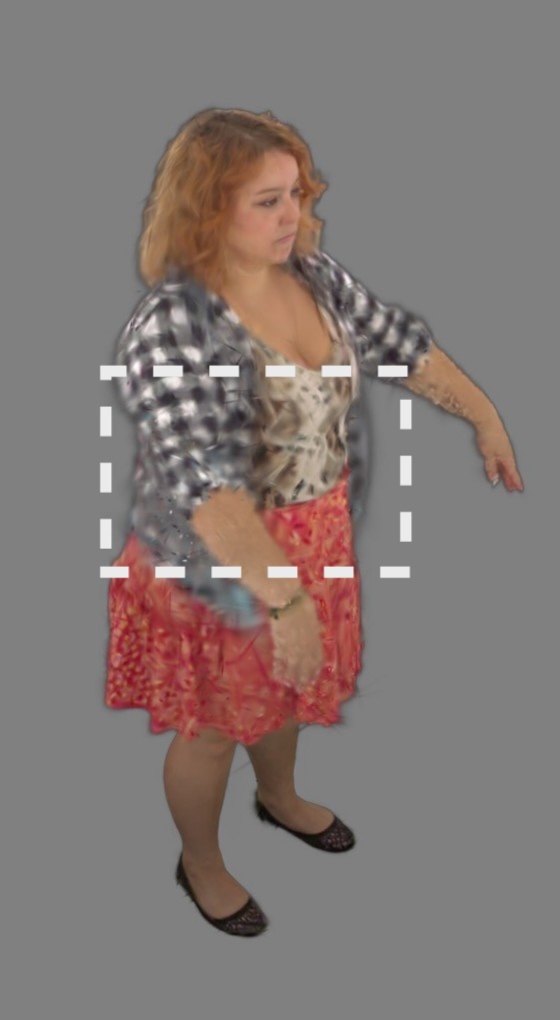} &
  \includegraphics[width=\imgw]{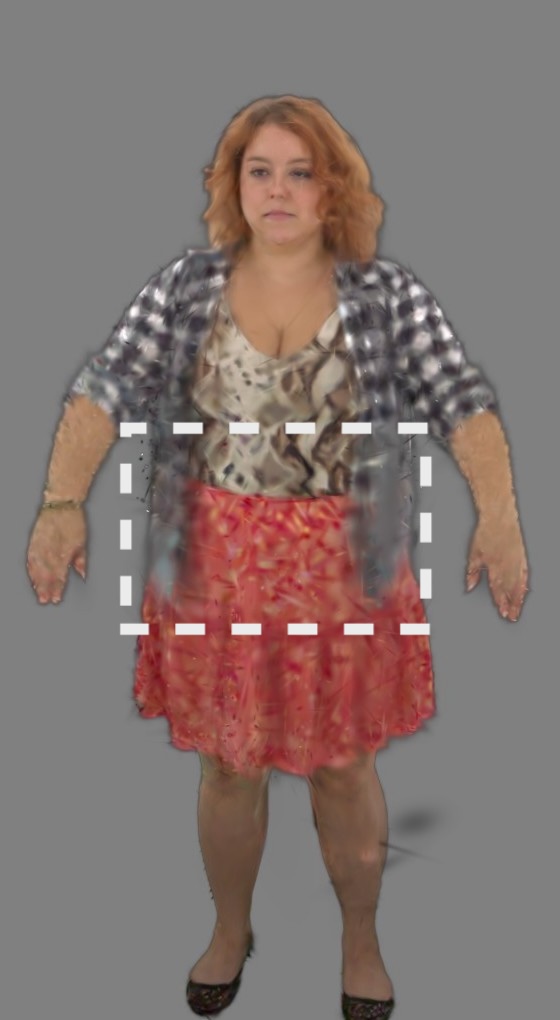} &
  \includegraphics[width=\imgw]{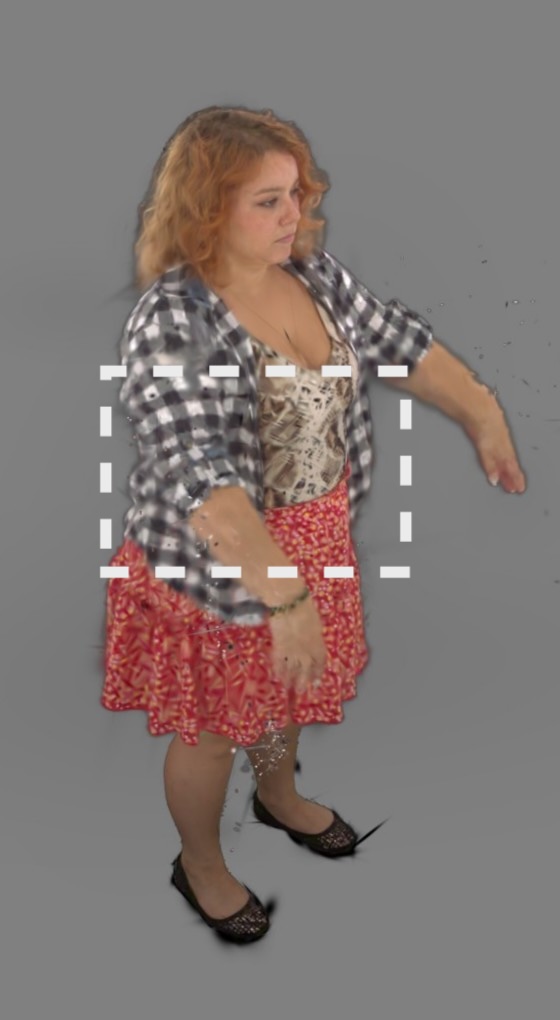} &
  \includegraphics[width=\imgw]{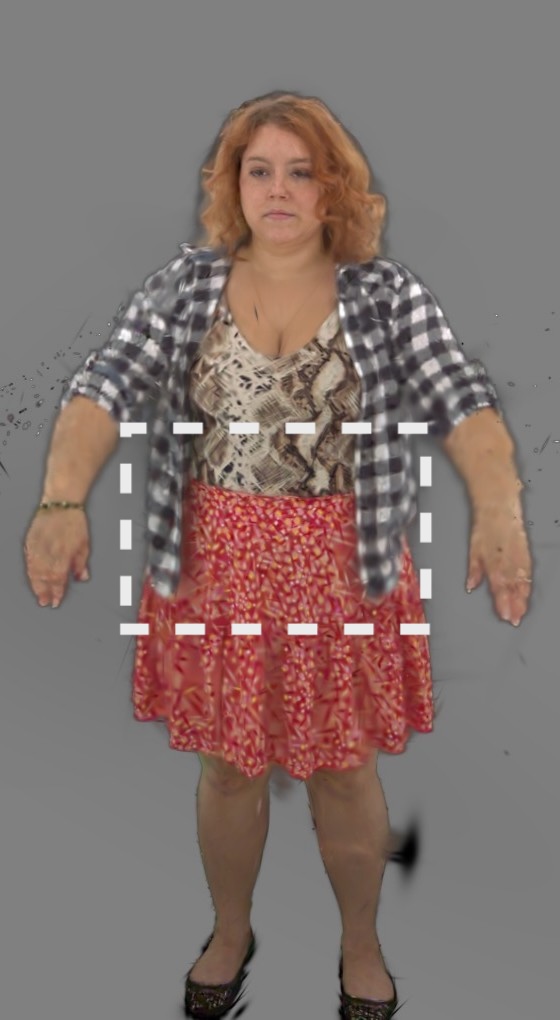} &
  \includegraphics[width=\imgw]{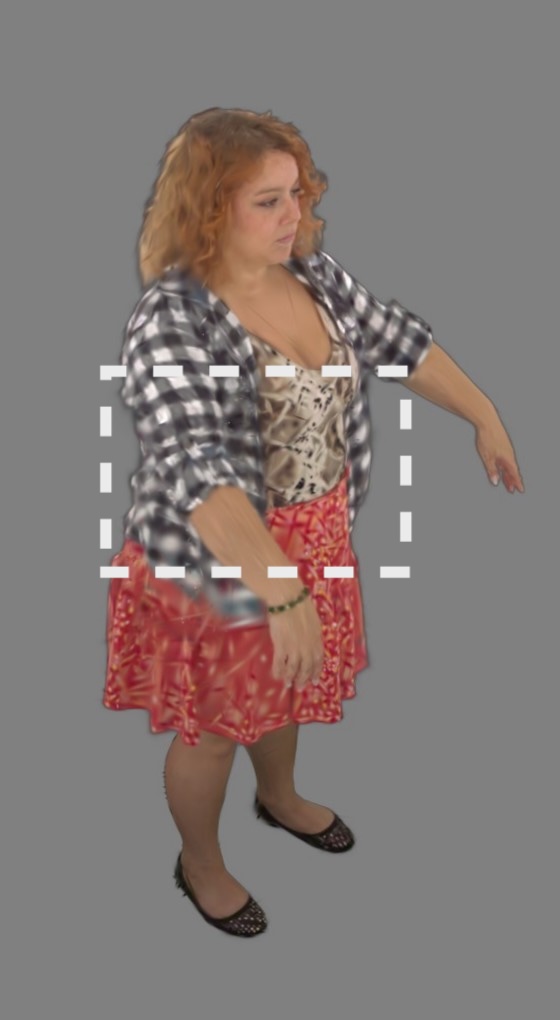} &
  \includegraphics[width=\imgw]{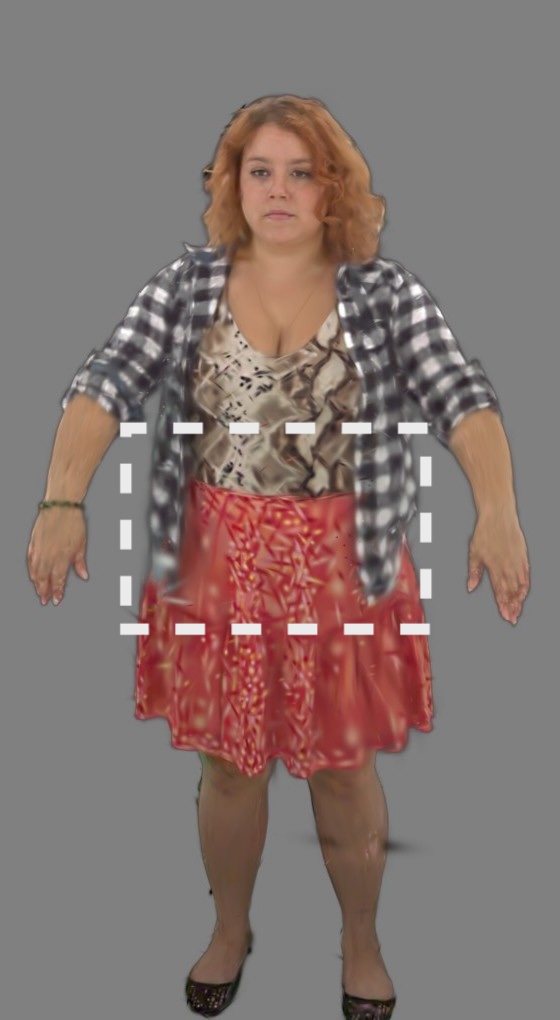} \\[-1.5pt]
  \includegraphics[width=\imgw]{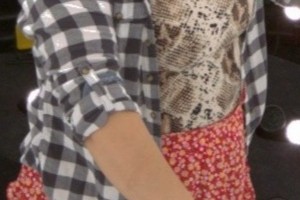} &
  \includegraphics[width=\imgw]{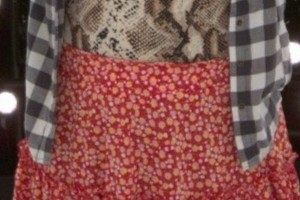} &
  \includegraphics[width=\imgw]{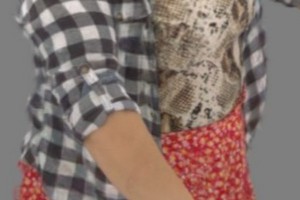} &
  \includegraphics[width=\imgw]{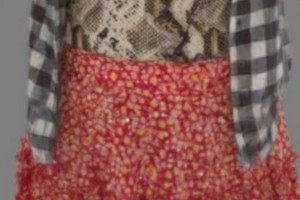} &
  \includegraphics[width=\imgw]{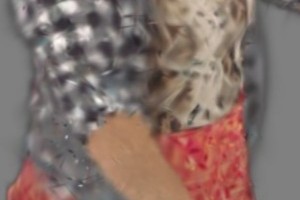} &
  \includegraphics[width=\imgw]{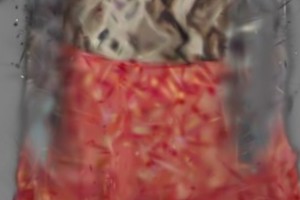} &
  \includegraphics[width=\imgw]{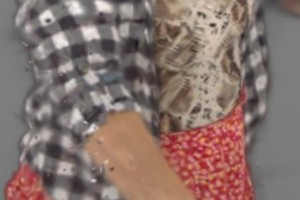} &
  \includegraphics[width=\imgw]{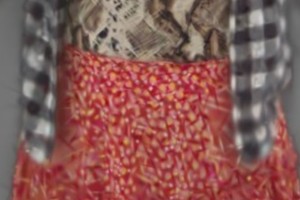} &
  \includegraphics[width=\imgw]{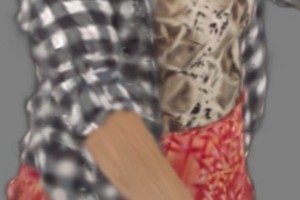} &
  \includegraphics[width=\imgw]{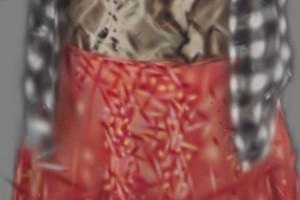} \\[6pt]
  \multicolumn{2}{c}{Reference Image} &
  \multicolumn{2}{c}{Ours} &
  \multicolumn{2}{c}{ToMiE~\cite{zhan2024tomiemodulargrowthenhanced}} &
  \multicolumn{2}{c}{Seq-Avatar~\cite{xu2025seqavatar}} &
  \multicolumn{2}{c}{$R^3$-Avatar~\cite{Zhan2025R3AvatarRA}} \\
  \end{tabular}
  \caption{\textbf{Qualitative comparison on novel-view synthesis in a reduced-data setting (trained on the first 100 frames).} While baseline methods improve and recover more fine-grained clothing patterns, our approach still produces higher-quality renderings.}
  \label{fig:novel_view_100frames}
  \end{figure*}

To test this hypothesis, we retrain all baselines using only the first 100 frames of each sequence; for a fair comparison, we also retrain our method with the same 100 frames per subject. The quantitative and qualitative novel-view results are reported in Table~\ref{Tab::Res-novel-view-100frames} and Figure~\ref{fig:novel_view_100frames}. Under this reduced-data setting, baseline quality improves noticeably and recovers more fine clothing patterns, close to their performance on public benchmarks. This supports our claim that the observed degradation is largely driven by increased clothing complexity in our capture. It's worth noting that our method still preserves more details and maintains higher rendering quality in this setting.

\begin{table}[t]
  \centering
  \caption{Average quantitative metrics on novel view synthesis (PSNR ($\uparrow$), SSIM ($\uparrow$), and LPIPS ($\downarrow$)) using first 100 frames in the training set.}
  \label{Tab::Res-novel-view-100frames}
  \begin{tabular}{c|ccc}
  \toprule
  Methods & PSNR ($\uparrow$) & SSIM ($\uparrow$) & LPIPS ($\downarrow$) \\
  \midrule
  ToMiE~\cite{zhan2024tomiemodulargrowthenhanced} & 30.40 & 0.934 & 0.079 \\
  Seq-Avatar~\cite{xu2025seqavatar} & 28.17 & 0.919 & 0.077  \\
  $R^3$-Avatar~\cite{Zhan2025R3AvatarRA} & 31.89 & 0.947 & 0.064 \\
  \midrule
  Ours & \textbf{32.32} & \textbf{0.981} & \textbf{0.052} \\
  \bottomrule
  \end{tabular}%
\end{table}

\section{Failure Cases}
\label{app:limitations}

\begin{figure}[h]
    \centering
    \setlength{\tabcolsep}{2pt}
    \begin{tabular}{cc}
        \includegraphics[width=0.15\linewidth]{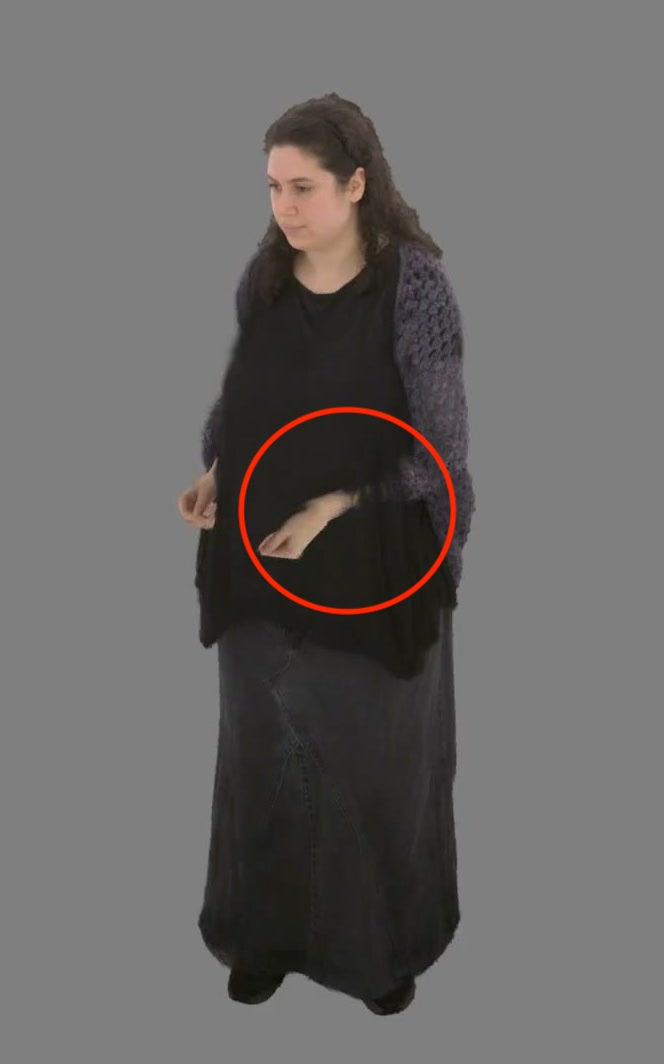} &
        \includegraphics[width=0.15\linewidth]{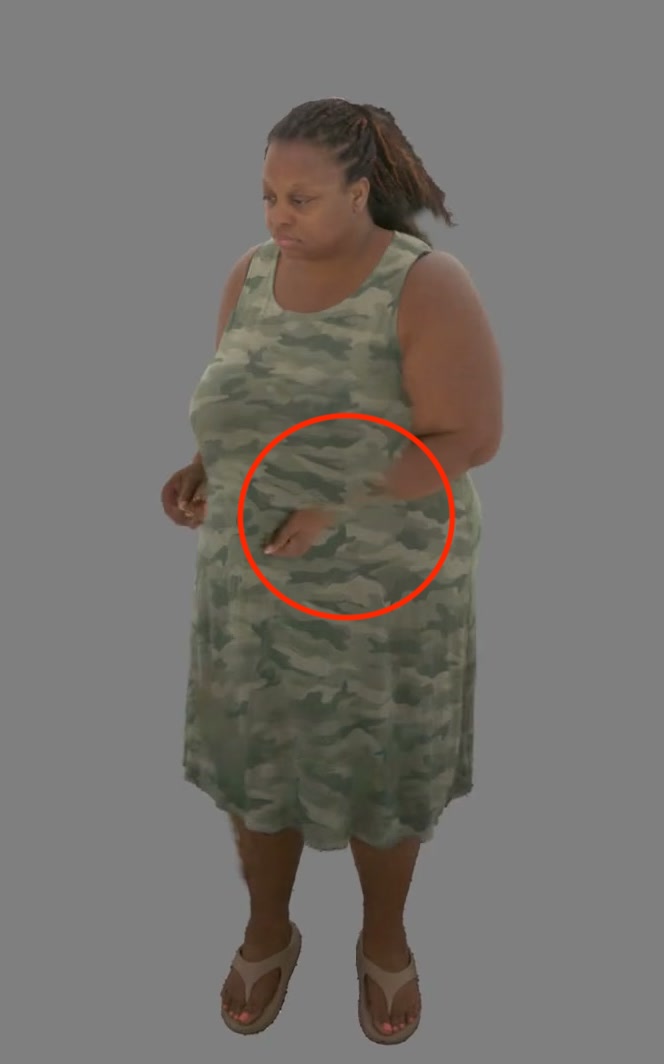} \\
    \end{tabular}
    \caption{\textbf{Failure cases.} Without an explicit contact prior, our method can produce body--clothing interpenetration and garment self-intersection on certain combinations of body shape, pose, and garment looseness.}
    \label{fig:failure_cases}
\end{figure}

We provide a concrete example for the contact-related limitation discussed in the main paper conclusion. Without an explicit contact prior, our model does not enforce hard constraints such as non-penetration; this is most visible in cross-driving, where transferring a fixed pose sequence to subjects with larger body shapes occasionally produces body--clothing interpenetration or garment self-intersection (Fig.~\ref{fig:failure_cases}).

\end{document}